\documentclass[journal]{new-aiaa}
\usepackage[utf8]{inputenc}
\usepackage{textcomp}
\usepackage{multirow}

\usepackage{graphicx}
\usepackage{float, caption, subcaption}
\usepackage{amsmath}
\usepackage{siunitx}
\usepackage{longtable,tabularx}
\usepackage{booktabs}
\usepackage{bm,stmaryrd,mathrsfs}
\usepackage[ruled,linesnumbered]{algorithm2e}
\newcommand\bbR{\mathbb{R}}

\newcommand{\bF}{\bm{F}}
\newcommand{\bn}{\bm{n}}

\newcommand{\bU}{\bm{U}}

\newcommand{\bV}{\bm{V}}

\newcommand\abs[1]{\lvert #1 \rvert}
\newcommand\norm[1]{\left\lvert\left\lvert #1 \right\rvert\right\rvert}

\newcommand{\DEIMI}{\bm{\mathcal{I}}}
\newcommand{\cddI}{\bm{\mathcal{\widetilde{I}}}}

\title{Machine learning enhanced real-time aerodynamic forces prediction based on sparse pressure sensor inputs}

\author{Junming Duan\footnote{Postdoctoral researcher, Chair of Computational Mathematics and Simulation Science, B\^atiment MA, Station 8, 1015 Lausanne}}
\affil{\'Ecole Polytechnique F\'ed\'erale de Lausanne, 1015 Lausanne, Switzerland}
\author{Qian Wang\footnote{Corresponding author. Assistant Professor, Mechanics Division, No.~10 East Xibeiwang Road, Beijing 100193, qian.wang@csrc.ac.cn}}
\affil{Beijing Computational Science Research Center, 100193 Beijing, China}
\author{Jan S. Hesthaven\footnote{Professor, Chair of Computational Mathematics and Simulation Science, B\^atiment MA, Station 8, 1015 Lausanne}}
\affil{\'Ecole Polytechnique F\'ed\'erale de Lausanne, 1015 Lausanne, Switzerland}

\begin{document}
	
\maketitle

% {\color{red}GUIDE:\\
% \url{https://www.aiaa.org/publications/journals/Journal-Author#prepare-your-manuscript}}

\begin{abstract}
%For autonomous navigation of unmanned aerial vehicles,
%accurate and efficient real-time unsteady aerodynamic prediction based on a small number of sensor inputs plays an important role.
%This paper proposes a machine learning-enhanced approach combined with discrete empirical interpolation method (DEIM) to predict aerodynamic coefficients.
%Pressure coefficients on the aircraft surface from numerical simulations serve as snapshots
%and are used to obtain a set of orthonormal modes.
%The first few most important sensor locations are selected by the DEIM utilizing the modes,
%then the pressure coefficients can be predicted based on real-time pressure sensor inputs at the selected locations.
%The DEIM model captures the main flow features,
%and a neural network is used to bridge the gap between the ground truth and the DEIM prediction,
%thus a more accurate machine learning-enhanced prediction model is obtained.
%The approach is tested on numerical and experimental dynamic stall data of a 2D NACA0015 airfoil,
%and numerical simulation data of the dynamic stall of a 3D drone.
Accurate prediction of aerodynamic forces in real-time is crucial for autonomous navigation of unmanned aerial vehicles (UAVs). This paper presents a data-driven aerodynamic force prediction model based on a small number of pressure sensors located on the surface of UAV.
The model is built on a linear term that can make a reasonably accurate prediction and a nonlinear correction for accuracy improvement.
The linear term is based on a reduced basis reconstruction of the surface pressure distribution, where the basis is extracted from numerical simulation data and the basis coefficients are determined by solving linear pressure reconstruction equations at a set of sensor locations. Sensor placement is optimized using the discrete empirical interpolation method (DEIM). Aerodynamic forces are computed by integrating the reconstructed surface pressure distribution. 
The nonlinear term is an artificial neural network (NN) that is trained to bridge the gap between the ground truth and the DEIM prediction, especially in the scenario where the DEIM model is constructed from simulation data with limited fidelity.
A large network is not necessary for accurate correction as the linear model already captures the main dynamics of the surface pressure field, thus yielding an efficient DEIM+NN aerodynamic force prediction model.
The model is tested on numerical and experimental dynamic stall data of a 2D NACA0015 airfoil, and numerical simulation data of dynamic stall of a 3D drone.
Numerical results demonstrate that the machine learning enhanced model can make fast and accurate predictions of aerodynamic forces using only a few pressure sensors, even for the NACA0015 case in which the simulations do not agree well with the wind tunnel experiments.
Furthermore, the model is robust to noise.
\end{abstract}

\section*{Nomenclature}

{\renewcommand\arraystretch{1.0}
\noindent\begin{longtable*}{@{}l @{\quad=\quad} l@{}}
$A$  & amplitude of pitching movement, \unit{deg} \\
$A_{\text{ref}}$ & reference area, \unit{m^2}\\
$\bm{b}$  & coefficient vector of reduced basis  \\
$C_d$ & drag coefficient $\frac{\text{drag force}}{\frac12\rho_\infty V_\infty^2 A_{\text{ref}}}$ \\
$C_l$ & lift coefficient $\frac{\text{lift force}}{\frac12\rho_\infty V_\infty^2 A_{\text{ref}}}$ \\
$C_p$ & pressure coefficient $\frac{p-p_{\infty}}{\frac12\rho_\infty V_\infty^2}$ \\
$\bm{C}_p$ & vector of pressure coefficients \\
$\overline{\bm{C}_p}$ & reference vector value of pressure coefficients \\
$f$  & pitching frequency, \unit{Hz} \\
$\bF$  & force on aircraft in body frame, \unit{N} \\
$\DEIMI$ & indices corresponding to the selected sensor locations \\
$\cddI$ & candidate indices for DEIM selection \\
$L(\ \cdot\ ; \Theta)$ & fully-connected layer \\
$m$  & number of candidate sensor locations \\
$M$  & number of snapshots \\
$\bm{M}^{\tt s}_F, \bm{M}^{0}_F$  & pre-computed matrices for DEIM prediction \\
$\bn$  & normal unit vector pointing into the aircraft in body frame \\
$N$  & number of available pressure coefficients on aircraft \\
$n_b$  & number of reduced bases \\
$\bn_d$  & unit vector in drag direction  \\
$\bn_l$  & unit vector in lift direction  \\
$n_L$  & number of layers in neural network \\
$n_s$  & number of selected sensors \\
$p$  & pressure, \unit{N} \\
$p_\infty$  & reference pressure, \qty{1e5}{N} \\
$\bm{R}$  & reconstruction matrix in DEIM \\
$Re$  & Reynolds number \\
$S$  & surface area, \unit{m^2} \\
$\mathbb{S}$  & snapshot matrix of pressure coefficients \\
$\overline{\mathbb{S}}$  & reference matrix of snapshot matrix \\
$t$  & time, \unit{s} \\
$\bU$  & reduced basis matrix \\
$V_\infty$  & freestream velocity, \unit{m/s} \\

\multicolumn{2}{@{}l}{Greek symbols}\\
$\alpha$  & angle of attack, \unit{deg} \\
$\alpha_0$  & initial angle of attack, \unit{deg} \\
$\alpha(\epsilon_{d, \infty}^z)$ & angle of attack corresponding to $\epsilon_{d,\infty}^z$, \unit{deg}, $z={\tt DEIM}, {\tt NN}$ \\
$\alpha(\epsilon_{l, \infty}^z)$ & angle of attack corresponding to $\epsilon_{l,\infty}^z$, \unit{deg}, $z={\tt DEIM}, {\tt NN}$ \\
$\epsilon_{d}^z$ & $\ell^2$ error in drag coefficient, $z={\tt DEIM}, {\tt NN}$ \\
$\epsilon_{d,\infty}^z$ & $\ell^\infty$ error in drag coefficient, $z={\tt DEIM}, {\tt NN}$ \\
$\epsilon_{l}^z$ & $\ell^2$ error in lift coefficient, $z={\tt DEIM}, {\tt NN}$ \\
$\epsilon_{l,\infty}^z$ & $\ell^\infty$ error in lift coefficient, $z={\tt DEIM}, {\tt NN}$ \\
$\epsilon_{\tt proj}$ & projection error on the reduced space \\
% $\mu$ & dynamic viscosity, kg/(m$\cdot$ s) \\
$\nu_\infty$ & freestream kinematic viscosity, \unit{m^2/s} \\
$\varphi$ & activation function \\
$\rho_\infty$  & freestream density, \unit{kg/m^3} \\
$\sigma$ & singular value \\
$\Theta$ & weights and biases in neural network \\

\multicolumn{2}{@{}l}{Superscripts}\\
$\tt DEIM$ & signifies DEIM \\
$\tt exper$ & signifies experiment \\
$\tt NN$ & signifies using neural network \\
$\tt rb$ & signifies reduced basis approximation \\
$\tt s$ & signifies sensor \\
$\tt URANS$ & signifies URANS \\

\multicolumn{2}{@{}l}{Subscripts}\\
$i$ & index of pressure coefficient locations \\
$j$ & index of layers in neural network \\
$k$ & index of singular values \\
$\ell$ & index of DEIM interpolation points \\
$\text{testing}$ & testing set \\
$\text{training}$ & training set \\
\end{longtable*}}

\section{Introduction}\label{sec:Introduction}
\lettrine{A}{utonomous} navigation of unmanned aerial vehicles (UAVs) is of great importance in real-world applications.
Global navigation satellite system (GNSS) aided inertial navigation system (INS) is the most popular navigation system for small UAVs \cite{Bryson_2015_UAV_InCollection}.
The INS contains inertial measurement units (IMUs) which are used to measure linear acceleration and angular rates of rotation,
based on accelerometers and gyroscopes.
From this, position, velocity and attitude can be obtained by integration,
but accuracy is limited due to an accumulation of errors in long-time integration, known as drift.
Filters in the INS provide a way to fuse the observations from the GNSS,
such as time, position, and velocity,
to control the drift.

In scenarios when GNSS outages occur,
e.g., UAVs passing through tunnels, high buildings, or forests,
the drift cannot be controlled and navigation errors increase fast \cite{Lau_2013_Inertial_IToAaES, Bryson_2015_UAV_InCollection},
leading to situations where UAVs may even cause danger to objects on the ground during long GNSS outages.
There is some past work that attempts to improve navigation performance during GNSS outages.
Advanced techniques were used to improve INS error modeling \cite{Nassar_2005_Wavelet_JoN, Nassar_2006_combined_GS, Noureldin_2009_Performance_VTITo},
but the improvements were limited.
Additional sensors were also adopted to aid the navigation system \cite{Yun_2013_IMU/Vision/Lidar_InProceedings,Madany_2013_Modelling_InProceedings,UijtdeHaag_2014_Flight_InProceedings},
adding extra costs and complexity to the system,
as well as platform dependence.
A vehicle dynamic model (VDM) is a mathematical model describing the dynamics of the platform,
constructed based on the physical laws of motion.
It does not rely on extra sensors,
while takes available quantities such as navigation states, UAV parameters, wind velocity, and control commands as input to compute the aerodynamics loads and then predict navigation states at the next time step that can be further combined with other observations by using filters.
VDMs have been integrated into navigation systems \cite{Koifman_1999_Inertial_IToCST, Bryson_2004_Vehicle_InProceedings, Vasconcelos_2010_Embedded_CEP, Crocoll_2014_Model_N, Crocoll_2014_Quadrotor_InProceedings, Khaghani_2016_Autonomous_N}
to improve navigation performance.
Compared to the conventional INS-GNSS system,
the VDM-INS-GNSS in \cite{Khaghani_2018_Assessment_RaAS} was shown to give better navigation performance during GNSS outage,
where the aerodynamic loads are computed using linear and quadratic polynomials of which the coefficients need to be calibrated for each specific UAV.

When the dynamics become highly nonlinear, e.g., UAVs at high angles of attack or in turbulent flows,
it is very difficult for the conventional VDM to work effectively,
as the aerodynamic models may be too simple to capture nonlinear dynamics.
One growing interest is to use distributed airflow sensors to aid the navigation and control system \cite{Fei_2007_Aircraft_SMaS, Shen_2013_Pitch_JoA, Magar_2016_Aerodynamic_B&B, Wood_2019_Distributed_JoA, Mark_2019_Review_JoA},
inspired by flying animals,
where many different sensors are distributed on the bodies and aerodynamic surfaces.
Artificial neural networks (NNs) were used in \cite{Fei_2007_Aircraft_SMaS, Magar_2016_Aerodynamic_B&B, Wood_2019_Distributed_JoA, AraujoEstrada_2021_Aerodynamic_JoA} to learn the dynamics based on sensor inputs.
The model in \cite{AraujoEstrada_2021_Aerodynamic_JoA} was shown to work also in the dynamic stall region,
where the dynamics are highly nonlinear and difficult to capture.
In the study of the navigation system based on distributed sensors,
it is often not practical or feasible to put too many sensors on the UAVs,
and a large NN is necessary if the sensor locations are not well selected.
Therefore, the locations of the sensors must be carefully optimized.
The sensors are usually placed near the leading edge \cite{AraujoEstrada_2017_Bio_InProceedings} as the pressure gradient is the highest in this part.
For more discussions on sensor arrangements, the reader is referred to the review article \cite{Mark_2019_Review_JoA} and references therein.

Aerodynamic forces can be also computed in real-time by integrating a fast surface flow field reconstruction obtained through reduced-order modeling and sensor measurements \cite{Gong_2021_Optimal_NEaD,Zhao_2022_Sparse_EiF}.
For a high Reynolds number flow around an aircraft, the contribution of viscous stress to lift and drag forces can be neglected as compared to pressure.
Therefore, a surface pressure reconstruction is sufficient for accurate aerodynamic force estimation.
The surface pressure can be approximated by a linear combination of reduced basis functions that are extracted from a collection of snapshots via principle component analysis methods such as proper orthogonal decomposition (POD) \cite{Berkooz_1993_proper_ARoFM}. 
The basis coefficients are determined by solving linear pressure reconstruction equations at a set of sensor locations.
Sensor placement can be optimized by algorithms such as gappy POD \cite{Willcox_2006_Unsteady_C&f}, discrete empirical interpolation method (DEIM) \cite{Chaturantabut_2010_Nonlinear_SJoSC}, generalized empirical interpolation method (GEIM) \cite{Argaud_2018_Sensor_JoCP}, and particle swarm optimization (PSO) \cite{Zhao_2022_Sparse_EiF}.
High resolution in space of the snapshots or basis functions is required for accurate computation of aerodynamic forces as it is based on the integration of the reduced basis approximation.
However, it is often very difficult and expensive to put a large number of sensors on the surface of an aircraft.
Therefore, snapshots are collected from computational fluid dynamics (CFD) simulations. For vortex-dominated flow phenomena such as dynamic stall, expensive high-fidelity simulations are needed to capture small-scale flow structures and offer good agreement with experiments.
Towards an efficient VDM for UAVs, it is necessary to develop algorithms to accurately predict the aerodynamic forces using an inaccurate basis and sparse sensor measurements.

This paper proposes a systematic approach to construct a GNSS-free surrogate aerodynamic model to predict the aerodynamic coefficients based on sparse pressure sensor inputs,
employing data fusion from numerical simulations and experiments.
The number of sensors is limited, especially in 3D,
so we generally cannot use experimental data to select the sensor locations and compute the aerodynamic coefficients by integration on the surface.
Usually, there are plenty of numerical simulation data for different aerodynamic states,
called snapshots, available on the whole surface of the UAVs,
thus the sensor locations can be selected by 
discrete empirical interpolation method (DEIM) \cite{Chaturantabut_2010_Nonlinear_SJoSC} using the pressure coefficients of the numerical snapshots.
The real-time pressure sensor values can then be collected at the selected locations,
and the surface flow fields and aerodynamic coefficients can be obtained from the DEIM model based on such given sensor inputs.
As the DEIM model uses a linear approximation,
it is not accurate enough when the number of sensors is small or the dynamics are highly nonlinear.
Therefore, an artificial NN is employed to correct the aerodynamic coefficients from the DEIM model with the help of the experimental data.
To be specific, the NN learns the map from the pressure sensor inputs to the error in the aerodynamic force prediction,
and the size of the NN is small so the training and online prediction can be very efficient.
In other words, the linear model obtained by the DEIM provides a rough approximation of the aerodynamic coefficients,
and the NN models the unresolved parts, which is calibrated by the experimental data,
leading to an accurate and efficient model.
The proposed approach is platform-independent and GNSS-free, without additional costs and weights on the UAVs. The test results show that the DEIM+NN model is capable of making fast and accurate aerodynamic force predictions,
and not sensitive to noise in the pressure sensor inputs.

This paper is organized as follows.
Section \ref{sec:Location} presents the surface pressure reconstruction and sensor placement optimization based on the DEIM.
Section \ref{sec:ML} describes the proposed machine learning-enhanced approach for the prediction of the aerodynamic coefficients.
The approach is applied to the dynamic stall of a 2D NACA0015 airfoil and a 3D drone in Section \ref{sec:Result},
and concluding remarks are given in Section \ref{sec:Conclusion}.

\section{Surface pressure reconstruction based on discrete empirical interpolation method (DEIM)}\label{sec:Location}
Assume that the vector of pressure coefficients $\bm{C}_p\in\bbR^{N}$ are measured at $N$ locations on the surface of the aircraft,
and the matrix $\bm{S}$ consists of the corresponding surface area scaled normal vectors
\begin{equation*}
    \bm{S} = \dfrac{1}{A_{\text{ref}}}
    [\bn_{1} S_1, \cdots, \bn_{N} S_N]
    \in\bbR^{3\times N},
\end{equation*}
where $\bn_i$ is unit normal vector, and $S_i$ is surface area.
Notice that in 2D, the last component of $\bn_i$ is zero.
Then the force coefficients in the body frame due to pressure can be computed by numerical integration
\begin{equation}\label{eq:Force}
    \bF = \bm{S}\bm{C}_p \in \bbR^{3}.
\end{equation}
The lift and drag coefficients can be obtained by
\begin{equation}\label{eq:ClCd}
    C_l = \bF\cdot \bn_l,\quad
    C_d = \bF\cdot \bn_d.
\end{equation}
The sensors are usually sparse on the aircraft,
in the sense that it is impractical to put too many sensors on the UAVs.
Thus it is of interest to use a small number of surface pressure measurements to predict the surface pressure field,
and compute the aerodynamic coefficients.

This work chooses to predict the surface pressure coefficients via a reduced basis approximation 
\begin{equation*}
    \bm{C}_p \approx \bm{C}^{\tt rb}_p= \overline{\bm{C}_p}+ \sum_{i=1}^{n_b} b_i \bm{u}_i= \overline{\bm{C}_p}+ \bU\bm{b},
\end{equation*}
where $\bU=\left[\bm{u}_1, \bm{u}_2, \cdots, \bm{u}_{n_b}\right] \in\bbR^{N\times n_b}$ and $\bm{b}=\left[b_1,b_2, \cdots, b_{n_b}\right]^\mathrm{T} \in \bbR^{n_b}$.
The reduced basis is extracted from a collection of snapshots $\mathbb{S} \in\bbR^{N\times M}$ obtained through numerical simulations, with each column being an instantaneous pressure coefficient vector.
The reference value $\overline{\bm{C}_p}$ is taken as the average of the columns of $\mathbb{S}$ in this paper.
The basis functions can be obtained through proper orthogonal decomposition (POD) \cite{Berkooz_1993_proper_ARoFM} based on a singular value decomposition (SVD)
\begin{equation*}
    \mathbb{S} - \overline{\mathbb{S}} = \widetilde{\bU}\Sigma\widetilde{\bV}^\mathrm{T},\quad
    \Sigma = \text{diag}\{\sigma_1,\cdots,\sigma_r\},
\end{equation*}
where $\overline{\mathbb{S}}= [\overline{\bm{C}_p},\cdots,\overline{\bm{C}_p}]\in \bbR^{N \times M}$,
$\sigma_1\geqslant\cdots\geqslant\sigma_r\geqslant 0$, $r=\min\{N,M\}$.
The orthonormal reduced basis functions $\bU$ are the first $n_b$ columns of $\widetilde{\bU}$.
Given a series of pressure sensor values $\bm{C}_p^{\tt s}\in\bbR^{n_s}$ at locations $\DEIMI=\left\{\mathcal{I}_1, \cdots, \mathcal{I}_{n_s}\right\}\subset\{1,2,\cdots,N\}$, the basis coefficients $\bm{b}$ can be determined by solving a linear system
\begin{equation*}
	\overline{\bm{C}_p}\left(\DEIMI\right) + \bU\left(\DEIMI,:\right) \bm{b} = \bm{C}_p^{\tt s},
\end{equation*}
where MATLAB notation is used.
In this paper, the number of sensors is taken as the number of the basis functions, i.e., $n_s=n_b$, resulting in a square linear system.
Therefore, the basis coefficients are $\bm{b}=  \bU^{-1}(\DEIMI,:)\left(\bm{C}_p^{\tt s}-\overline{\bm{C}_p}\left(\DEIMI\right)\right)$ and the predicted surface pressure coefficients are
\begin{equation}\label{eq:RB_DEIM}
    \bm{C}^{\tt rb}_p = \overline{\bm{C}_p} + \bU\bU^{-1}(\DEIMI,:)\left(\bm{C}_p^{\tt s}-\overline{\bm{C}_p}\left(\DEIMI\right)\right) = \overline{\bm{C}_p} + \bm{R}\left(\bm{C}_p^{\tt s}-\overline{\bm{C}_p}\left(\DEIMI\right)\right),
\end{equation}
where $\bm{R} = \bU\bU^{-1}(\DEIMI,:) \in \bbR^{N\times n_s}$.
Substituting \eqref{eq:RB_DEIM} into \eqref{eq:Force}, the aerodynamic force coefficients are obtained as
\begin{equation}\label{eq:DEIM_force}
    \bF = \bm{M}^{\tt s}_F \bm{C}_p^{\tt s} + \bm{M}^{0}_F,
\end{equation}
where $\bm{M}^{\tt s}_F =\bm{S}\bm{R} \in \bbR^{3 \times n_s}$ and $\bm{M}^{0}_F = \bm{S}\overline{\bm{C}_p}-\bm{S}\bm{R}\overline{\bm{C}_p}\left(\DEIMI\right) \in \bbR^{3}$.
One concludes from \eqref{eq:DEIM_force} that the overall computational cost of the prediction is $\mathcal{O}(6n_s)$,
as $\bm{M}^{\tt s}_F$ and $\bm{M}^{0}_F$ can be computed during an offline stage,
and real-time online prediction is fast when a small number of sensors are used.

The sensor locations $\DEIMI$ need to be optimized for surface pressure reconstruction accuracy. This work adopts the DEIM \cite{Chaturantabut_2010_Nonlinear_SJoSC} in Algorithm \ref{alg:DEIM} to select the sensor locations,
which is a greedy algorithm to find the most important interpolation points.
Here the $n_s$ sensors are selected from $m$ candidate locations.
 
\begin{algorithm}[H]
\SetAlgoLined
\KwIn{Orthonormal bases $\bU\in\bbR^{N\times n_s}$, candidate indices $\cddI=\{\mathcal{\widetilde{I}}_1, \cdots, \mathcal{\widetilde{I}}_{m}\}\subset\{1,2,\cdots,N\}$}
\KwOut{Selected indices $\DEIMI=\left\{\mathcal{I}_1, \cdots, \mathcal{I}_{n_s}\right\}\subset\cddI$}
$\mathcal{I}_1 = \mathop{\arg\max}\limits_{\cddI}\{\abs{\bU(\cddI,1)}\},
\DEIMI=[\mathcal{I}_1]$\;
\For{$\ell=2,\cdots,n_s$}{
	Solve $\bU(\DEIMI,:(\ell-1))\bm{c} = \bU(\DEIMI,\ell)$ for $\bm{c}$\;
	$\mathcal{I}_\ell = \mathop{\arg\max}\limits_{\cddI}
	\{ \abs{\bU(:,\ell) - \bU(:,:(\ell-1))\bm{c}} \}$\;
	$\DEIMI=[\bm{\mathcal{I}}, \mathcal{I}_\ell]$.
}
\caption{DEIM}
\label{alg:DEIM}
\end{algorithm}

%Then the pressure coefficients are predicted by
%$\bm{C}_p^{\tt DEIM} = \bm{R}\bm{C}_p^{\tt s} + \overline{\bm{C}_p}$,
%for given vector of pressure coefficients $\bm{C}_p^{\tt s}\in\bbR^{n_s}$ at the selected locations $\DEIMI$. 

\section{Aerodynamic coefficients prediction enhanced by machine learning}\label{sec:ML}

\begin{figure}[hbt!]
\centering
\includegraphics[width=0.6\textwidth]{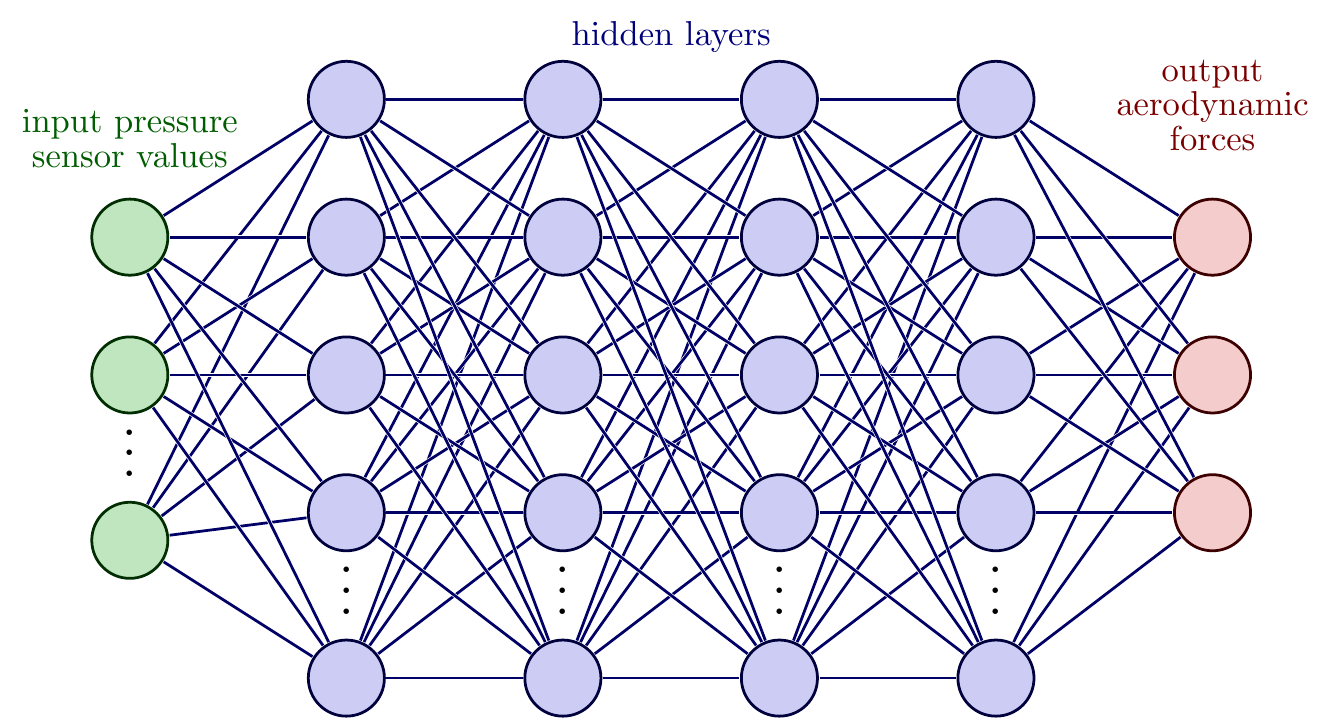}
\caption{A sketch of the fully-connected NN used in this work.}
\label{fig:fcnn}
\end{figure}

\begin{figure}[hbt!]
\centering
\includegraphics[width=\textwidth]{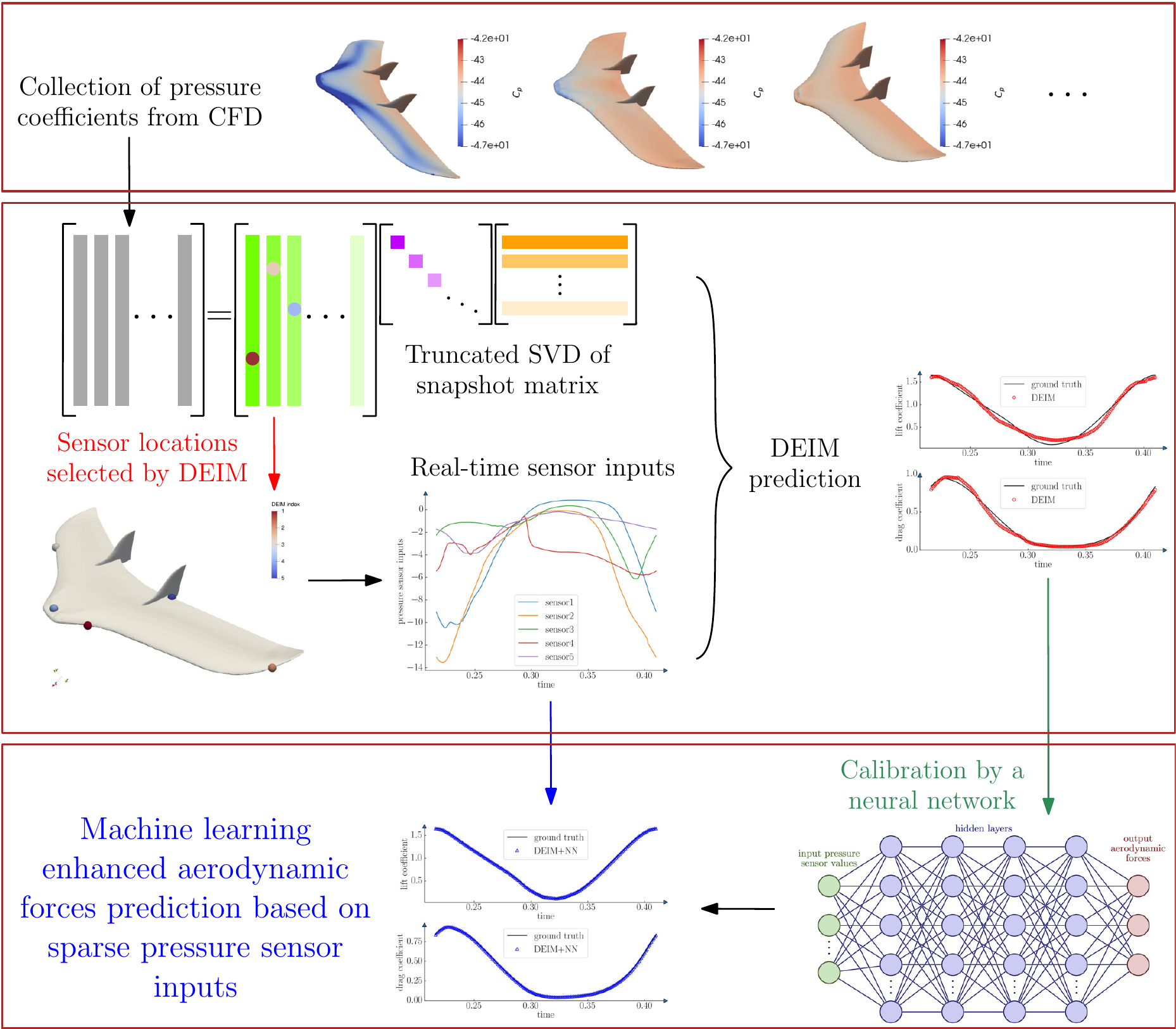}
\caption{The workflow of the proposed machine learning-enhanced aerodynamic forces prediction based on sparse pressure sensor inputs.}
\label{fig:workflow}
\end{figure}

The characteristics of the flow field can be highly nonlinear and very complex,
e.g., in the region of dynamic stall.
In such cases, the DEIM prediction is not efficient as a large number of bases need to be used, corresponding to many sensors,
since a linear subspace is employed in the reduced basis approximation.
The approximation ability of the NN has been exploited in many tasks due to its nonlinear nature,
and we propose to use an NN as a correction term
together with the DEIM prediction.
To be specific,
\begin{equation*}
    \bF = \bF^{\tt DEIM}
    + \text{NN}(\bm{C}_p^{\tt s};\Theta),
\end{equation*}
where the input of the NN is the pressure sensor values at the selected locations,
and the output is the difference in the aerodynamic force prediction coefficients between the ground truth $\bF$ and the DEIM prediction $\bF^{\tt DEIM}
    = \bm{M}^{\tt s}_F \bm{C}_p^{\tt s} + \bm{M}^{0}_F$.
Figure \ref{fig:fcnn} presents a sketch of the architecture of the fully-connected NN used in this work.
The NN can be expressed as
\begin{equation*}
    \text{NN} (\ \cdot\ ; \Theta) = L_{n_L}(\ \cdot\ ; \Theta_{n_L}) \circ 
    \varphi_{n_L-1}\left(L_{n_L-1}(\ \cdot\ ; \Theta_{n_L-1})\right) \circ \cdots \varphi_1\left(L_{1}(\ \cdot\ ; \Theta_{1})\right),
\end{equation*}
where $L_{j}$ is a fully-connected layer, $j=1,\cdots,n_L$, with the weights and biases $\Theta_{j}$.
The activation function $\varphi_j$ is chosen as ReLU in this paper.
% \begin{equation*}
%     y_j^{(k)} = \sigma^{(k)}\left(\sum_{z=1}^{n_{z-1}}
%     \bm{W}_{jz}^{(k)}y_{z}^{(k-1)} + b_{j}^{(k)}\right)
% \end{equation*}
As the main dynamics has been captured by the DEIM model and the input dimension $n_s$ is low,
the size of the NN is not large, which means that the offline training of the NN is efficient and the online evaluation is fast.
It should be mentioned that the ground truth is obtained from numerical or experimental data in this paper,
and the NN correction term can be viewed as a closure for the complex behavior between the DEIM prediction and ground truth,
e.g., viscous forces, deviation of the numerical simulations, uncertainty in the experiments, or other missing effects.
The NN is trained to minimize the error in the training set
\begin{equation*}
    \Theta = \arg\min\sum\limits_{\substack{t\in t_{\tt training}\\ f\in f_{\tt training}}}
        \norm{\bF^{\tt DEIM}(t, f)
	+ \text{NN}(\bm{C}_{p}^{\tt s}(t, f);\Theta)
	- \bF(t, f)}_2^2.
\end{equation*}
Our implementation of the NN is built on the PyTorch library \cite{Paszke_2019_PyTorch_InProceedings}.
In all the tests, the mini-batch ADAM optimizer with an initial learning rate of $0.001$ is adopted to train the NN,
and the StepLR scheduler with step size $50$ and decay rate $0.95$ is used,
so that the learning rate is $\lambda=0.001\times 0.95^{\lfloor n/50\rfloor}$, where $n$ is the number of epochs.
An early stopping technique is used to avoid overfitting,
i.e., the training is terminated if the error in the validation set has not improved for $100$ epochs.
We also use the weight decay technique implemented in PyTorch as a regularization.
Notice that the force coefficients in the body frame are used to train the NN,
so that the angle of attack is not included in the input of the NN.
The final outputs are transformed based on the angle of attack to obtain the lift and drag coefficients
\begin{equation*}
    C_l^{\tt NN} = \left(\bF^{\tt DEIM}
    + \text{NN}(\bm{C}_p^{\tt s};\Theta)\right)\cdot\bn_l,\quad
    C_d^{\tt NN} = \left(\bF^{\tt DEIM}
    + \text{NN}(\bm{C}_p^{\tt s};\Theta)\right)\cdot\bn_d.
\end{equation*}
The entire workflow is shown in Fig. \ref{fig:workflow}.

\section{Applications}\label{sec:Result}

\subsection{Numerical simulation setup}

\begin{figure}[hbt!]
    \centering
    \includegraphics[width=0.5\linewidth]{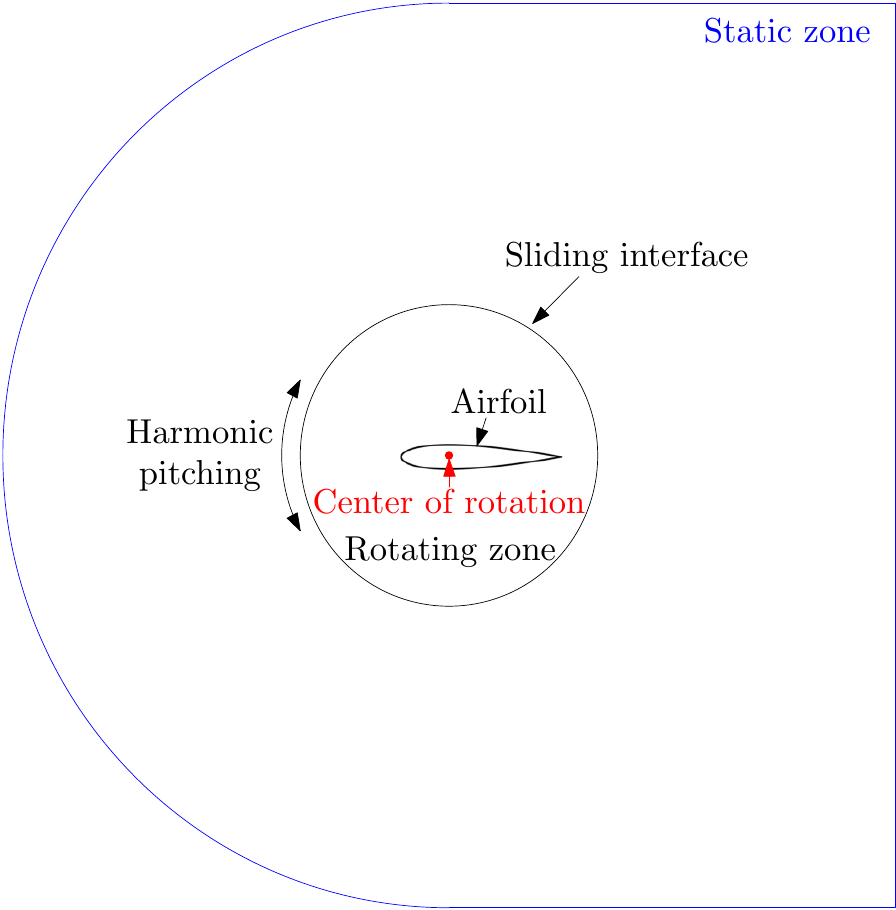}
    \caption{A sketch of the numerical simulation of the 2D NACA0015 airfoil.}
    \label{fig:2DAirfoil_simulation_sketch}
\end{figure}

In this work, the numerical snapshots of the transient aerodynamic flows are computed by using the open-source CFD solver OpenFOAM \cite{_2022_OpenFOAM} (v2112),
which solves the incompressible unsteady Reynolds-Averaged Navier–Stokes equations (URANS) in an arbitrary Lagrangian-Eulerian (ALE) framework,
with the $k$-$\omega$-Shear Stress Transport (SST) turbulence model \cite{Menter_2003_Ten_Thamt} as a closure.
The cyclic arbitrary mesh interface (AMI) is used to model the sliding interface between the static zone and the rotating zone,
where the airfoil or drone is located at the center of the rotating zone.
The PIMPLE solver is adopted for the transient simulations,
while the flow fields are initialized by using the steady-state solutions from the SIMPLE solver.
The movement of the 2D airfoil and 3D drone is modeled by the harmonic pitching
\begin{equation*}
    \alpha(t) = \alpha_0 + A\sin(2\pi f t).
\end{equation*}
A sketch of the numerical simulation of the 2D NACA0015 airfoil is shown in Fig. \ref{fig:2DAirfoil_simulation_sketch}

% combines the Pressure-Implicit with Splitting of Operators (PISO) and Semi-Implicit Method for Pressure-Linked Equations (SIMPLE)
% \begin{align}
%     &\nabla\cdot(\bu - \bu_g) = 0, \\
%     &\pd{\bu}{t} + \nabla\cdot[(\bu - \bu_g)\bu] = -\dfrac{1}{\rho}\nabla p + \nabla\cdot(\nu_{\text {eff}}\nabla \bu),
% \end{align}
% where $\bu_g$ is the grid velocity.
% The laminar kinematic viscosity $\nu$ is combined with the turbulent kinematic viscosity $\nu_t$, which accounts for the turbulent stresses arising from the Reynolds-averaged eddy viscosity turbulence model, yielding the effective kinematic viscosity $\nu_{\text{eff}}$.
% The well-known $k$-$\omega$ shear stress transport ($k$-$\omega$-SST) model \cite{Menter2003ten} is used for the closure of turbulence quantities.
% For convenience, the model is written in the following form
% \begin{align}
%     &\pd{k}{t} + \nabla\cdot[(\bu - \bu_g)k]
%     = \dfrac{1}{\rho}P_k - \beta^{*}\omega k
%     + \nabla\cdot[(\nu + \nu_t\alpha_k)\nabla k],\\
%     &\pd{\omega}{t} + \nabla\cdot[(\bu - \bu_g)\omega]
%     = \dfrac{\rho C_1 P}{\nu_t}
%     - C_2\omega^2
%     + \nabla\cdot[(\nu + \nu_t\alpha_\omega)\nabla\omega]
%     + \dfrac{2\alpha_\epsilon(1-F_1)}{\omega}\nabla k \cdot \nabla\omega,
% \end{align}
% with the blending function $F_1$,
% which is used to switch between
% the $k$-$\omega$ model inside the boundary layer
% and the $k$-$\epsilon$ model away from the surface.

\subsection{2D NACA0015 airfoil}
The approach is first verified by predicting the aerodynamic coefficients in dynamic stall of a 2D NACA0015 airfoil.
The experimental data were provided by He et al. \cite{He_2020_Stall_AJ},
collected based on a recirculating wind tunnel with an open jet test section.
This paper only uses the data in the pitching movement of the airfoil,
while the experiment in \cite{He_2020_Stall_AJ} also considered the flapping of the trailing edge.
It should be mentioned that due to the use of the open jet test section,
the experimental aerodynamic coefficients must be corrected to recover the case of a full wind tunnel.
Two cases are considered,
where the DEIM models are built on the pressure coefficients from the experiment and URANS simulation, respectively.
The former is used to verify the effectiveness of using NN as a correction term,
while the latter follows the proposed approach based on data fusion from the numerical simulation and experiment.
Note that the input pressure coefficients do not contain viscous forces,
so the NN is also used to model the viscous effects.
In the second case, the sensor locations and DEIM model are obtained from the URANS simulation data,
while the prediction takes the experimental pressure coefficients as input,
thus the NN also models the deviation between the URANS simulation and the experiment.

\begin{figure}[hbt!]
\centering
    \begin{subfigure}[t]{0.49\textwidth}
        \centering
        \includegraphics[width=0.765\linewidth]{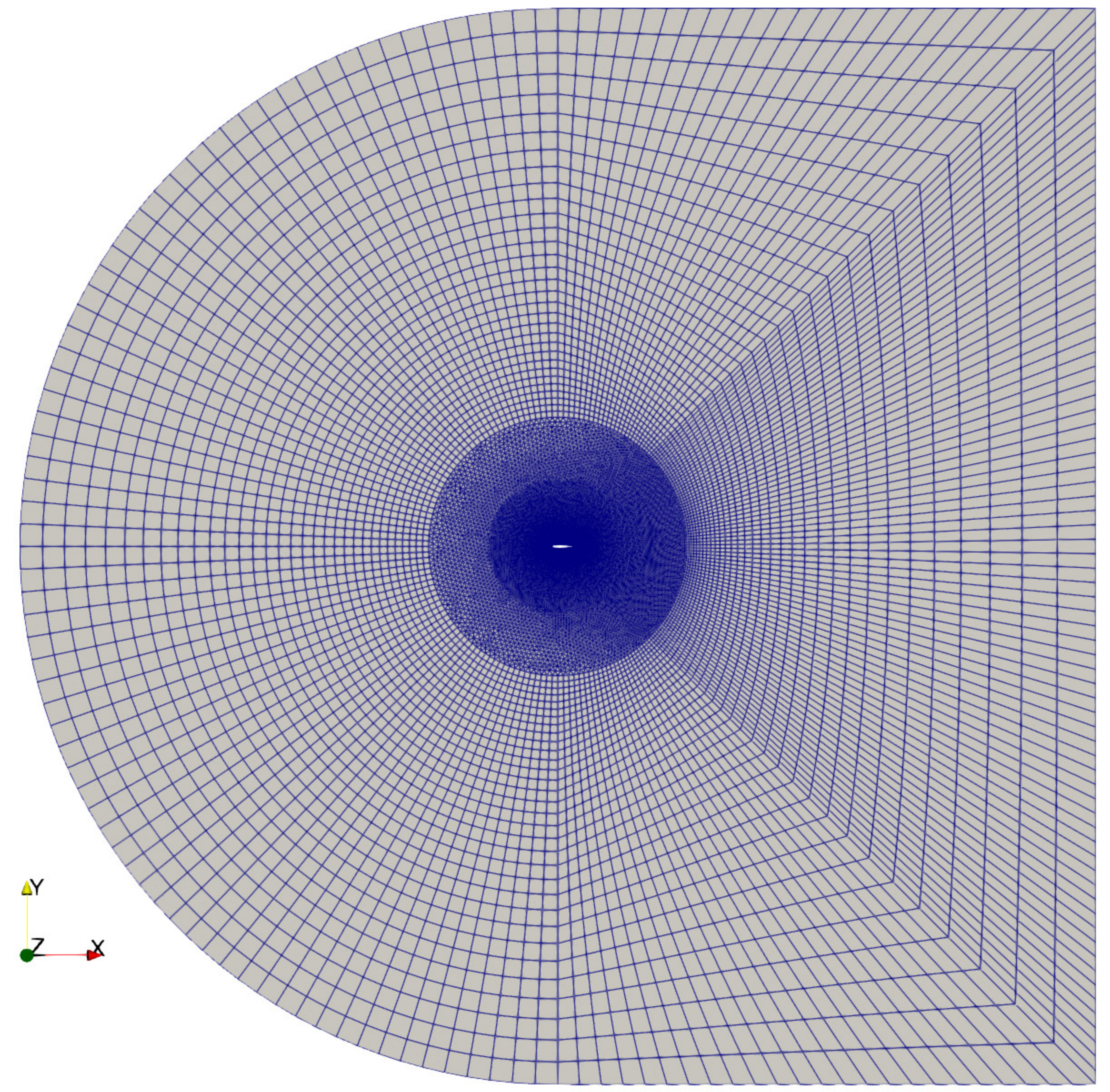}
        \caption{The whole computational domain.}
    \end{subfigure}
    \begin{subfigure}[t]{0.49\textwidth}
        \centering
        \includegraphics[width=1.0\linewidth]{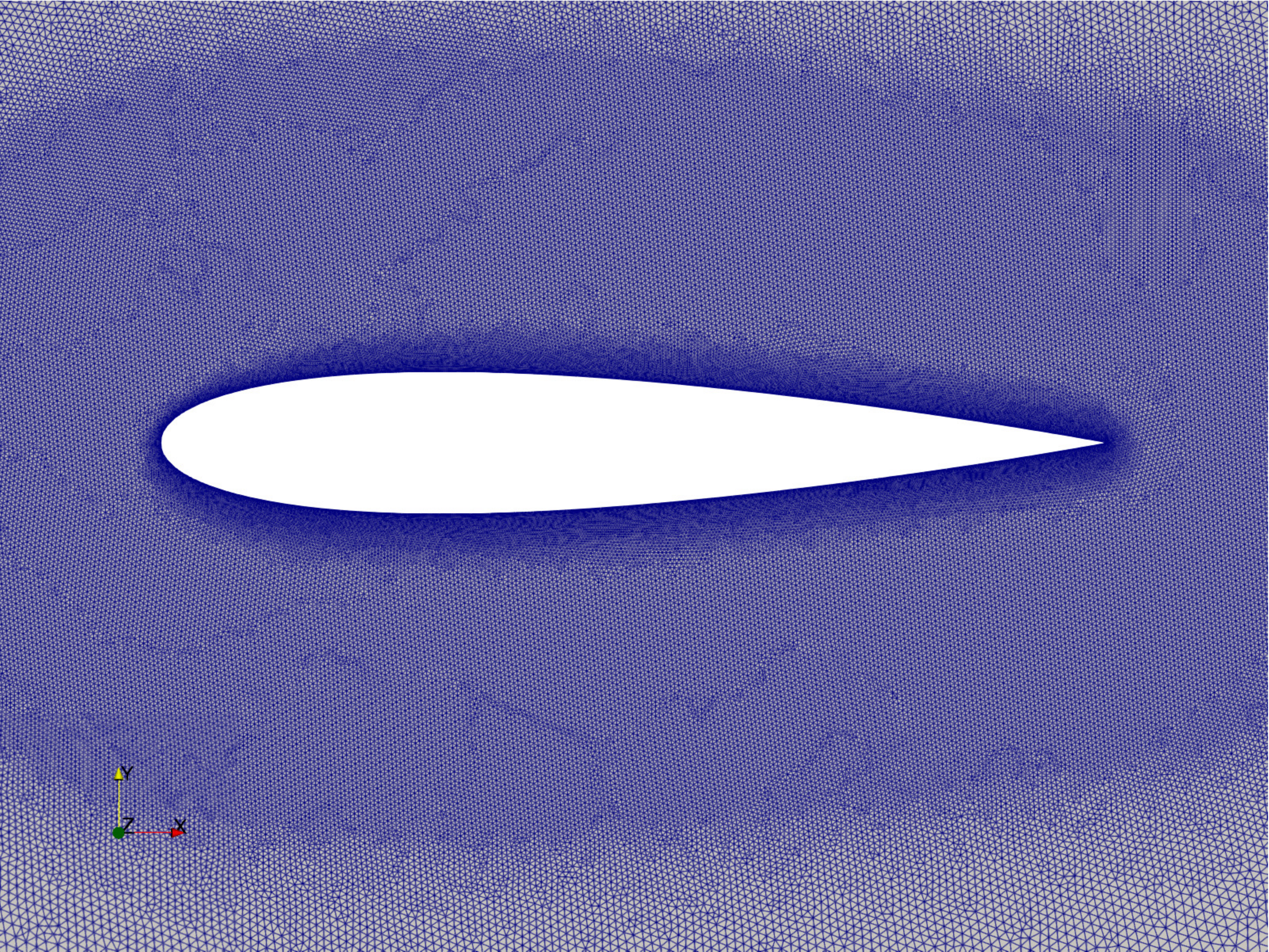}
        \caption{Close view of the mesh.}
    \end{subfigure}

    \begin{subfigure}[t]{0.49\textwidth}
        \centering
        \includegraphics[width=1.0\linewidth]{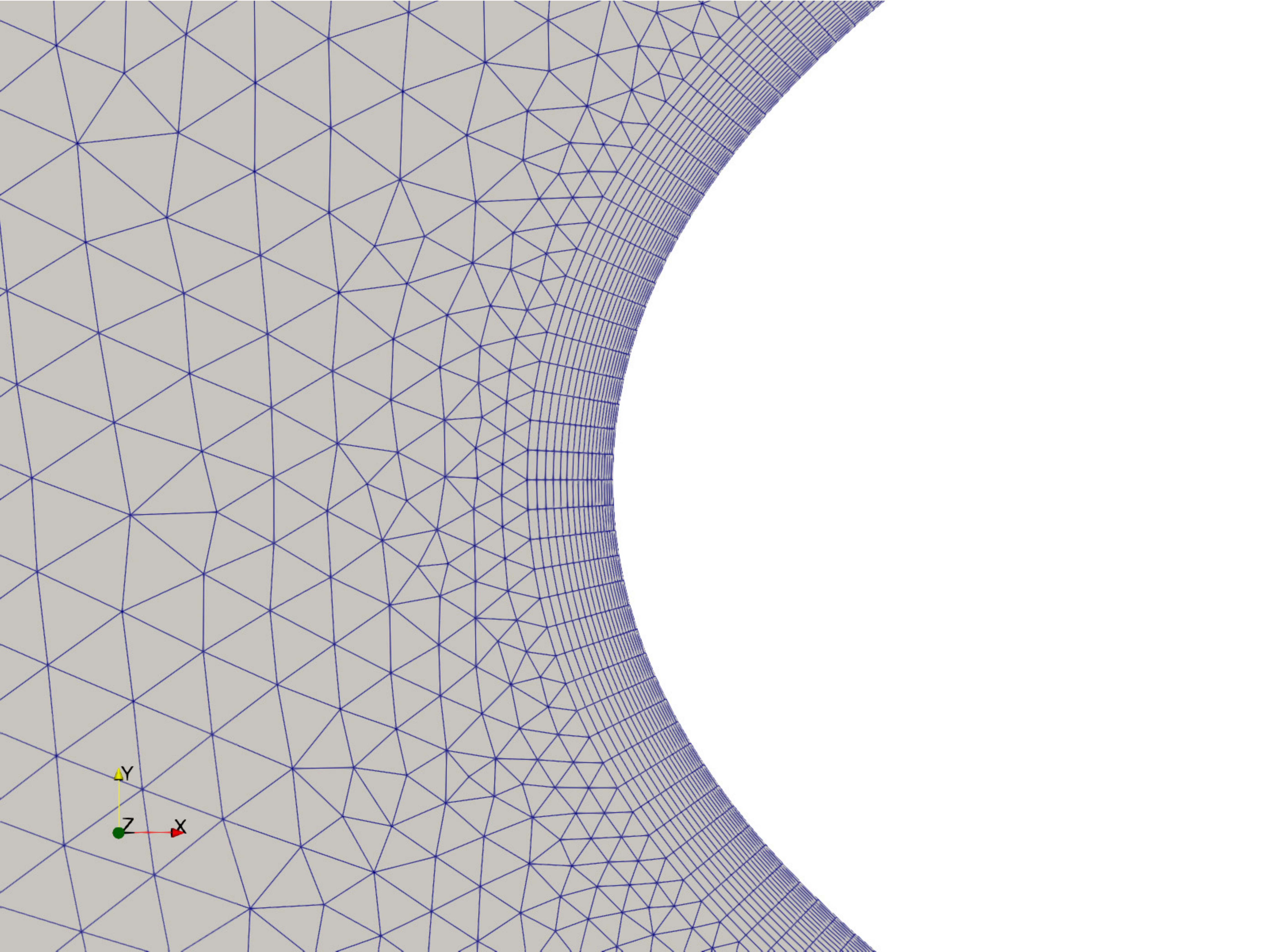}
        \caption{Further close view of the mesh.}
    \end{subfigure}
    \begin{subfigure}[t]{0.49\textwidth}
        \centering
        \includegraphics[width=1.0\linewidth]{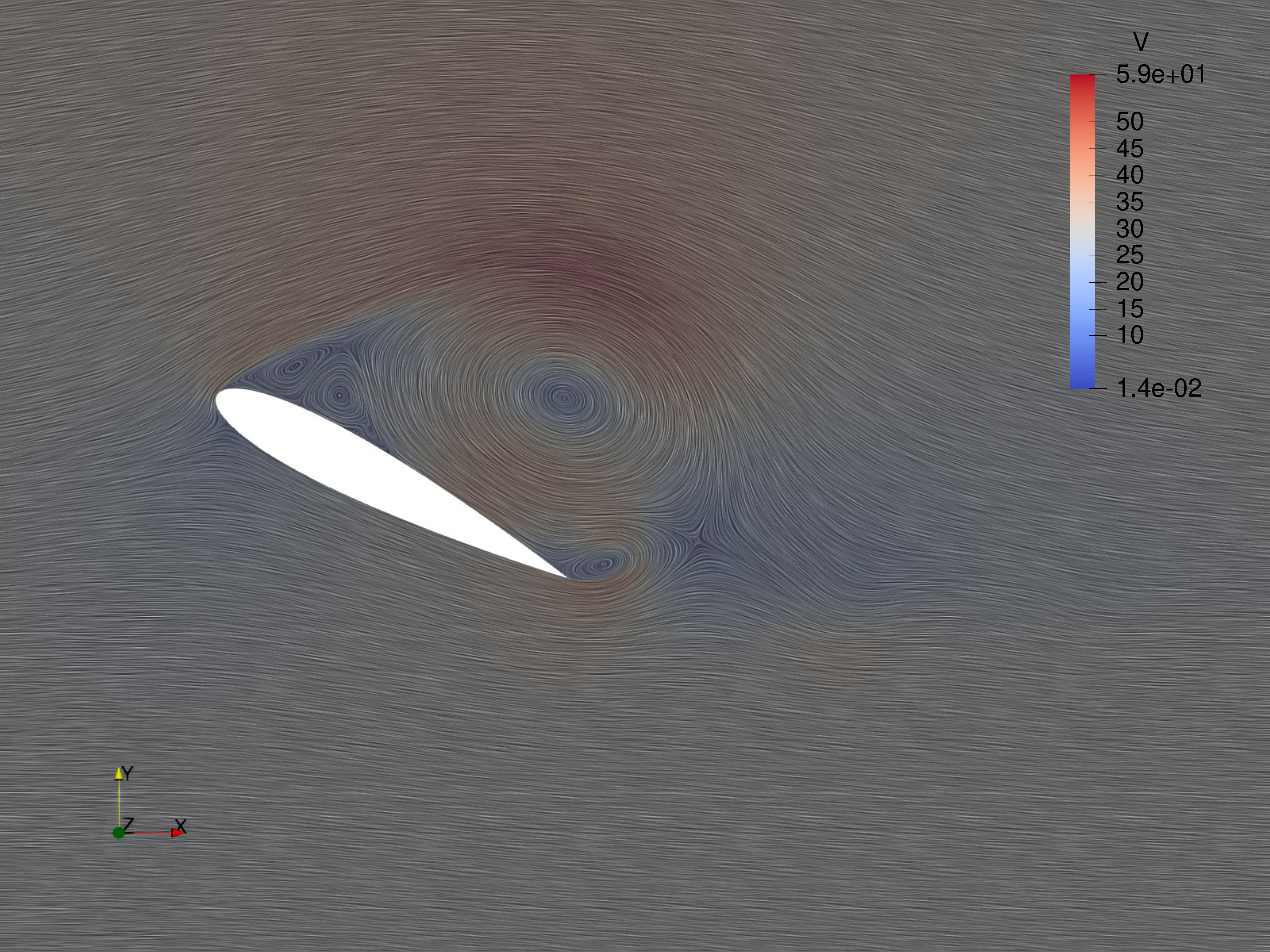}
        \caption{Visualization using surface line integral convolution (LIC) colored by the magnitude of the velocity.}
    \end{subfigure}
\caption{Computational mesh and URANS results with $f = \qty{3.979}{Hz}$ at $t = \qty{0.3}{s}$.}
\label{fig:2DAirfoil_mesh_flow_field}
\end{figure}

In this test, the parameters in the pitching movement are chosen as $\alpha_0 = \qty{20}{deg}$, $A = \qty{8}{deg}$,
and the freestream conditions are $\rho_\infty = \qty{1.146}{kg/m^3}$, $V_\infty = \qty{30}{m/s}$, $\nu_\infty = \qty{1.655e-5}{m^2/s}$.
The reference surface area is $A_{\text{ref}} = \qty{0.0225}{m^2}$ with the chord length $\qty{0.3}{m}$,
so that the Reynolds number based on the chord length is $Re=5.4\times 10^{5}$. %Re = V*c/\nu
The pitching frequencies in the experiment are $f = 0.796$, $2.387$, $3.183$, $4.775$, $1.592$, $3.979$ \unit{Hz},
where the first $4$ frequencies are used for training the NN,
and the last two are used for validation and testing, respectively.
In the first case, the training frequencies are used to build the DEIM model.
In the second case, the URANS results with $10$ pitching frequencies uniform in $0.5$ to \qty{5}{Hz} are used to obtain the DEIM model.
The computational mesh is generated using Gmsh \cite{Geuzaine_2009_Gmsh_IJfNMiE},
consisting of about $2.35\times 10^{5}$ cells and $8.38\times 10^{5}$ faces,
as shown in Fig. \ref{fig:2DAirfoil_mesh_flow_field}.
The visualization of the velocity field with a specific frequency and time using surface line integral convolution (LIC) is also presented,
which clearly shows that the complex flow separation happens near the upper surface and trailing edge of the airfoil.

\begin{figure}[hbt!]
\centering
    \begin{subfigure}[t]{0.33\textwidth}
        \centering
        \includegraphics[width=1.0\linewidth]{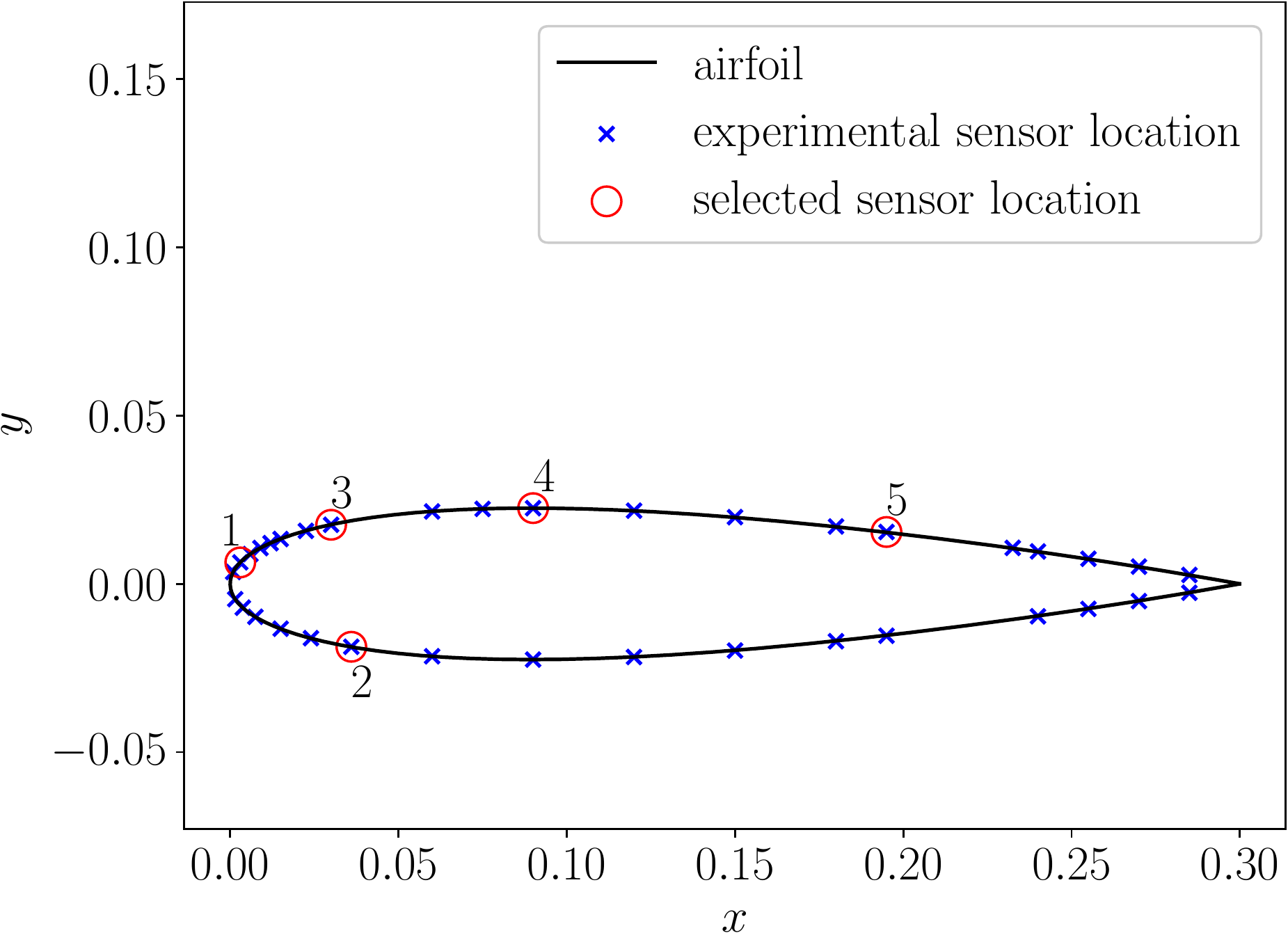}
         \caption{$n_s=5$}
    \end{subfigure}
    \begin{subfigure}[t]{0.33\textwidth}
        \centering
        \includegraphics[width=1.0\linewidth]{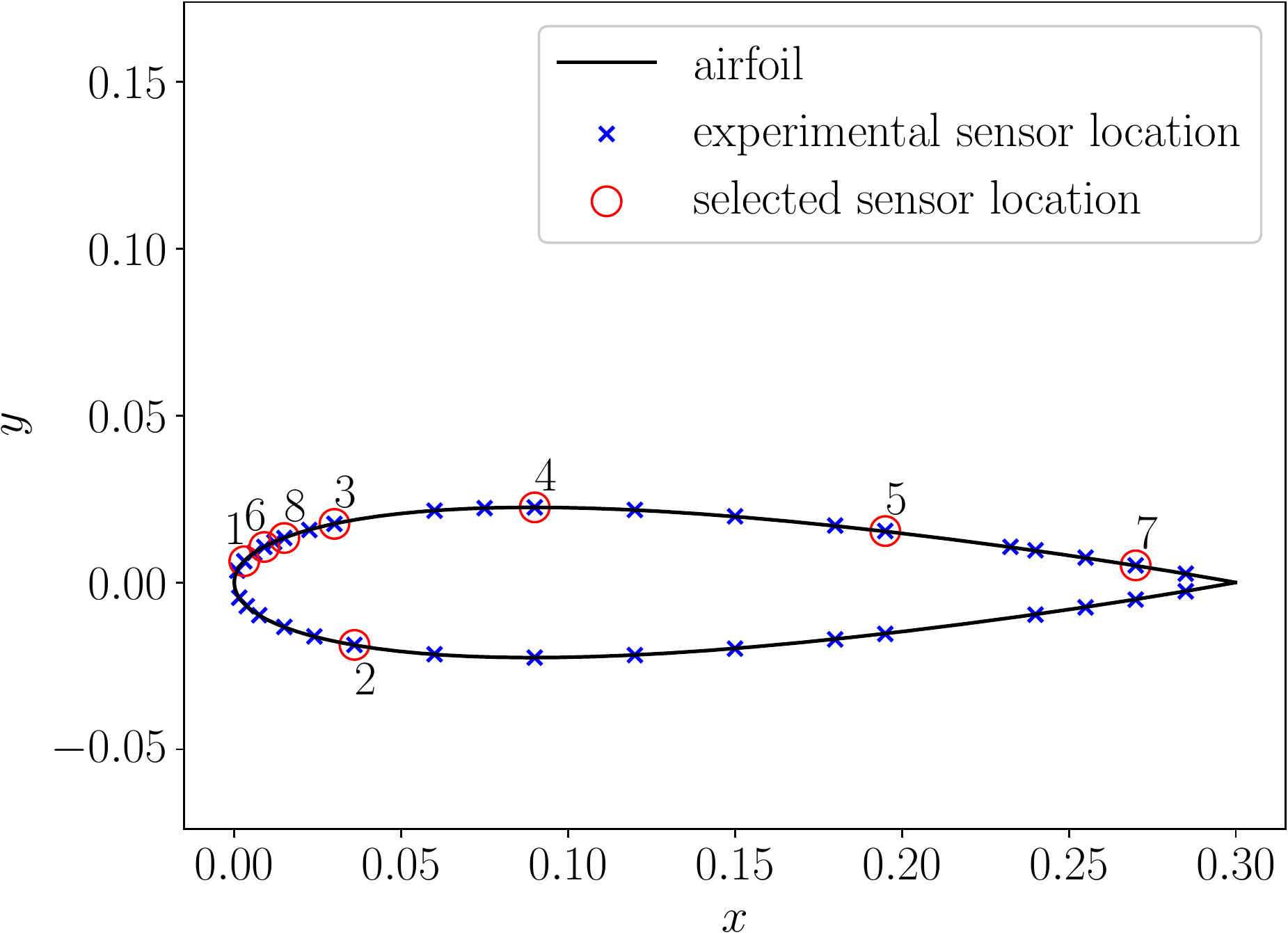}
         \caption{$n_s=8$}
    \end{subfigure}
    \begin{subfigure}[t]{0.33\textwidth}
        \centering
        \includegraphics[width=1.0\linewidth]{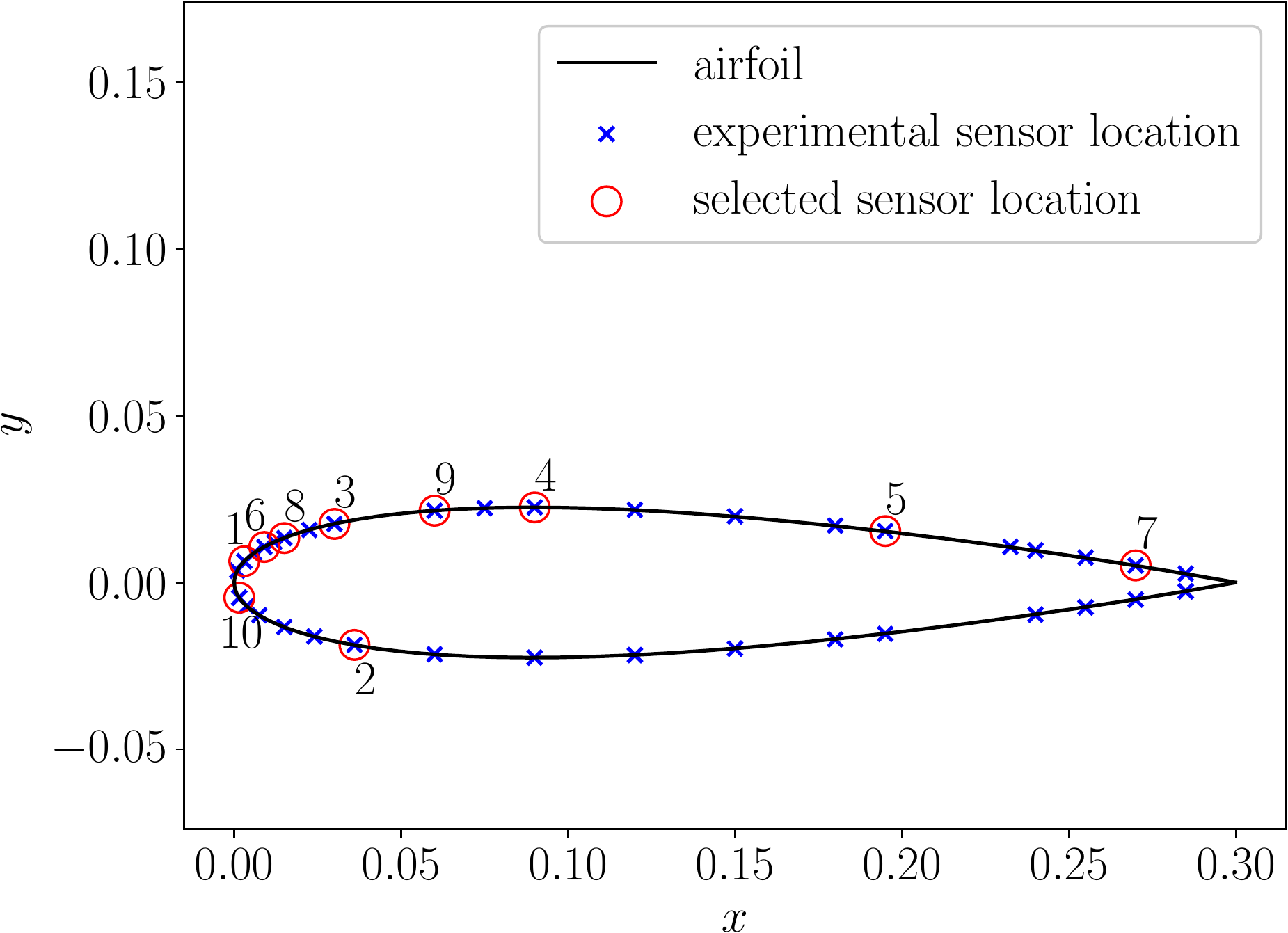}
         \caption{$n_s=10$}
    \end{subfigure}
\caption{Sensor locations on the airfoil selected by the DEIM based on the experimental data.}
\label{fig:2DAirfoil_exper_DEIM_locations}
\end{figure}

\begin{figure}[hbt!]
\centering
    \begin{subfigure}[t]{0.33\textwidth}
        \centering
        \includegraphics[width=1.0\linewidth]{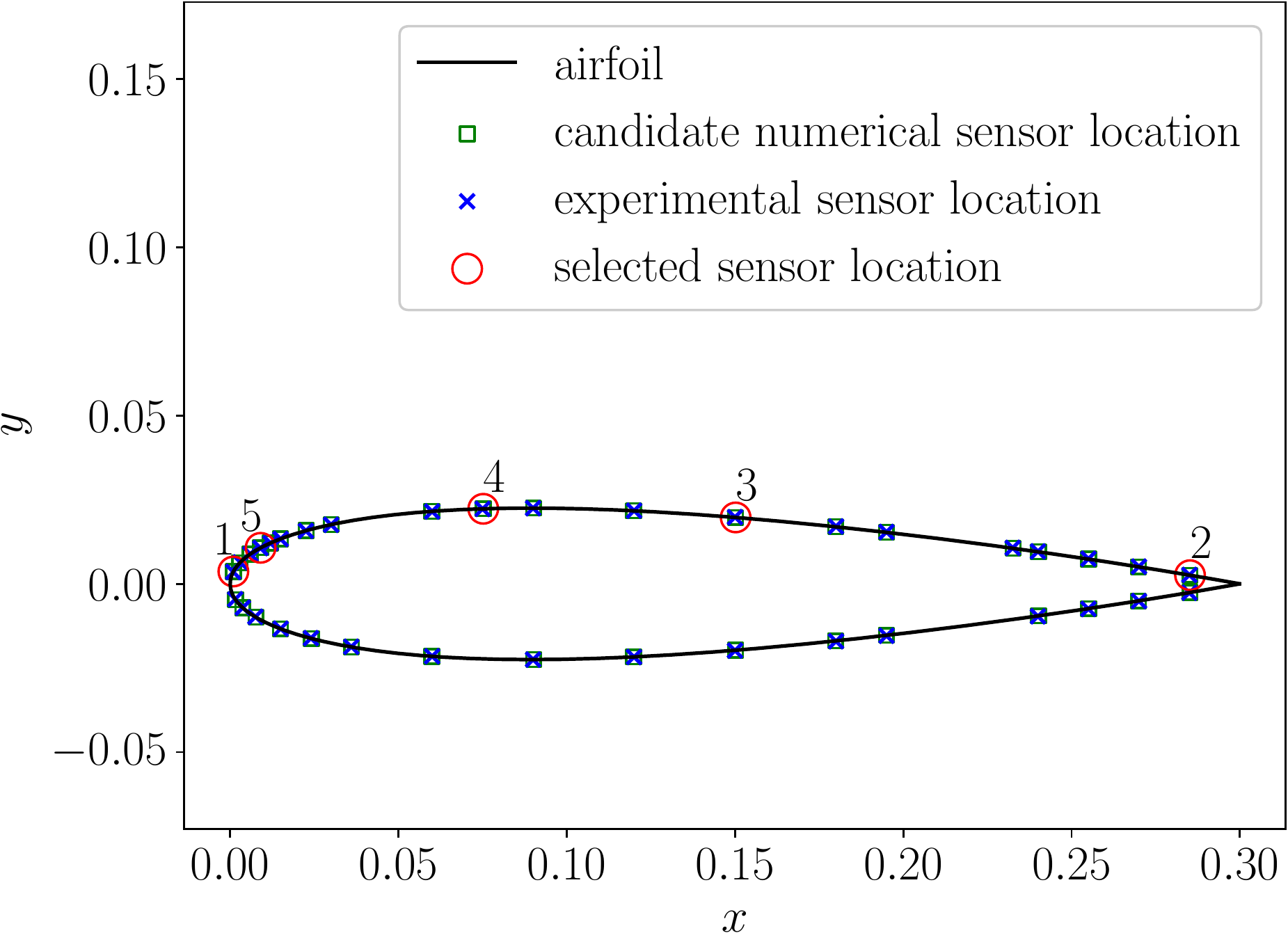}
         \caption{$n_s=5$}
    \end{subfigure}
    \begin{subfigure}[t]{0.33\textwidth}
        \centering
        \includegraphics[width=1.0\linewidth]{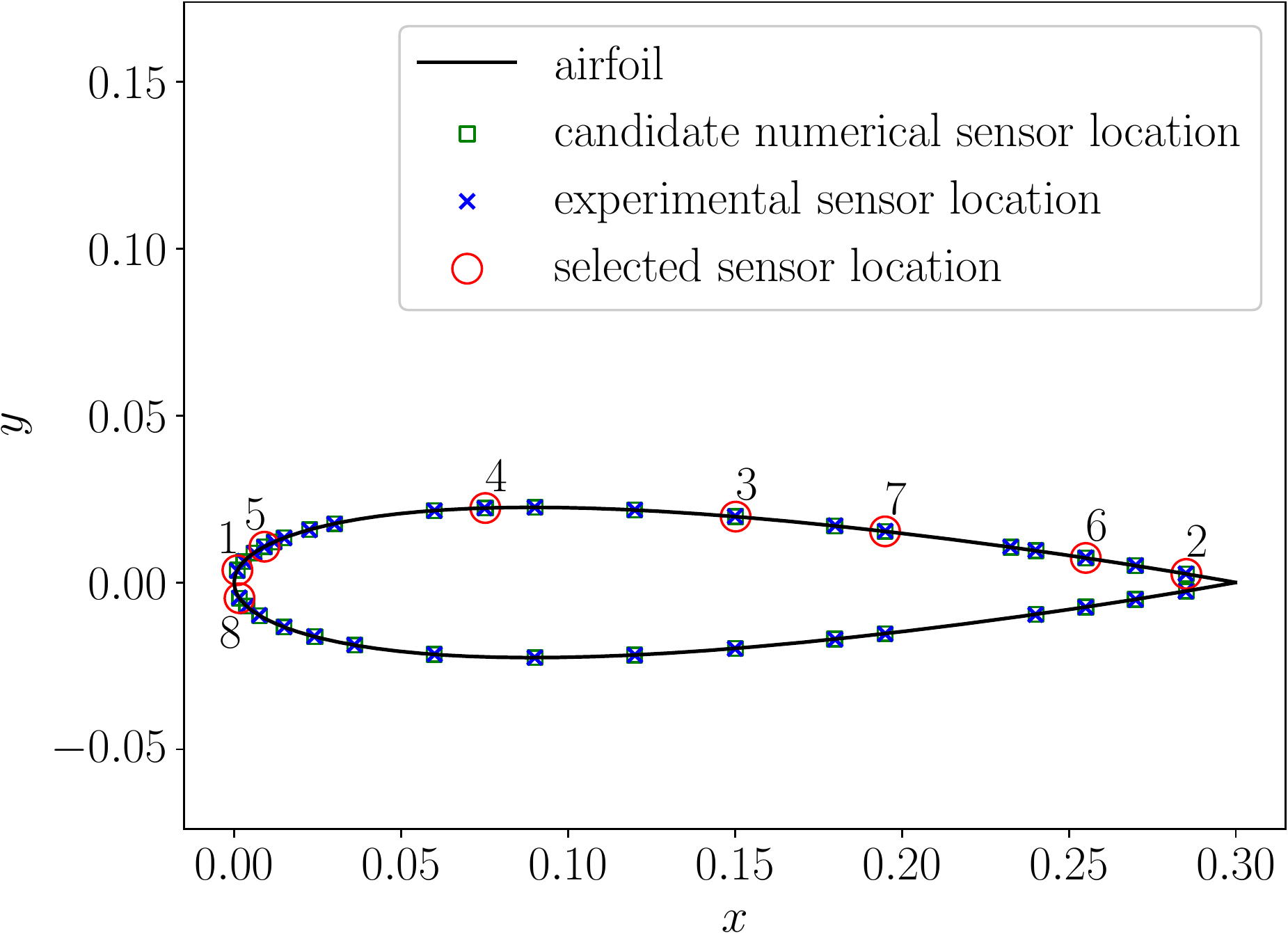}
         \caption{$n_s=8$}
    \end{subfigure}
    \begin{subfigure}[t]{0.33\textwidth}
        \centering
        \includegraphics[width=1.0\linewidth]{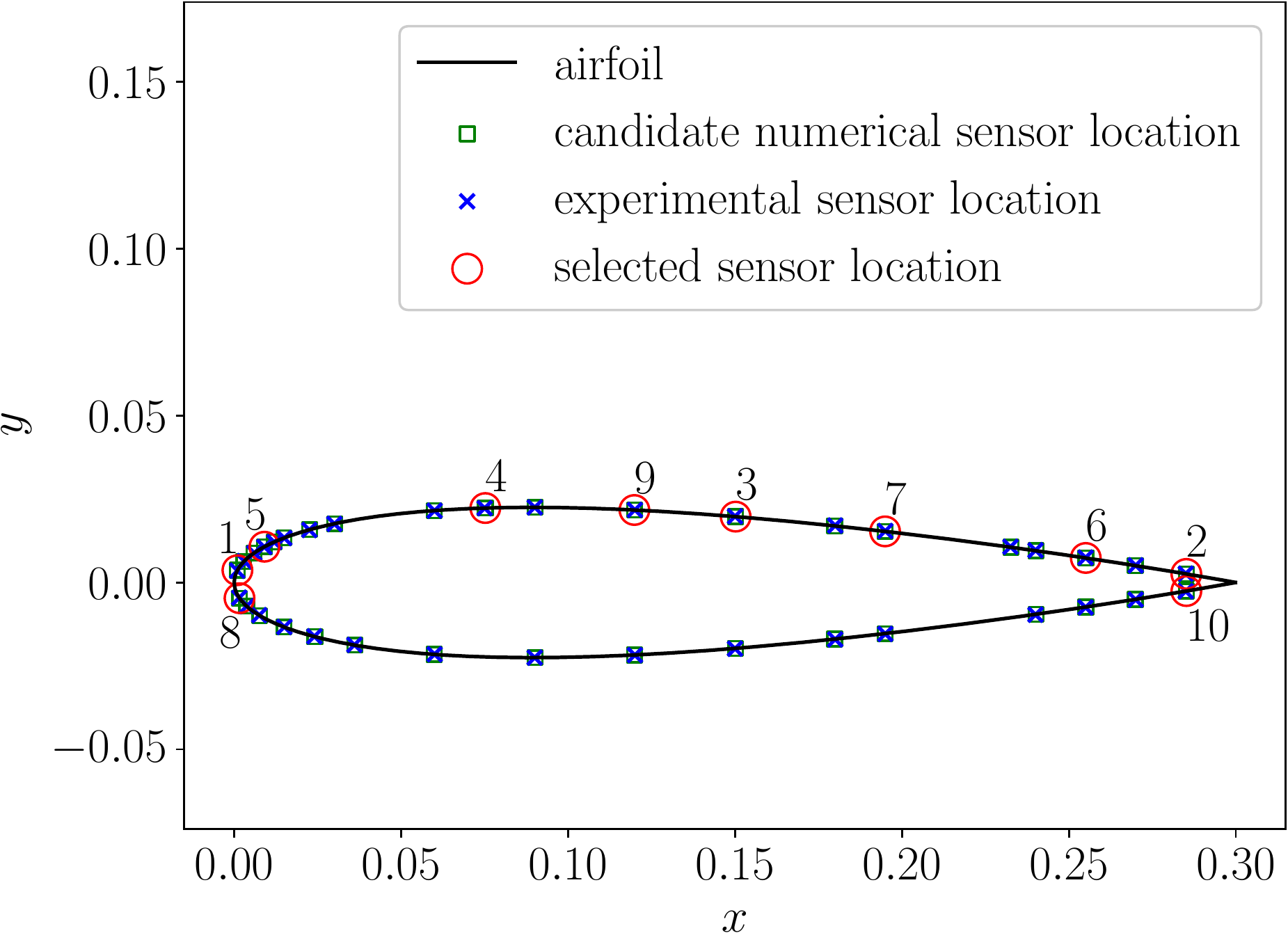}
         \caption{$n_s=10$}
    \end{subfigure}
\caption{Sensor locations on the airfoil selected by the DEIM based on the URANS data.}
\label{fig:2DAirfoil_numer_DEIM_locations}
\end{figure}

There are $36$ sensor locations on the airfoil in the experiment,
which serve as the candidate locations to be selected by the DEIM based on the experimental data in the first case,
see Fig. \ref{fig:2DAirfoil_exper_DEIM_locations} with $n_s = 5,8,10$.
In the second case, the mesh centers, among a total of $1316$ surface elements, nearest to the experimental sensor locations are used as the candidate locations,
and the sensor locations are selected by the DEIM based on the URANS data,
shown in Fig. \ref{fig:2DAirfoil_numer_DEIM_locations} with $n_s = 5,8,10$.
In the figures, the indices are the order of importance of the locations resulting from the selection in the DEIM.
The selected locations predominantly lie on the upper surface of the airfoil
and concentrate near the leading and trailing edges,
which is consistent with the flow characteristics in Fig. \ref{fig:2DAirfoil_mesh_flow_field}.

\begin{figure}[hbt!]
    \centering
    \begin{subfigure}[t]{0.49\textwidth}
        \centering
        \includegraphics[width=\linewidth]{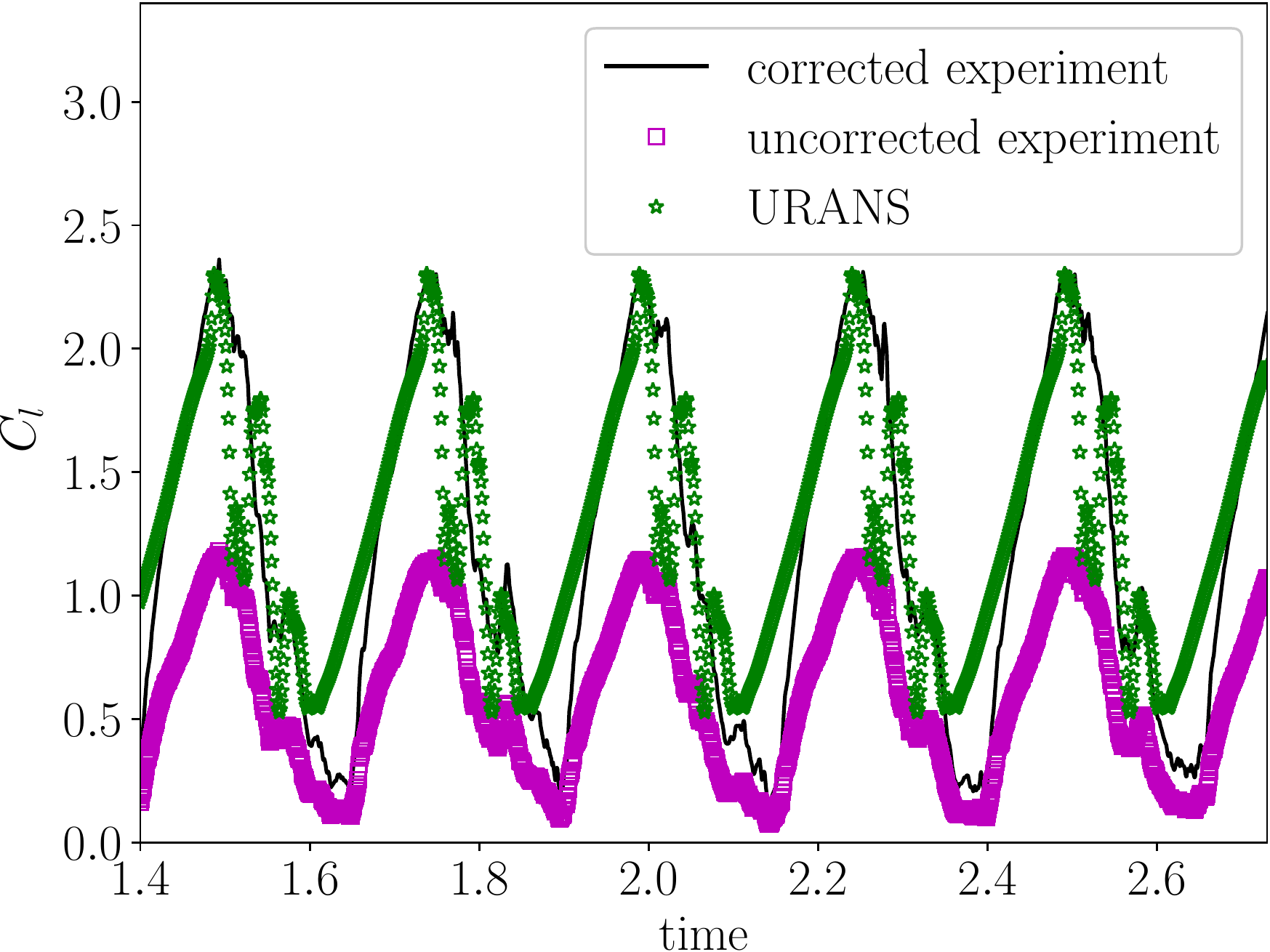}
        \caption{$C_l$}
    \end{subfigure}
    \begin{subfigure}[t]{0.49\textwidth}
        \centering
        \includegraphics[width=\linewidth]{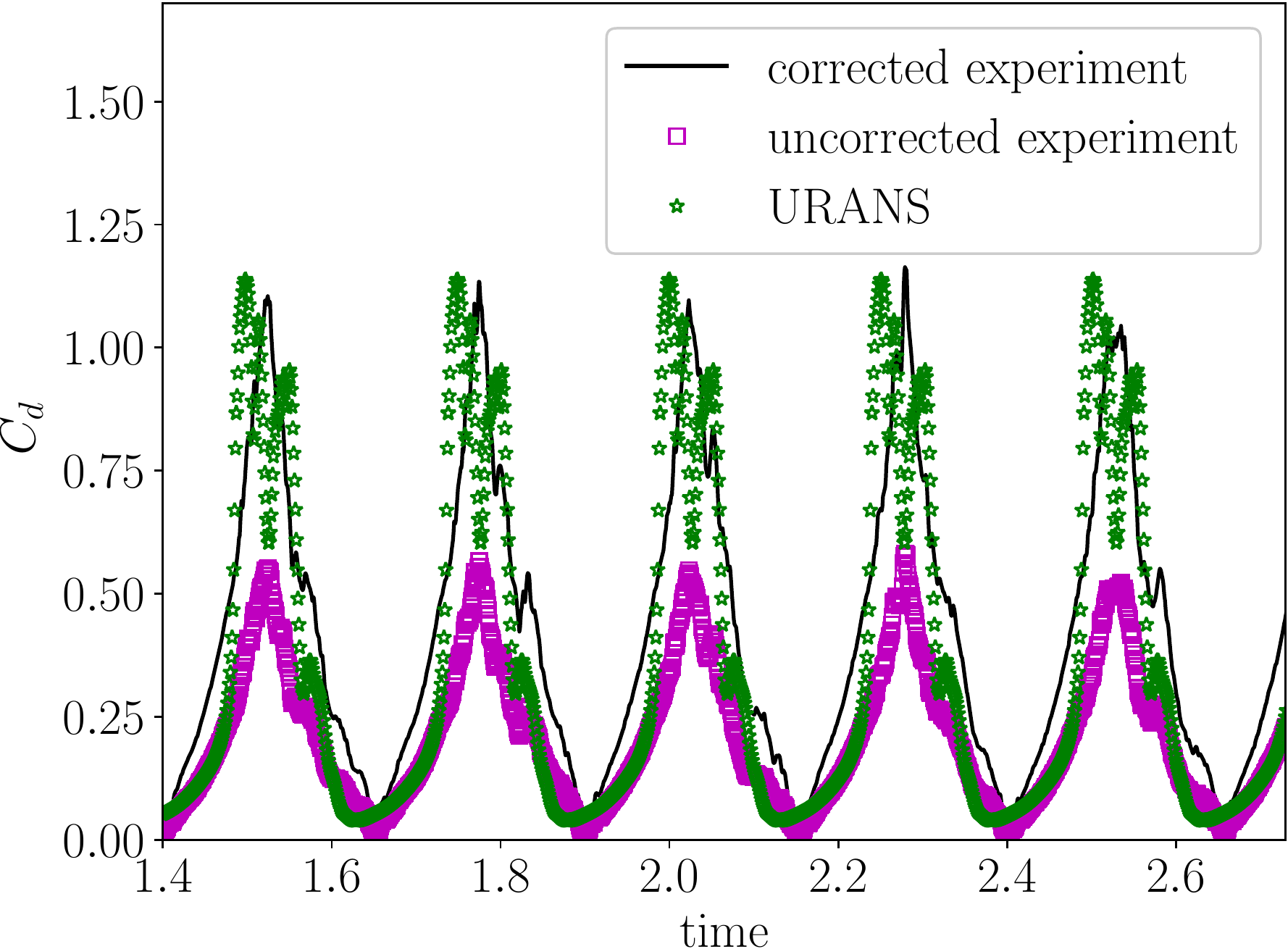}
        \caption{$C_d$}
    \end{subfigure}
    \caption{2D airfoil: Comparison of the corrected and uncorrected experimental, and URANS aerodynamic coefficients.}
\label{fig:2DAirfoil_numer_exper_cmp}
\end{figure}

\begin{figure}[hbt!]
    \centering
    \begin{subfigure}[t]{0.49\textwidth}
        \centering
        \includegraphics[width=\linewidth]{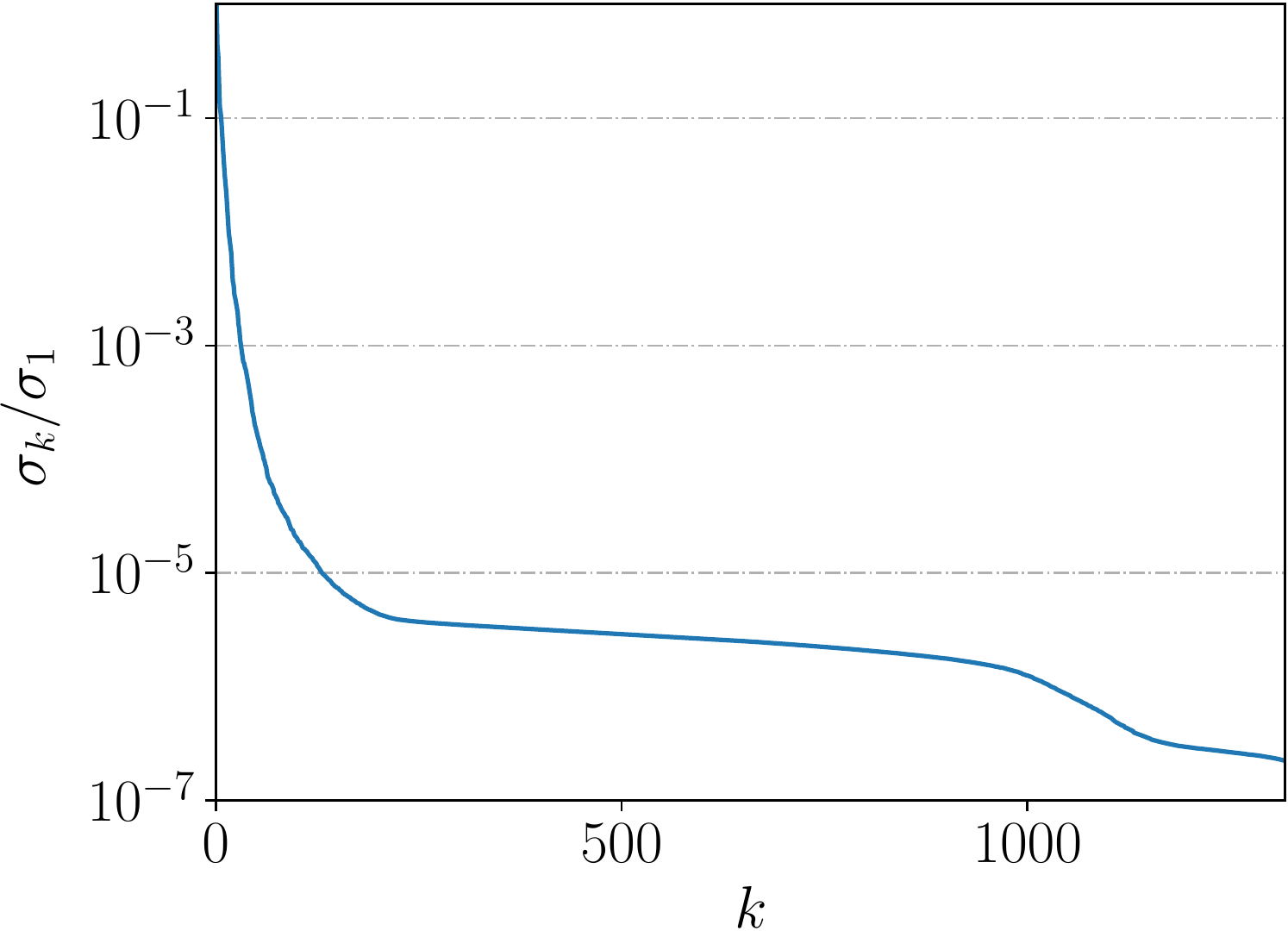}
        \caption{Scaled singular values}
    \end{subfigure}
    \begin{subfigure}[t]{0.49\textwidth}
        \centering
        \includegraphics[width=\linewidth]{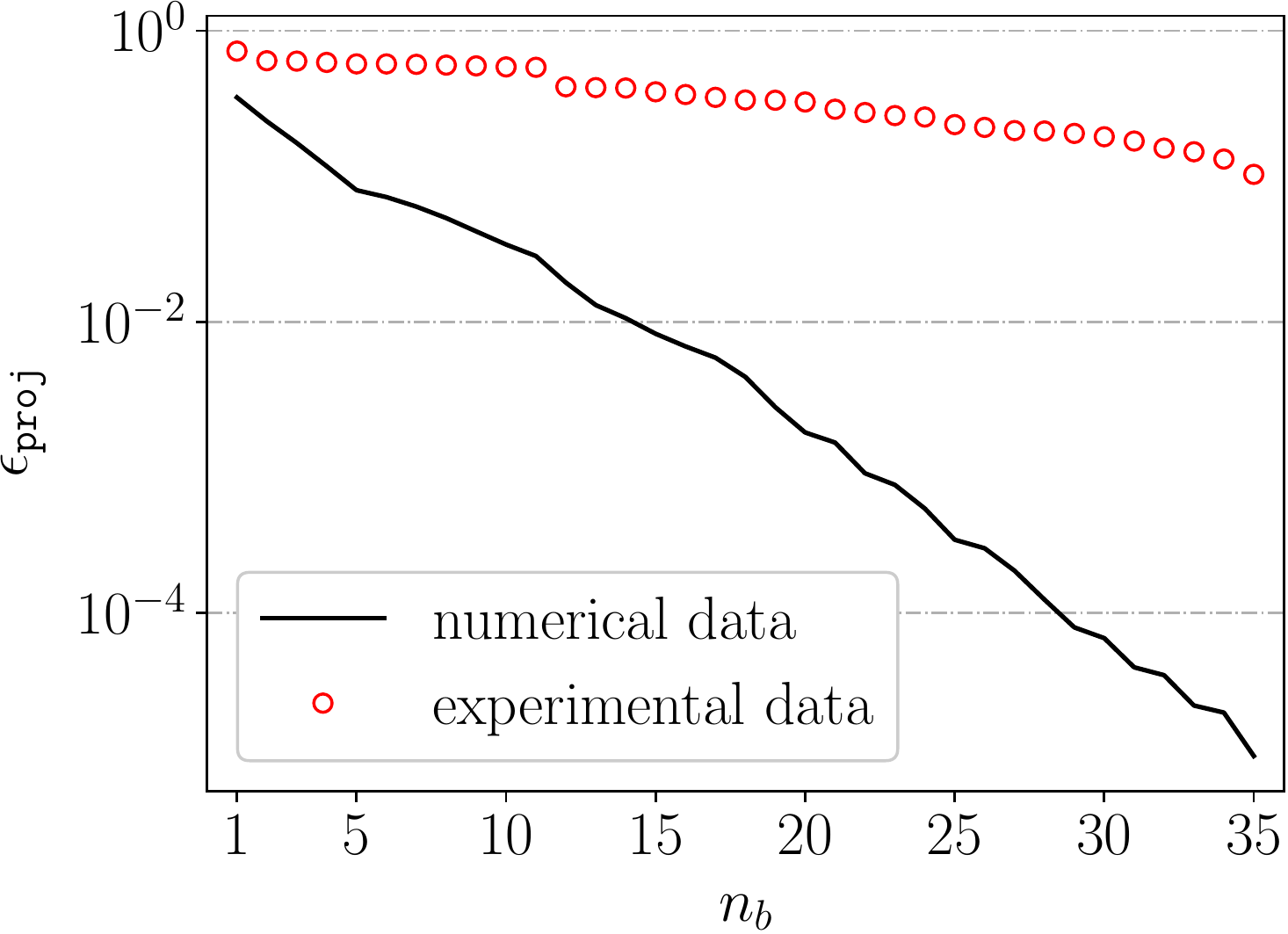}
        \caption{Projection errors}
    \end{subfigure}
    \caption{2D airfoil: The scaled singular values and the projection errors $\epsilon_{\tt proj}$ w.r.t. $n_b$ of the URANS and experimental data on the reduced space generated by the URANS data.}
\label{fig:2DAirfoil_svd_projection_error}
\end{figure}

The comparisons of the corrected and uncorrected experimental, and URANS aerodynamic coefficients are shown in Fig. \ref{fig:2DAirfoil_numer_exper_cmp}.
One observes that the URANS results can capture the main flow features,
but deviate from the corrected experimental results.
To further examine whether the reduced subspace generated by the URANS simulation can represent the main features of the surface pressure field,
the projection errors $\epsilon_{\tt proj}$ of the experimental data onto the reduced space are computed
\begin{equation*}
    \epsilon_{\tt proj} = \dfrac{1}{N_{t_{\tt exper}}N_{f_{\tt exper}}}
    \sum\limits_{\substack{t\in t_{\tt exper}\\ f\in f_{\tt exper}}}
    \dfrac{\norm{\left(I_{n_s} - \widehat{\bU}\widehat{\bU}^\mathrm{T}\right)\left(\bm{C}_p^{\tt exper}(t,f)-\overline{\bm{C}_p}(\bm{\mathcal{I}})\right)}_2}
    {\norm{\bm{C}_p^{\tt exper}(t,f)-\overline{\bm{C}_p}(\bm{\mathcal{I}})}_2},
\end{equation*}
where $\widehat{\bU}\in\bbR^{m\times n_s}$ are the orthonormal reduced basis functions obtained by performing the Gram-Schmidt method on $\bU(\bm{\mathcal{I}},:)$.
The scaled singular values in the SVD of the URANS data are plotted in Fig. \ref{fig:2DAirfoil_svd_projection_error},
which shows that a larger basis improves the approximation of the numerical snapshots as expected.
One also observes from Fig. \ref{fig:2DAirfoil_svd_projection_error} that the projection errors decay as $n_b$ increases,
reflecting that a larger reduced basis can express the experimental data better.

\begin{figure}[hbt!]
    \centering
    \begin{subfigure}[t]{0.49\textwidth}
        \centering
        \includegraphics[width=\linewidth]{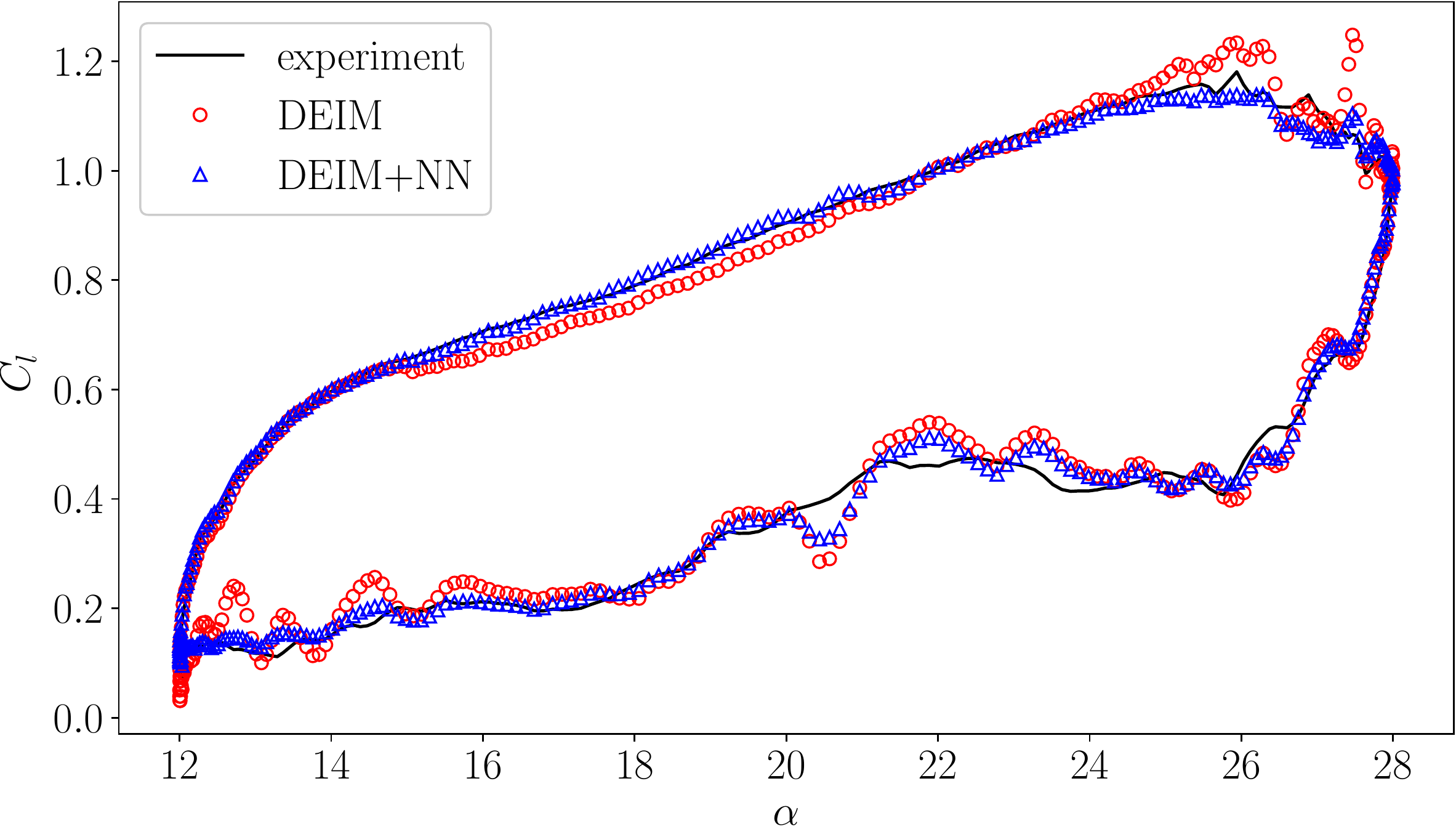}
        \caption{$n_s=5$, $C_l$}
    \end{subfigure}
    \begin{subfigure}[t]{0.49\textwidth}
        \centering
        \includegraphics[width=\linewidth]{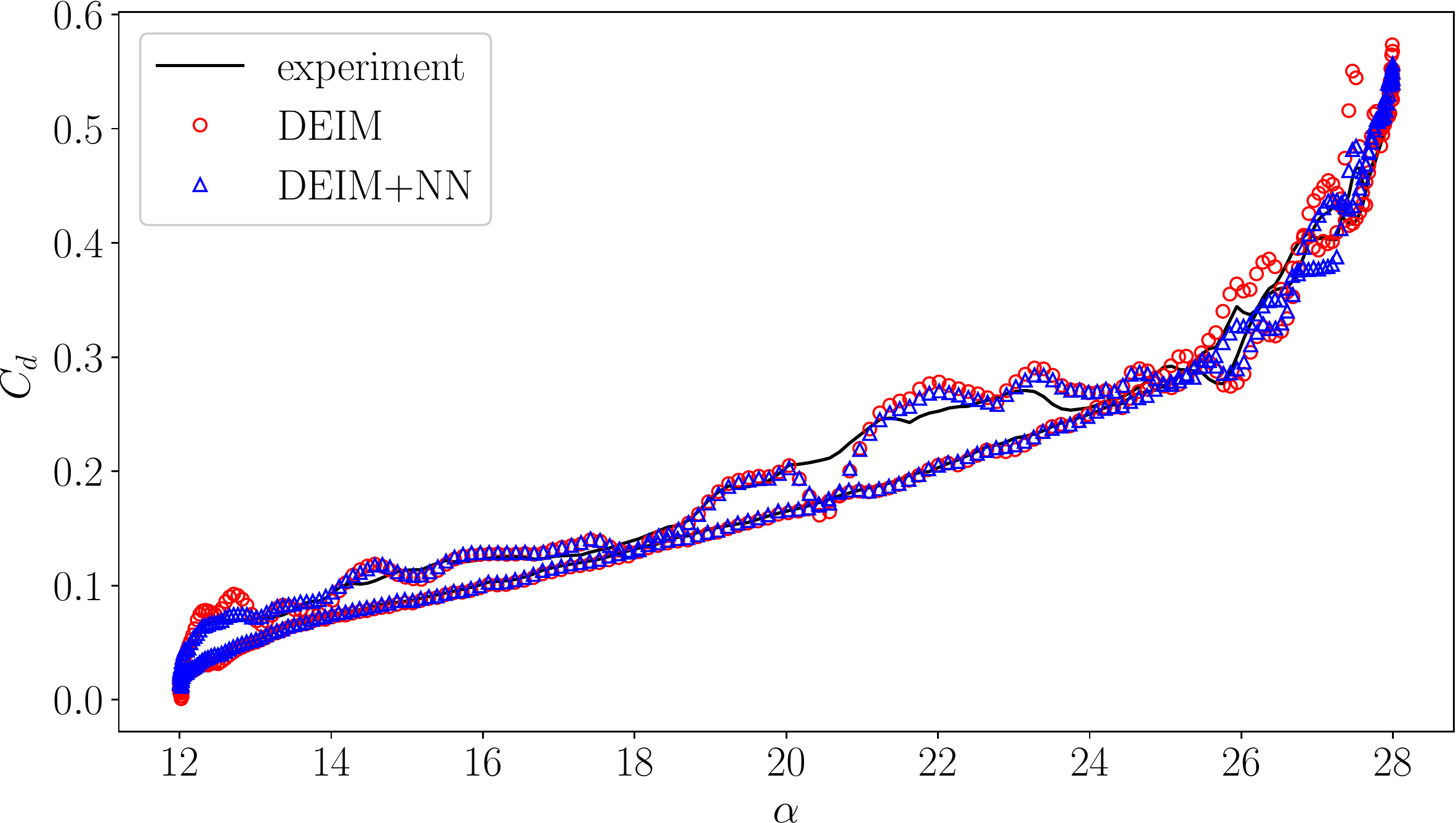}
        \caption{$n_s=5$, $C_d$}
    \end{subfigure}

    \begin{subfigure}[t]{0.49\textwidth}
        \centering
        \includegraphics[width=\linewidth]{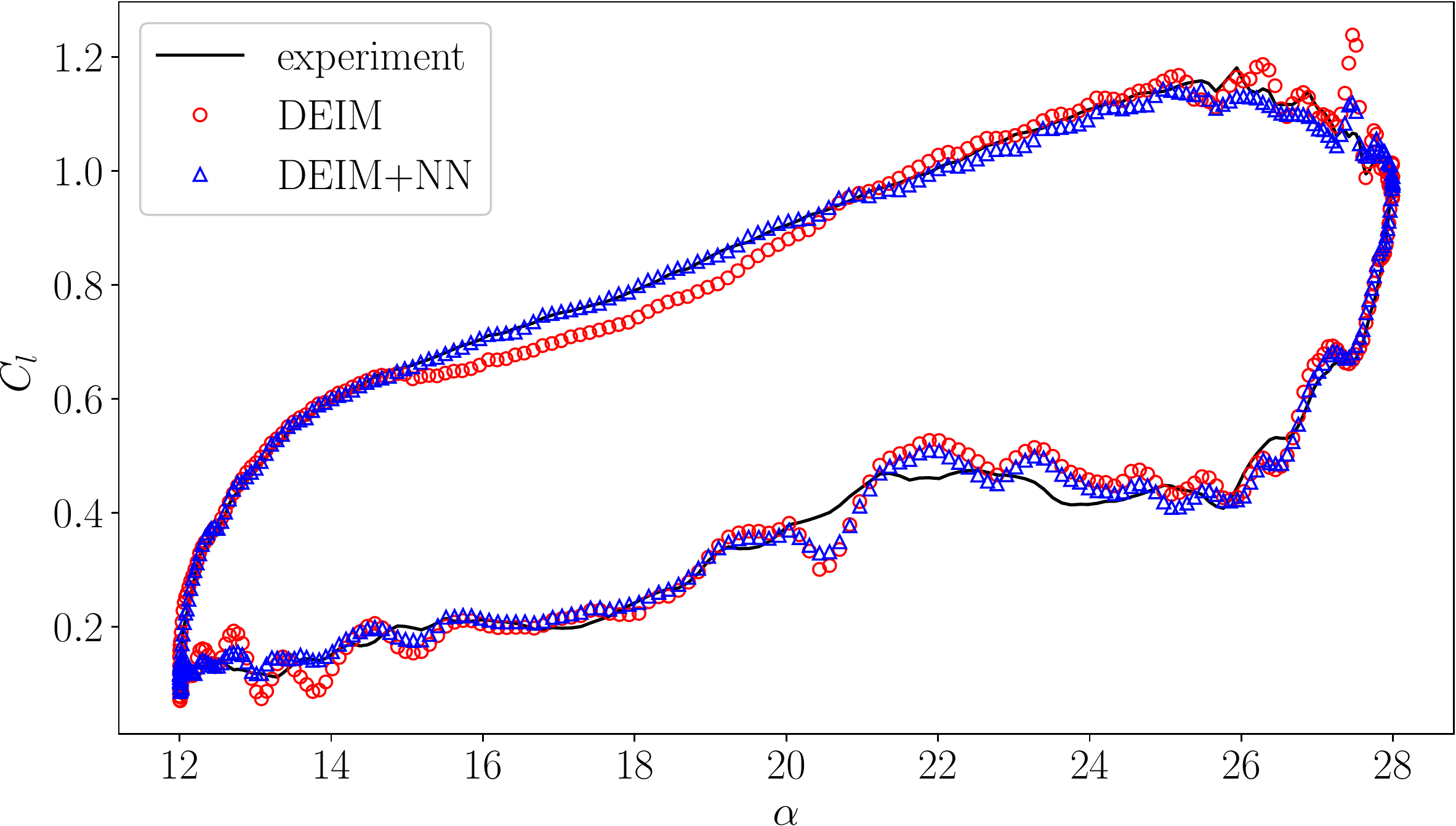}
        \caption{$n_s=8$, $C_l$}
    \end{subfigure}
    \begin{subfigure}[t]{0.49\textwidth}
        \centering
        \includegraphics[width=\linewidth]{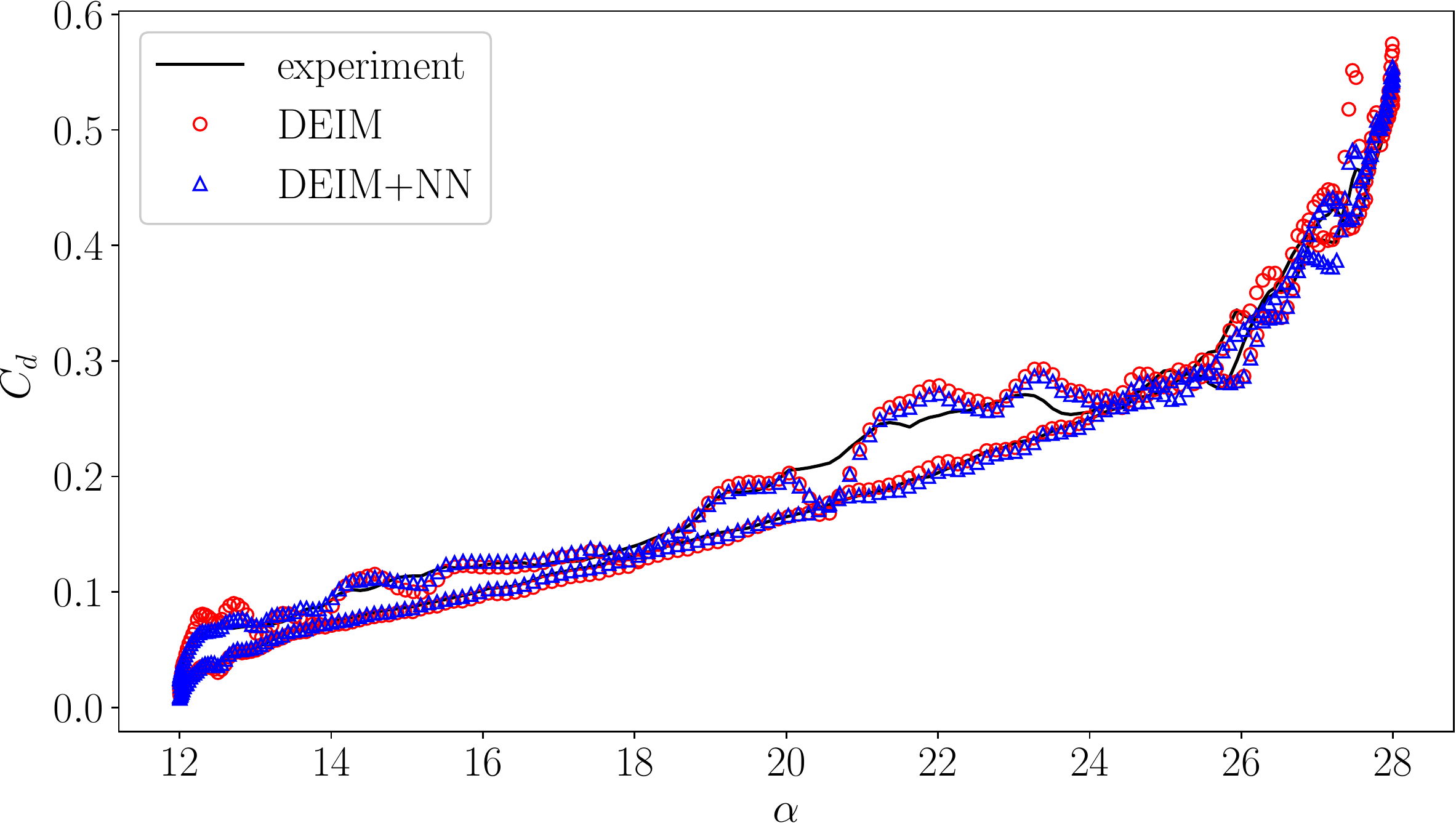}
        \caption{$n_s=8$, $C_d$}
    \end{subfigure}

    \begin{subfigure}[t]{0.49\textwidth}
        \centering
        \includegraphics[width=\linewidth]{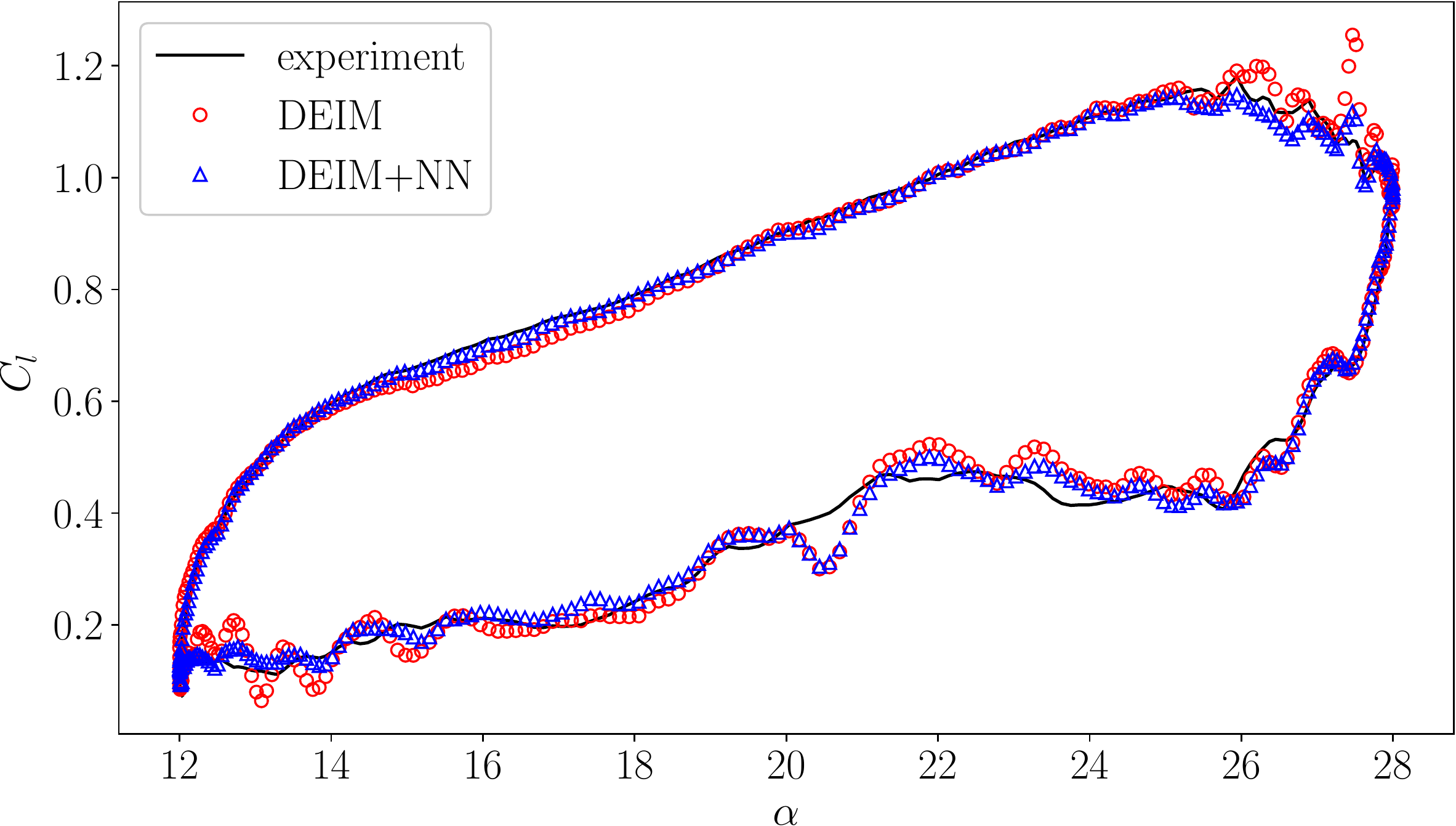}
        \caption{$n_s=10$, $C_l$}
    \end{subfigure}
    \begin{subfigure}[t]{0.49\textwidth}
        \centering
        \includegraphics[width=\linewidth]{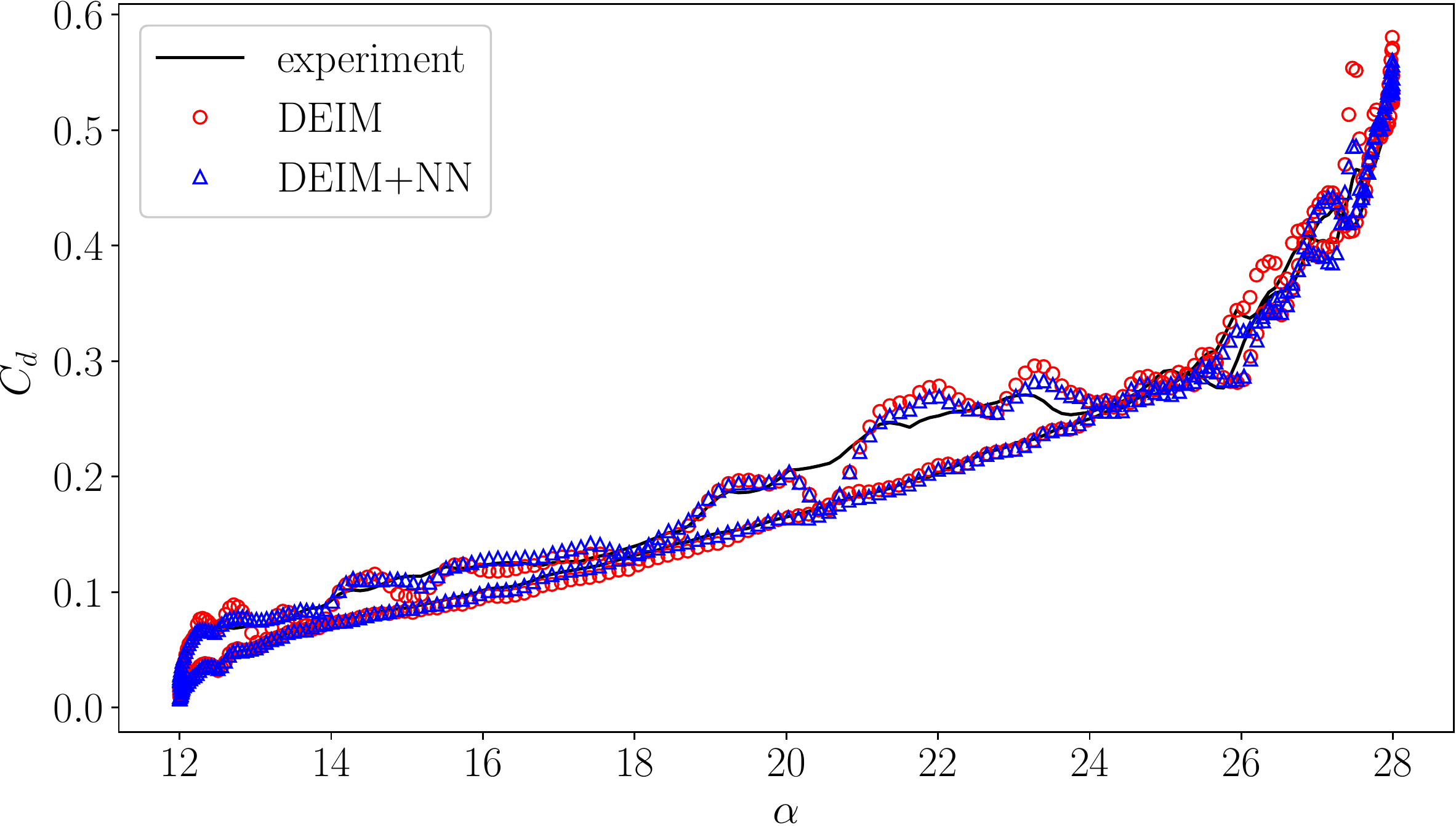}
        \caption{$n_s=10$, $C_d$}
    \end{subfigure}
    \caption{2D airfoil: $C_l,C_d$ w.r.t. $\alpha$ for the testing experimental data without noise in the pressure sensor inputs.
    The DEIM models are based on the experimental data.}
    \label{fig:2DAirfoil_exper_DEIM_location_aero_coeff_aoa_noise=0}
\end{figure}

\begin{figure}[hbt!]
    \centering
    \begin{subfigure}[t]{0.49\textwidth}
        \centering
        \includegraphics[width=\linewidth]{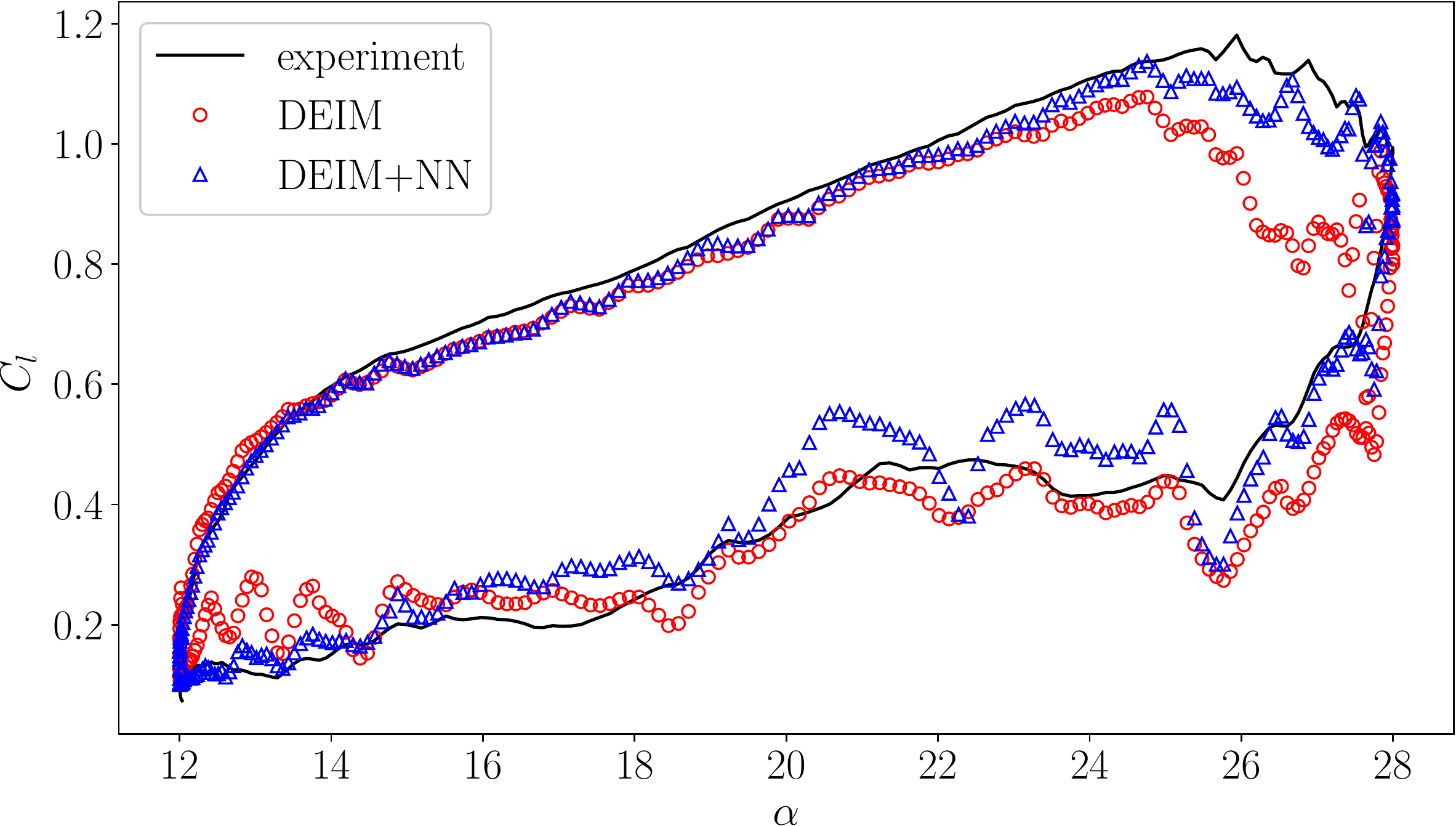}
        \caption{$n_s=5$, $C_l$}
    \end{subfigure}
    \begin{subfigure}[t]{0.49\textwidth}
        \centering
        \includegraphics[width=\linewidth]{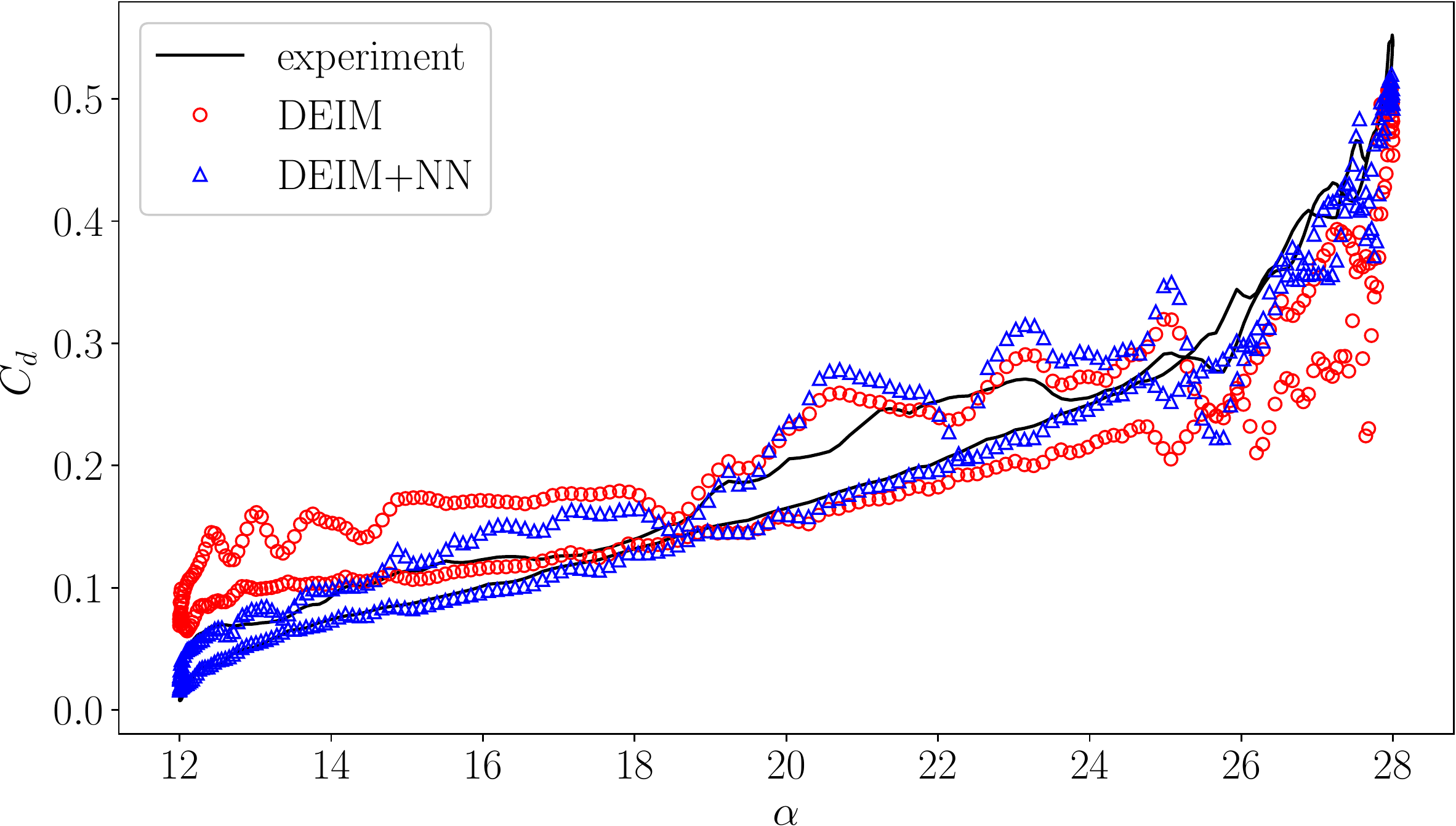}
        \caption{$n_s=5$, $C_d$}
    \end{subfigure}

    \begin{subfigure}[t]{0.49\textwidth}
        \centering
        \includegraphics[width=\linewidth]{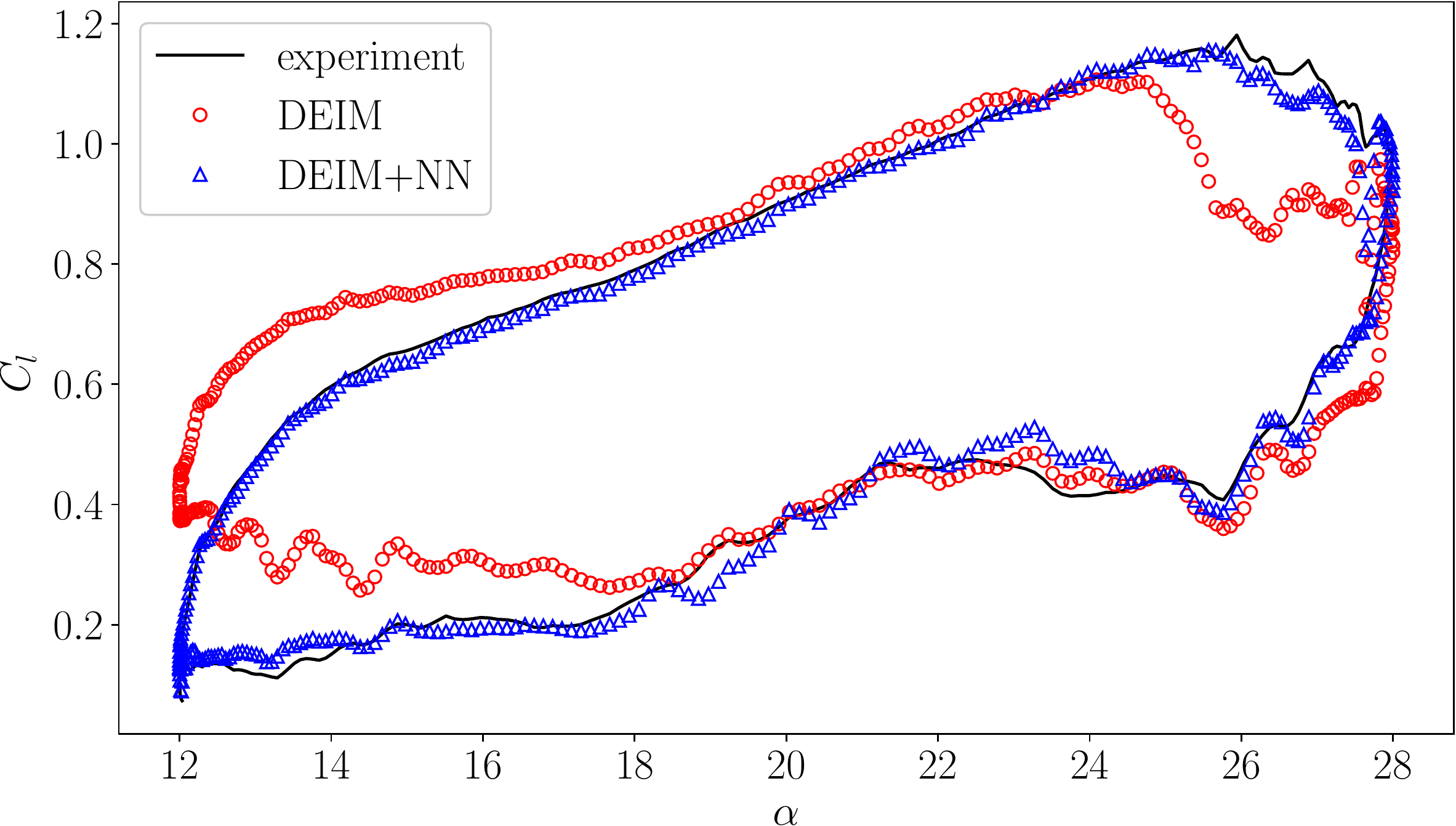}
        \caption{$n_s=8$, $C_l$}
    \end{subfigure}
    \begin{subfigure}[t]{0.49\textwidth}
        \centering
        \includegraphics[width=\linewidth]{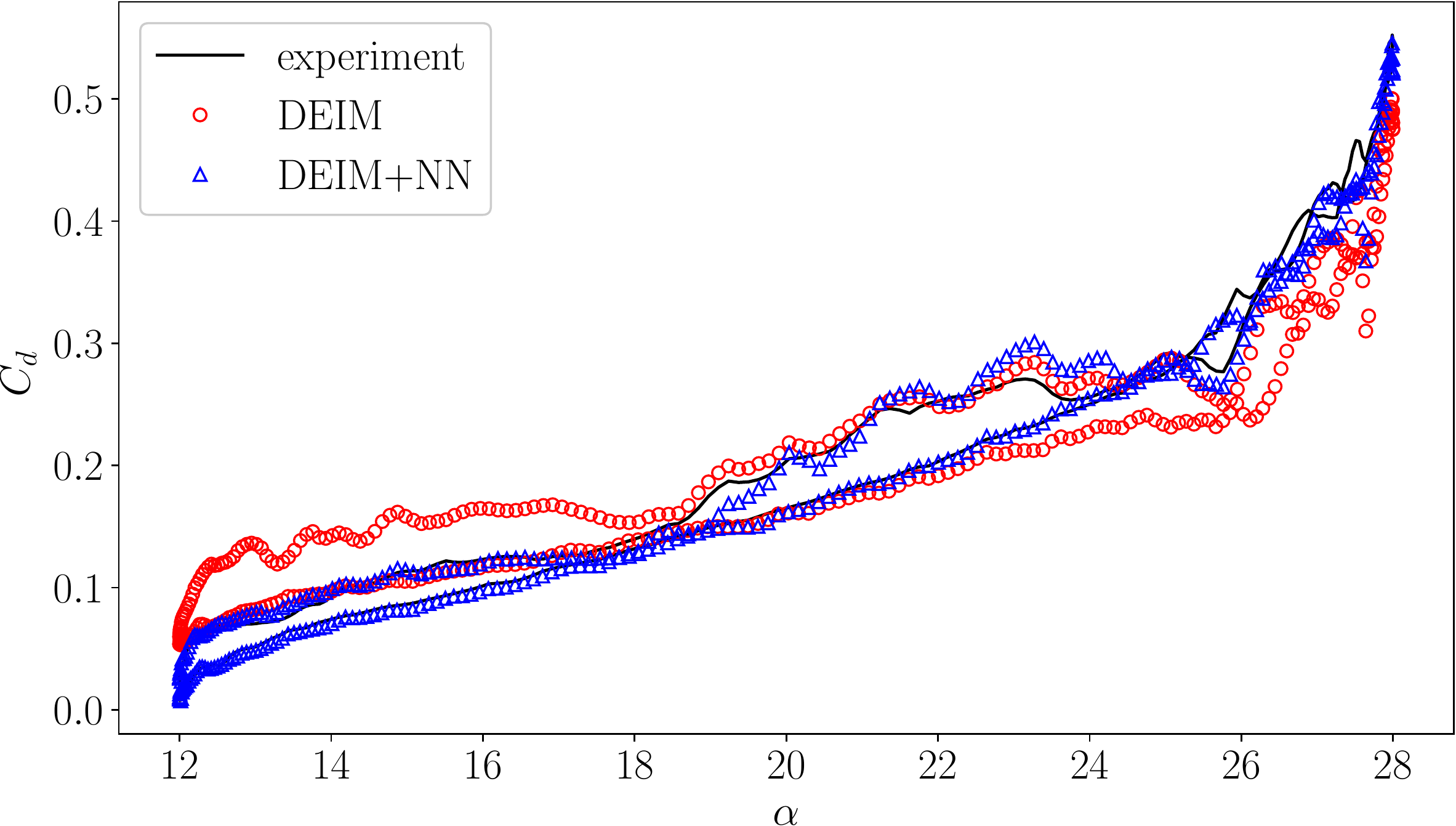}
        \caption{$n_s=8$, $C_d$}
    \end{subfigure}

    \begin{subfigure}[t]{0.49\textwidth}
        \centering
        \includegraphics[width=\linewidth]{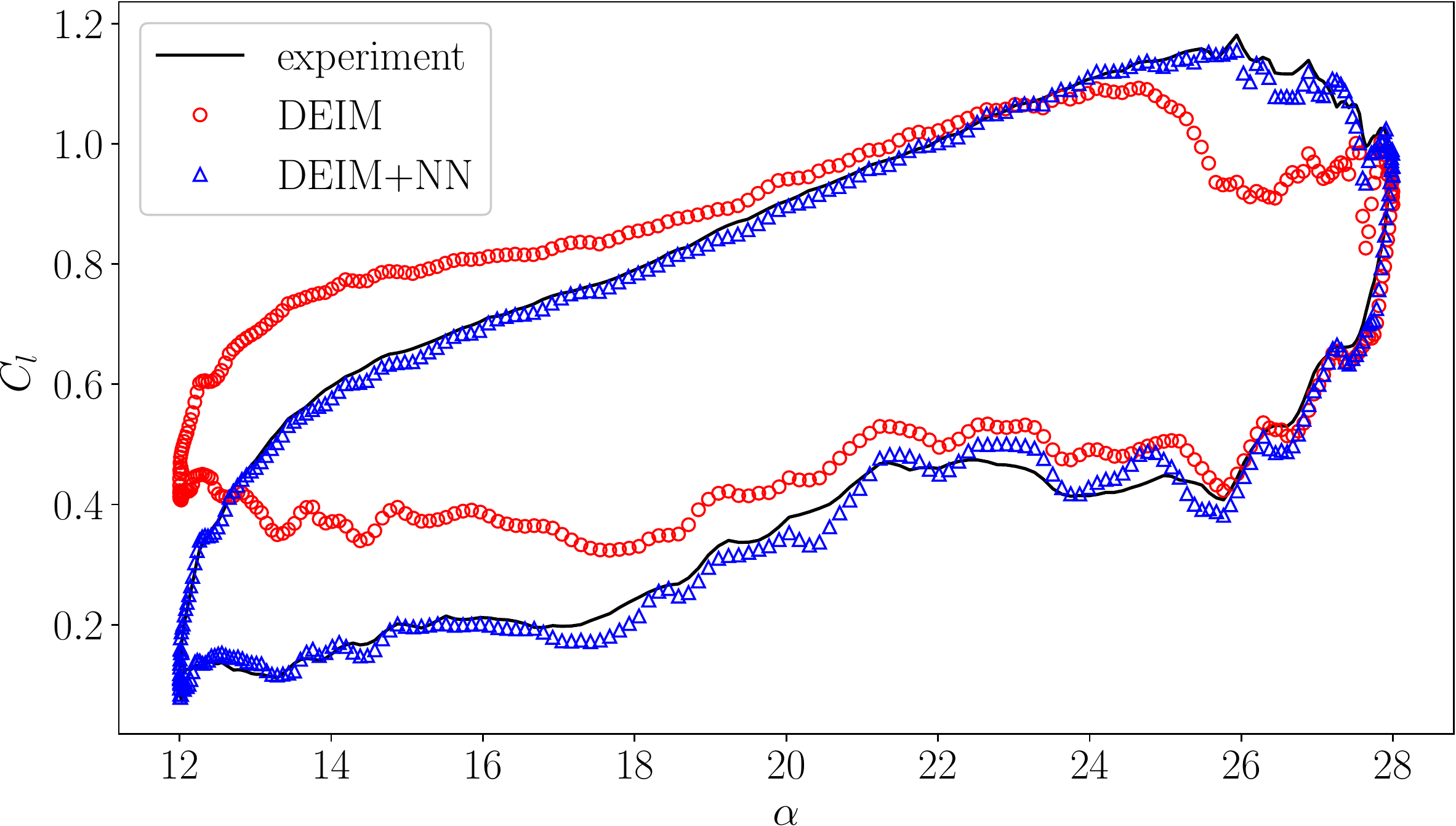}
        \caption{$n_s=10$, $C_l$}
    \end{subfigure}
    \begin{subfigure}[t]{0.49\textwidth}
        \centering
        \includegraphics[width=\linewidth]{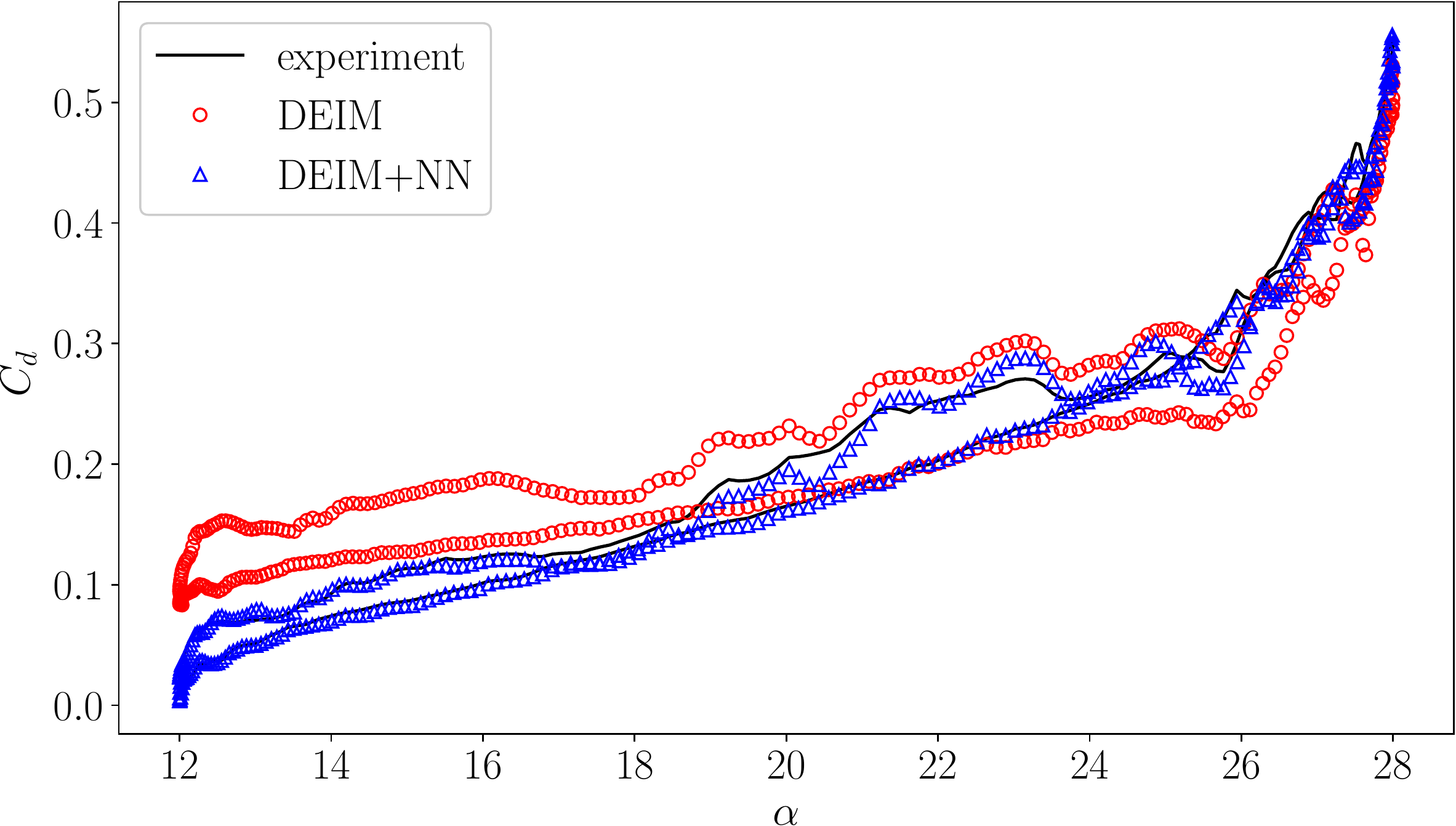}
        \caption{$n_s=10$, $C_d$}
    \end{subfigure}
    \caption{2D airfoil: $C_l,C_d$ w.r.t. $\alpha$ for the testing experimental data without noise in the pressure sensor inputs.
    The DEIM models are based on the URANS data.}
    \label{fig:2DAirfoil_numer_DEIM_location_aero_coeff_aoa_noise=0}
\end{figure}

\begin{figure}[hbt!]
    \centering
    \begin{subfigure}[t]{0.49\textwidth}
        \centering
        \includegraphics[width=\linewidth]{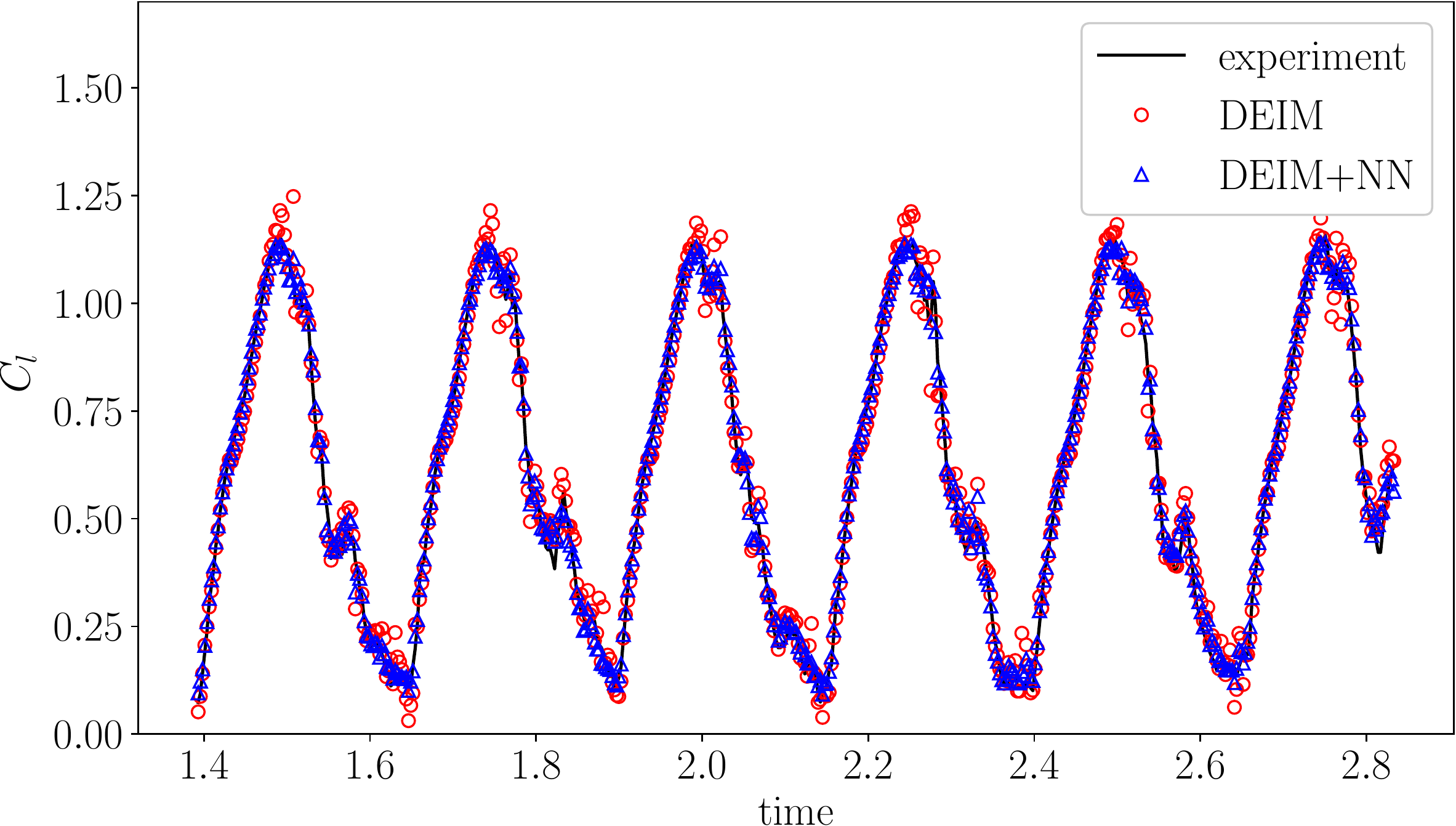}
        \caption{$n_s=5$, $C_l$}
    \end{subfigure}
    \begin{subfigure}[t]{0.49\textwidth}
        \centering
        \includegraphics[width=\linewidth]{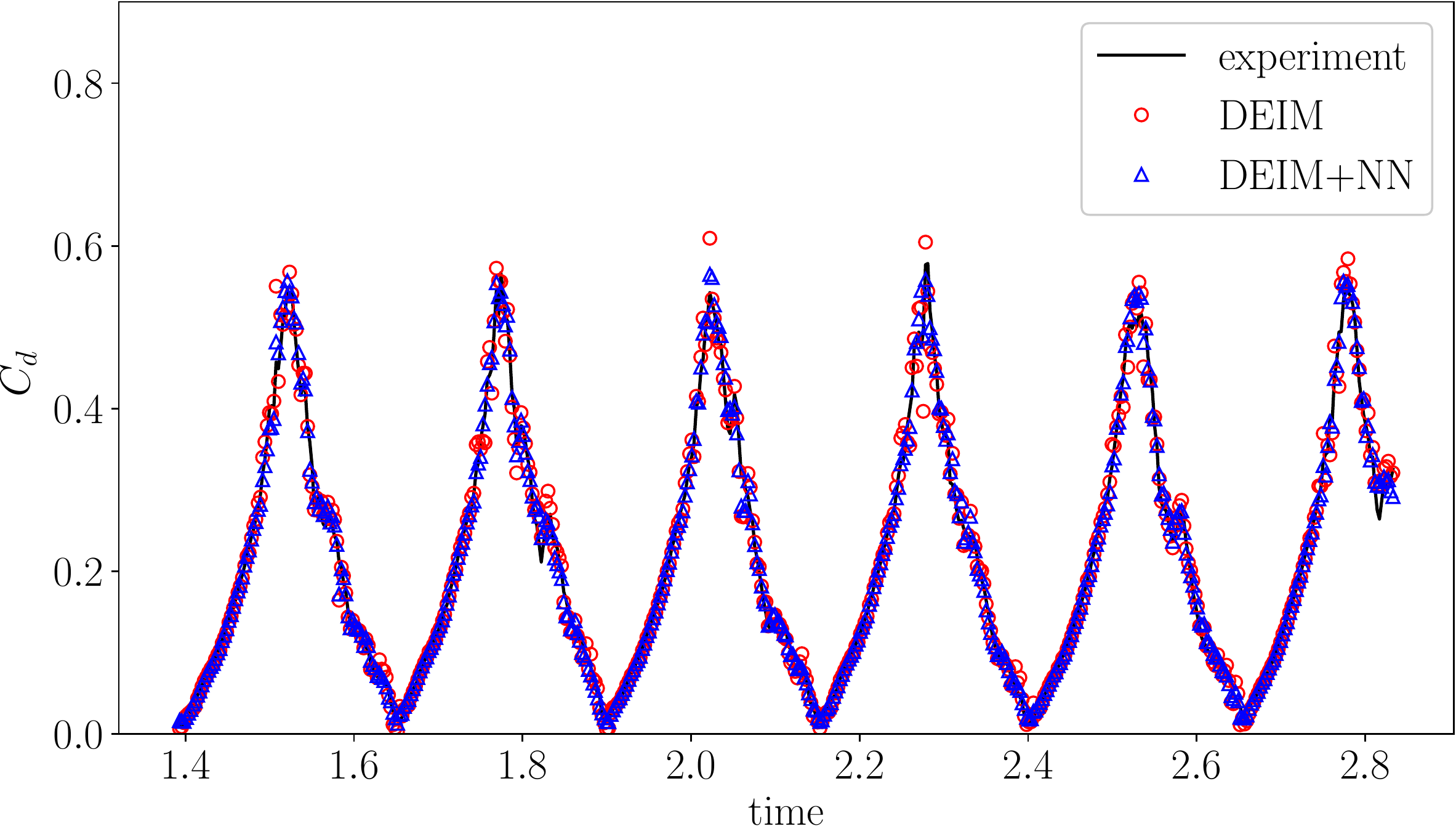}
        \caption{$n_s=5$, $C_d$}
    \end{subfigure}

    \begin{subfigure}[t]{0.49\textwidth}
        \centering
        \includegraphics[width=\linewidth]{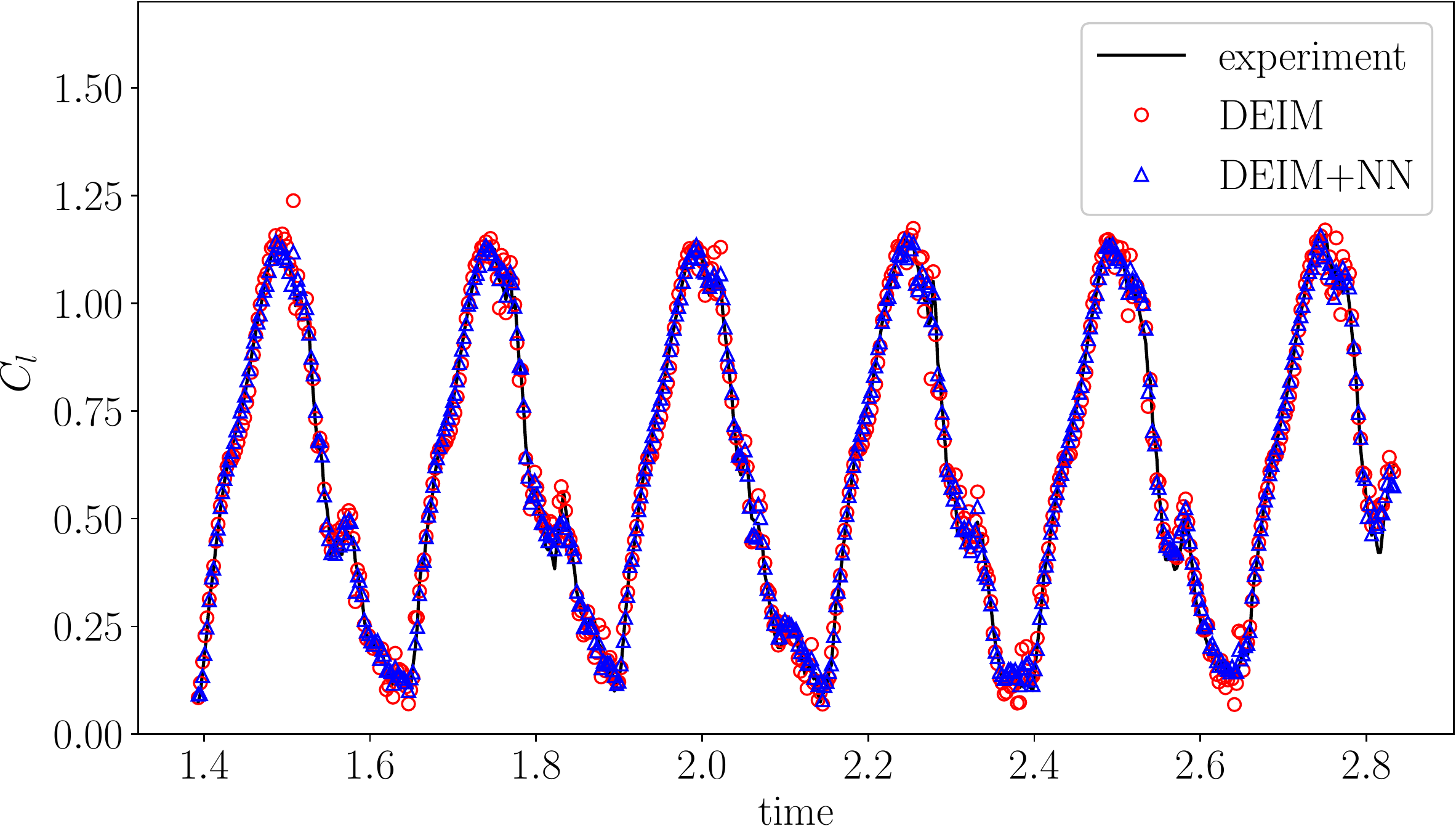}
        \caption{$n_s=8$, $C_l$}
    \end{subfigure}
    \begin{subfigure}[t]{0.49\textwidth}
        \centering
        \includegraphics[width=\linewidth]{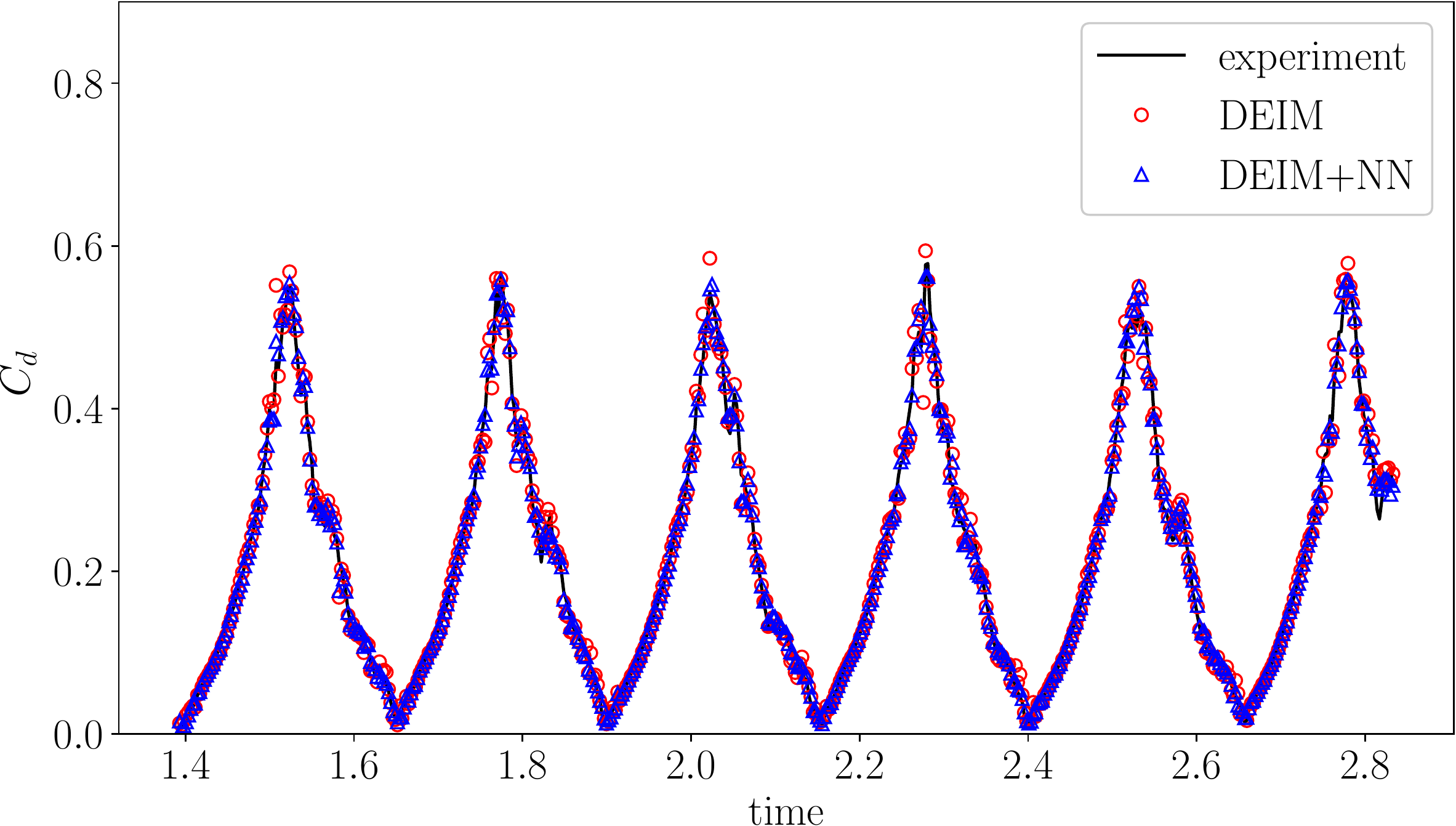}
        \caption{$n_s=8$, $C_d$}
    \end{subfigure}

    \begin{subfigure}[t]{0.49\textwidth}
        \centering
        \includegraphics[width=\linewidth]{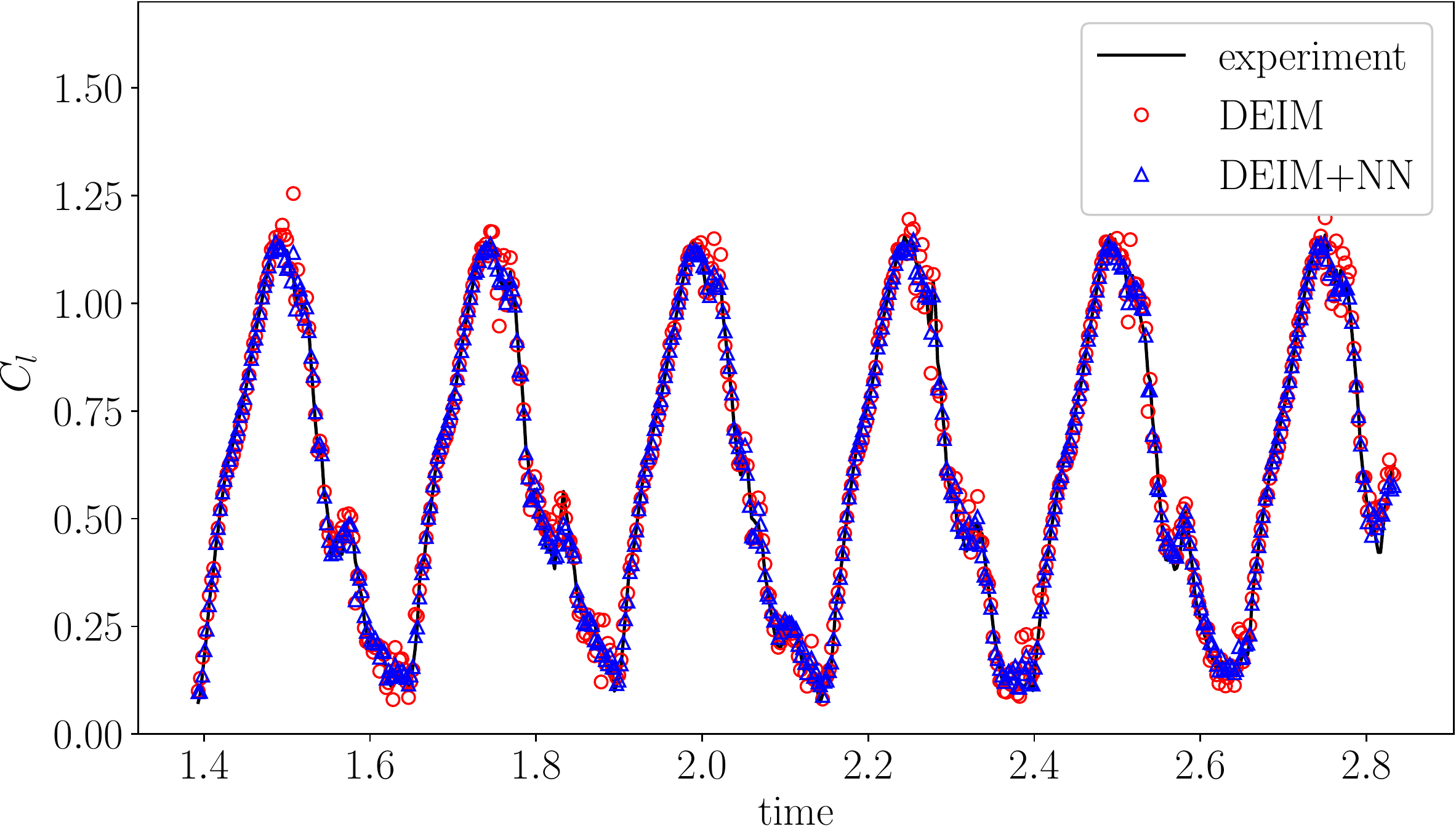}
        \caption{$n_s=10$, $C_l$}
    \end{subfigure}
    \begin{subfigure}[t]{0.49\textwidth}
        \centering
        \includegraphics[width=\linewidth]{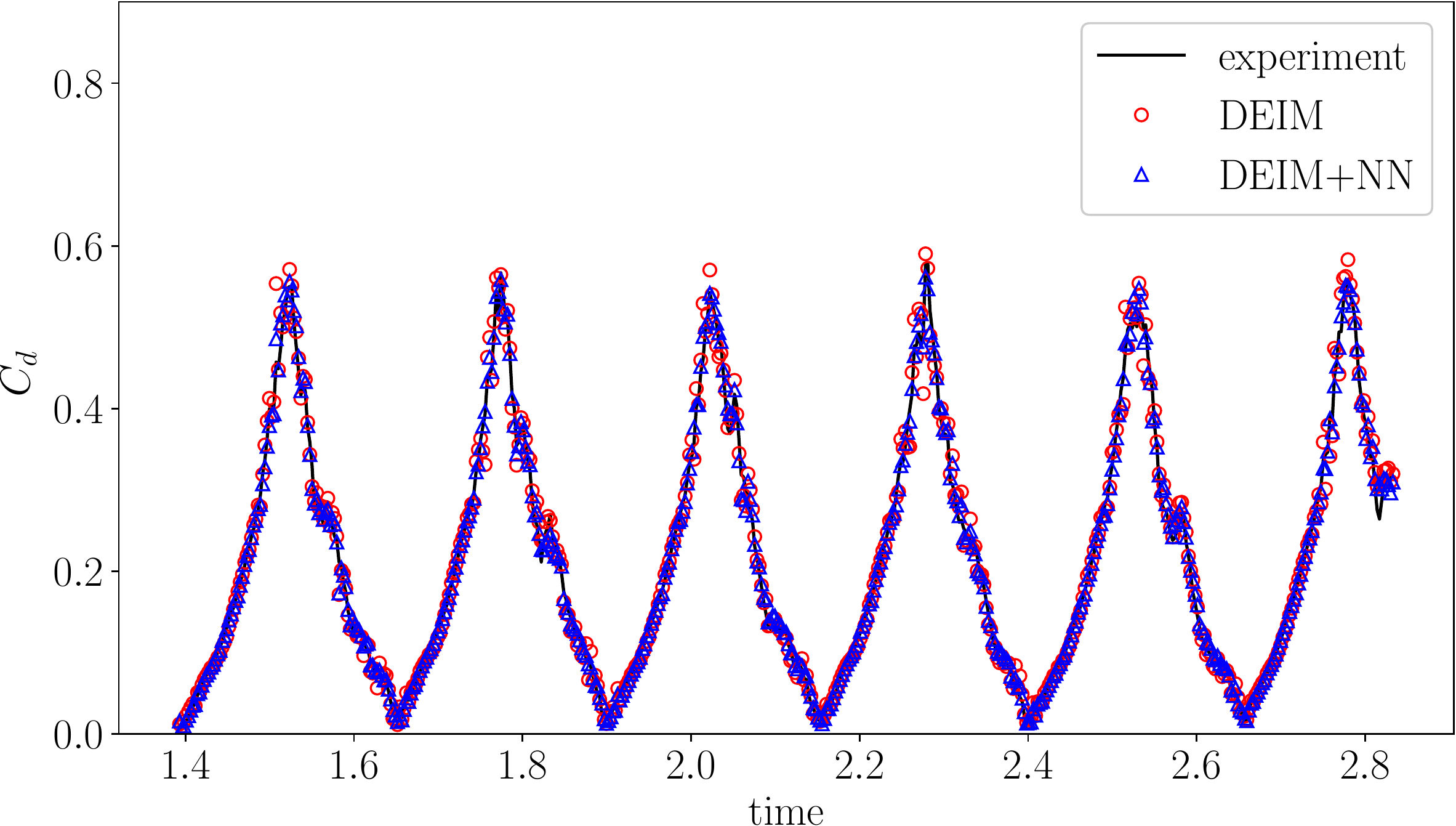}
        \caption{$n_s=10$, $C_d$}
    \end{subfigure}
    \caption{2D airfoil: $C_l,C_d$ w.r.t. time for the testing experimental data without noise in the pressure sensor inputs.
    The DEIM models are based on the experimental data.}
    \label{fig:2DAirfoil_exper_DEIM_location_aero_coeff_time_noise=0}
\end{figure}

\begin{figure}[hbt!]
    \centering
    \begin{subfigure}[t]{0.49\textwidth}
        \centering
        \includegraphics[width=\linewidth]{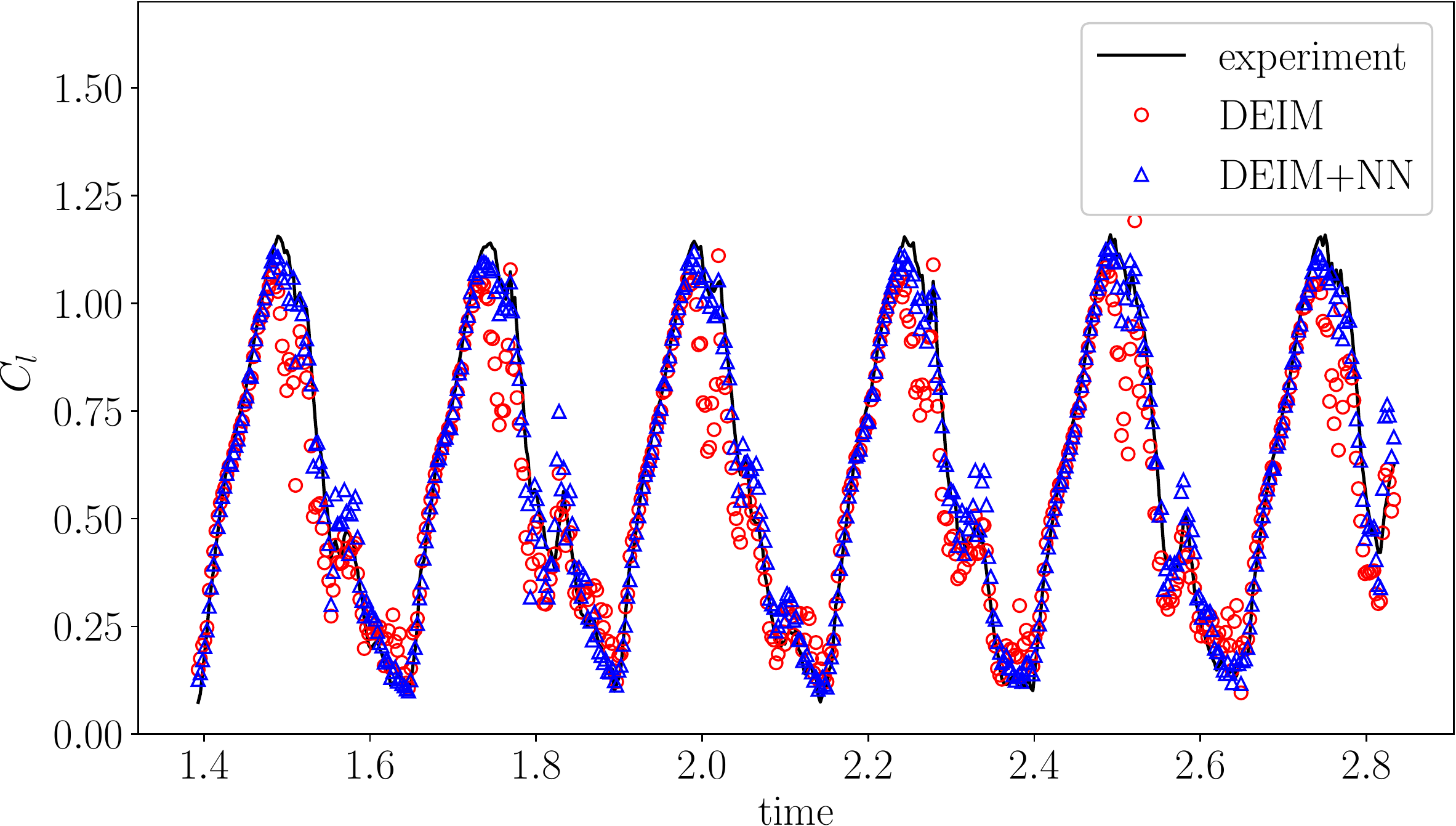}
        \caption{$n_s=5$, $C_l$}
    \end{subfigure}
    \begin{subfigure}[t]{0.49\textwidth}
        \centering
        \includegraphics[width=\linewidth]{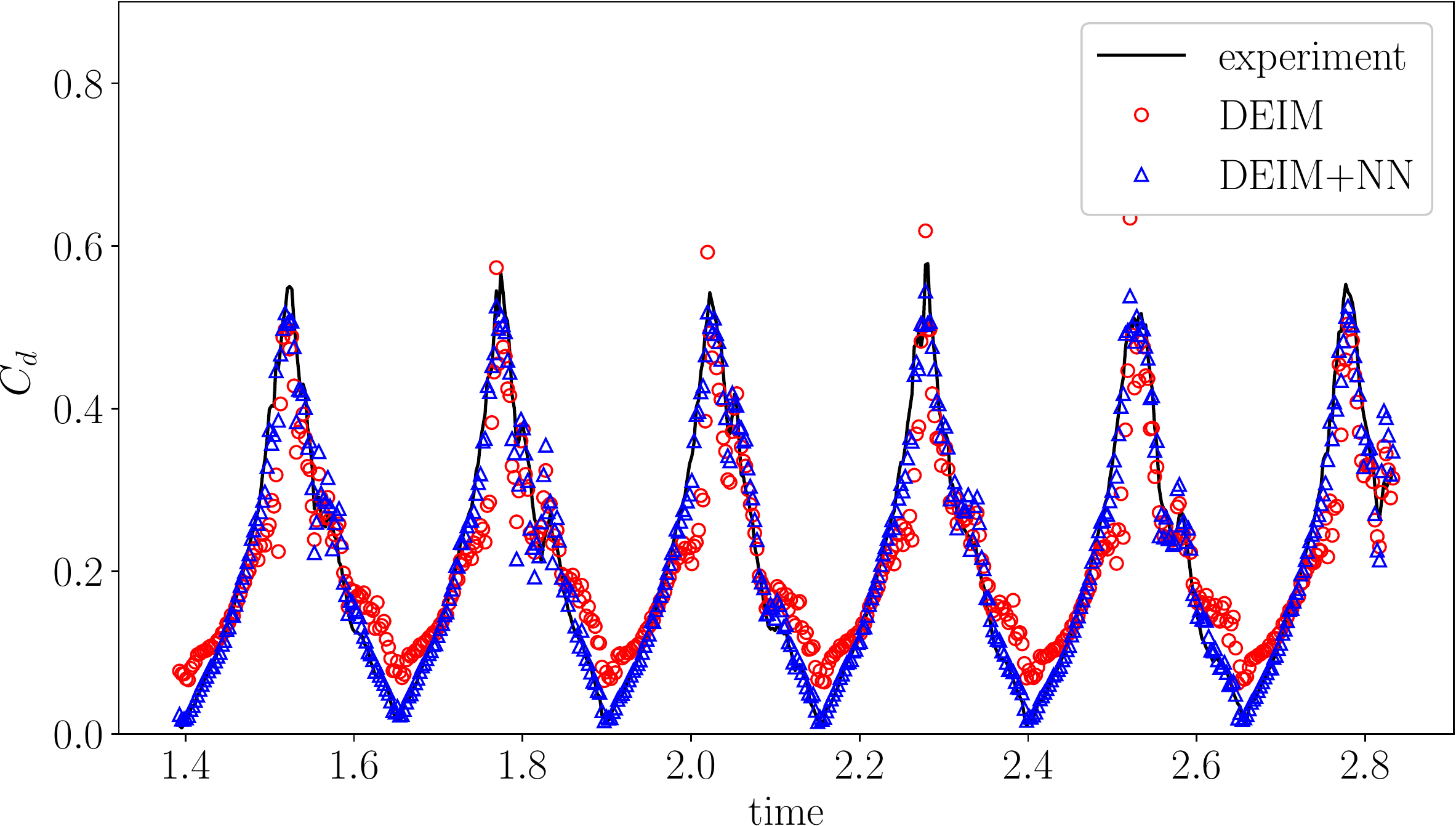}
        \caption{$n_s=5$, $C_d$}
    \end{subfigure}

    \begin{subfigure}[t]{0.49\textwidth}
        \centering
        \includegraphics[width=\linewidth]{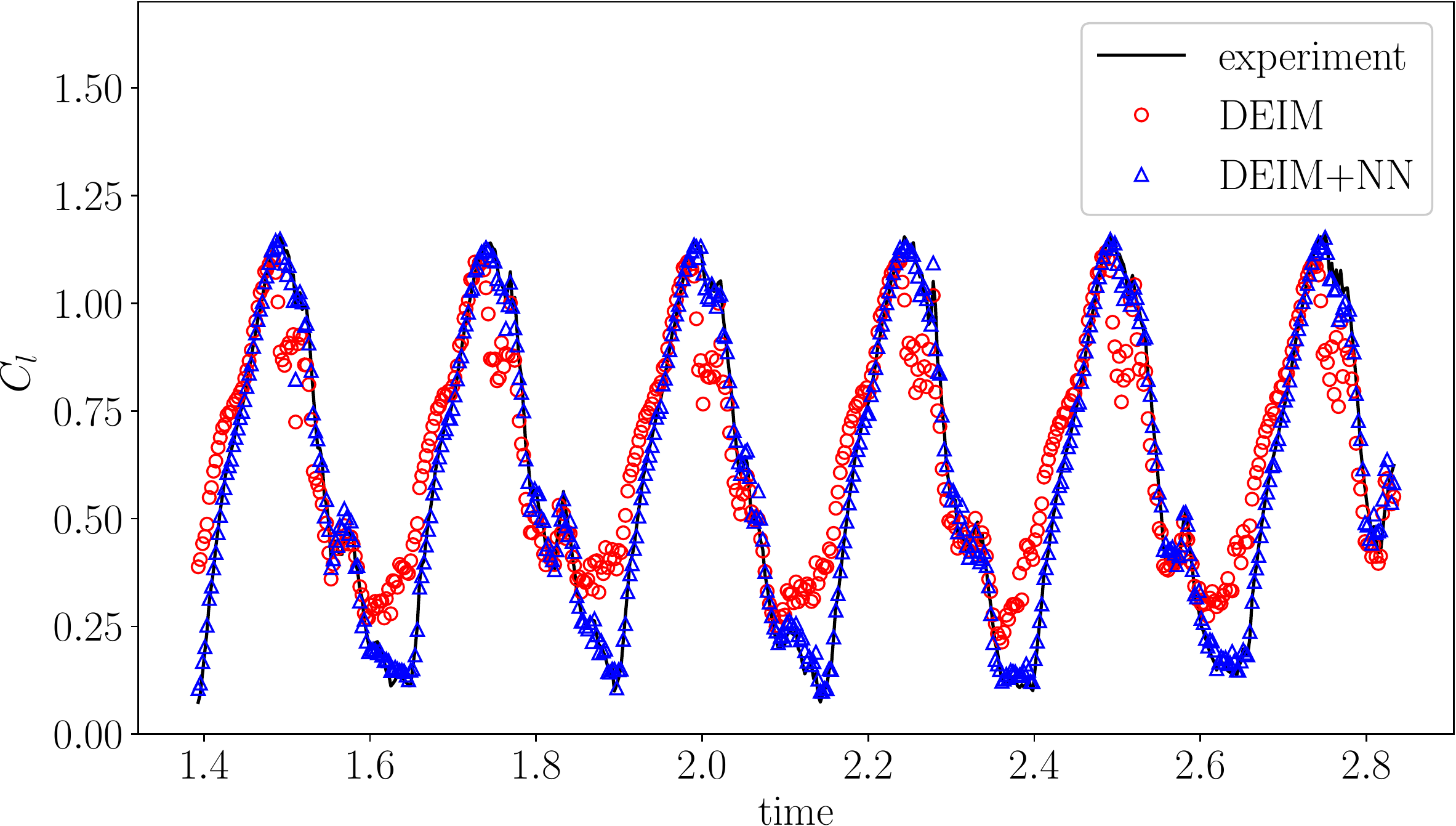}
        \caption{$n_s=8$, $C_l$}
    \end{subfigure}
    \begin{subfigure}[t]{0.49\textwidth}
        \centering
        \includegraphics[width=\linewidth]{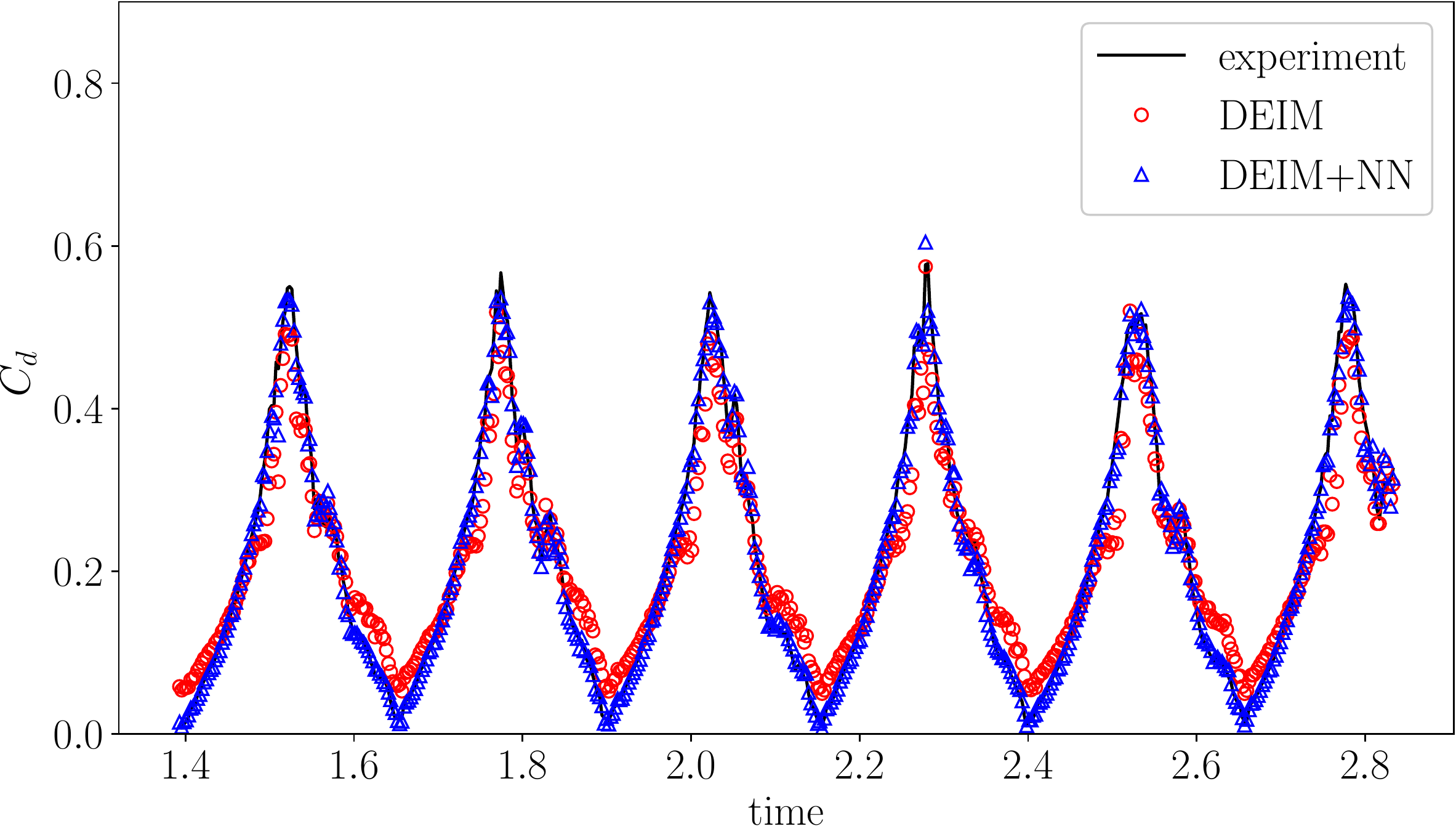}
        \caption{$n_s=8$, $C_d$}
    \end{subfigure}

    \begin{subfigure}[t]{0.49\textwidth}
        \centering
        \includegraphics[width=\linewidth]{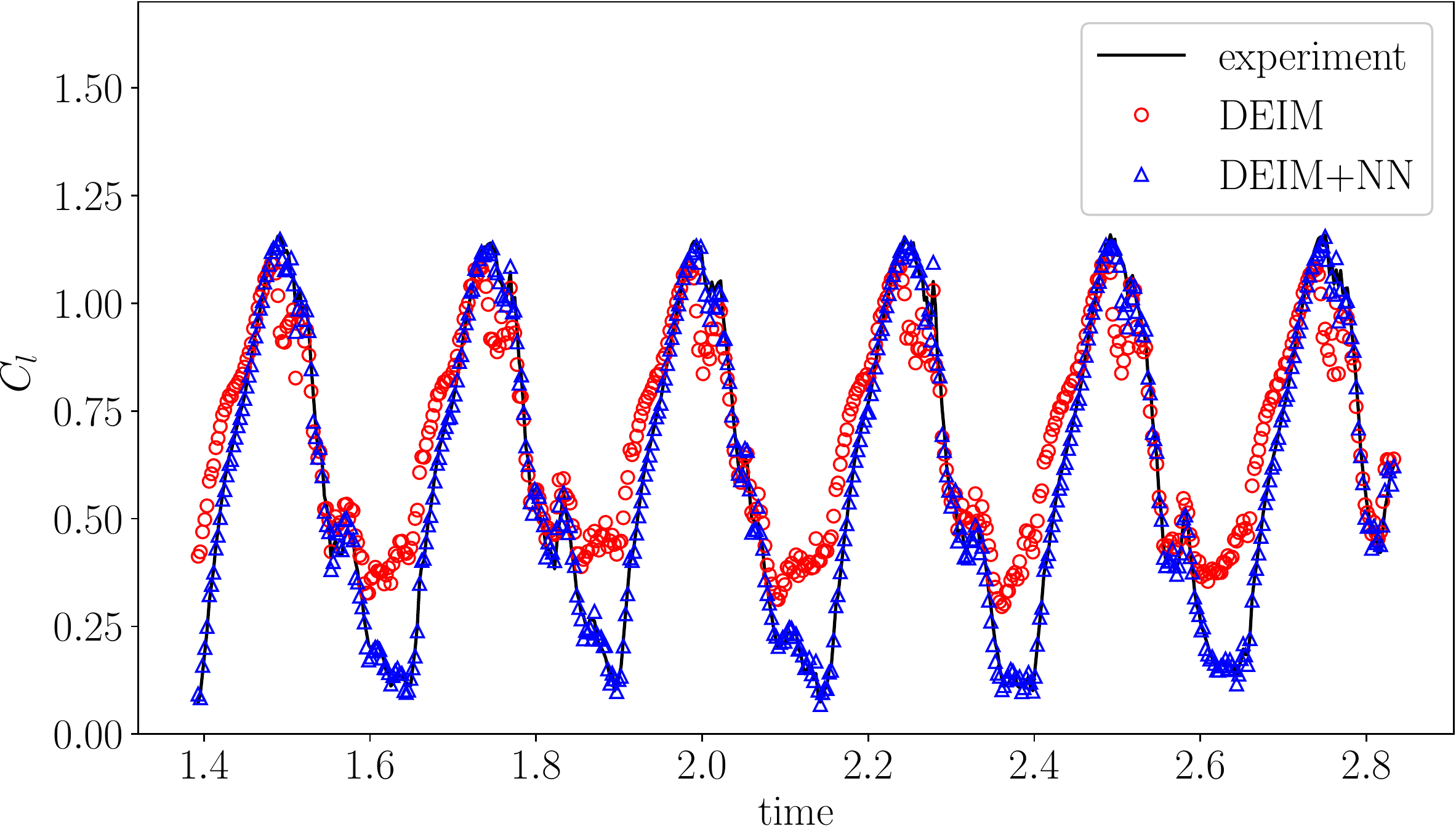}
        \caption{$n_s=10$, $C_l$}
    \end{subfigure}
    \begin{subfigure}[t]{0.49\textwidth}
        \centering
        \includegraphics[width=\linewidth]{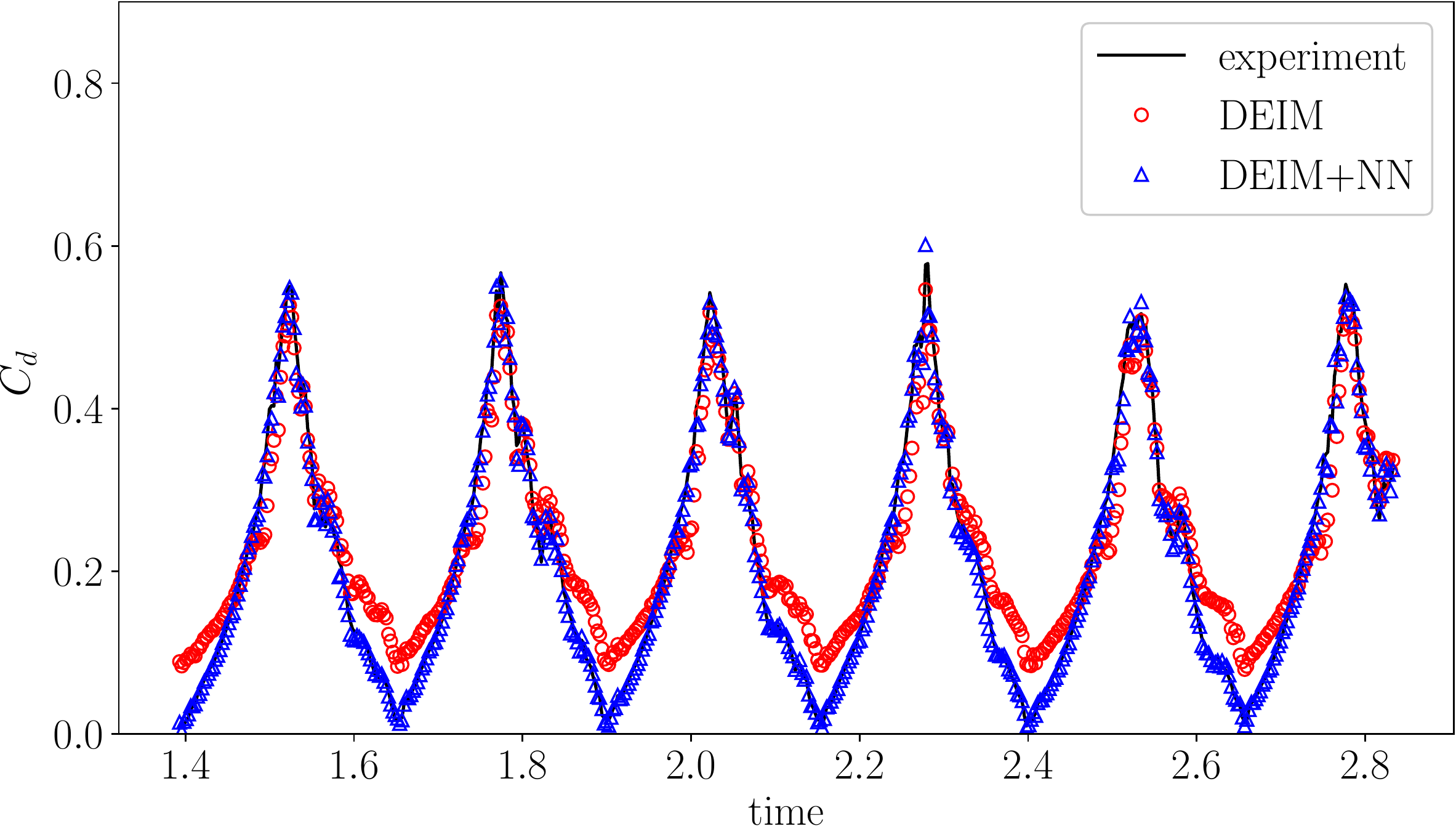}
        \caption{$n_s=10$, $C_d$}
    \end{subfigure}
    \caption{2D airfoil: $C_l,C_d$ w.r.t. time for the testing experimental data without noise in the pressure sensor inputs.
    The DEIM models are based on the URANS data.}
    \label{fig:2DAirfoil_numer_DEIM_location_aero_coeff_time_noise=0}
\end{figure}

To find a preferred architecture of the NN,
we perform a grid search with $2$, $3$, $4$ hidden layers,
$10$, $20$, $30$, $40$ neurons in each layer,
$10^{-5}$, $10^{-6}$, $10^{-7}$ weight decay,
with a mini-batch size of $64$ in the training.
When the DEIM model is built on the URANS data, the best NN architecture is obtained when $n_s=10$,
consisting of $2$ hidden layers with $10$ neurons in each layer,
and a weight decay as $10^{-5}$.
Figures \ref{fig:2DAirfoil_exper_DEIM_location_aero_coeff_aoa_noise=0}-\ref{fig:2DAirfoil_numer_DEIM_location_aero_coeff_aoa_noise=0} show the lift and drag coefficients $C_l, C_d$ with respect to the angle of attack $\alpha$ for different $n_s$ in two cases,
where only one whole period is presented.
Figures \ref{fig:2DAirfoil_exper_DEIM_location_aero_coeff_time_noise=0}-\ref{fig:2DAirfoil_numer_DEIM_location_aero_coeff_time_noise=0} plot the evolution of $C_l, C_d$ with respect to time.
The maximal $C_l$ appears at $\alpha\approx \qty{26}{deg}$.
When the airfoil is near the dynamic stall region with $\alpha$ in $\left[24,28\right] \text{deg}$,
the errors of the DEIM prediction become larger.
The lift and drag coefficients $C_l, C_d$ from the DEIM prediction deviate from the experimental data,
especially near the minimal and maximal $\alpha$,
and increasing $n_s$ does not improve the results.
It is observed that the predictions from the DEIM models based on the experimental data are more accurate, as expected.
The URANS data based DEIM model can capture the main dynamics, so it makes sense to calibrate the model by adding a correction term.
The results obtained by adding the NN corrections are much closer to the experimental data in both cases,
especially near the minimum of $C_l, C_d$, see Figs. \ref{fig:2DAirfoil_exper_DEIM_location_aero_coeff_time_noise=0}-\ref{fig:2DAirfoil_numer_DEIM_location_aero_coeff_time_noise=0}.
This verifies the effectiveness of the NN correction term and our proposed approach employing the data fusion from the numerical simulation and experiment.

The following $\ell^2$ and $\ell^\infty$ errors in the lift and drag coefficients
for testing times $t_{\tt testing}$ and frequencies $f_{\tt testing}$
are evaluated,
\begin{align*}
    &\epsilon_{a}^z=
    \sqrt{\dfrac{1}{N_{t_{\tt testing}}N_{f_{\tt testing}}}
    \sum\limits_{\substack{t\in t_{\tt testing}\\ f\in f_{\tt testing}}}
	\left|C_a^{\tt exper}(t, f)
		- C_a^z(t, f)\right|^2},\\
    &\epsilon_{a,\infty}^z=\max\limits_{\substack{t\in t_{\tt testing}\\ f\in f_{\tt testing}}}
    \left|C_a^{\tt exper}(t, f)
		- C_a^z(t, f)\right|,\\
    &\alpha(\epsilon_{a, \infty}^z)=\mathop{\arg\max}\limits_{\substack{t\in t_{\tt testing}\\ f\in f_{\tt testing}}}
    \left|C_a^{\tt exper}(t, f)
		- C_a^z(t, f)\right|,
\end{align*}
where $N_{t_{\tt testing}}$ and $N_{f_{\tt testing}}$ are the numbers of the testing times and frequencies, respectively,
with $a=l, d$ and $z={\tt DEIM}, {\tt NN}$.
The errors are listed in Tables \ref{tab:2DAirfoil_numer_DEIM_location_lift_err}-\ref{tab:2DAirfoil_numer_DEIM_location_drag_err}.
The results show that the DEIM combined with the NN gives more accurate predictions than only using the DEIM in both cases,
indicating that the NN correction term can accurately bridge the gap between the DEIM prediction and the ground truth.
Although the DEIM models based on experimental data are more accurate,
the number of sensors in 3D is limited so the experimental data cannot be used to obtain optimal sensor locations.
One observes that in the second case, the smallest errors in $C_l$ and $C_d$ with the NN correction are \num{2.21e-2} and \num{1.07e-2}, respectively,
and the DEIM+NN gives about $7$ and $4$ times more accurate $C_l$ and $C_d$ than DEIM, respectively.
The online CPU time costs of the DEIM and NN parts are recorded in Table \ref{tab:CPU_times},
highlighting the high computational efficiency of the proposed model.

\begin{table}[hbt!]
\caption{2D airfoil: the $\ell^2$ errors, $\ell^\infty$ errors in $C_l$, and the angles of attack corresponding to the $\ell^\infty$ errors for different $n_s$.}
\label{tab:2DAirfoil_numer_DEIM_location_lift_err}
\centering
\begin{tabular}{cccccccc}
\hline\hline
& & \multicolumn{3}{c}{DEIM} & \multicolumn{3}{c}{DEIM+NN} \\
\cmidrule(lr){3-5}\cmidrule(lr){6-8}
& $n_s$ & $\epsilon_{l}^{\tt DEIM}$ & $\epsilon_{l,\infty}^{\tt DEIM}$ & $\alpha(\epsilon_{l,\infty}^{\tt DEIM})$ & $\epsilon_{l}^{\tt NN}$ & $\epsilon_{l,\infty}^{\tt NN}$ & $\alpha(\epsilon_{l,\infty}^{\tt NN})$ \\
\cline{2-8}
& & \multicolumn{6}{c}{Experimental data based DEIM model}\\
\hline
\multirow{3}{*}{without noise} & $5$ & \num{3.97e-02} & \num{1.82e-01} & \qty{27.5}{deg} & \num{2.35e-02} & \num{9.97e-02} & \qty{28.0}{deg}\\
 % \hline
& $8$ & \num{3.49e-02} & \num{1.72e-01} & \qty{27.5}{deg} & \num{2.26e-02} & \num{9.46e-02} & \qty{24.6}{deg}\\
% \hline
& $10$ & \num{3.41e-02} & \num{1.89e-01} & \qty{27.5}{deg} & \num{2.17e-02} & \num{9.41e-02} & \qty{24.6}{deg}\\
\hline
\multirow{3}{*}{$1.5\%$ noise} & $5$ & \num{4.02e-02} & \num{1.80e-01} & \qty{27.5}{deg} & \num{2.41e-02} & \num{9.72e-02} & \qty{28.0}{deg}\\
 % \hline
& $8$ & \num{3.56e-02} & \num{1.71e-01} & \qty{27.5}{deg} & \num{2.34e-02} & \num{9.50e-02} & \qty{24.6}{deg}\\
% \hline
& $10$ & \num{3.45e-02} & \num{1.87e-01} & \qty{27.5}{deg} & \num{2.23e-02} & \num{9.60e-02} & \qty{24.8}{deg}\\
\hline
& & \multicolumn{6}{c}{URANS data based DEIM model}\\
\hline
\multirow{3}{*}{without noise} & $5$ & \num{1.14e-01} & \num{4.83e-01} & \qty{27.5}{deg} & \num{5.37e-02} & \num{2.89e-01} & \qty{21.9}{deg}\\
 % \hline
& $8$ & \num{1.45e-01} & \num{3.29e-01} & \qty{12.3}{deg} & \num{2.55e-02} & \num{1.71e-01} & \qty{27.6}{deg}\\
% \hline
& $10$ & \num{1.56e-01} & \num{3.57e-01} & \qty{12.4}{deg} & \num{2.21e-02} & \num{1.22e-01} & \qty{27.0}{deg}\\
\hline
\multirow{3}{*}{$1.5\%$ noise} & $5$ & \num{1.14e-01} & \num{4.79e-01} & \qty{27.6}{deg} & \num{5.41e-02} & \num{2.87e-01} & \qty{21.9}{deg}\\
 % \hline
& $8$ & \num{1.45e-01} & \num{3.31e-01} & \qty{27.6}{deg} & \num{2.59e-02} & \num{1.72e-01} & \qty{27.6}{deg}\\
% \hline
& $10$ & \num{1.56e-01} & \num{3.56e-01} & \qty{12.4}{deg} & \num{2.24e-02} & \num{1.22e-01} & \qty{27.0}{deg}\\
\hline\hline
\end{tabular}
\end{table}

\begin{table}[hbt!]
\caption{2D airfoil: the $\ell^2$ errors, $\ell^\infty$ errors in $C_d$, and the angles of attack corresponding to the $\ell^\infty$ errors for different $n_s$.}
\label{tab:2DAirfoil_numer_DEIM_location_drag_err}
\centering
\begin{tabular}{cccccccc}
\hline\hline
& & \multicolumn{3}{c}{DEIM} & \multicolumn{3}{c}{DEIM+NN} \\
\cmidrule(lr){3-5}\cmidrule(lr){6-8}
& $n_s$ & $\epsilon_{d}^{\tt DEIM}$ & $\epsilon_{d,\infty}^{\tt DEIM}$ & $\alpha(\epsilon_{d,\infty}^{\tt DEIM})$ & $\epsilon_{d}^{\tt NN}$ & $\epsilon_{d,\infty}^{\tt NN}$ & $\alpha(\epsilon_{d,\infty}^{\tt NN})$ \\
\cline{2-8}
& & \multicolumn{6}{c}{Experimental data based DEIM model}\\
\hline
\multirow{3}{*}{without noise} & $5$ & \num{1.60e-02} & \num{9.64e-02} & \qty{28.0}{deg} & \num{1.16e-02} & \num{6.98e-02} & \qty{28.0}{deg}\\
 % \hline
& $8$ & \num{1.41e-02} & \num{9.43e-02} & \qty{27.5}{deg} & \num{9.69e-03} & \num{4.97e-02} & \qty{28.0}{deg}\\
% \hline
& $10$ & \num{1.47e-02} & \num{9.65e-02} & \qty{27.5}{deg} & \num{9.60e-03} & \num{4.46e-02} & \qty{24.5}{deg}\\
\hline
\multirow{3}{*}{$1.5\%$ noise} & $5$ & \num{1.61e-02} & \num{1.01e-01} & \qty{28.0}{deg} & \num{1.17e-02} & \num{6.88e-02} & \qty{28.0}{deg}\\
 % \hline
& $8$ & \num{1.43e-02} & \num{9.13e-02} & \qty{27.5}{deg} & \num{9.90e-03} & \num{4.87e-02} & \qty{28.0}{deg}\\
% \hline
& $10$ & \num{1.48e-02} & \num{9.36e-02} & \qty{27.5}{deg} & \num{9.82e-03} & \num{4.49e-02} & \qty{24.5}{deg}\\
\hline
& & \multicolumn{6}{c}{URANS data based DEIM model}\\
\hline
\multirow{3}{*}{without noise} & $5$ & \num{5.87e-02} & \num{2.49e-01} & \qty{27.5}{deg} & \num{2.40e-02} & \num{1.39e-01} & \qty{27.1}{deg}\\
 % \hline
& $8$ & \num{4.40e-02} & \num{1.66e-01} & \qty{27.6}{deg} & \num{1.13e-02} & \num{8.14e-02} & \qty{27.6}{deg}\\
% \hline
& $10$ & \num{4.83e-02} & \num{1.07e-01} & \qty{26.4}{deg} & \num{1.07e-02} & \num{6.26e-02} & \qty{27.9}{deg}\\
\hline
\multirow{3}{*}{$1.5\%$ noise} & $5$ & \num{5.87e-02} & \num{2.48e-01} & \qty{27.5}{deg} & \num{2.41e-02} & \num{1.34e-01} & \qty{27.1}{deg}\\
 % \hline
& $8$ & \num{4.40e-02} & \num{1.68e-01} & \qty{27.6}{deg} & \num{1.14e-02} & \num{8.18e-02} & \qty{27.6}{deg}\\
% \hline
& $10$ & \num{4.83e-02} & \num{1.07e-01} & \qty{26.4}{deg} & \num{1.07e-02} & \num{6.44e-02} & \qty{27.9}{deg}\\
\hline\hline
\end{tabular}
\end{table}

\begin{figure}[hbt!]
    \centering
    \begin{subfigure}[t]{0.49\textwidth}
        \centering
        \includegraphics[width=\linewidth]{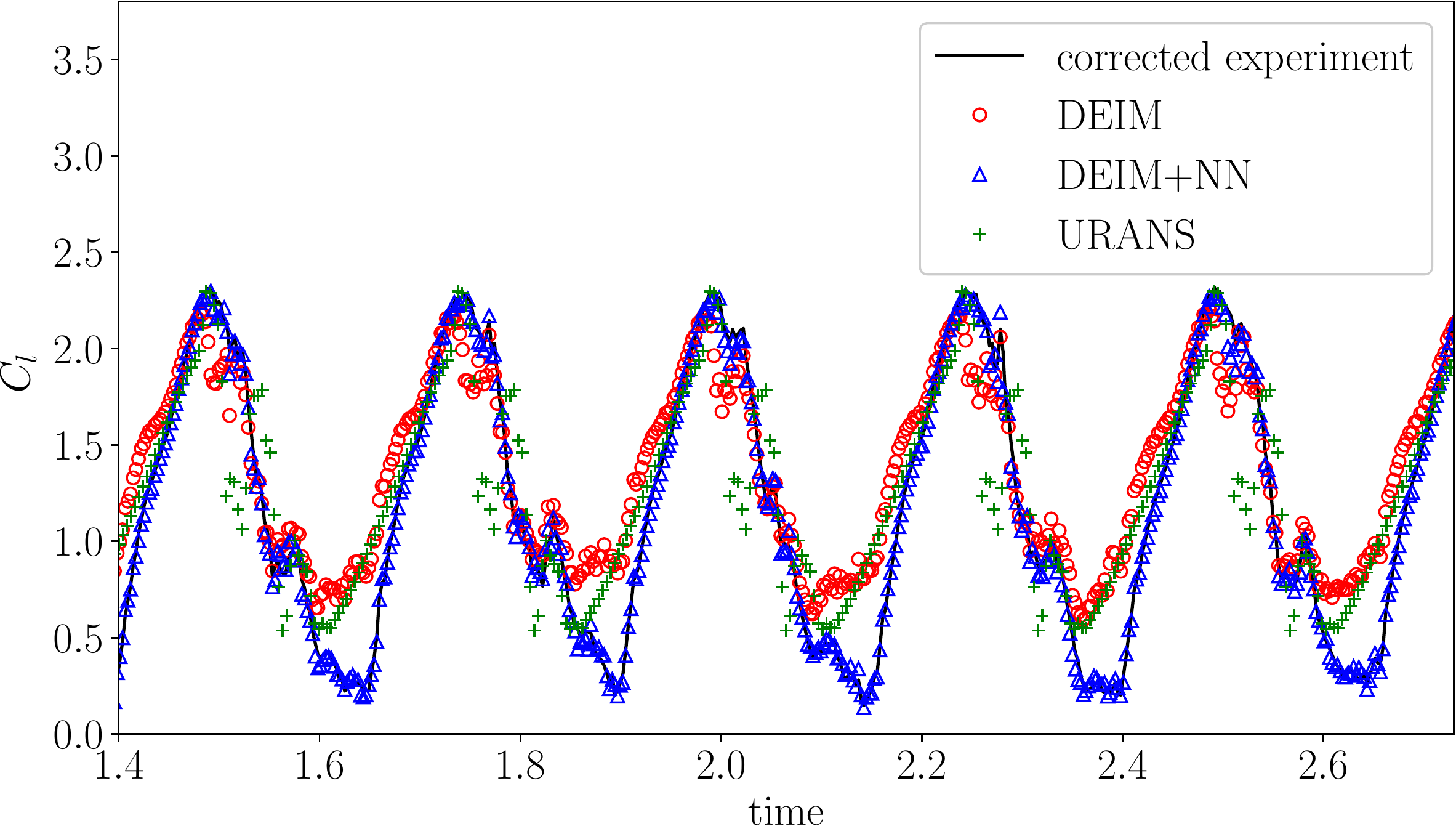}
        \caption{$n_s=10$, corrected $C_l$}
    \end{subfigure}
    \begin{subfigure}[t]{0.49\textwidth}
        \centering
        \includegraphics[width=\linewidth]{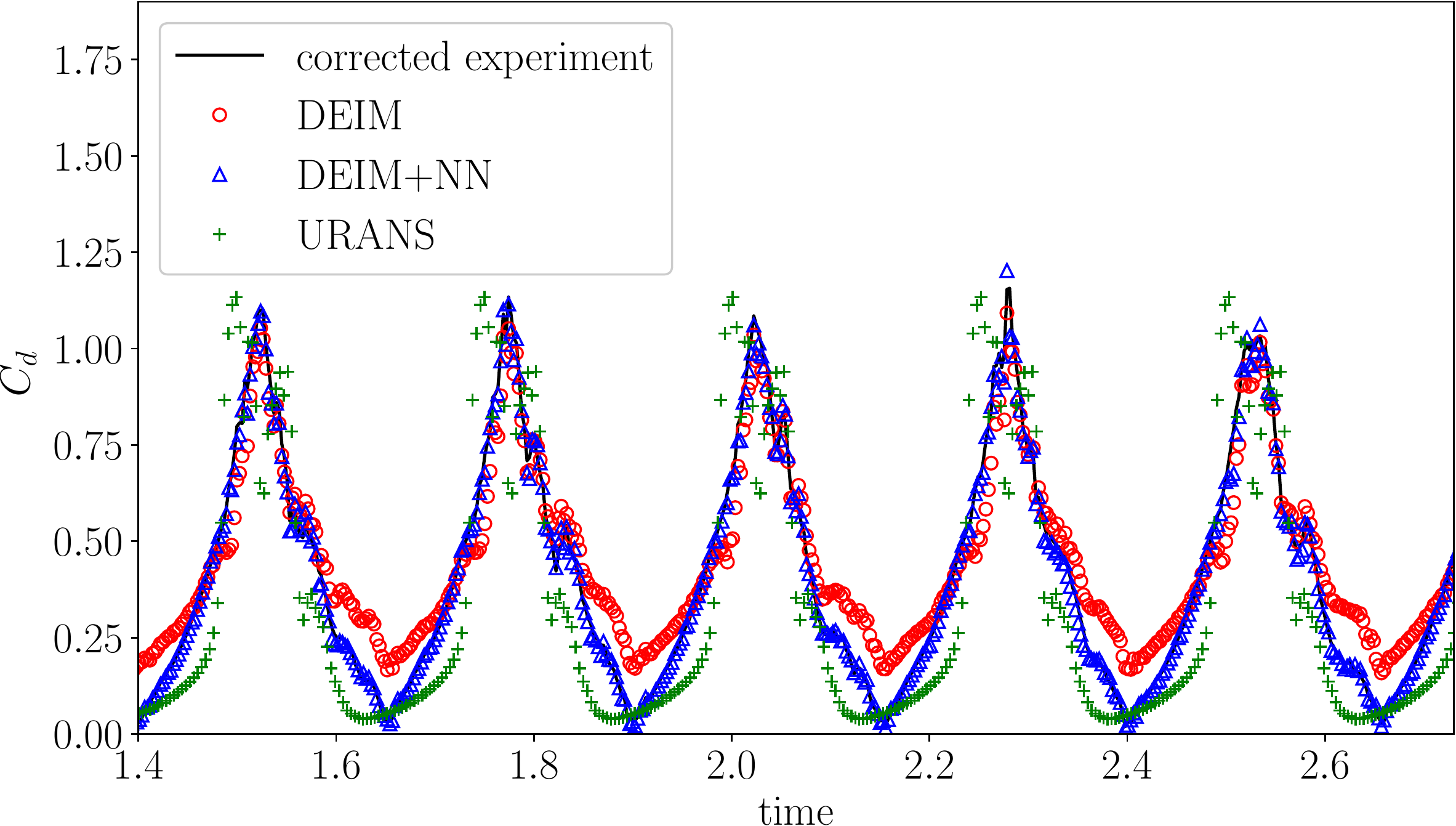}
        \caption{$n_s=10$, corrected $C_d$}
    \end{subfigure}
    \caption{2D airfoil: The corrected $C_l, C_d$ w.r.t. time for the testing experimental data without noise in the pressure sensor inputs.
    The DEIM model is based on the URANS data.}
    \label{fig:2DAirfoil_numer_DEIM_location_aero_coeff_time_noise=0_corrected}
\end{figure}

The predicted lift and drag coefficients $C_l, C_d$ are also corrected and compared with the URANS results, shown in Fig. \ref{fig:2DAirfoil_numer_DEIM_location_aero_coeff_time_noise=0_corrected}.
One observes that the URANS and DEIM prediction deviate from the corrected experimental results,
while the DEIM+NN gives accurate results.

\begin{figure}[hbt!]
    \centering
    \begin{subfigure}[t]{0.49\textwidth}
        \centering
        \includegraphics[width=\linewidth]{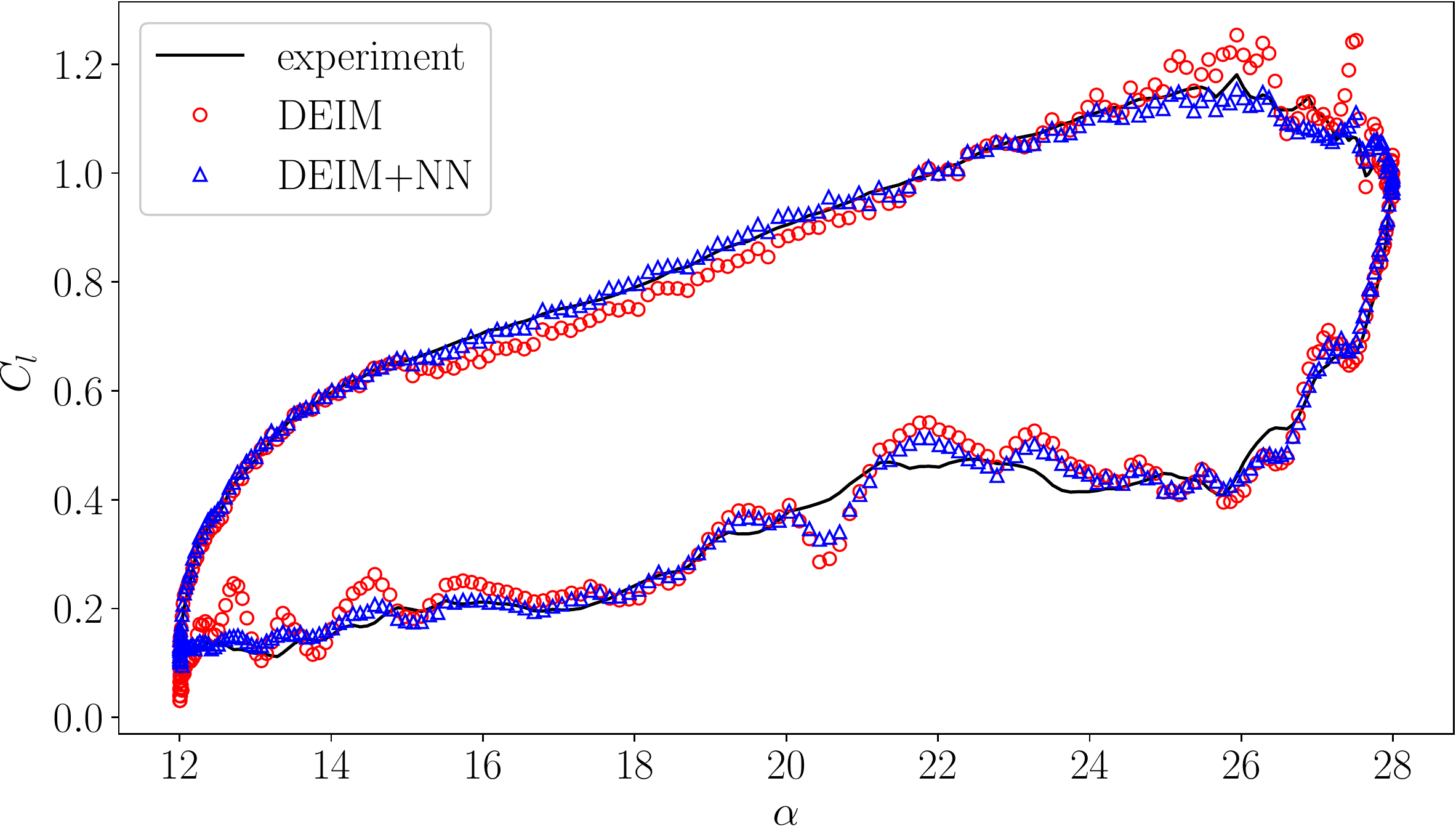}
        \caption{$n_s=5$, $C_l$}
    \end{subfigure}
    \begin{subfigure}[t]{0.49\textwidth}
        \centering
        \includegraphics[width=\linewidth]{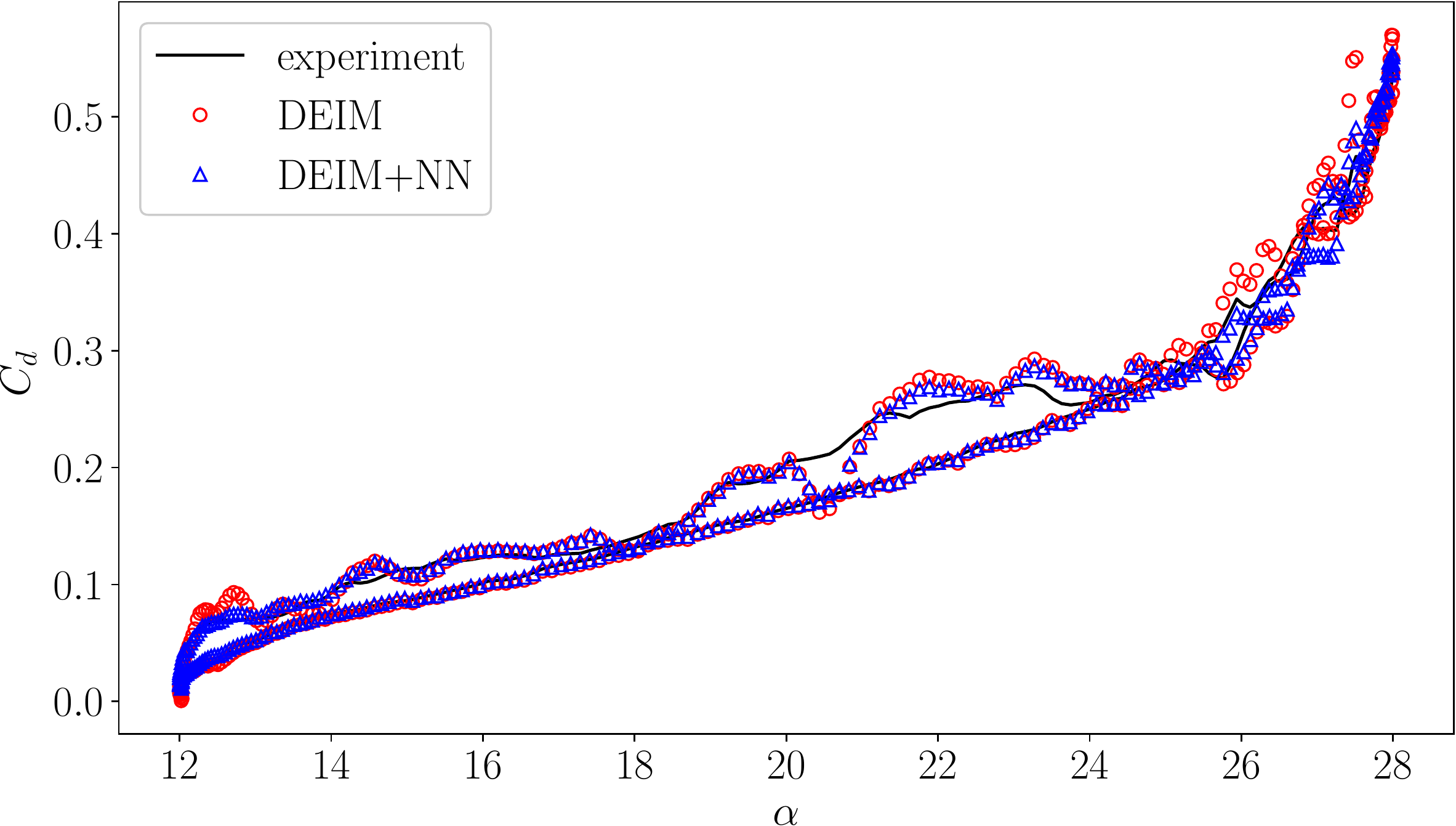}
        \caption{$n_s=5$, $C_d$}
    \end{subfigure}

    \begin{subfigure}[t]{0.49\textwidth}
        \centering
        \includegraphics[width=\linewidth]{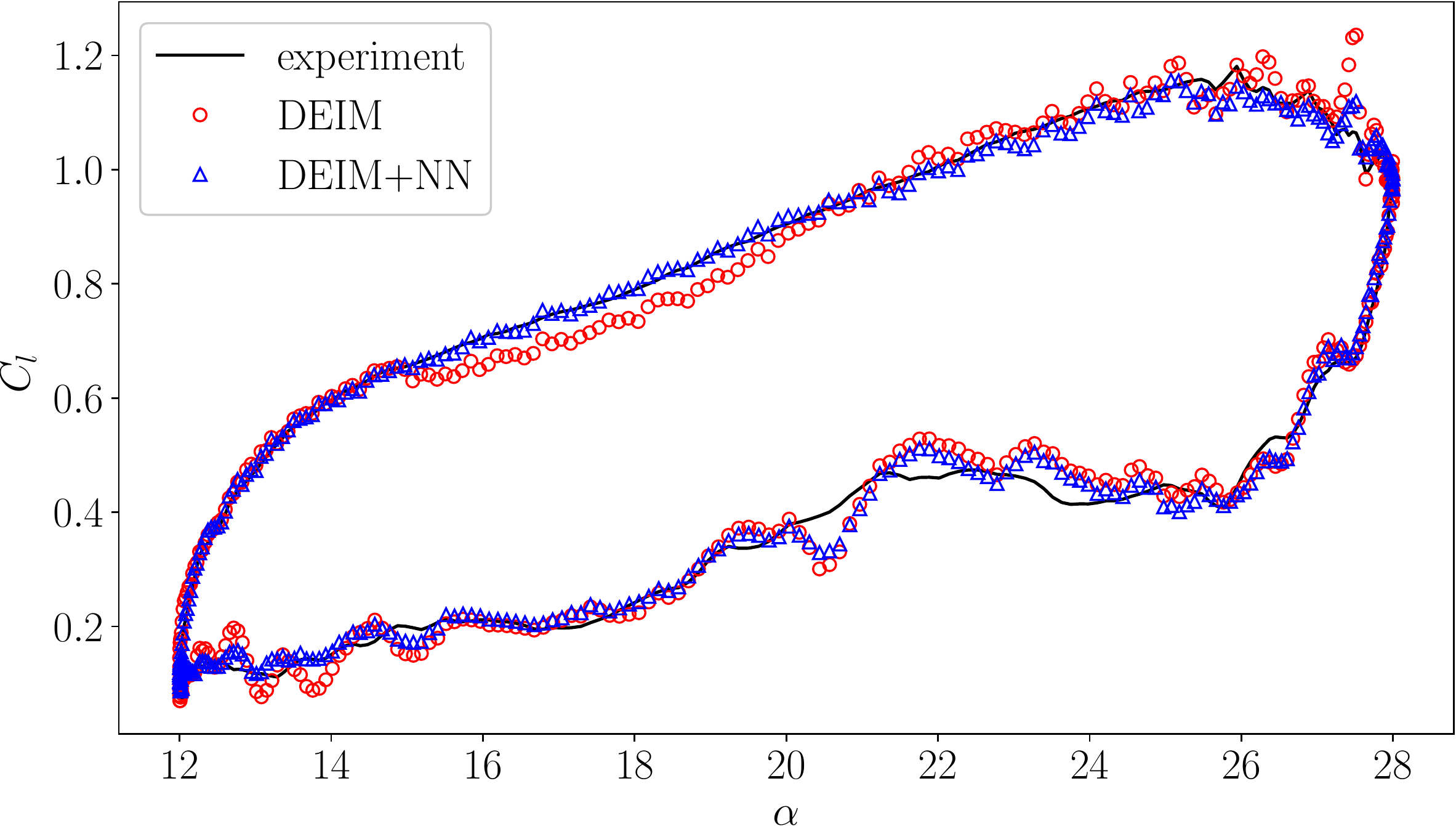}
        \caption{$n_s=8$, $C_l$}
    \end{subfigure}
    \begin{subfigure}[t]{0.49\textwidth}
        \centering
        \includegraphics[width=\linewidth]{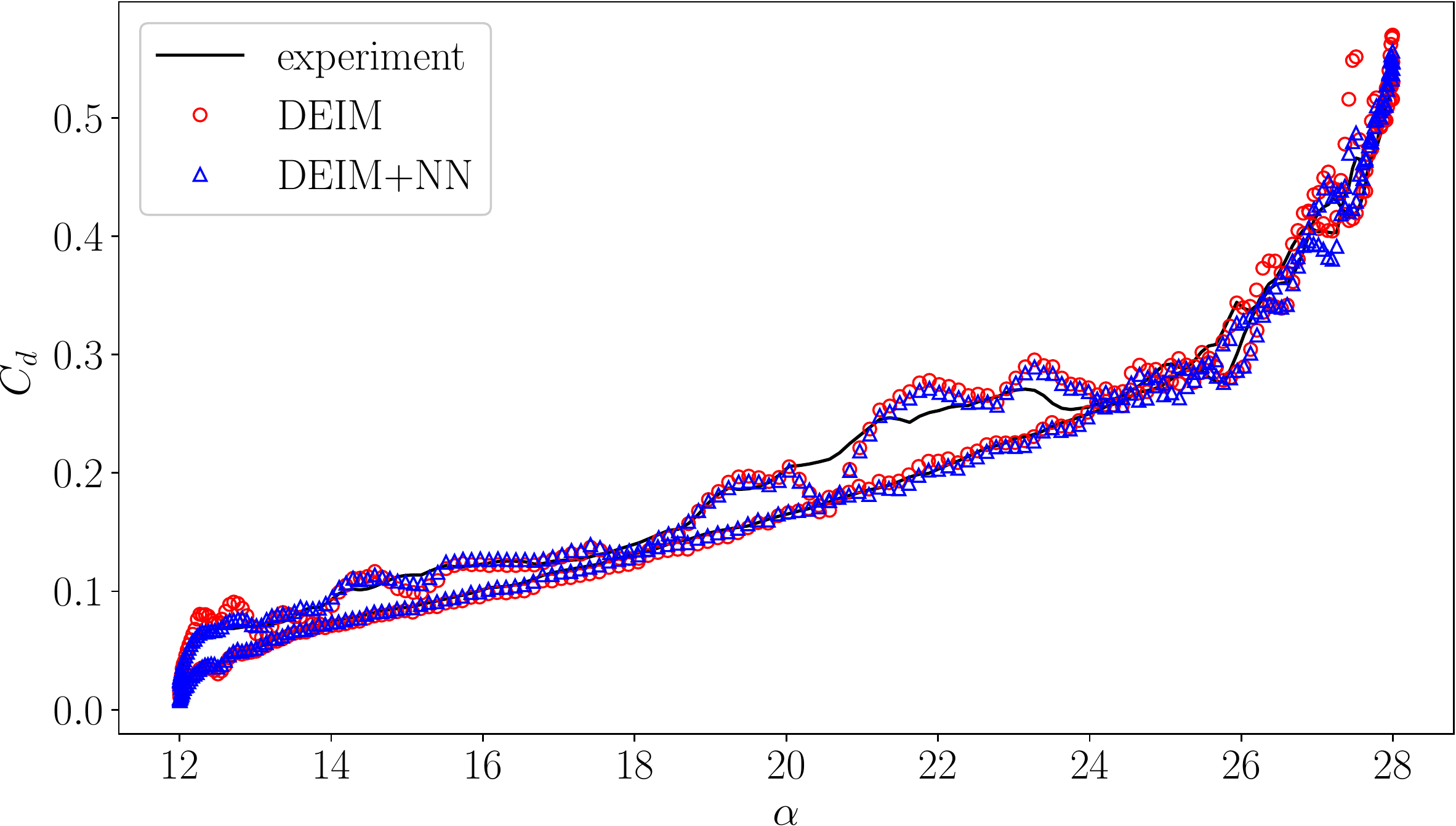}
        \caption{$n_s=8$, $C_d$}
    \end{subfigure}

    \begin{subfigure}[t]{0.49\textwidth}
        \centering
        \includegraphics[width=\linewidth]{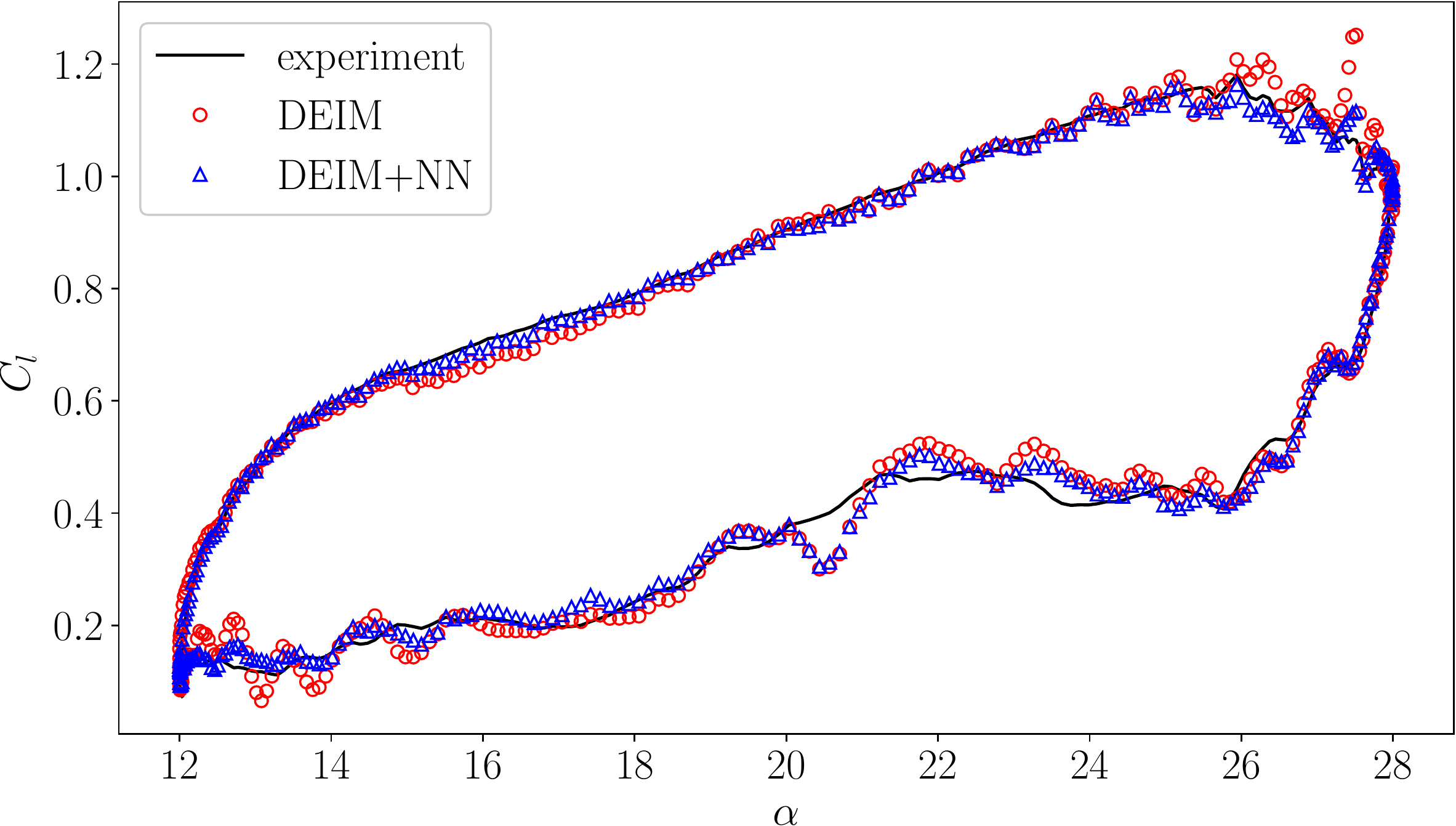}
        \caption{$n_s=10$, $C_l$}
    \end{subfigure}
    \begin{subfigure}[t]{0.49\textwidth}
        \centering
        \includegraphics[width=\linewidth]{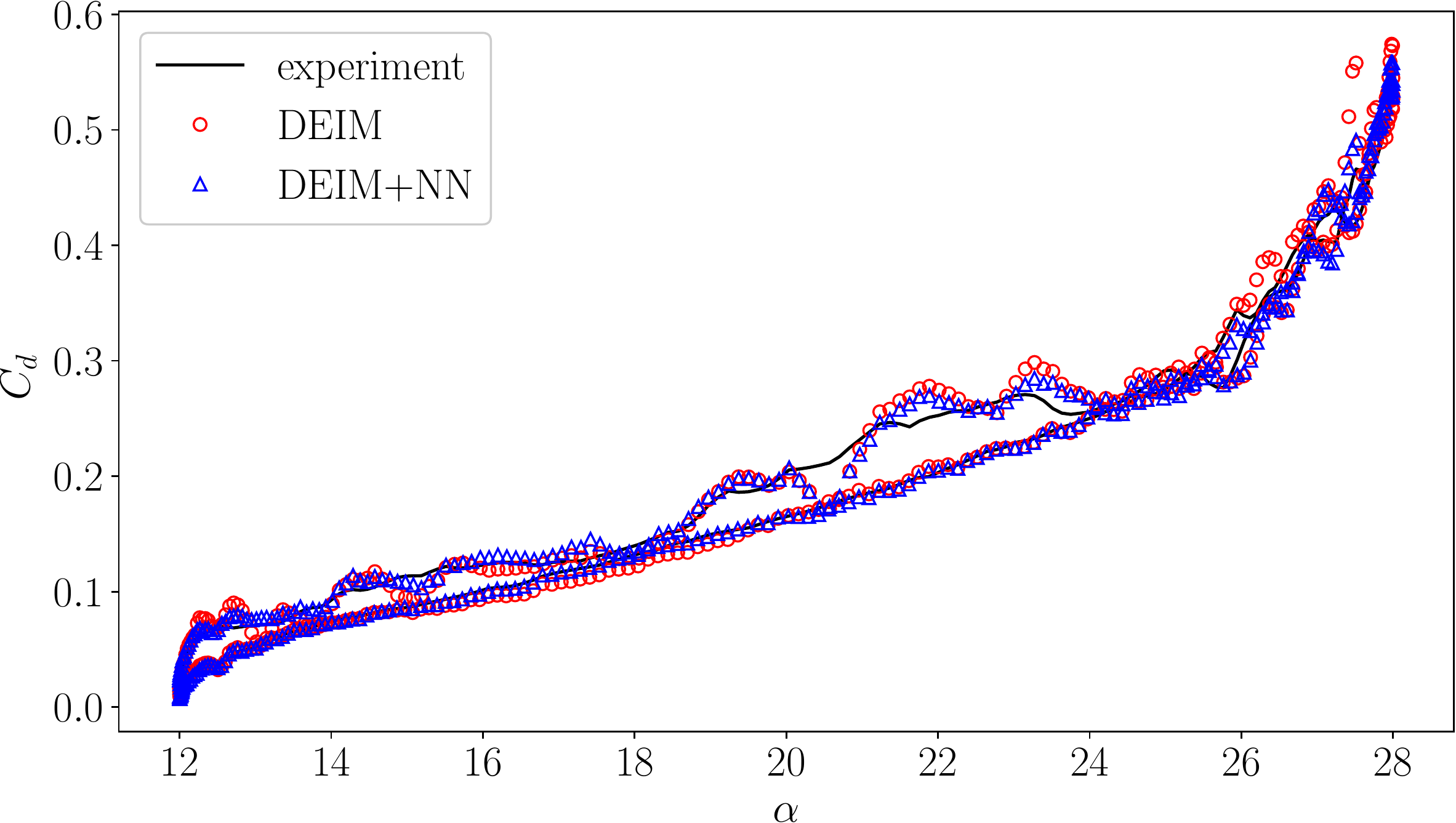}
        \caption{$n_s=10$, $C_d$}
    \end{subfigure}
    \caption{2D airfoil: $C_l,C_d$ w.r.t. $\alpha$ for the testing experimental data with $1.5\%$ noise in the pressure sensor inputs.
    The DEIM models are based on the experimental data.}
    \label{fig:2DAirfoil_exper_DEIM_location_aero_coeff_aoa_noise=1.5}
\end{figure}

\begin{figure}[hbt!]
    \centering
    \begin{subfigure}[t]{0.49\textwidth}
        \centering
        \includegraphics[width=\linewidth]{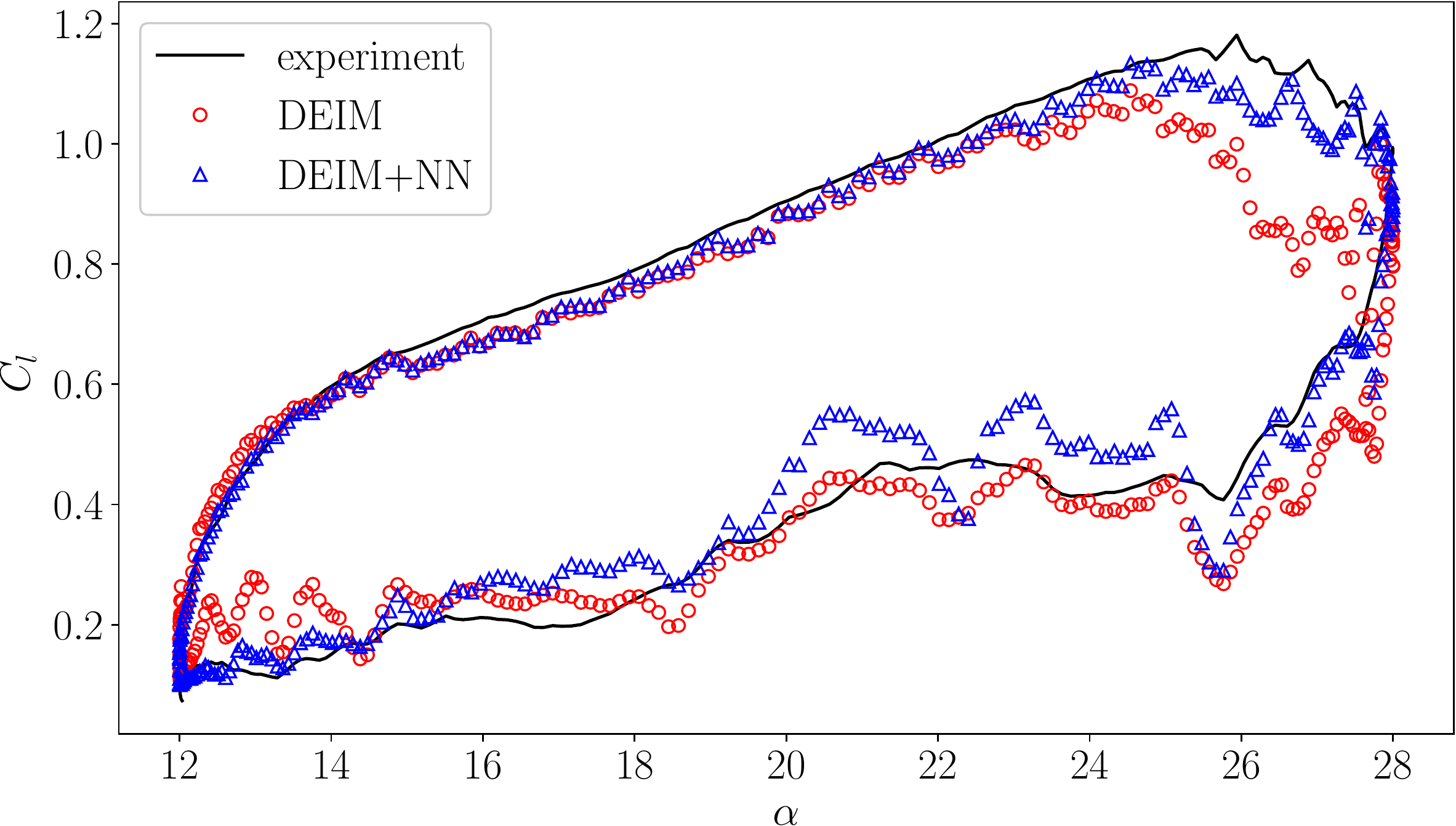}
        \caption{$n_s=5$, $C_l$}
    \end{subfigure}
    \begin{subfigure}[t]{0.49\textwidth}
        \centering
        \includegraphics[width=\linewidth]{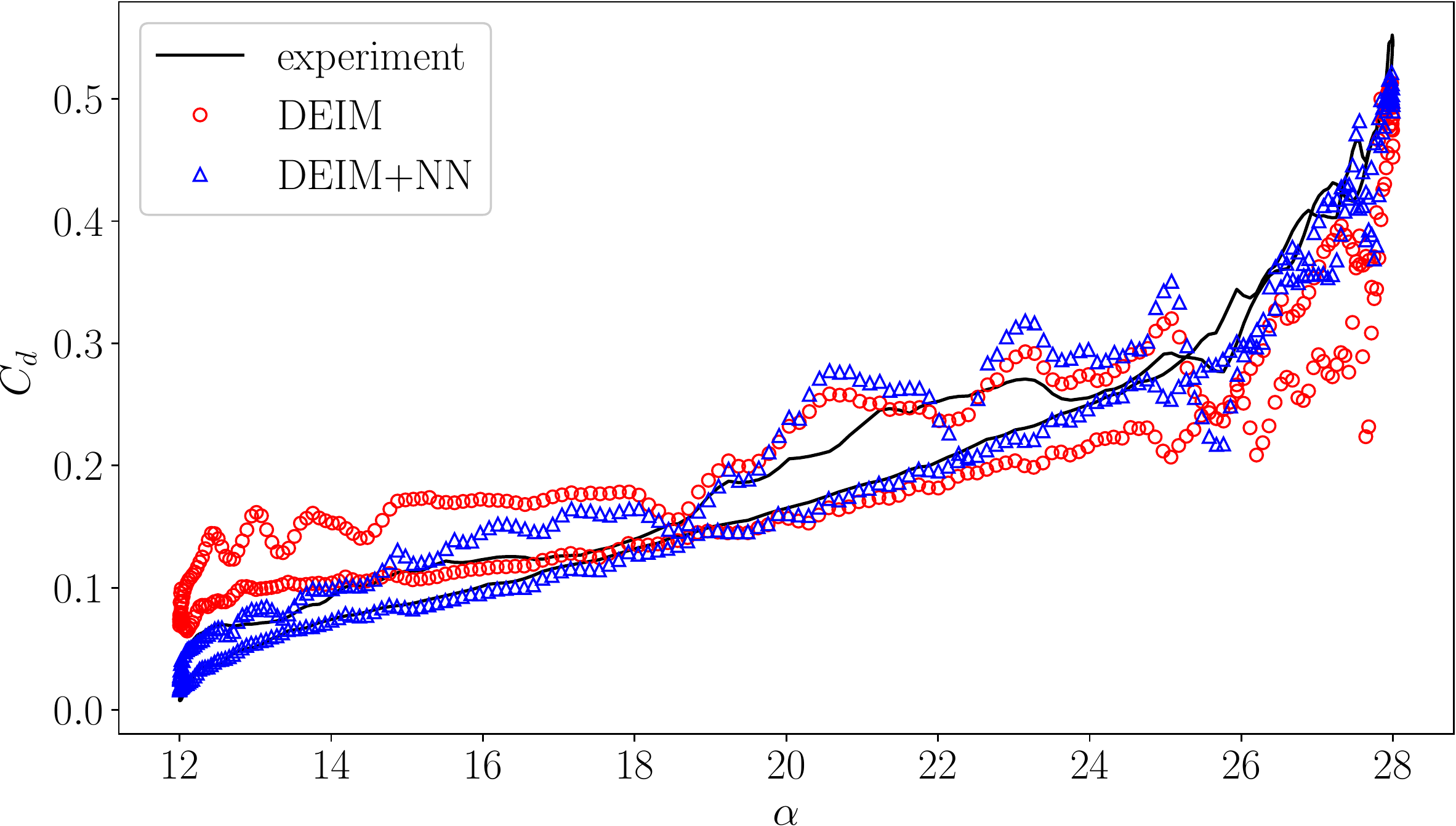}
        \caption{$n_s=5$, $C_d$}
    \end{subfigure}

    \begin{subfigure}[t]{0.49\textwidth}
        \centering
        \includegraphics[width=\linewidth]{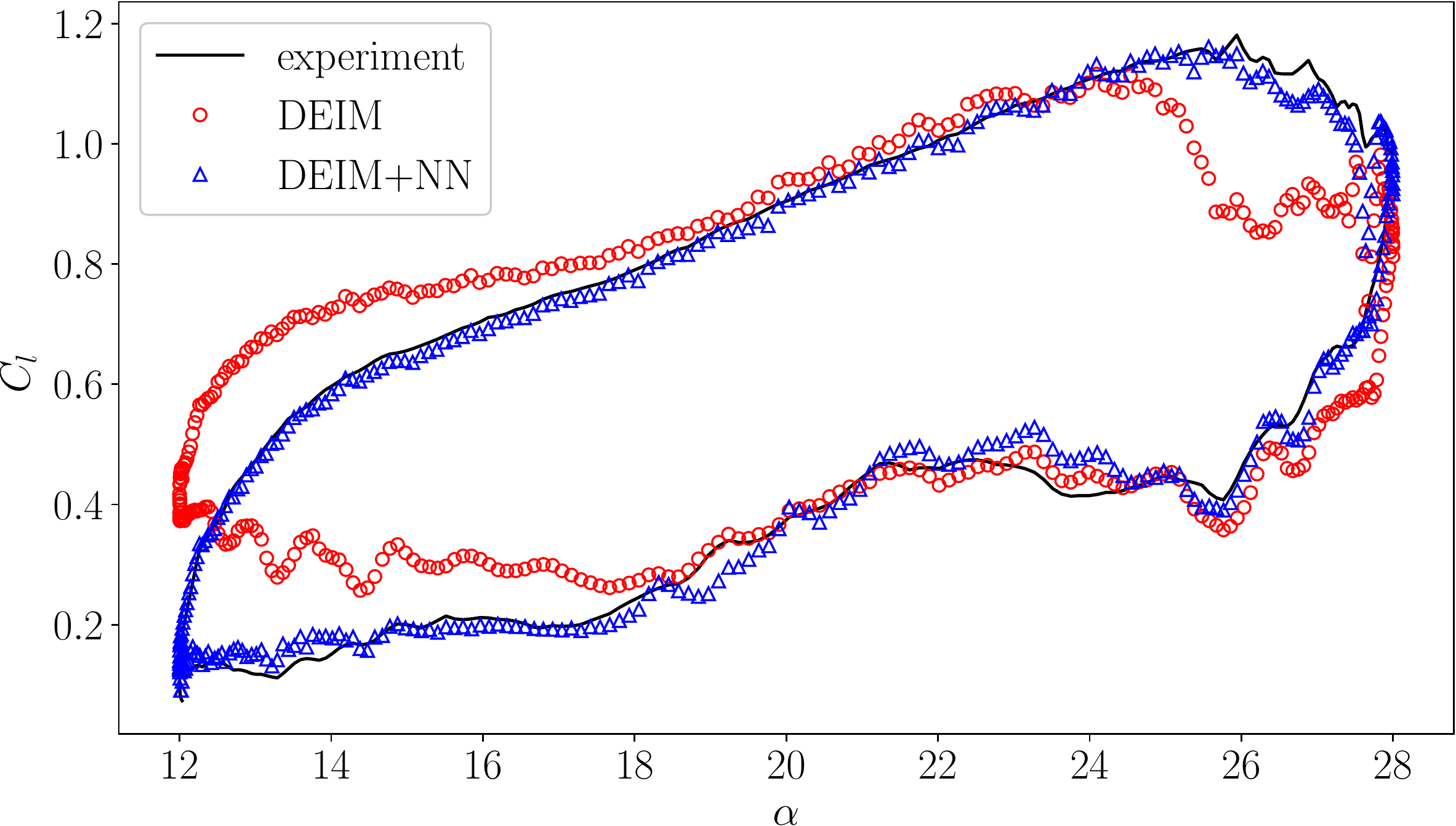}
        \caption{$n_s=8$, $C_l$}
    \end{subfigure}
    \begin{subfigure}[t]{0.49\textwidth}
        \centering
        \includegraphics[width=\linewidth]{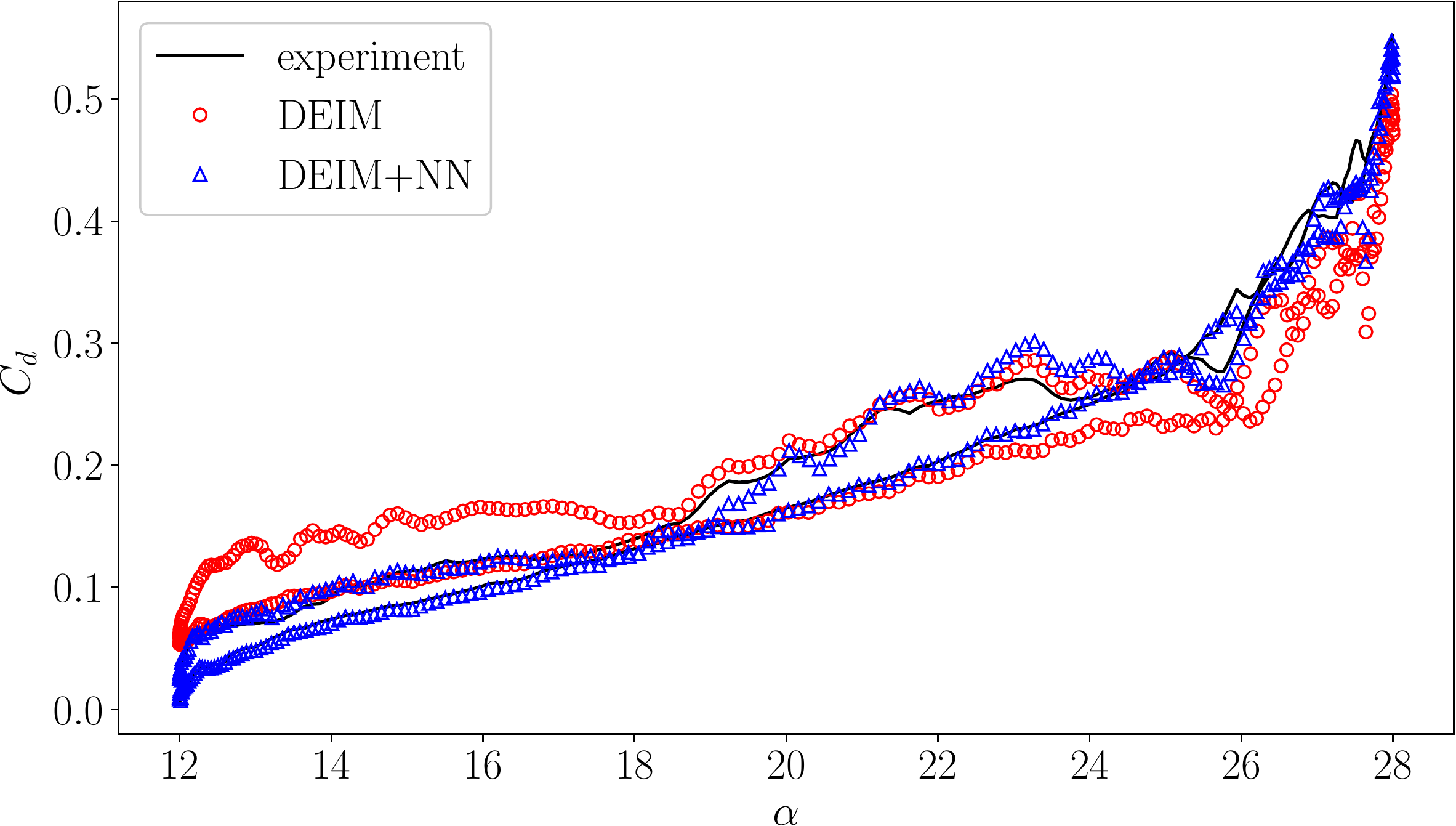}
        \caption{$n_s=8$, $C_d$}
    \end{subfigure}

    \begin{subfigure}[t]{0.49\textwidth}
        \centering
        \includegraphics[width=\linewidth]{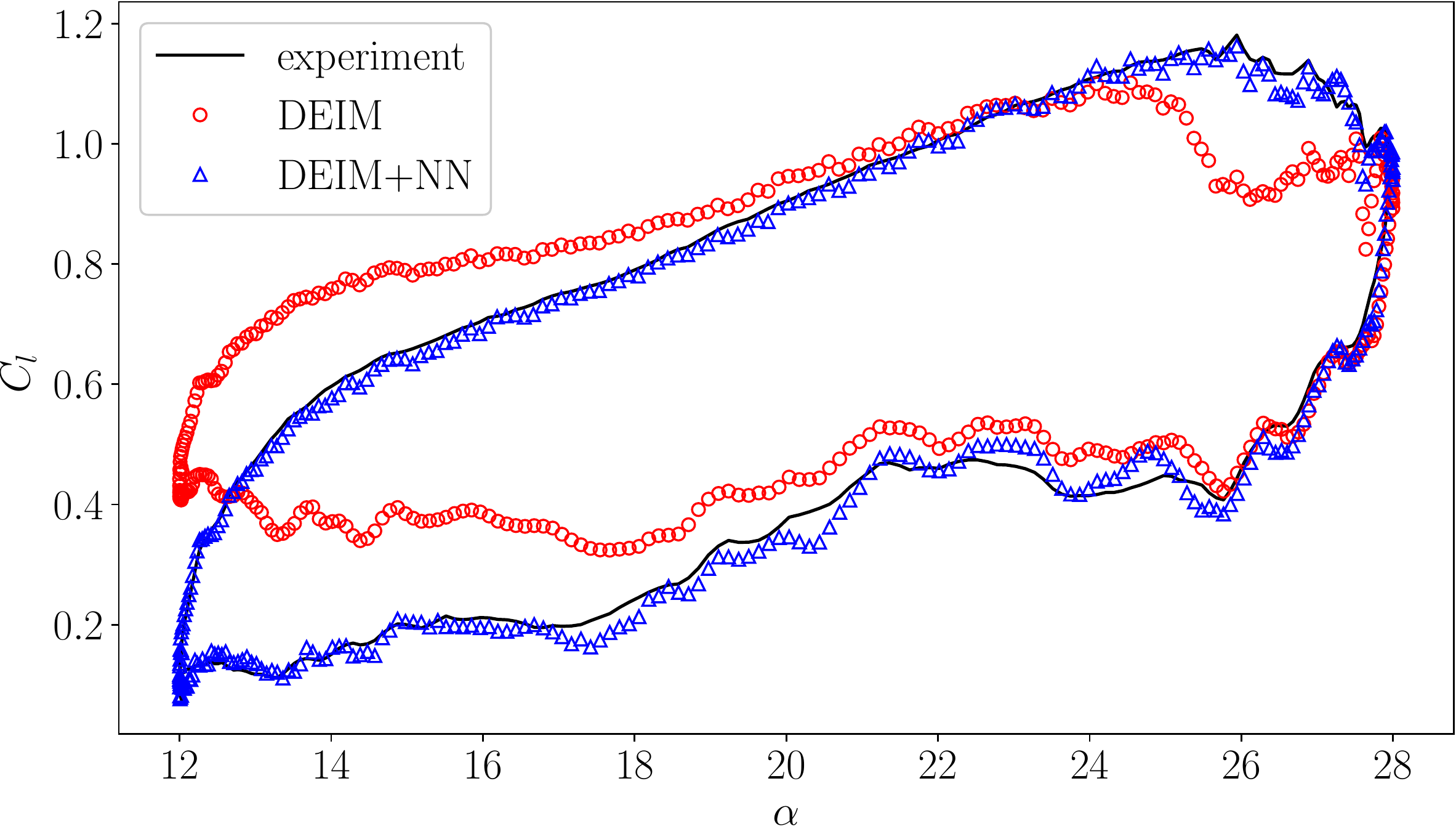}
        \caption{$n_s=10$, $C_l$}
    \end{subfigure}
    \begin{subfigure}[t]{0.49\textwidth}
        \centering
        \includegraphics[width=\linewidth]{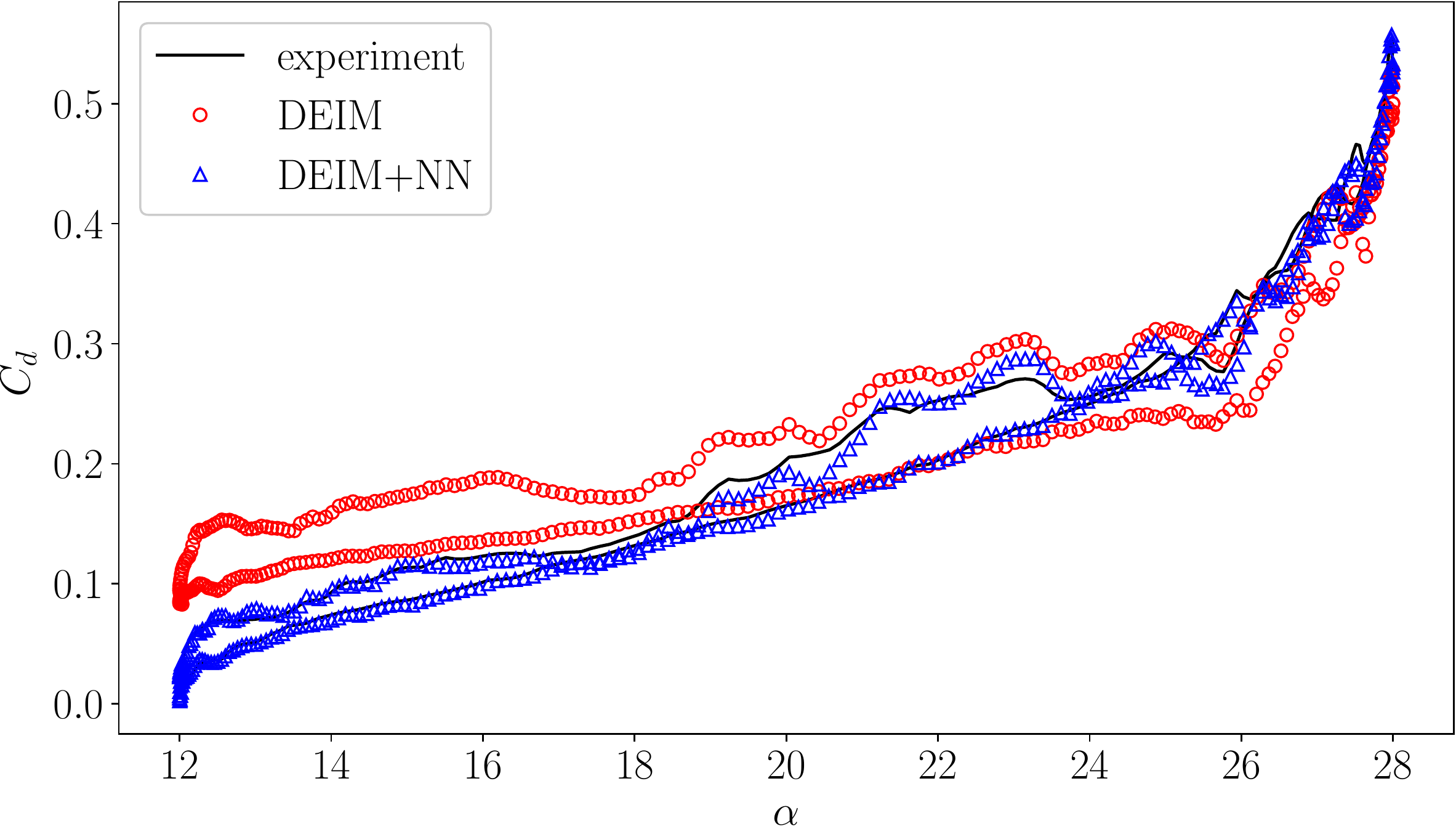}
        \caption{$n_s=10$, $C_d$}
    \end{subfigure}
    \caption{2D airfoil: $C_l,C_d$ w.r.t. $\alpha$ for the testing experimental data with $1.5\%$ noise in the pressure sensor inputs.
    The DEIM models are based on the URANS data.}
    \label{fig:2DAirfoil_numer_DEIM_location_aero_coeff_aoa_noise=1.5}
\end{figure}

\begin{figure}[hbt!]
    \centering
    \begin{subfigure}[t]{0.49\textwidth}
        \centering
        \includegraphics[width=\linewidth]{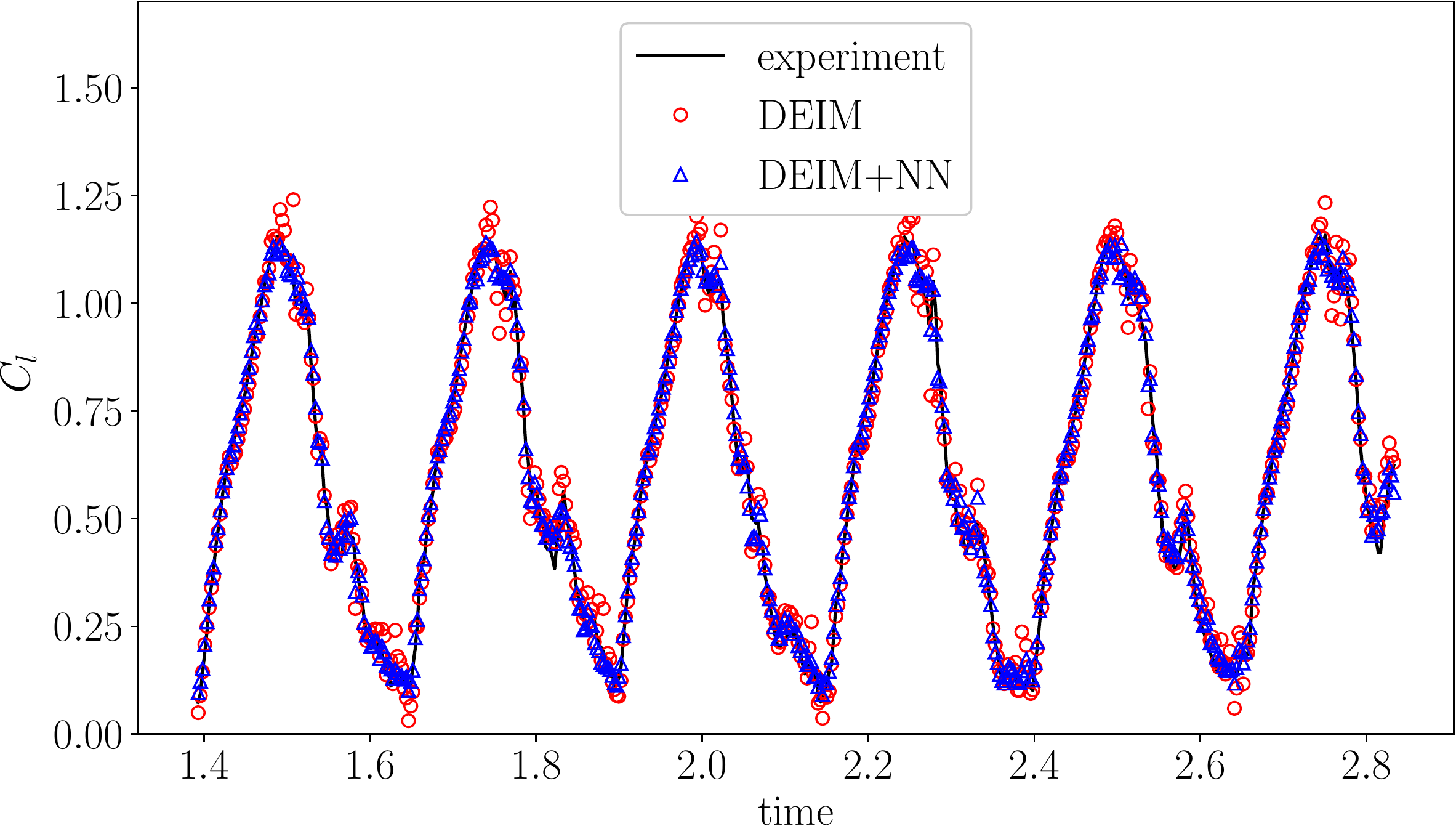}
        \caption{$n_s=5$, $C_l$}
    \end{subfigure}
    \begin{subfigure}[t]{0.49\textwidth}
        \centering
        \includegraphics[width=\linewidth]{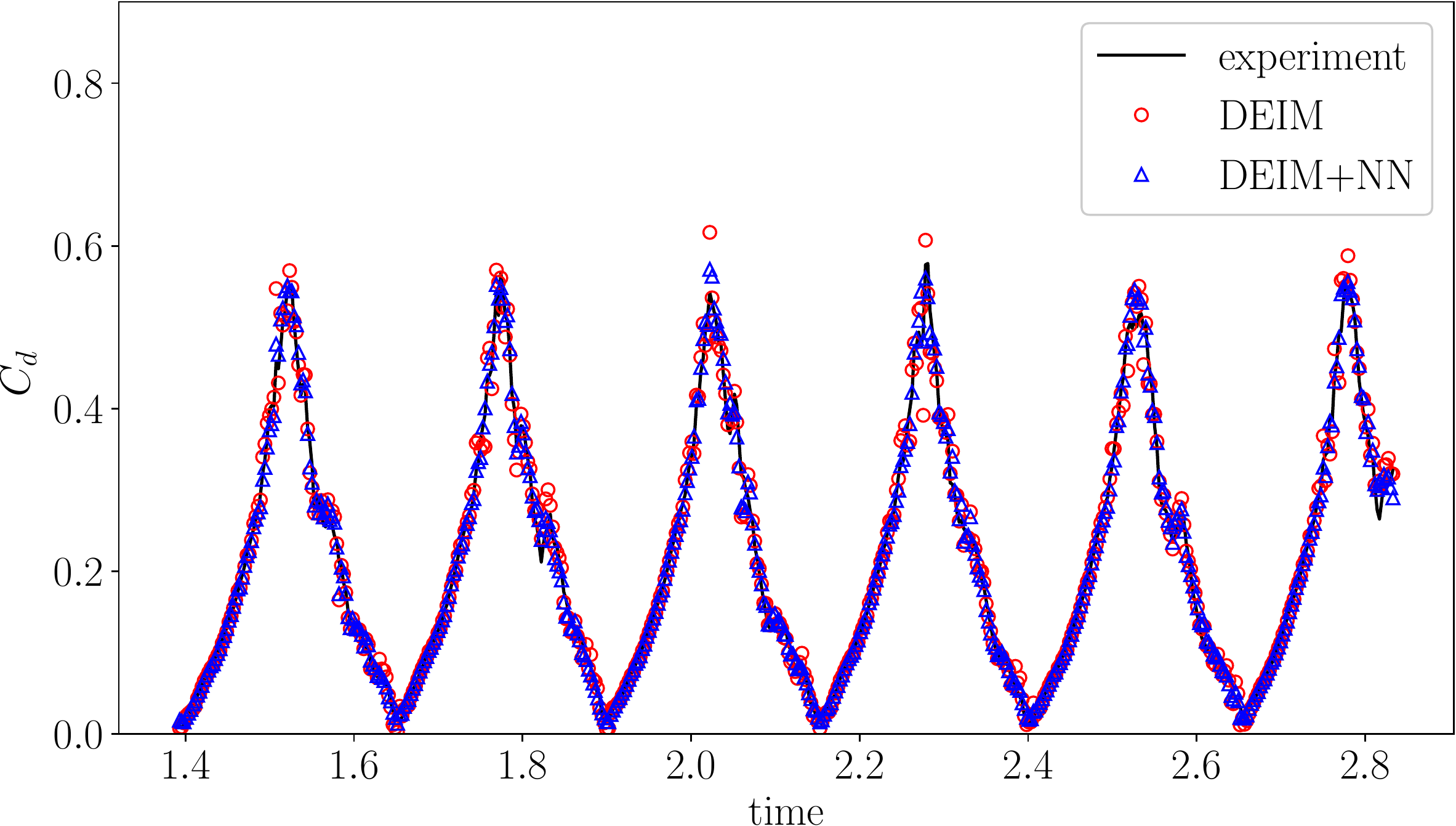}
        \caption{$n_s=5$, $C_d$}
    \end{subfigure}

    \begin{subfigure}[t]{0.49\textwidth}
        \centering
        \includegraphics[width=\linewidth]{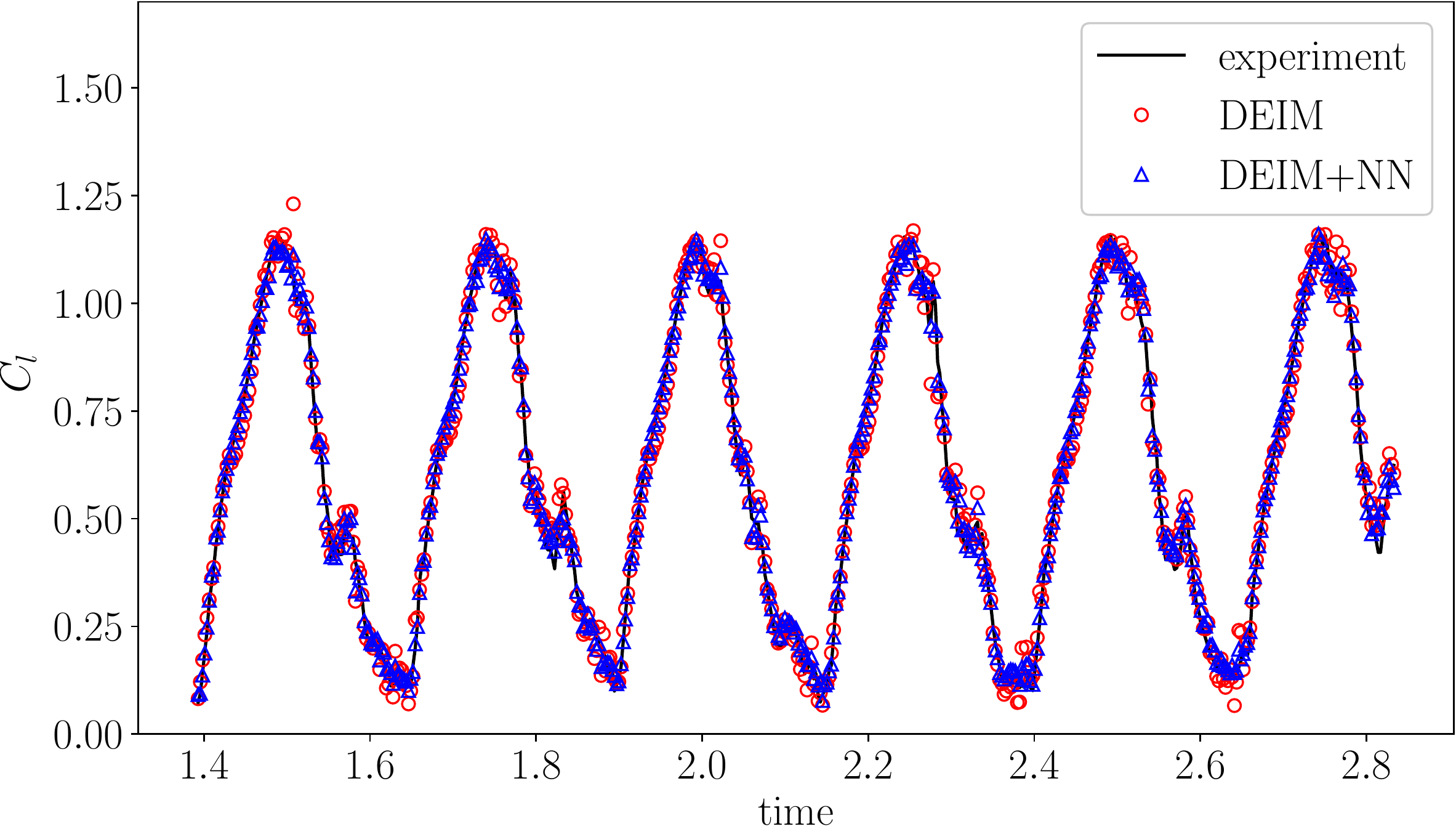}
        \caption{$n_s=8$, $C_l$}
    \end{subfigure}
    \begin{subfigure}[t]{0.49\textwidth}
        \centering
        \includegraphics[width=\linewidth]{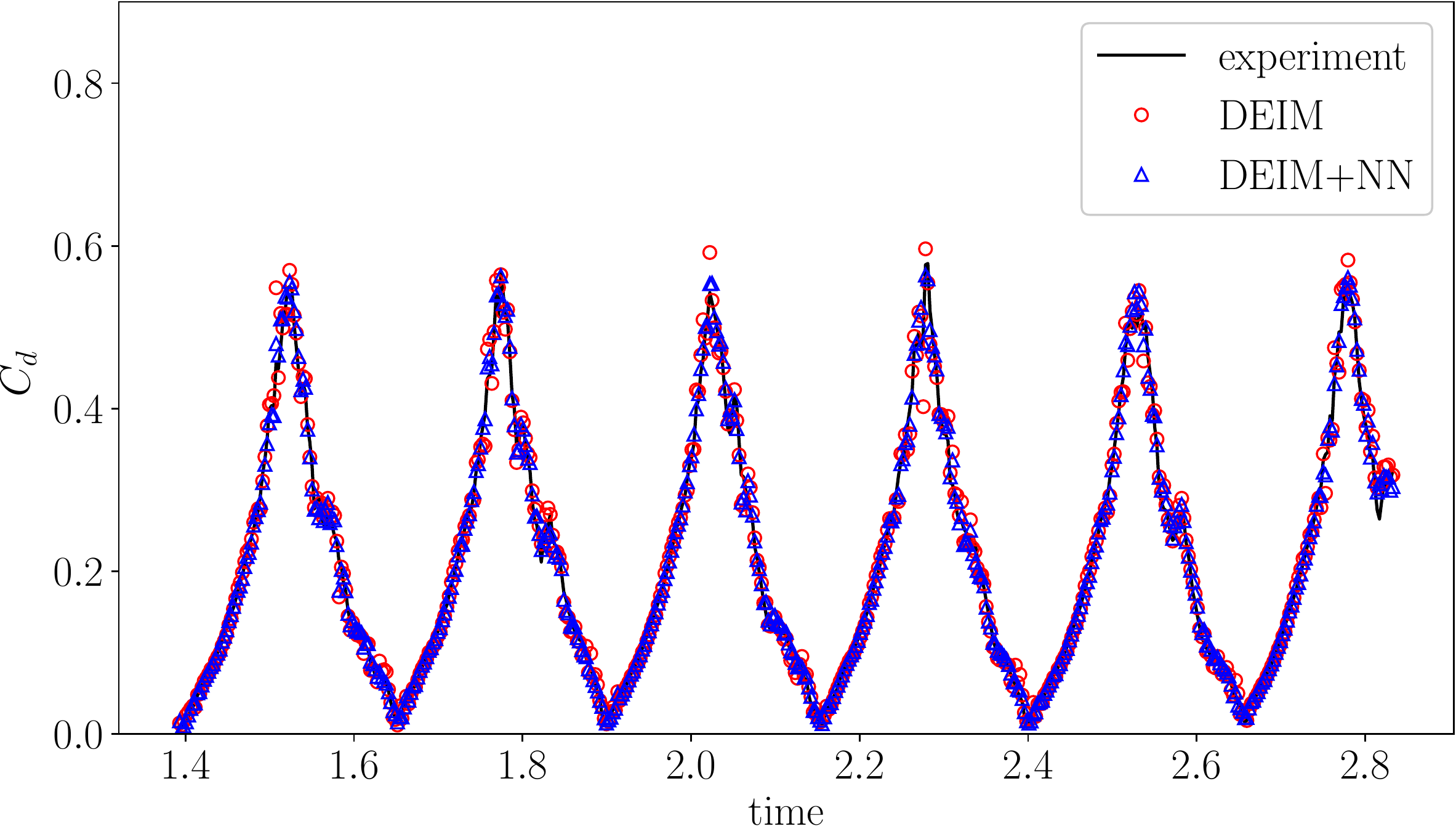}
        \caption{$n_s=8$, $C_d$}
    \end{subfigure}

    \begin{subfigure}[t]{0.49\textwidth}
        \centering
        \includegraphics[width=\linewidth]{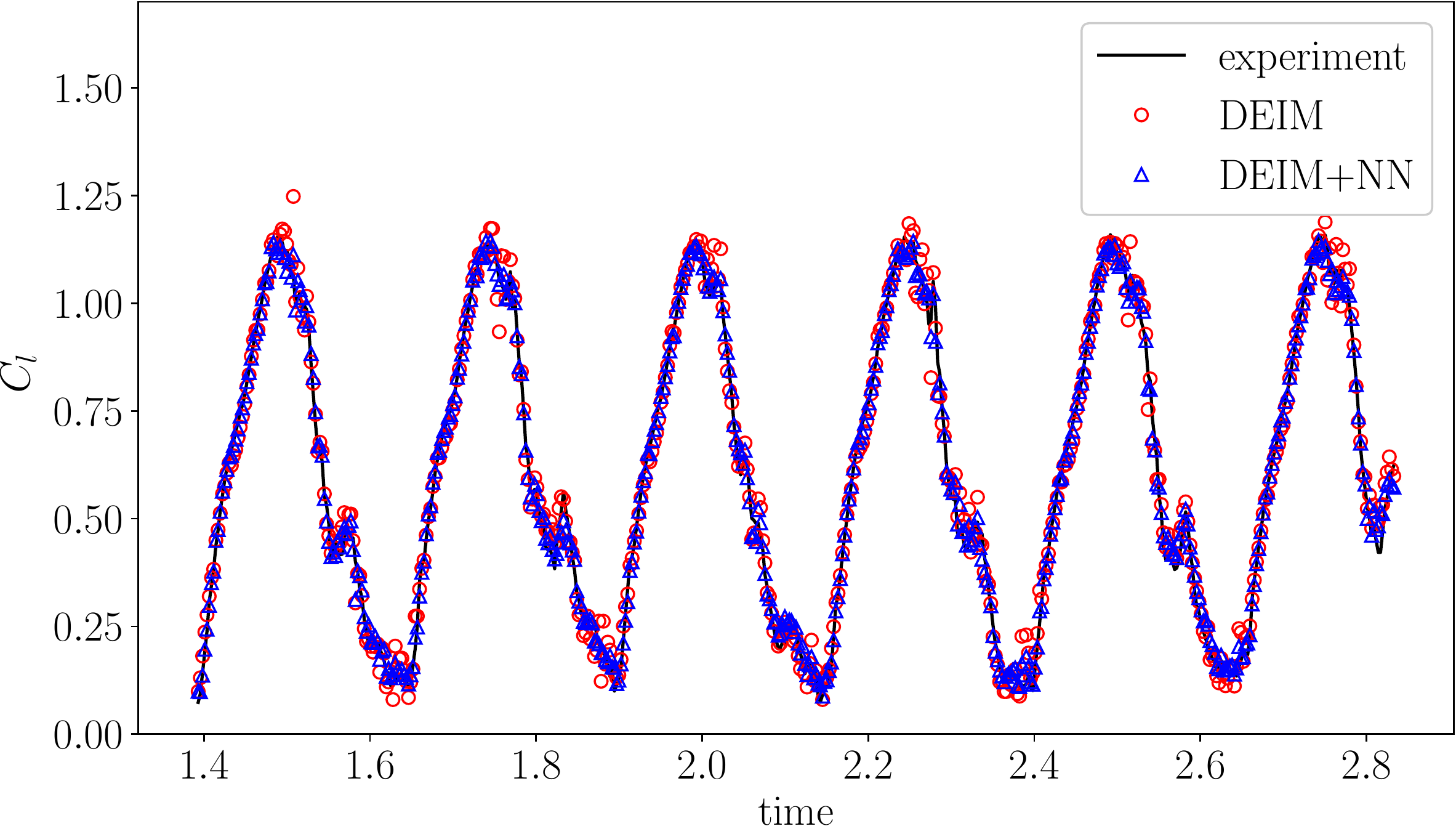}
        \caption{$n_s=10$, $C_l$}
    \end{subfigure}
    \begin{subfigure}[t]{0.49\textwidth}
        \centering
        \includegraphics[width=\linewidth]{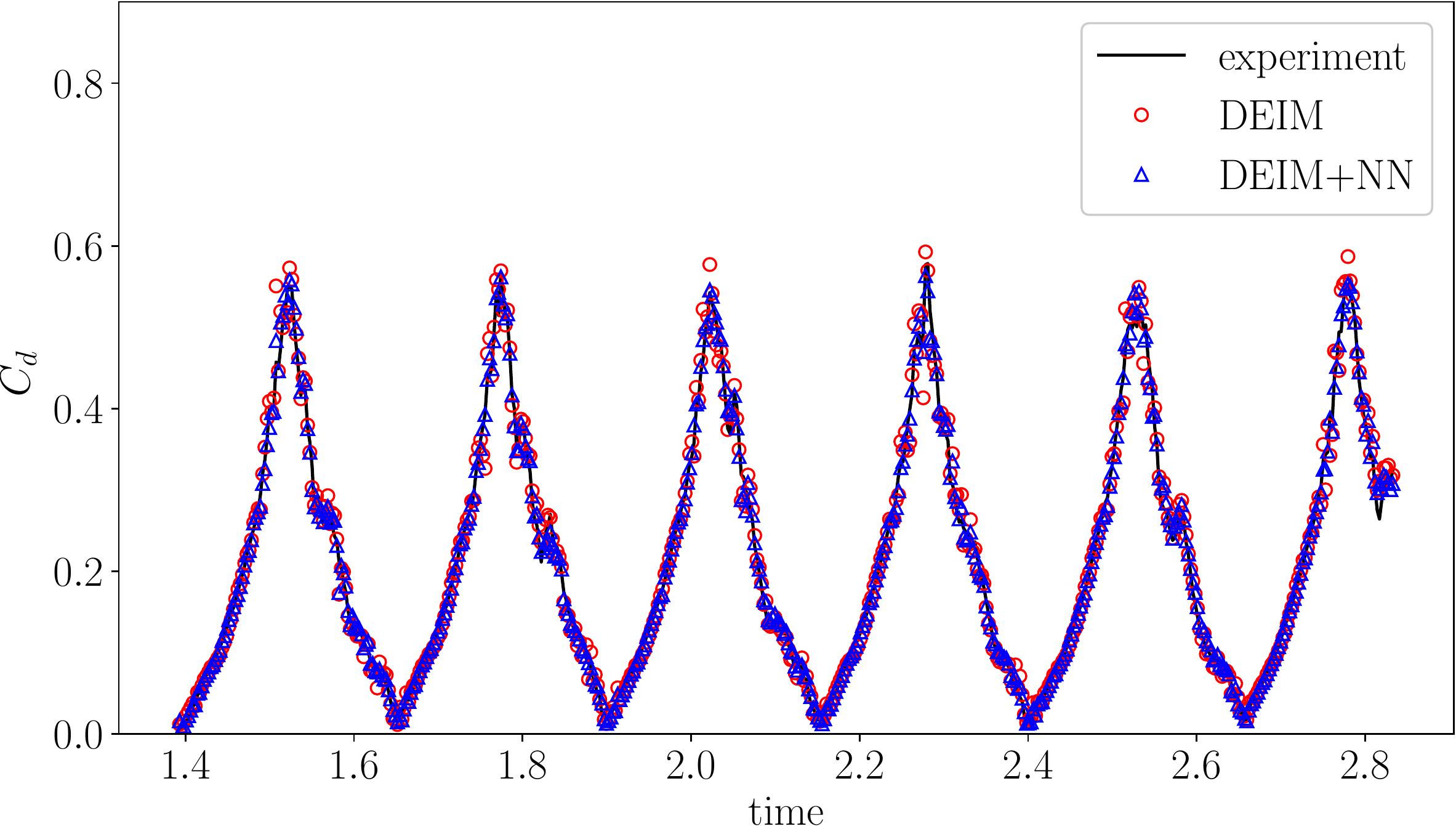}
        \caption{$n_s=10$, $C_d$}
    \end{subfigure}
    \caption{2D airfoil: $C_l,C_d$ w.r.t. time for the testing experimental data with $1.5\%$ noise in the pressure sensor inputs.
    The DEIM models are based on the experimental data.}
    \label{fig:2DAirfoil_exper_DEIM_location_aero_coeff_time_noise=1.5}
\end{figure}

\begin{figure}[hbt!]
    \centering
    \begin{subfigure}[t]{0.49\textwidth}
        \centering
        \includegraphics[width=\linewidth]{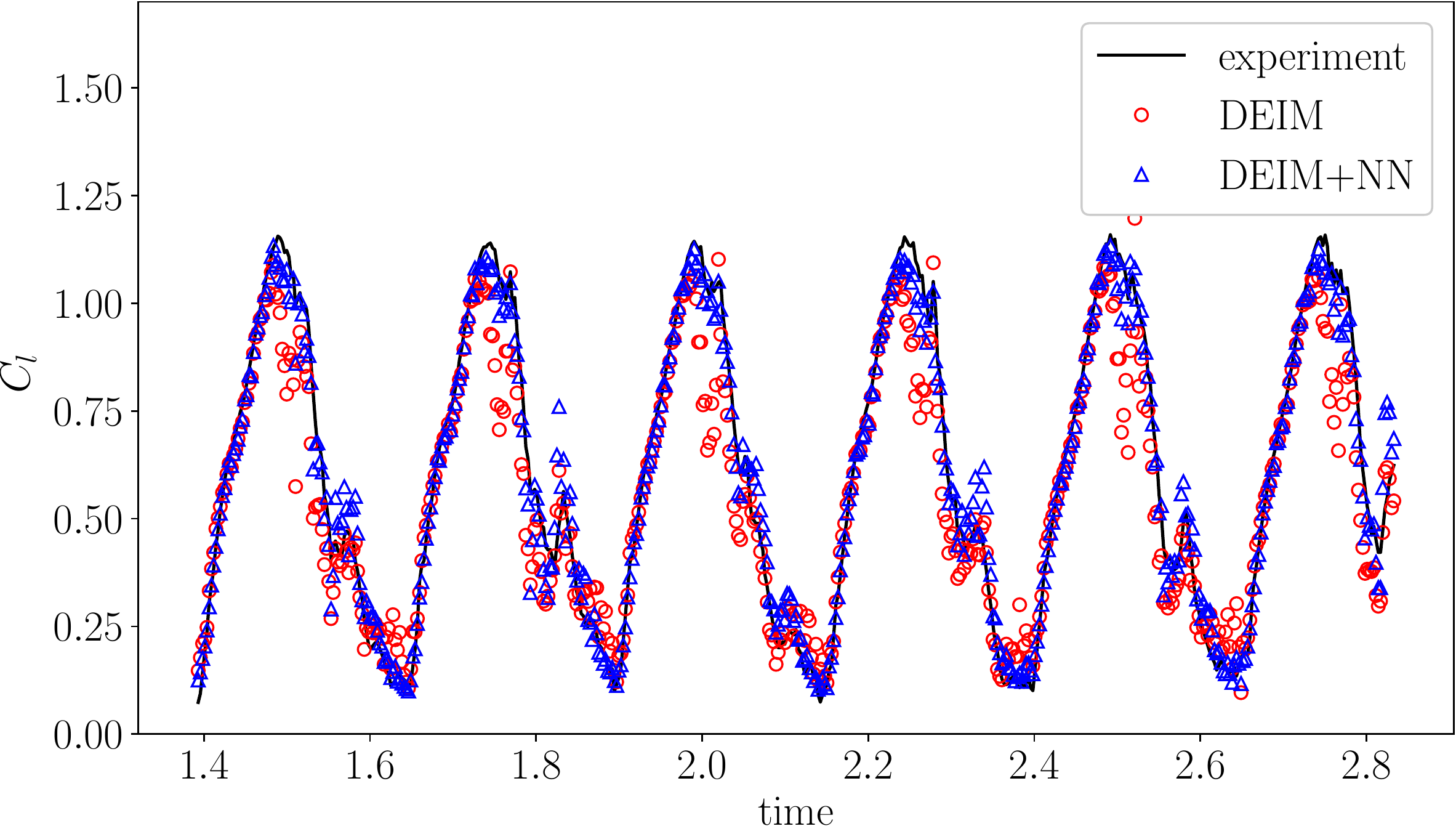}
        \caption{$n_s=5$, $C_l$}
    \end{subfigure}
    \begin{subfigure}[t]{0.49\textwidth}
        \centering
        \includegraphics[width=\linewidth]{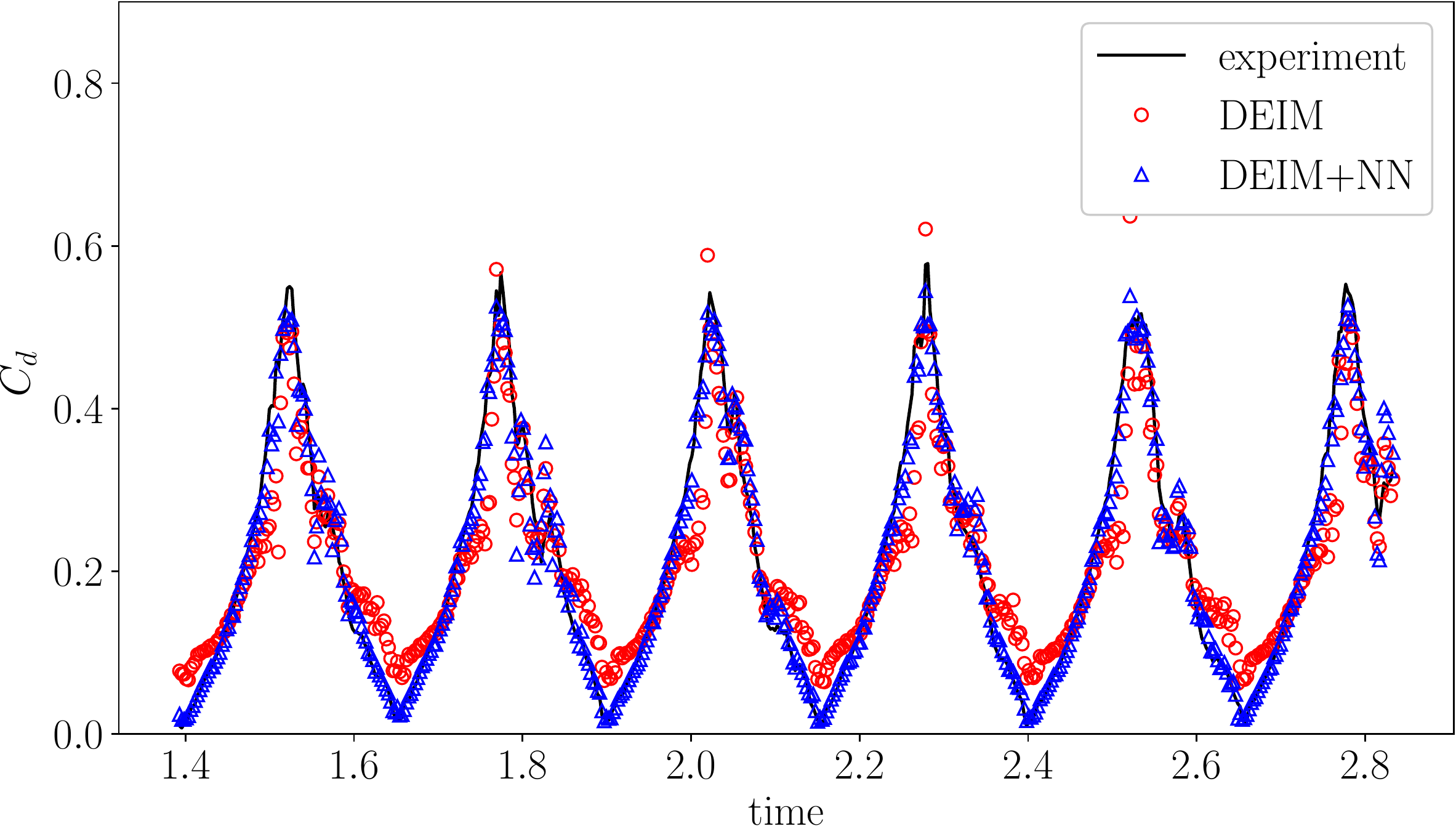}
        \caption{$n_s=5$, $C_d$}
    \end{subfigure}

    \begin{subfigure}[t]{0.49\textwidth}
        \centering
        \includegraphics[width=\linewidth]{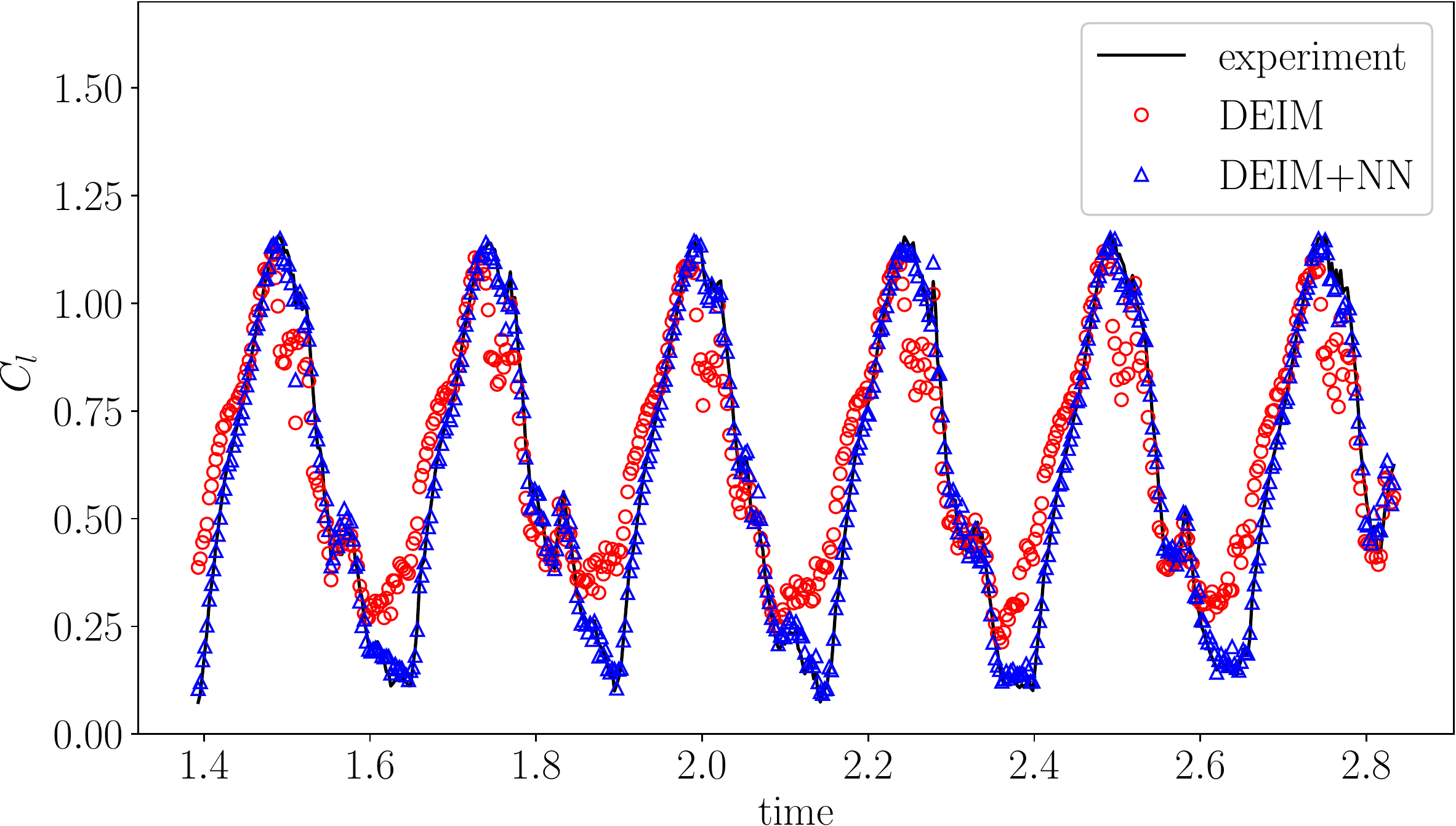}
        \caption{$n_s=8$, $C_l$}
    \end{subfigure}
    \begin{subfigure}[t]{0.49\textwidth}
        \centering
        \includegraphics[width=\linewidth]{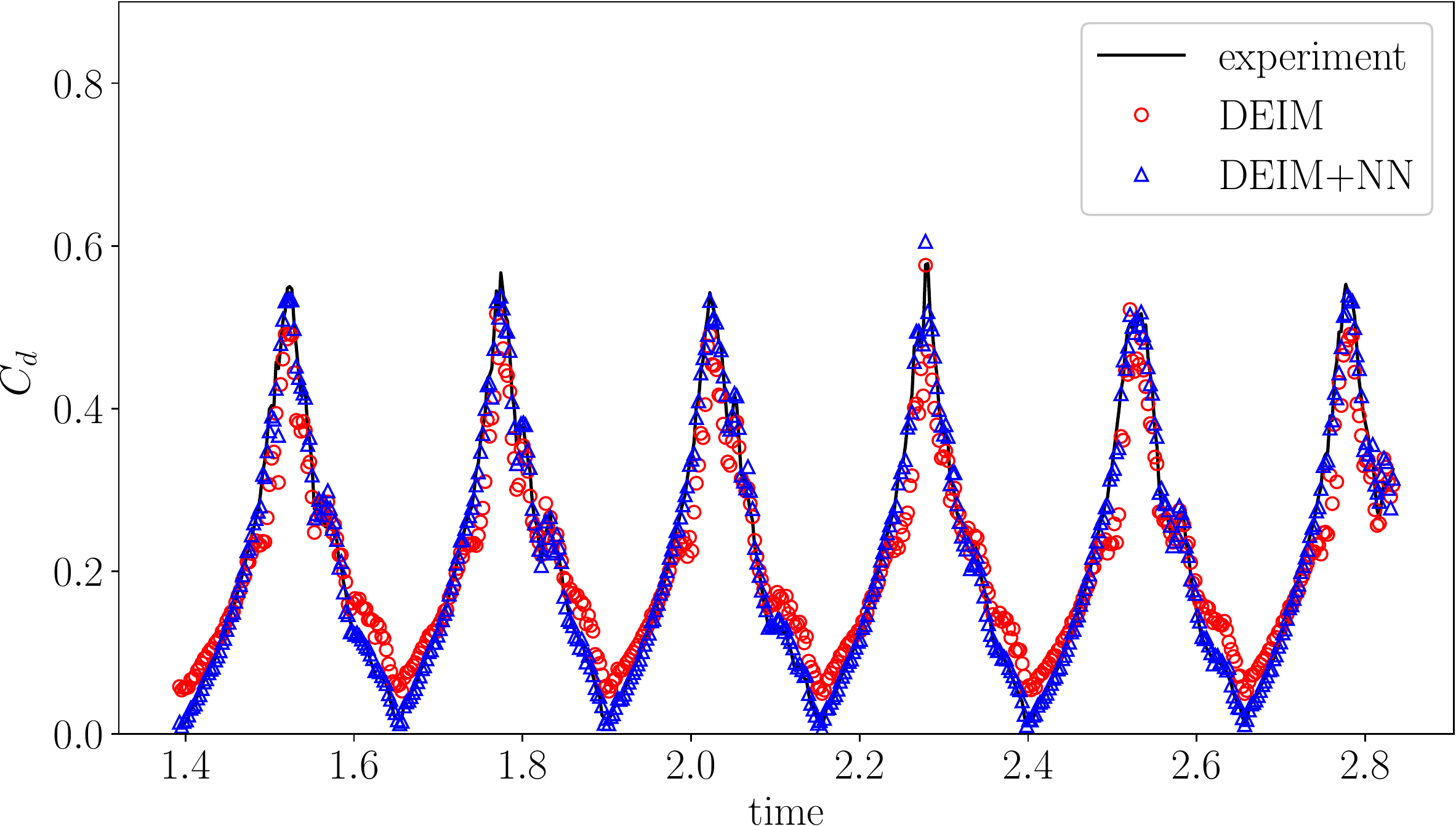}
        \caption{$n_s=8$, $C_d$}
    \end{subfigure}

    \begin{subfigure}[t]{0.49\textwidth}
        \centering
        \includegraphics[width=\linewidth]{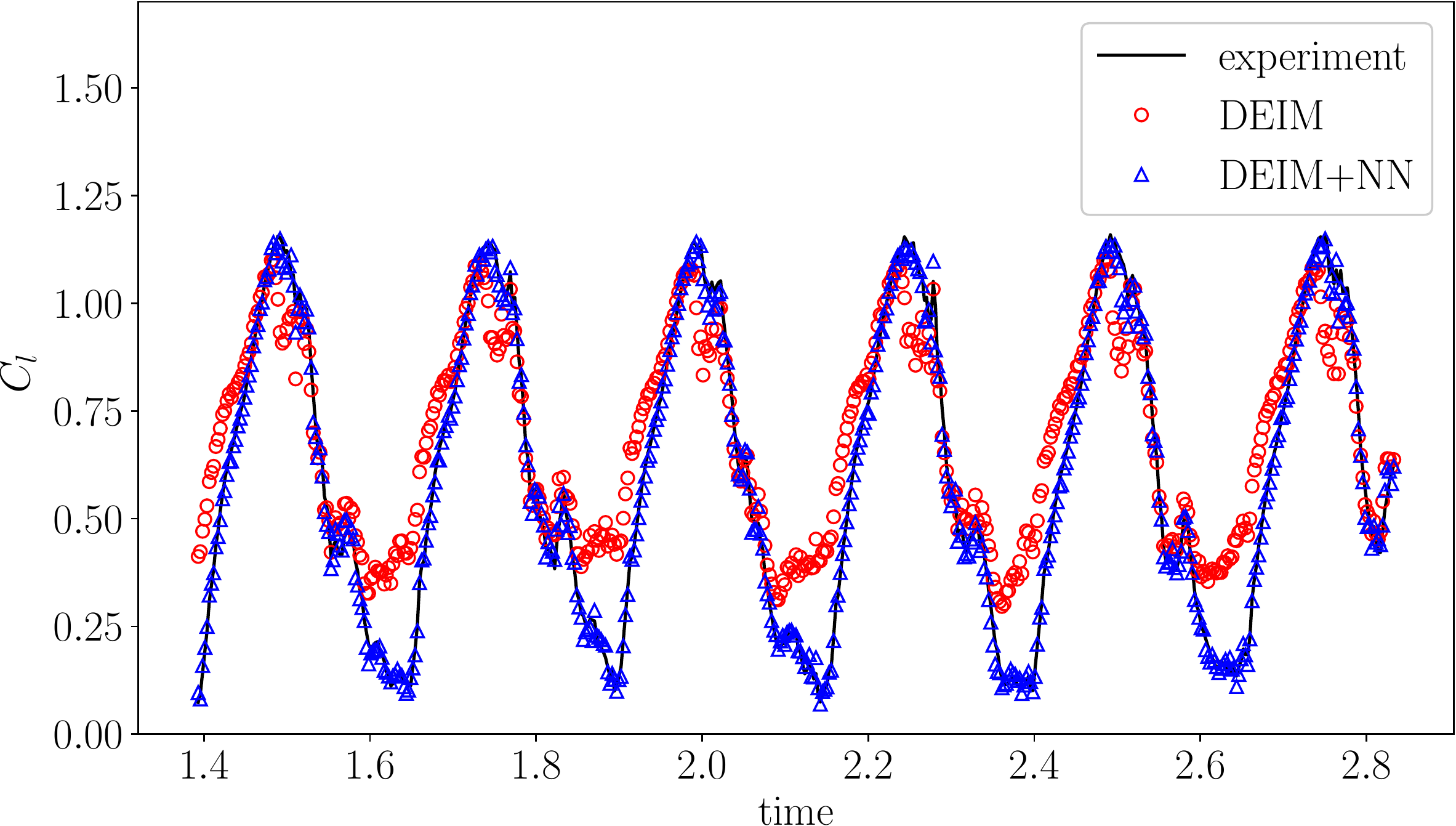}
        \caption{$n_s=10$, $C_l$}
    \end{subfigure}
    \begin{subfigure}[t]{0.49\textwidth}
        \centering
        \includegraphics[width=\linewidth]{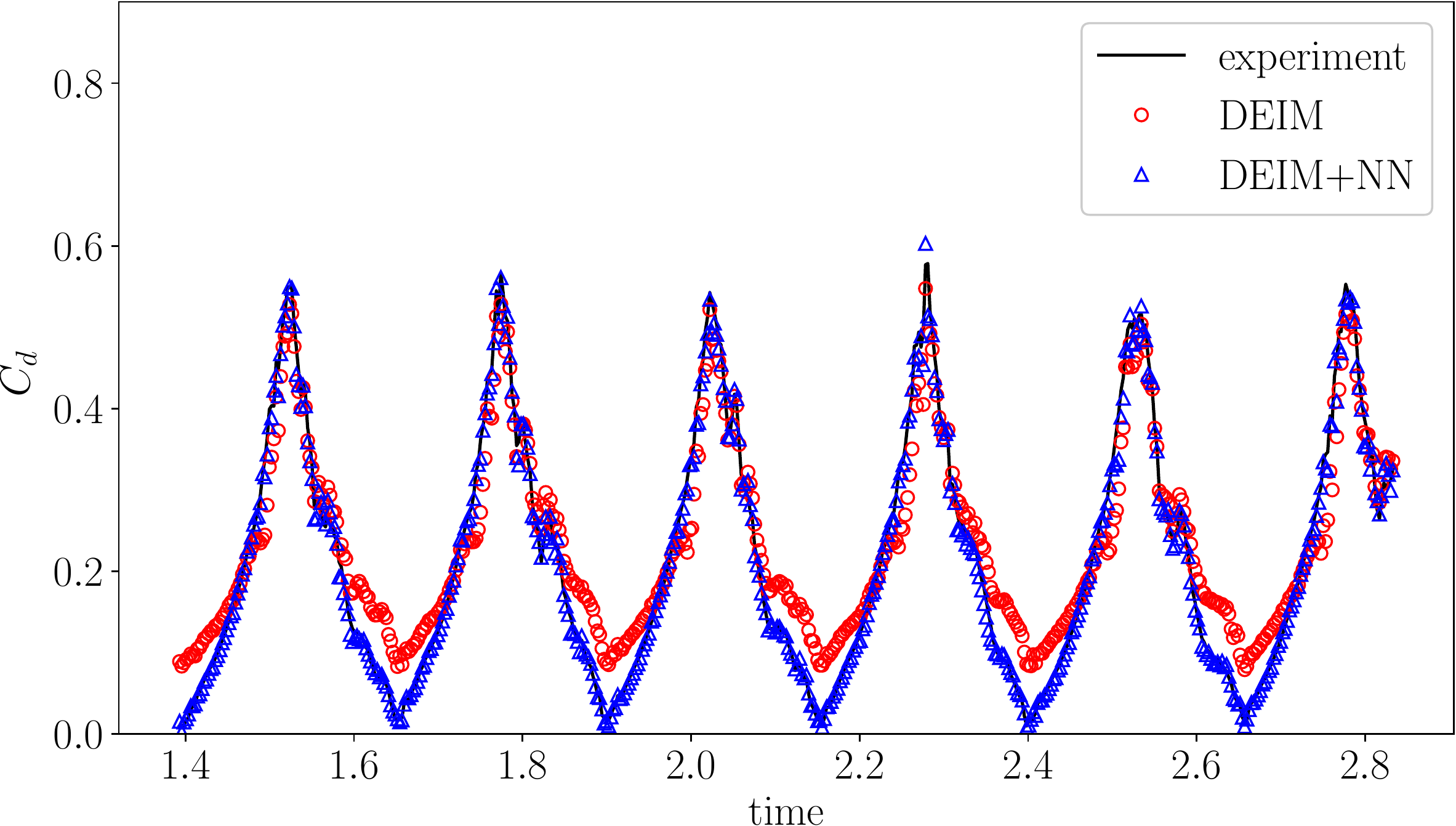}
        \caption{$n_s=10$, $C_d$}
    \end{subfigure}
    \caption{2D airfoil: $C_l,C_d$ w.r.t. time for the testing experimental data with $1.5\%$ noise in the pressure sensor inputs.
    The DEIM models are based on the URANS data.}
    \label{fig:2DAirfoil_numer_DEIM_location_aero_coeff_time_noise=1.5}
\end{figure}

To test the robustness of the proposed model,
predictions with $1.5\%$ noise in the pressure sensor inputs are considered \cite{Zhao_2021_Research_TaAML}.
The results are shown in Figs. \ref{fig:2DAirfoil_exper_DEIM_location_aero_coeff_aoa_noise=1.5}-\ref{fig:2DAirfoil_numer_DEIM_location_aero_coeff_time_noise=1.5},
which demonstrate that the model can still predict the aerodynamic coefficients well and is not influenced by the noise.
From Tables \ref{tab:2DAirfoil_numer_DEIM_location_lift_err}-\ref{tab:2DAirfoil_numer_DEIM_location_drag_err} one can also see that the smallest errors in $C_l$ and $C_d$ of the URANS data based DEIM prediction with NN correction are \num{2.24e-2} and \num{1.07e-2}, which are very close to the smallest errors of the predictions without noise, confirming that the model is not sensitive to noise.

\subsection{3D drone}
In this section, the DEIM+NN approach is used to predict the aerodynamic coefficients during dynamic stall of a 3D drone.
As there is no experimental data,
the pressure coefficients from the URANS simulation are used to select the sensor locations and build the DEIM model,
and then the numerical pressure coefficients at the selected locations serve as the sensor inputs.
It is worth mentioning that the sensor inputs do not consider the viscous forces,
so the NN also models the viscous effects.

\begin{figure}[hbt!]
    \centering
    \begin{subfigure}[t]{0.49\textwidth}
        \centering
        \includegraphics[width=\linewidth]{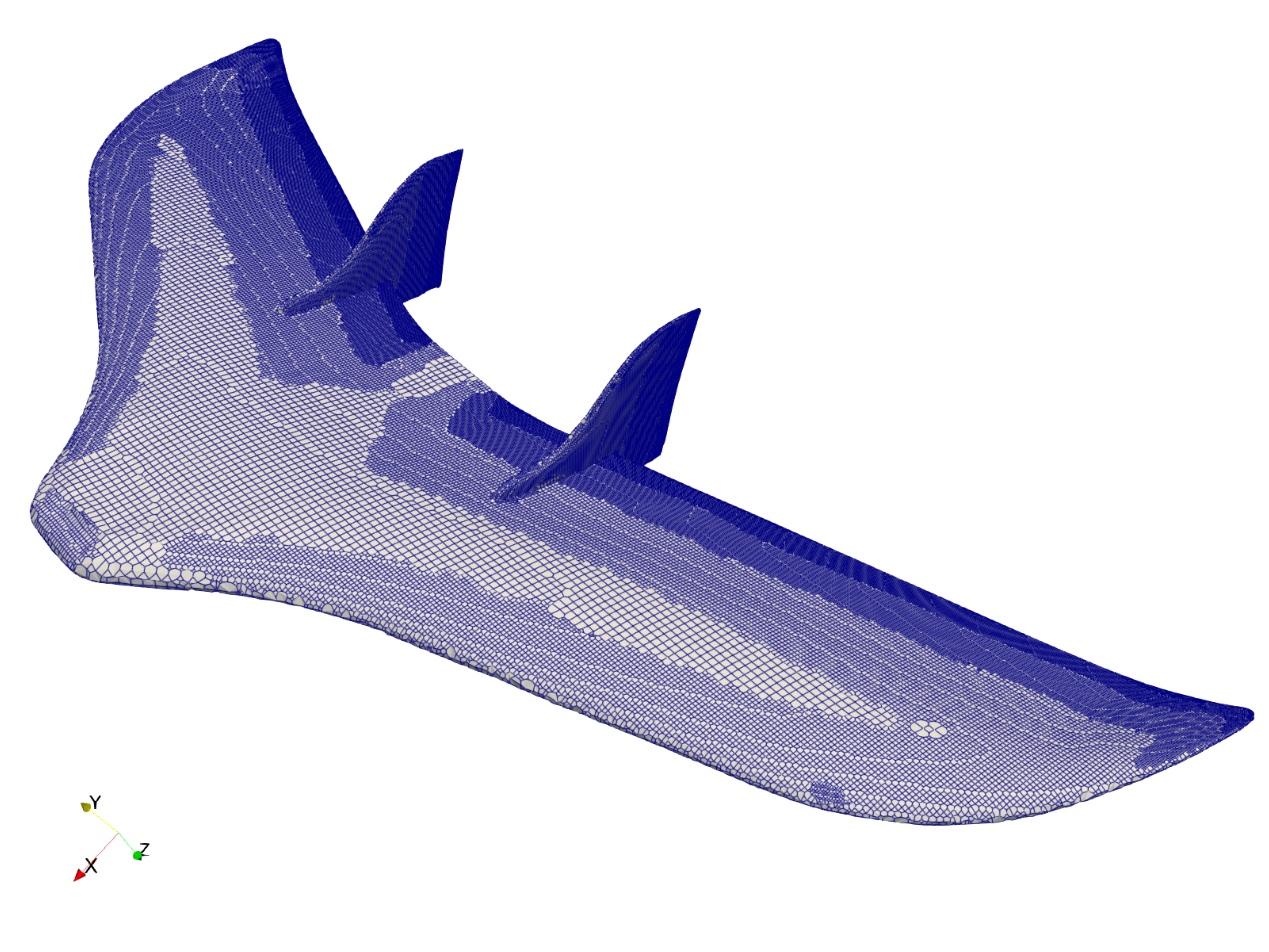}
        \caption{The surface mesh of the drone.}
    \end{subfigure}
    \begin{subfigure}[t]{0.49\textwidth}
        \centering
        \includegraphics[width=\linewidth]{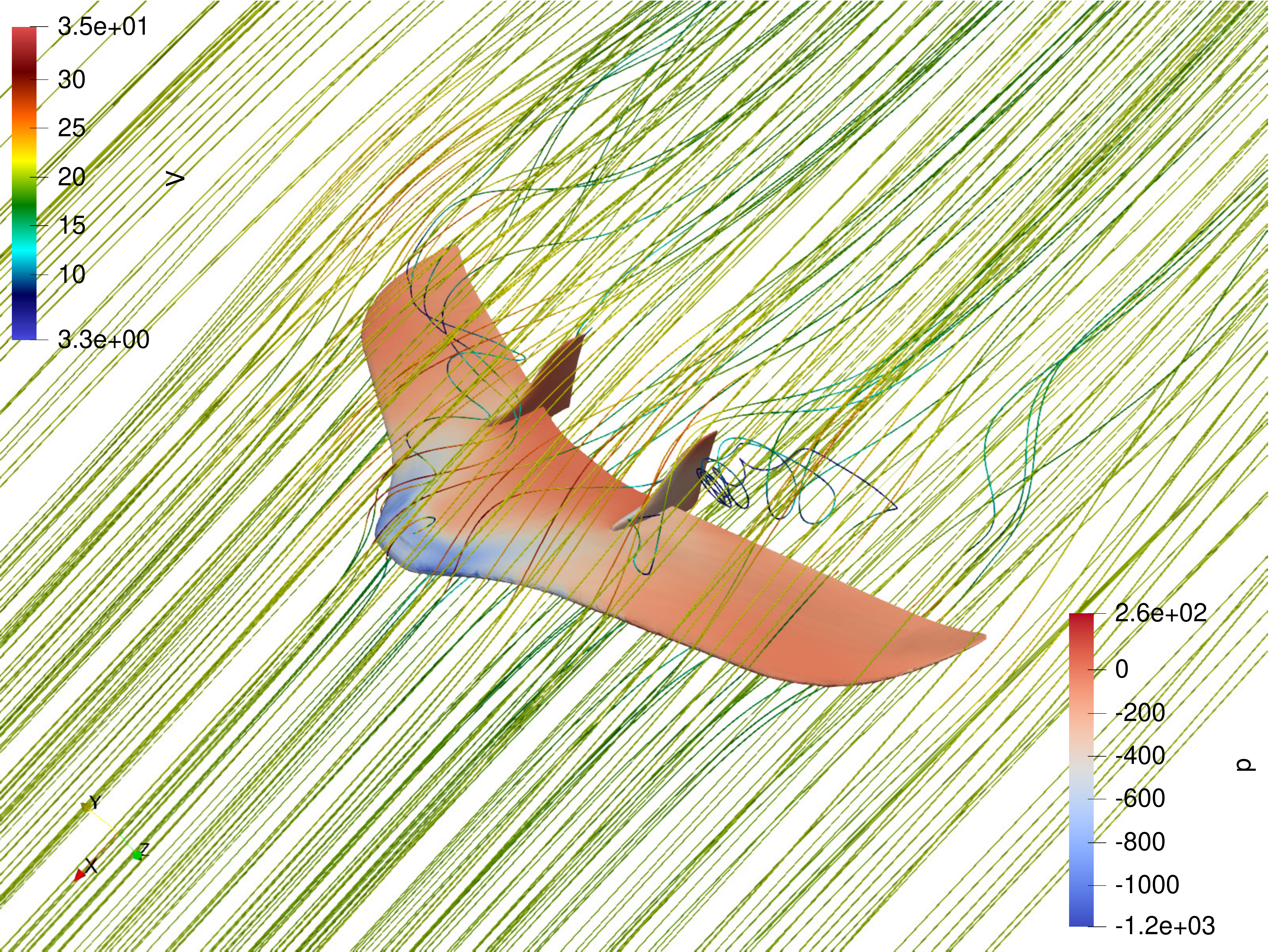}
        \caption{The surface pressure and streamlines colored by the magnitude of the velocity.}
    \end{subfigure}
\caption{Computational mesh and URANS results with $f = \qty{6.786}{Hz}$ at $t = \qty{0.2}{s}$.}
\label{fig:3DDrone_mesh_flow_field}
\end{figure}

The parameters in the pitching movement are chosen as $\alpha_0 = \qty{20}{deg}$, $A = \qty{15}{deg}$,
and the freestream conditions are $\rho_\infty = \qty{1.146}{kg/m^3}$, $V_\infty = \qty{20}{m/s}$, $\nu_\infty = \qty{1.655e-5}{m^2/s}$.
The reference surface area and chord length of the drone are $A_{\text{ref}} = \qty{0.39}{m^2}$ and $\qty{0.3}{m}$, respectively,
so the Reynolds number based on the chord length is $Re = 3.6\times 10^{5}$. %Re = V*c/\nu
The URANS simulation is performed with $9$ pitching frequencies uniform in $[4,8]$ for training,
and two random frequencies in the same domain for validation and testing, respectively.
The snappyHexMesh utility in OpenFOAM is used to generate the computational mesh,
consisting of about $3.58\times 10^{6}$ cells and $1.13\times 10^{7}$ faces.
The surface mesh of the drone is shown in Fig. \ref{fig:3DDrone_mesh_flow_field},
where the surface pressure field and streamlines are also presented for $f = \qty{6.786}{Hz}$ at $t = \qty{0.2}{s}$.

\begin{figure}[hbt!]
    \centering
    \includegraphics[width=0.4\linewidth]{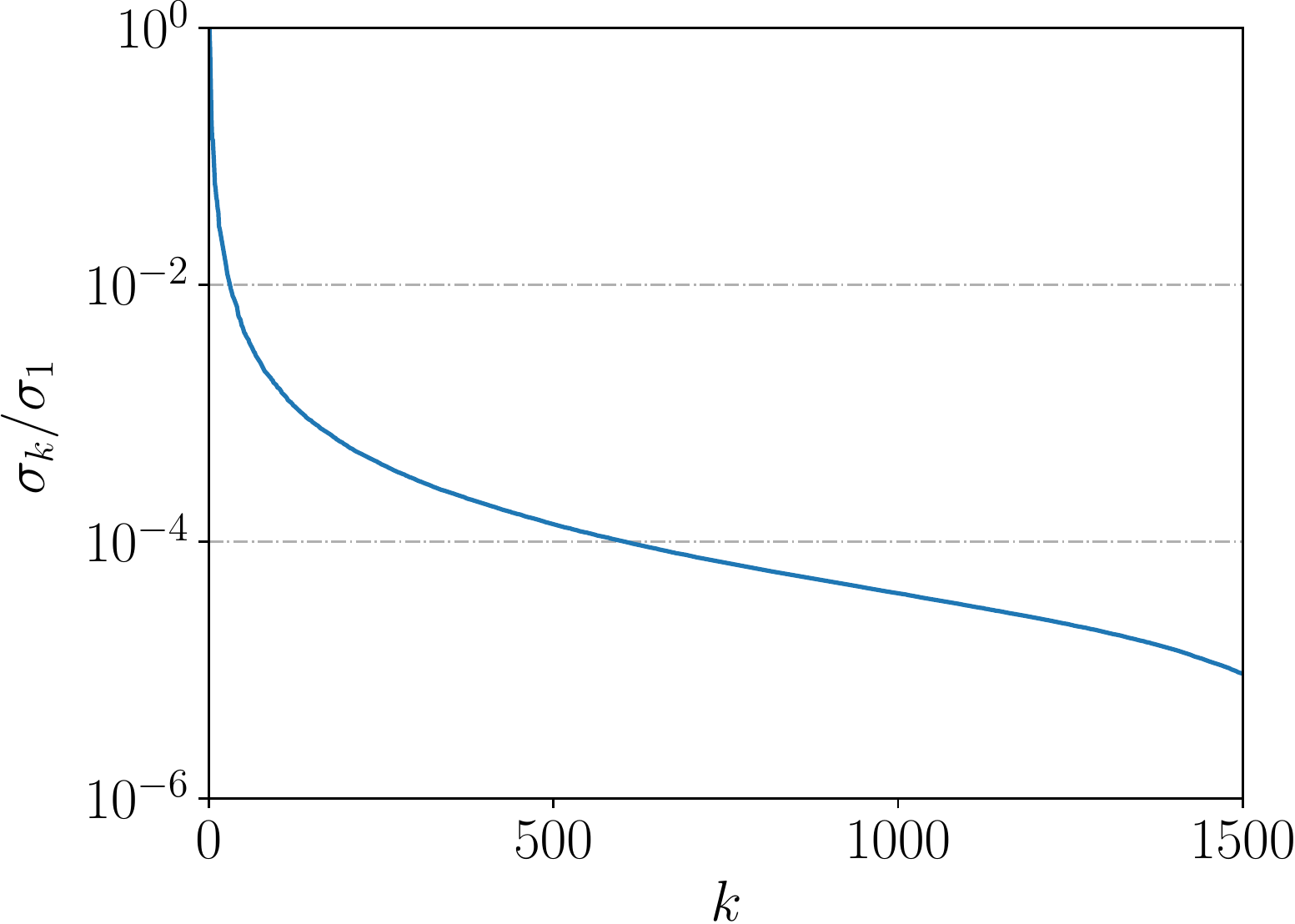}
    \caption{3D drone: The scaled singular values in the SVD.}
\label{fig:3DAirfoil_singular_value}
\end{figure}

The scaled singular values in the SVD are shown in Fig. \ref{fig:3DAirfoil_singular_value},
indicating that the linear subspace is not efficient in capturing the dynamics.

\begin{figure}[hbt!]
    \centering
    \begin{subfigure}[t]{0.33\textwidth}
        \centering
        \includegraphics[width=\linewidth]{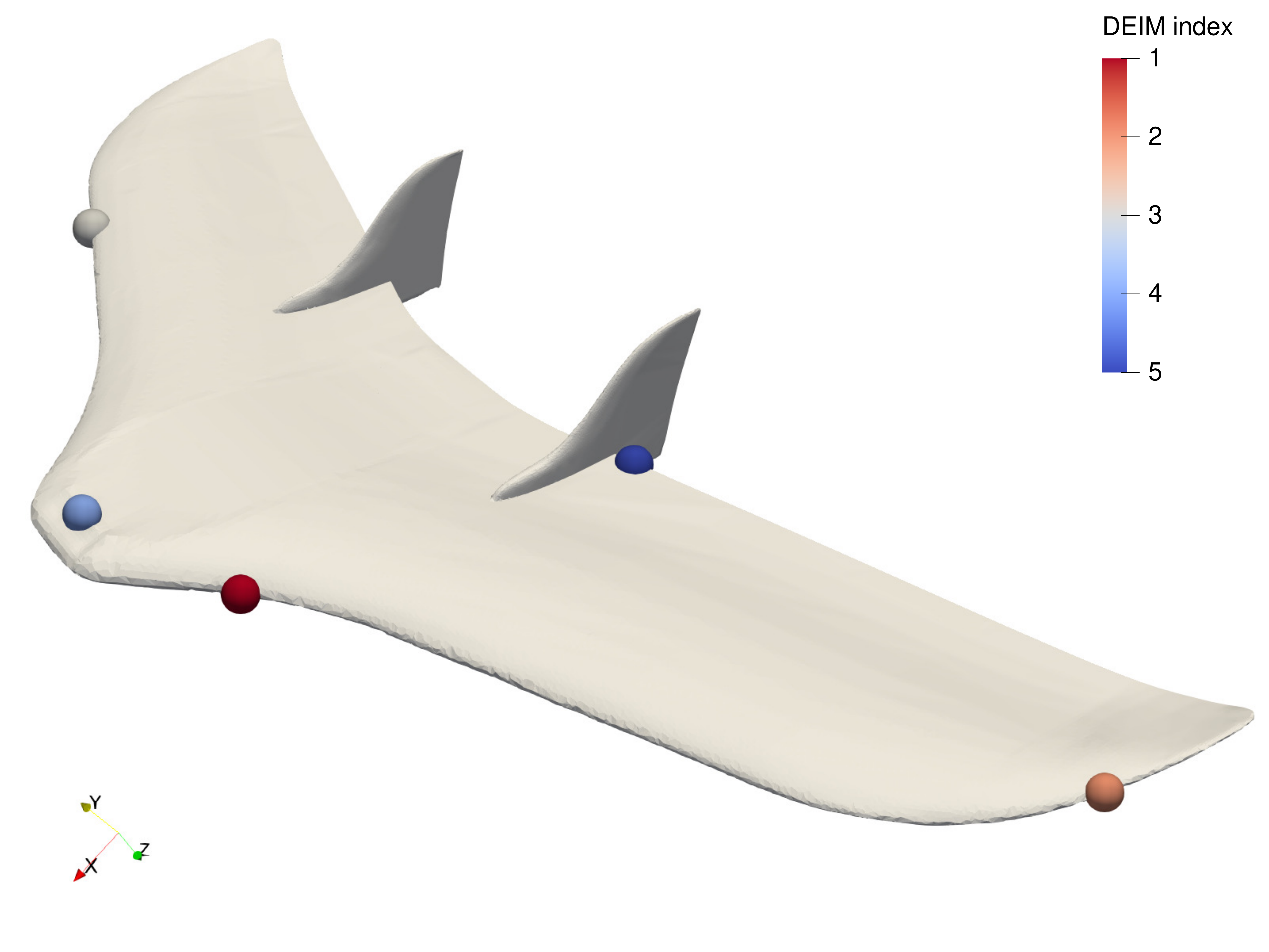}
        \caption{$n_s=5$}
    \end{subfigure}
    \begin{subfigure}[t]{0.33\textwidth}
        \centering
        \includegraphics[width=\linewidth]{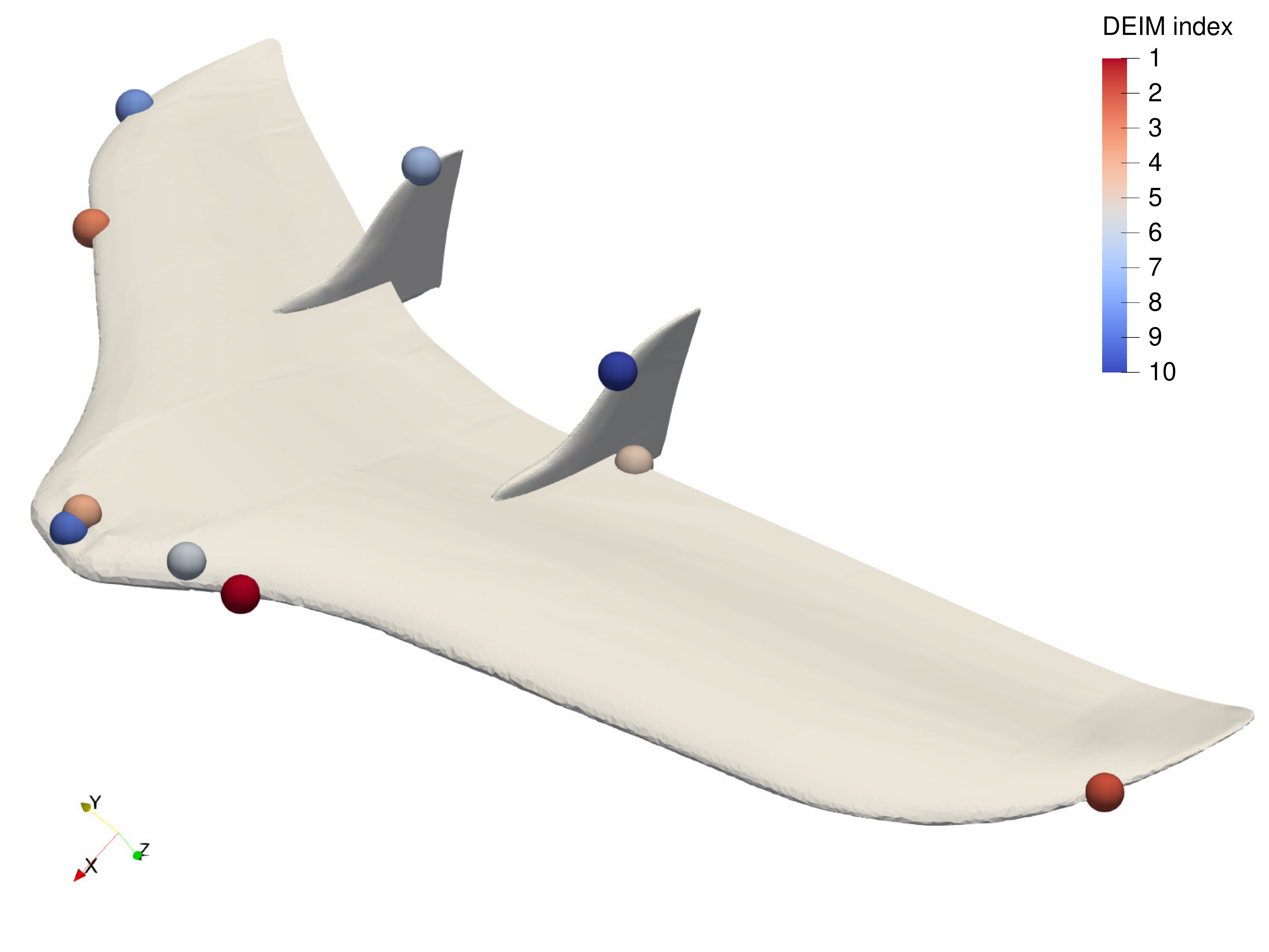}
        \caption{$n_s=10$}
    \end{subfigure}
    \begin{subfigure}[t]{0.33\textwidth}
        \centering
        \includegraphics[width=\linewidth]{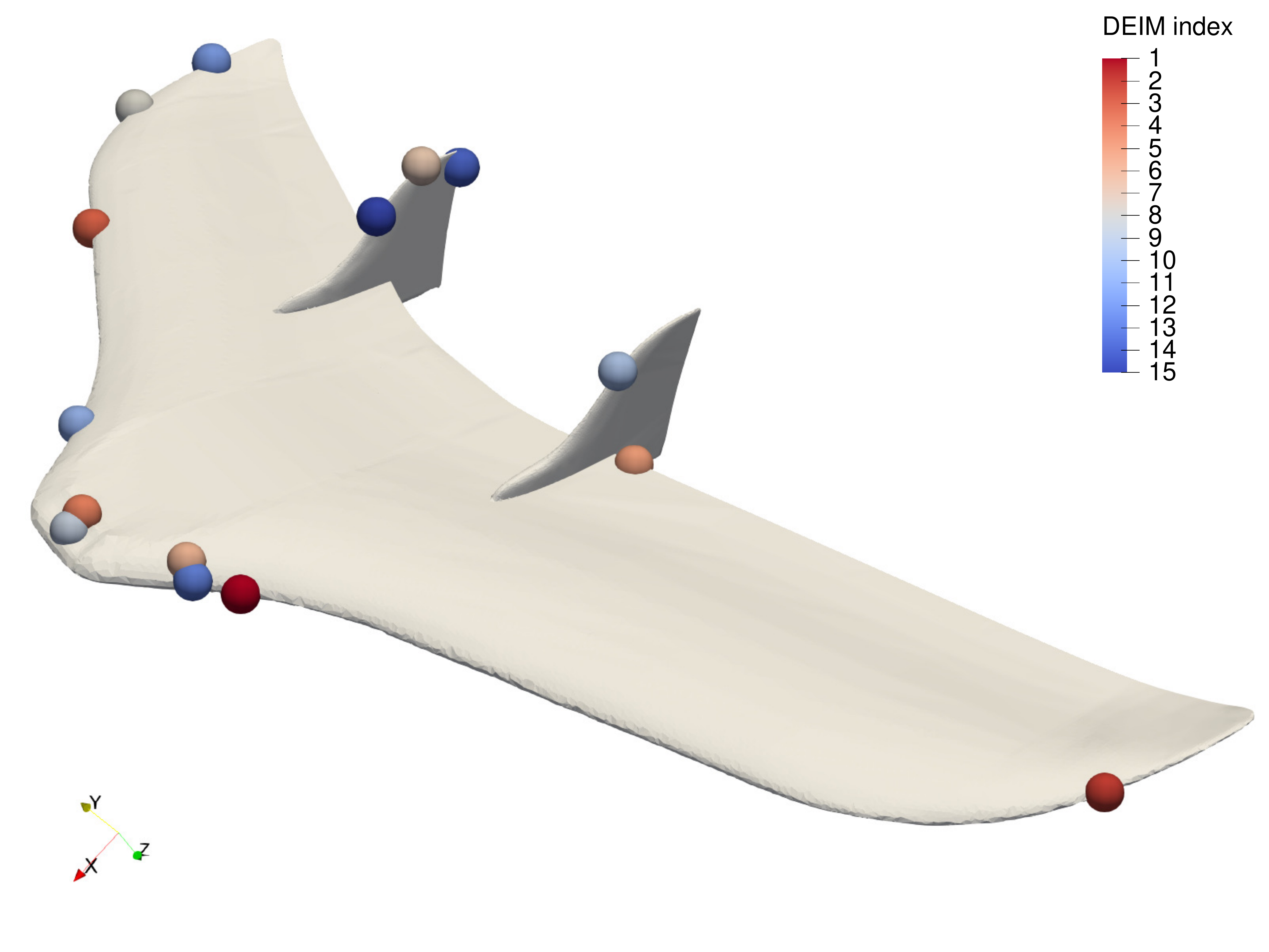}
        \caption{$n_s=15$}
    \end{subfigure}
\caption{Sensor locations selected by the DEIM on the drone.}
\label{fig:3DDrone_locations}
\end{figure}

In the test, all surface mesh centers ($299,771$ in total) in the computational mesh are taken as candidate locations,
and the DEIM is used to select $n_s$ locations.
Figure \ref{fig:3DDrone_locations} presents the selected sensor locations with $n_s=5, 10, 15$,
where the color corresponds to the order during the selection in the DEIM.
One observes that the selected locations lie on the upper surface,
and the locations on the leading edges are preferred, similar to the 2D airfoil case.

\begin{figure}[hbt!]
    \centering
    \begin{subfigure}[t]{0.49\textwidth}
        \centering
        \includegraphics[width=\linewidth]{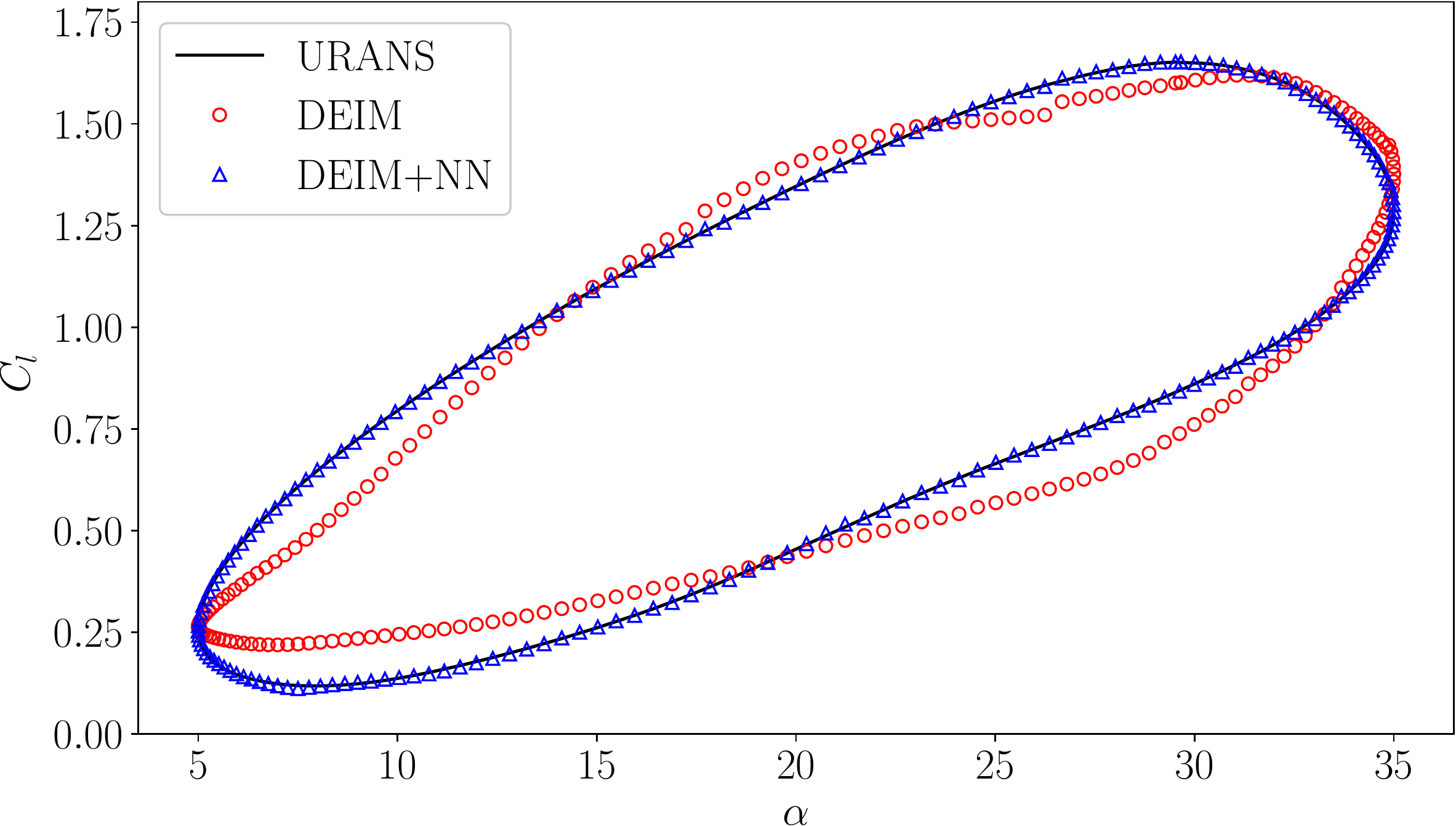}
        \caption{$n_s=5$, $C_l$}
    \end{subfigure}
    \begin{subfigure}[t]{0.49\textwidth}
        \centering
        \includegraphics[width=\linewidth]{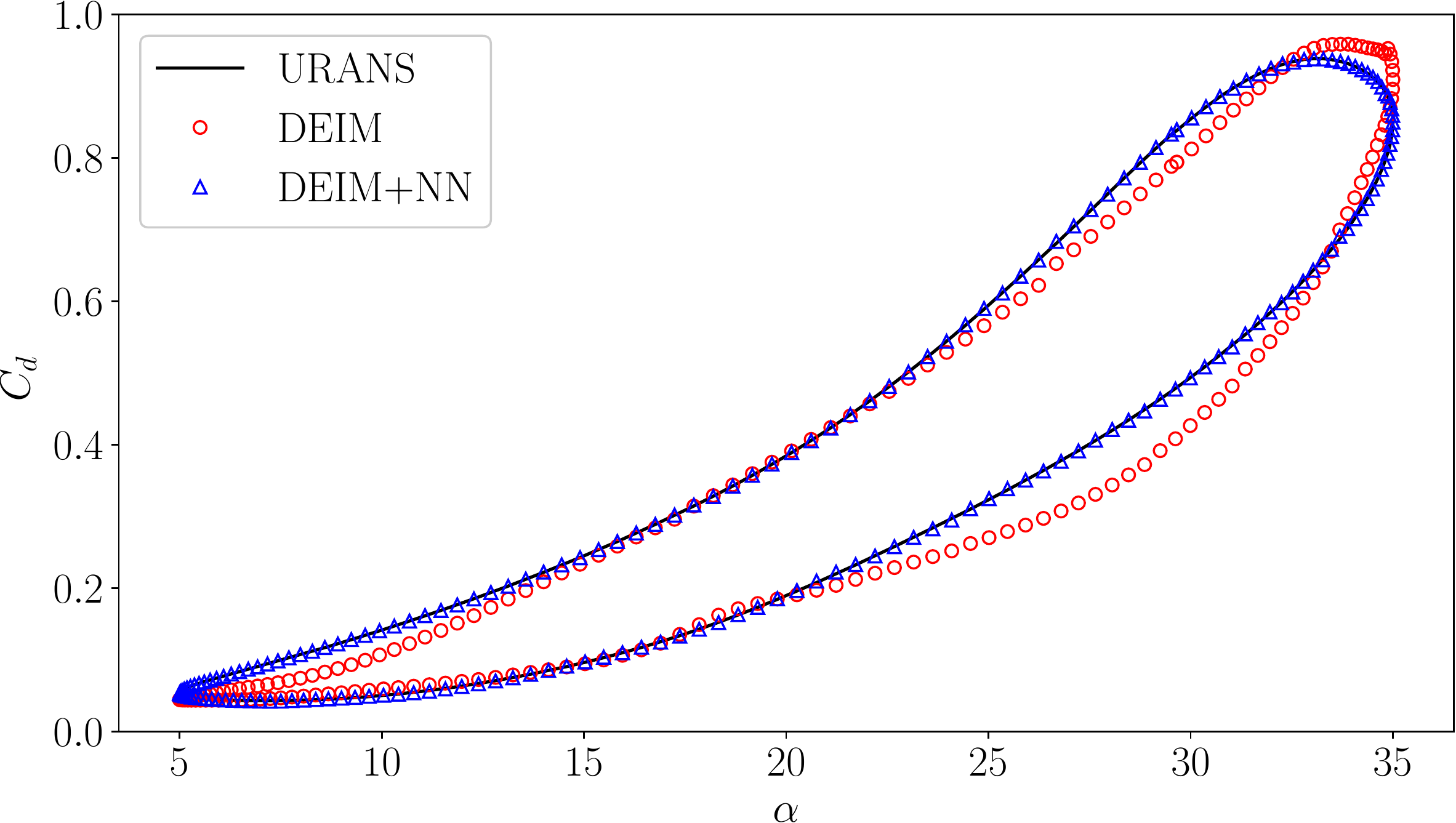}
        \caption{$n_s=5$, $C_d$}
    \end{subfigure}

    \begin{subfigure}[t]{0.49\textwidth}
        \centering
        \includegraphics[width=\linewidth]{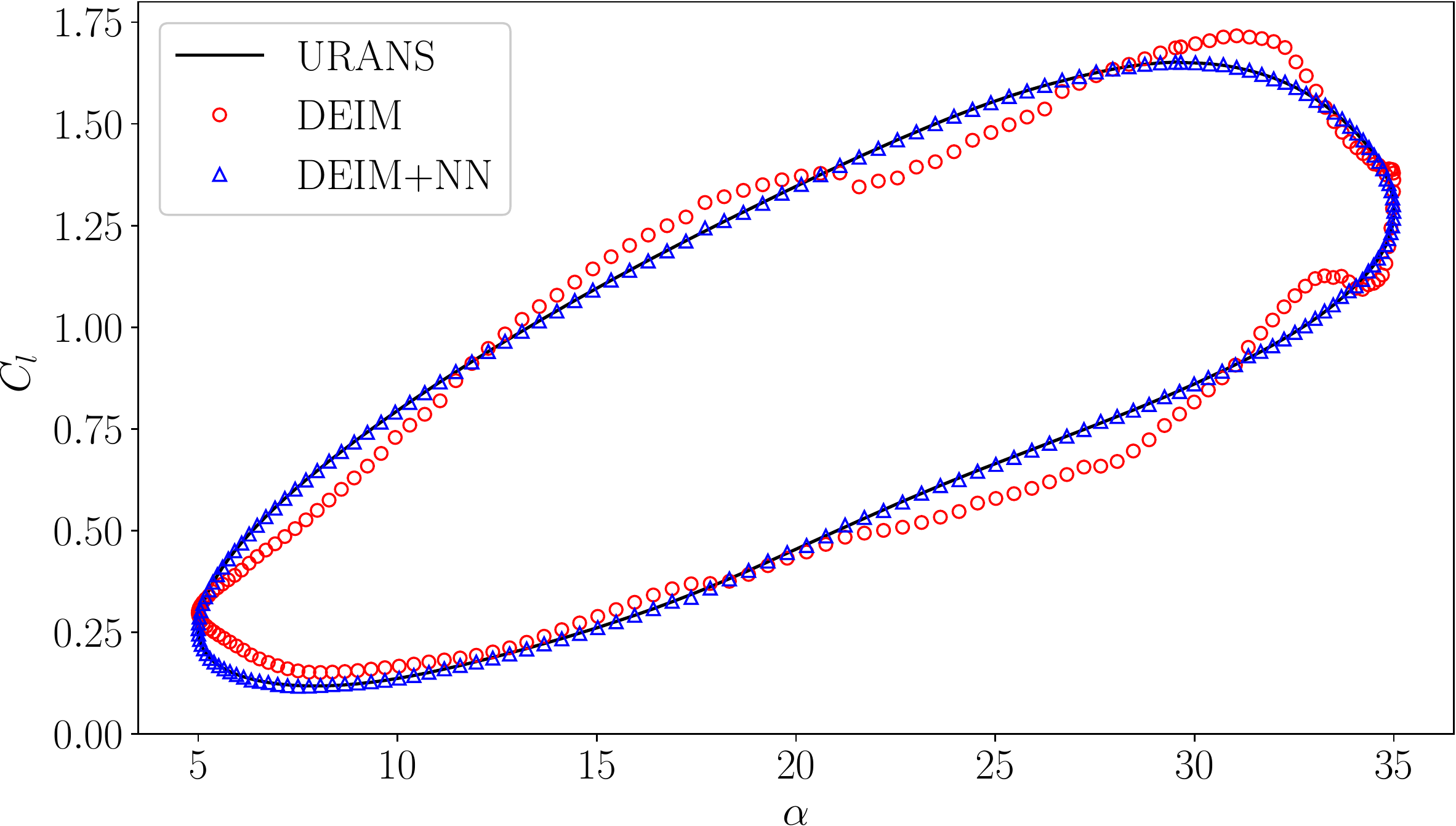}
        \caption{$n_s=10$, $C_l$}
    \end{subfigure}
    \begin{subfigure}[t]{0.49\textwidth}
        \centering
        \includegraphics[width=\linewidth]{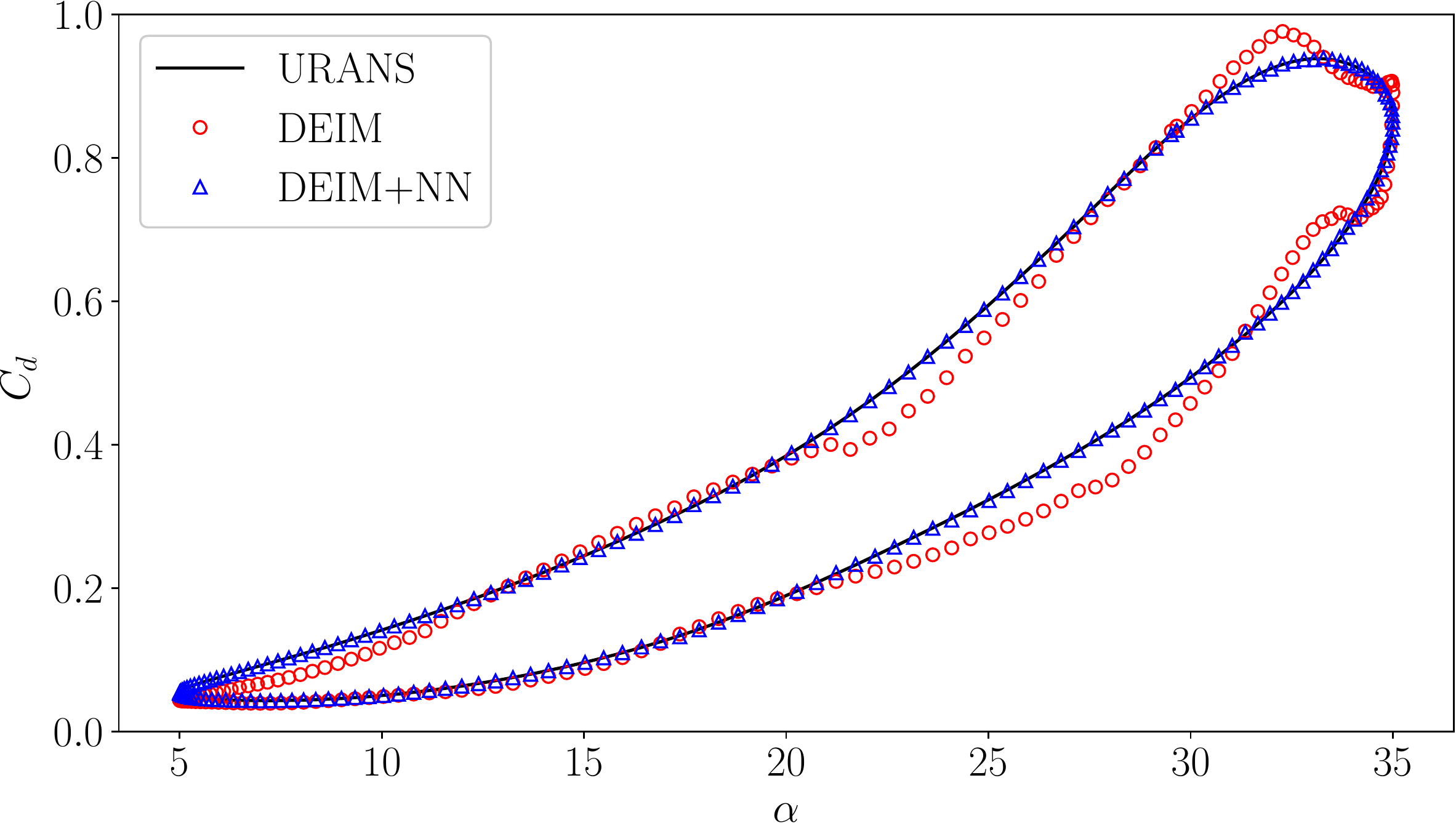}
        \caption{$n_s=10$, $C_d$}
    \end{subfigure}

    \begin{subfigure}[t]{0.49\textwidth}
        \centering
        \includegraphics[width=\linewidth]{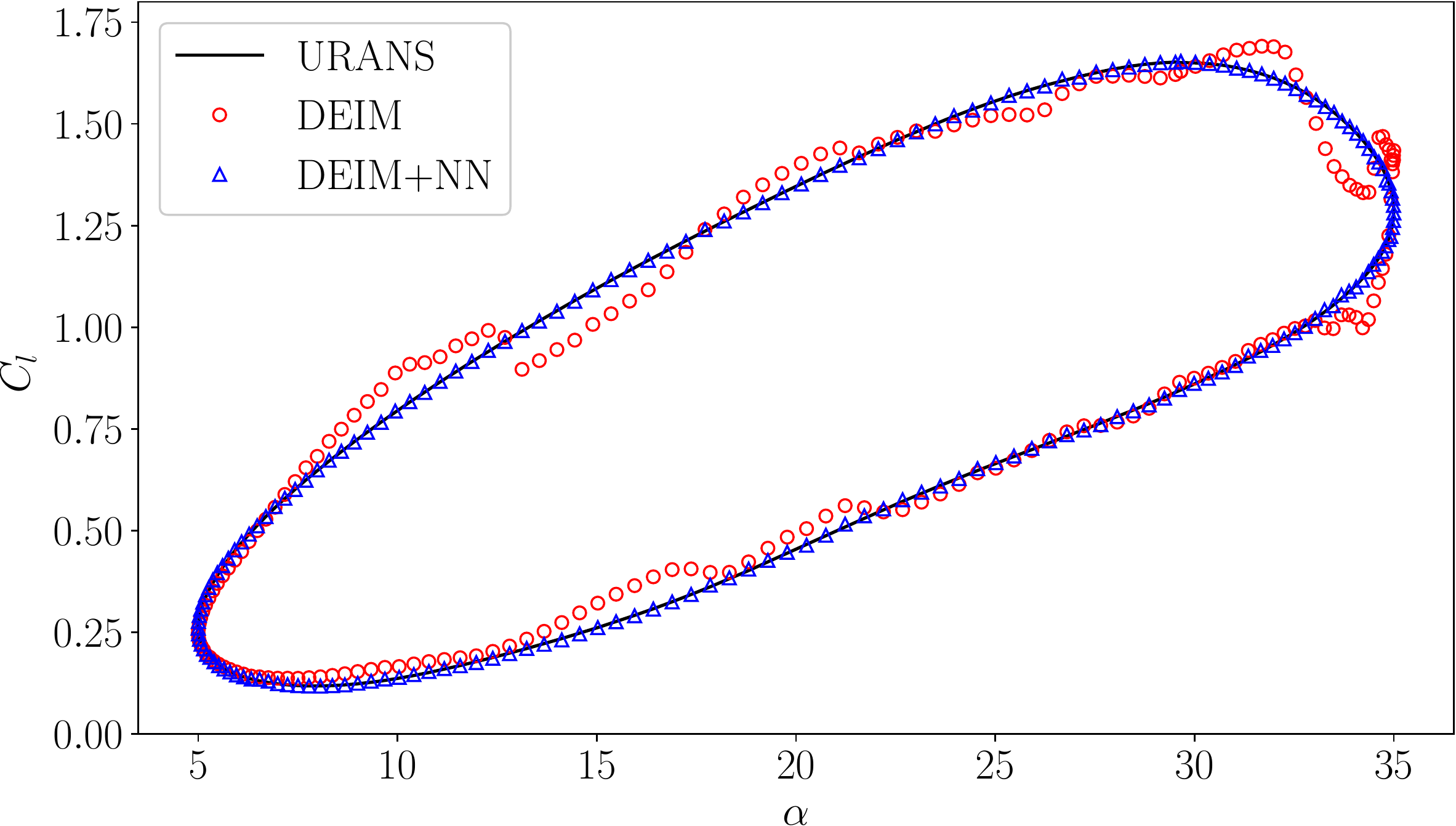}
        \caption{$n_s=15$, $C_l$}
    \end{subfigure}
    \begin{subfigure}[t]{0.49\textwidth}
        \centering
        \includegraphics[width=\linewidth]{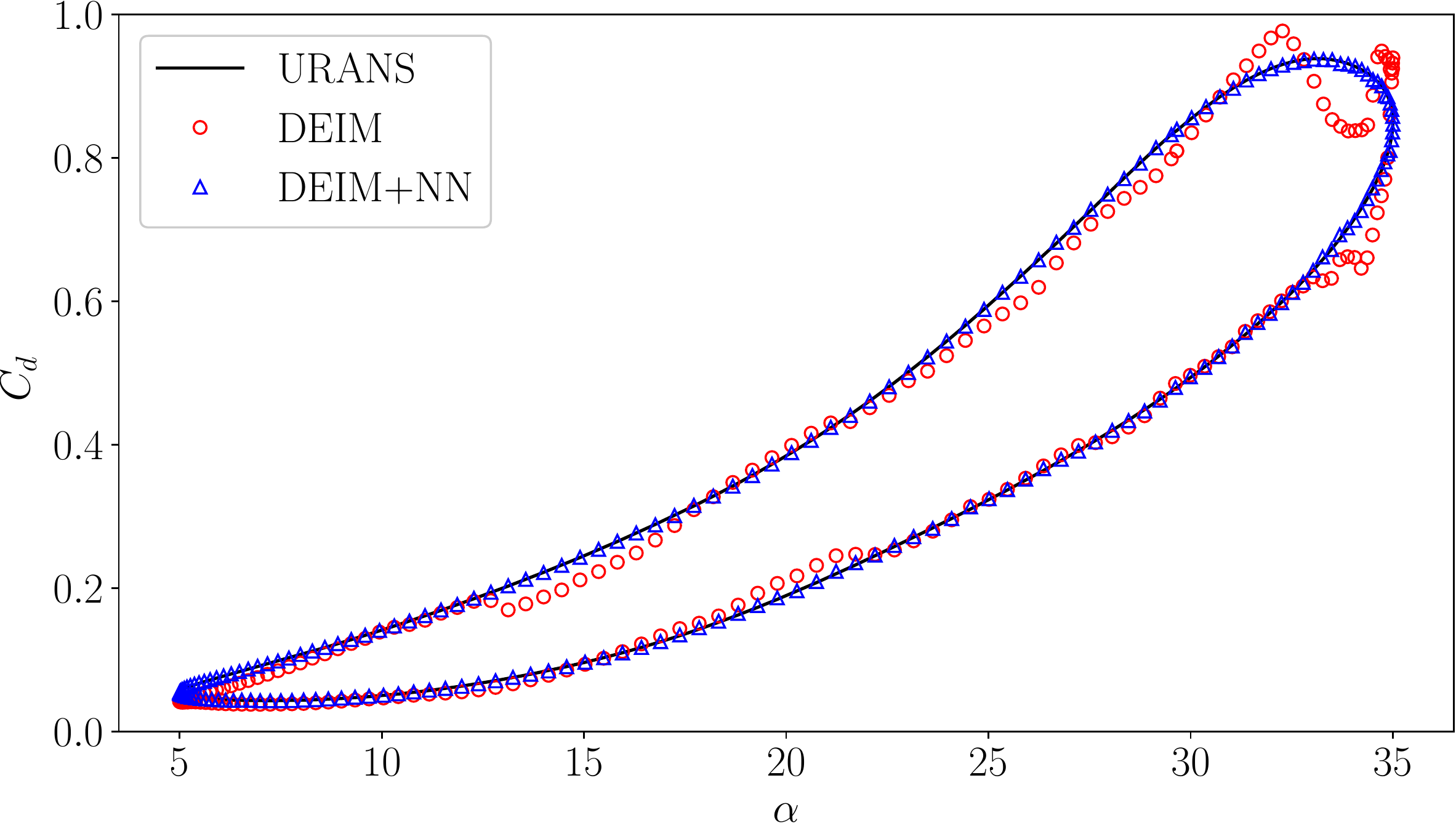}
        \caption{$n_s=15$, $C_d$}
    \end{subfigure}
\caption{3D drone: $C_l, C_d$ w.r.t. $\alpha$ for the testing data without noise in the pressure sensor inputs.
}
\label{fig:3DDrone_aero_coeff_aoa_noise=0}
\end{figure}

\begin{figure}[hbt!]
    \centering
    \begin{subfigure}[t]{0.49\textwidth}
        \centering
        \includegraphics[width=\linewidth]{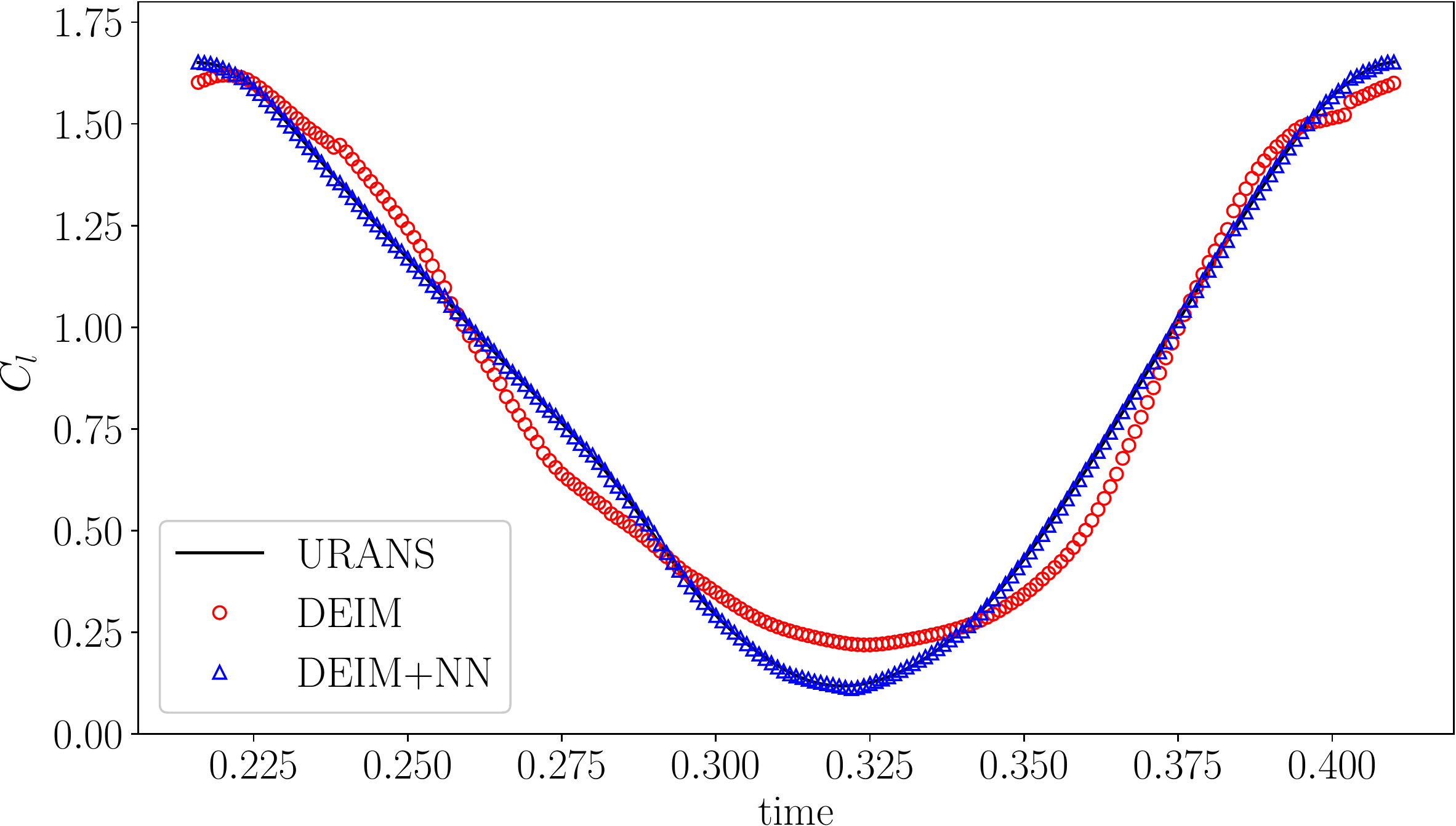}
        \caption{$n_s=5$, $C_l$}
    \end{subfigure}
    \begin{subfigure}[t]{0.49\textwidth}
        \centering
        \includegraphics[width=\linewidth]{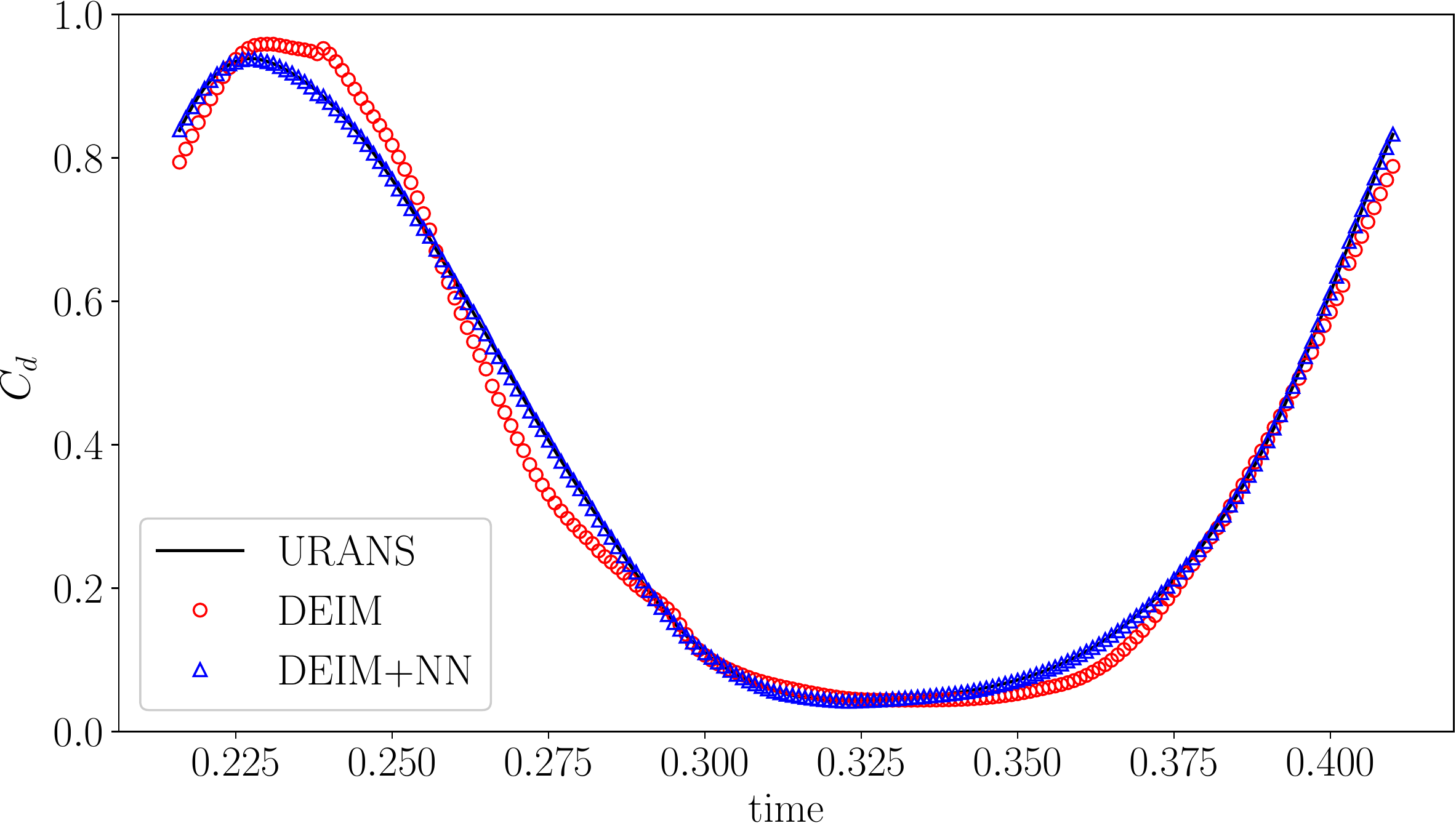}
        \caption{$n_s=5$, $C_d$}
    \end{subfigure}

    \begin{subfigure}[t]{0.49\textwidth}
        \centering
        \includegraphics[width=\linewidth]{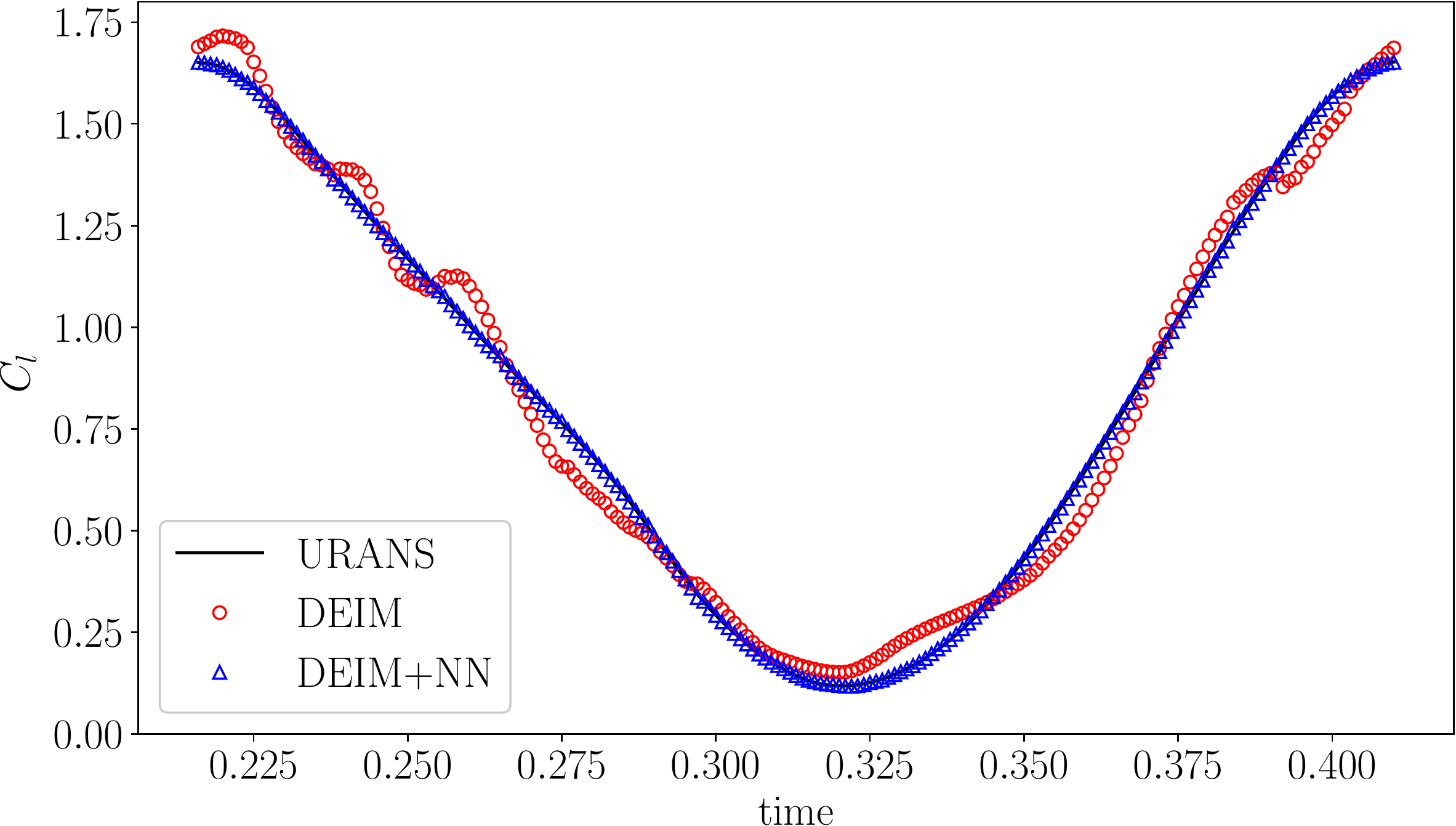}
        \caption{$n_s=10$, $C_l$}
    \end{subfigure}
    \begin{subfigure}[t]{0.49\textwidth}
        \centering
        \includegraphics[width=\linewidth]{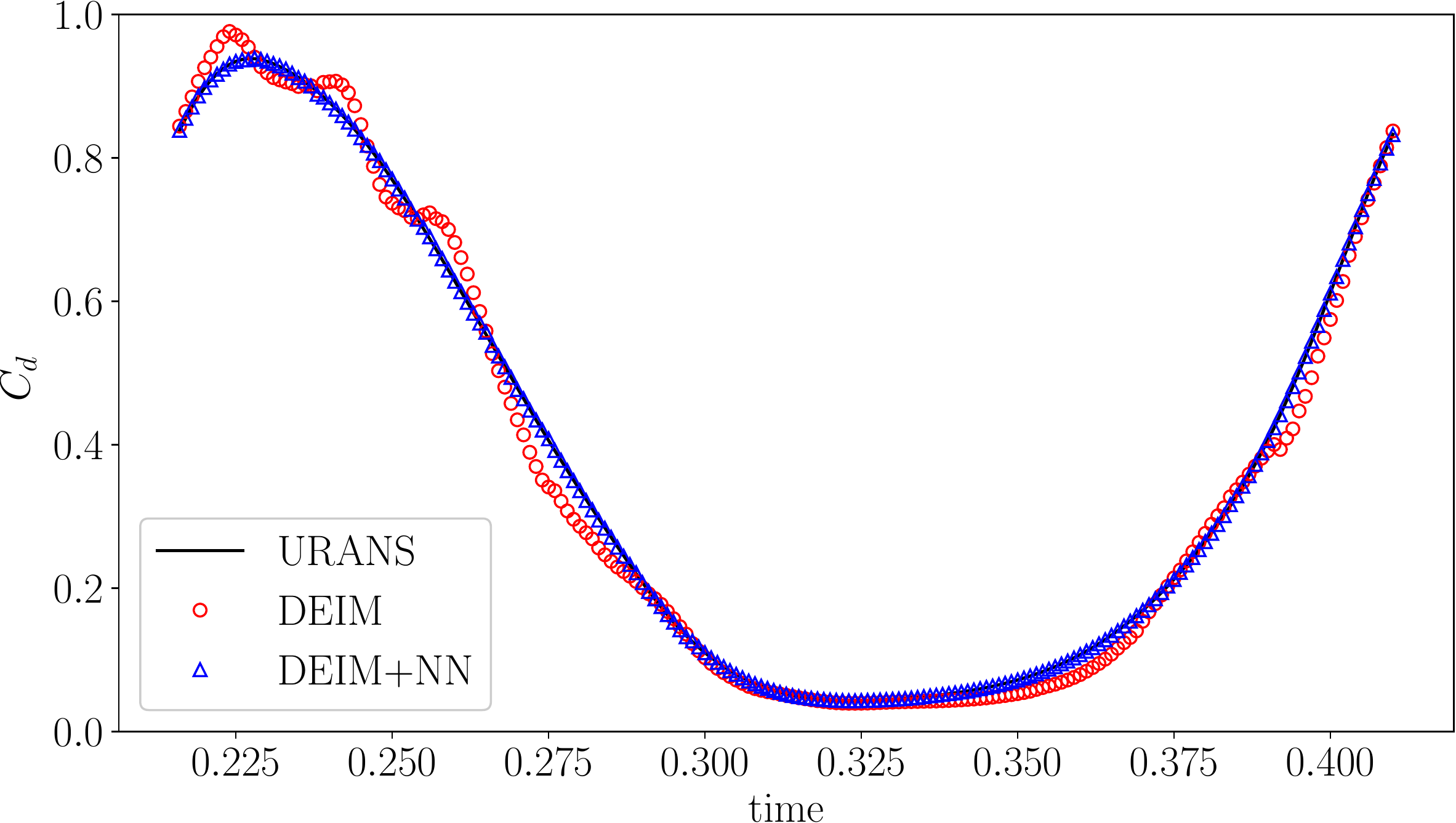}
        \caption{$n_s=10$, $C_d$}
    \end{subfigure}

    \begin{subfigure}[t]{0.49\textwidth}
        \centering
        \includegraphics[width=\linewidth]{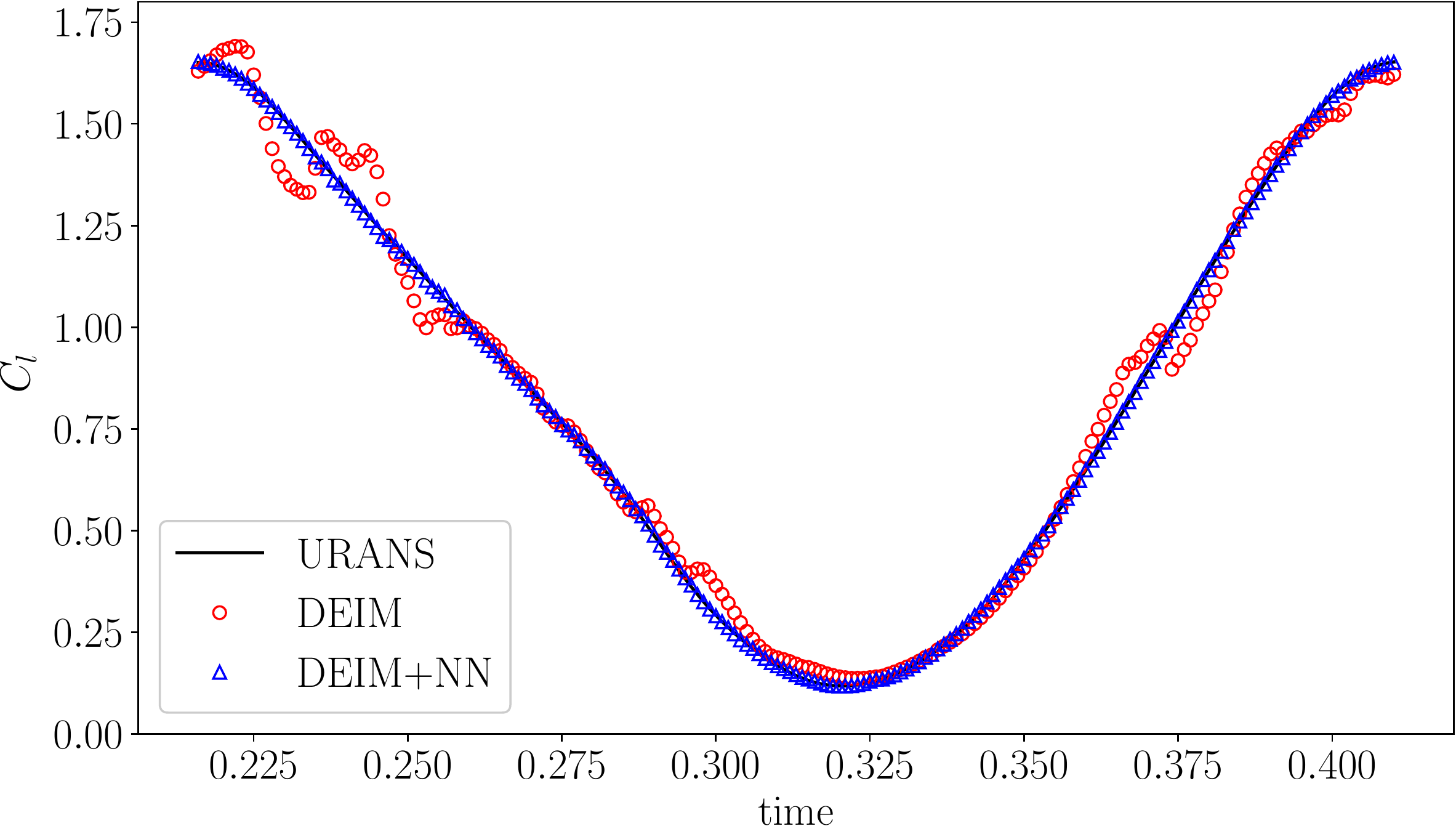}
        \caption{$n_s=15$, $C_l$}
    \end{subfigure}
    \begin{subfigure}[t]{0.49\textwidth}
        \centering
        \includegraphics[width=\linewidth]{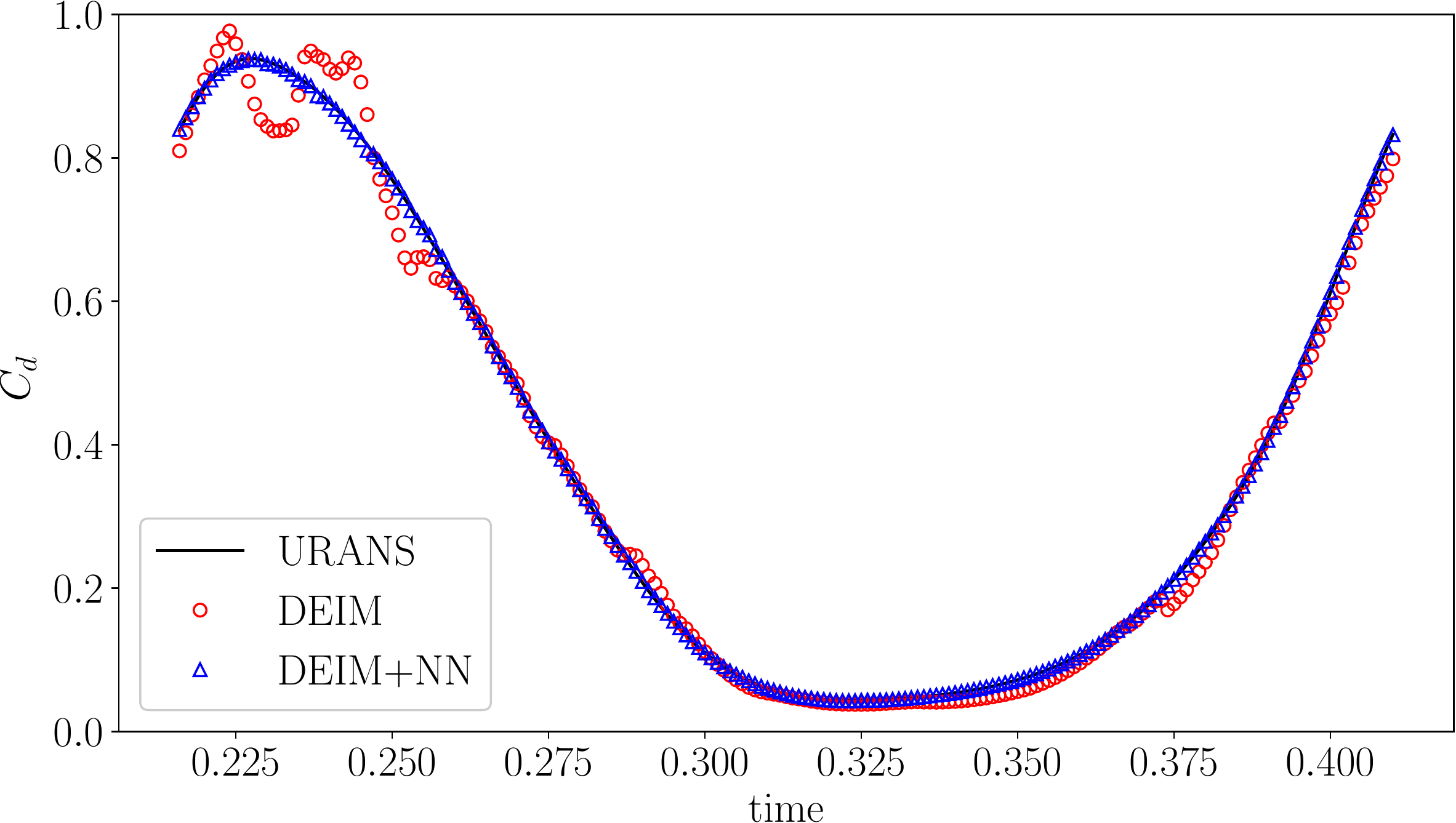}
        \caption{$n_s=15$, $C_d$}
    \end{subfigure}
\caption{3D drone: $C_l, C_d$ w.r.t. time for the testing data  without noise in the pressure sensor inputs.
}
\label{fig:3DDrone_aero_coeff_time_noise=0}
\end{figure}

\begin{figure}[hbt!]
    \centering
    \begin{subfigure}[t]{0.49\textwidth}
        \centering
        \includegraphics[width=\linewidth]{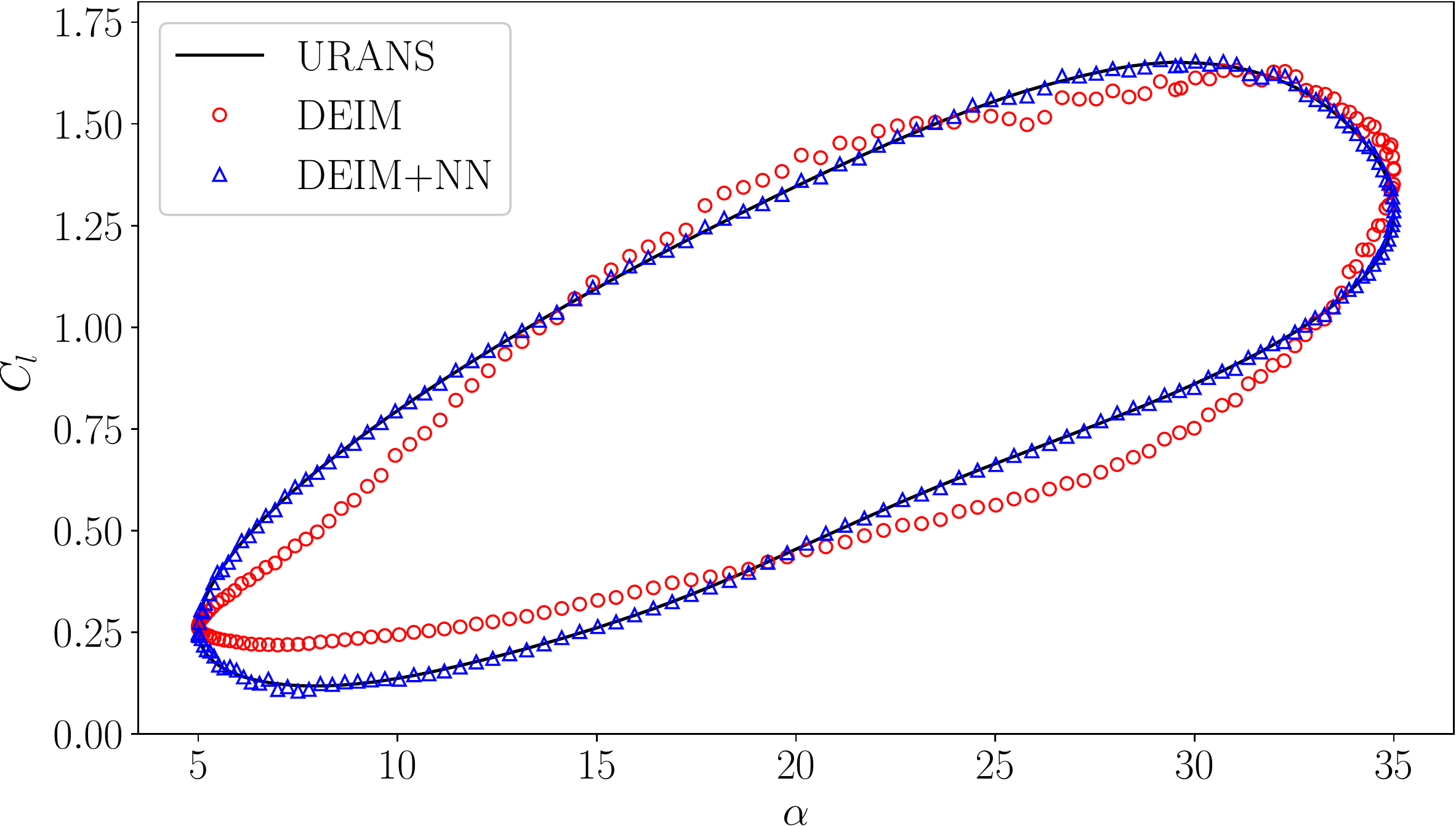}
        \caption{$n_s=5$, $C_l$}
    \end{subfigure}
    \begin{subfigure}[t]{0.49\textwidth}
        \centering
        \includegraphics[width=\linewidth]{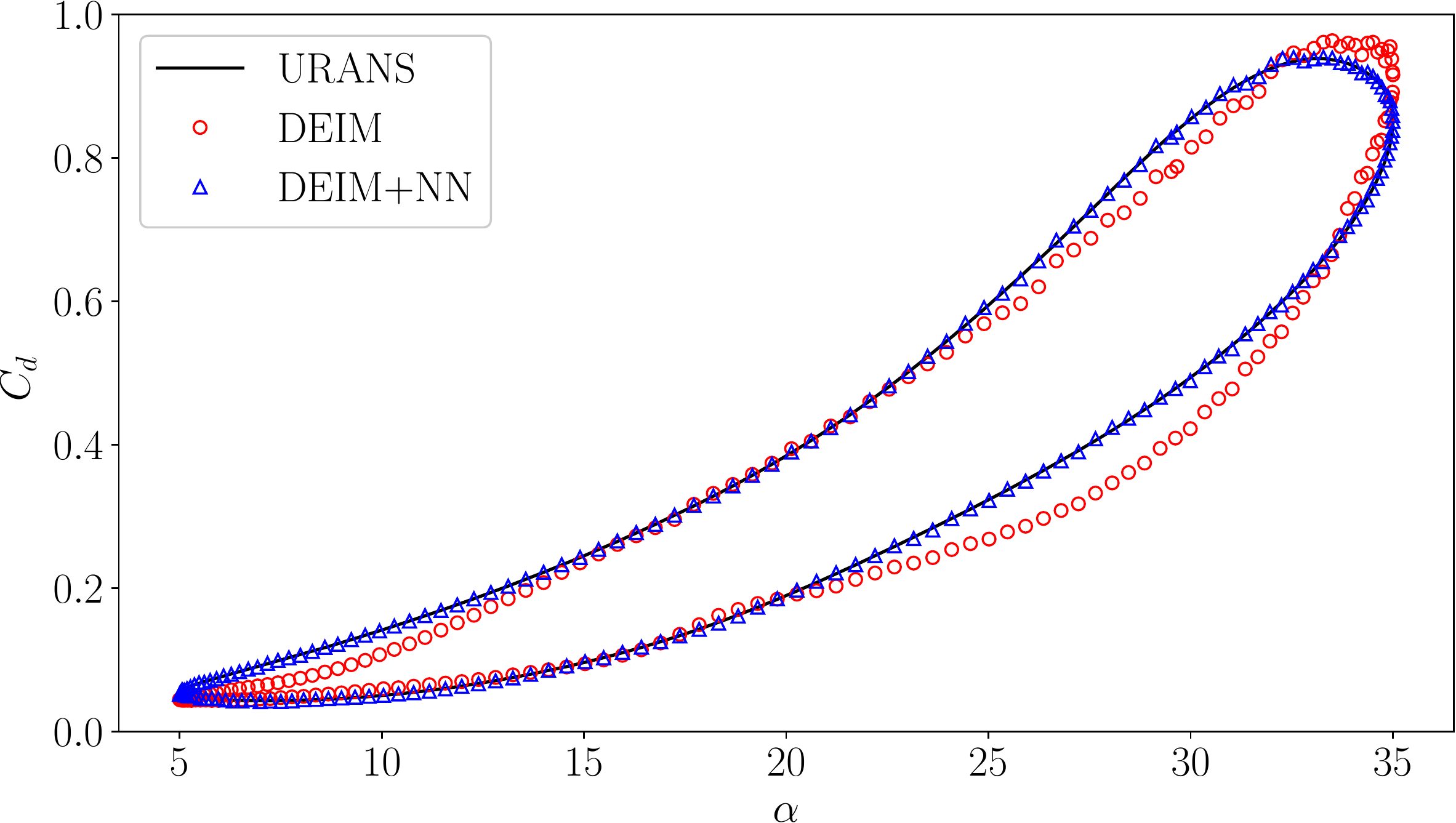}
        \caption{$n_s=5$, $C_d$}
    \end{subfigure}

    \begin{subfigure}[t]{0.49\textwidth}
        \centering
        \includegraphics[width=\linewidth]{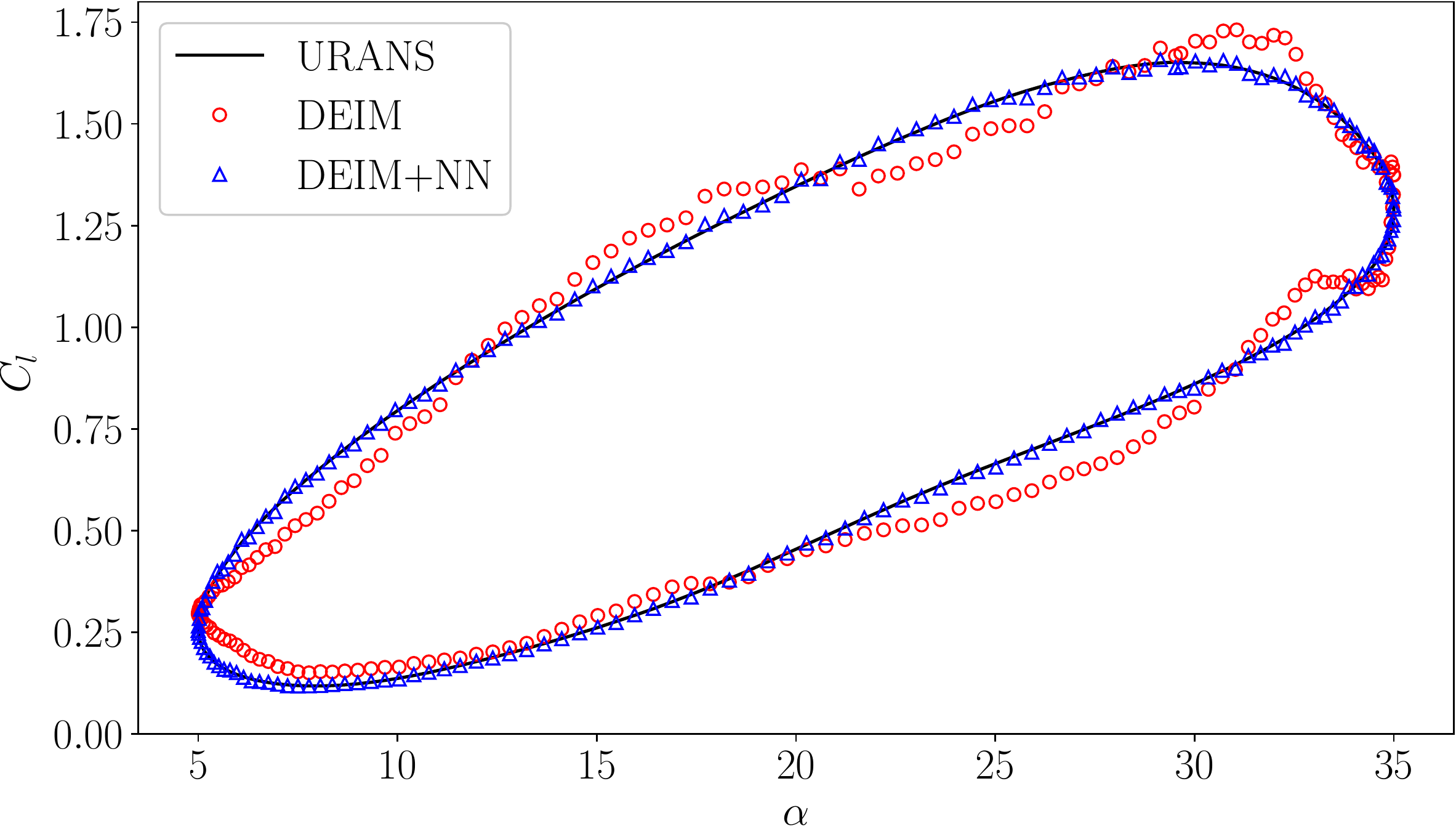}
        \caption{$n_s=10$, $C_l$}
    \end{subfigure}
    \begin{subfigure}[t]{0.49\textwidth}
        \centering
        \includegraphics[width=\linewidth]{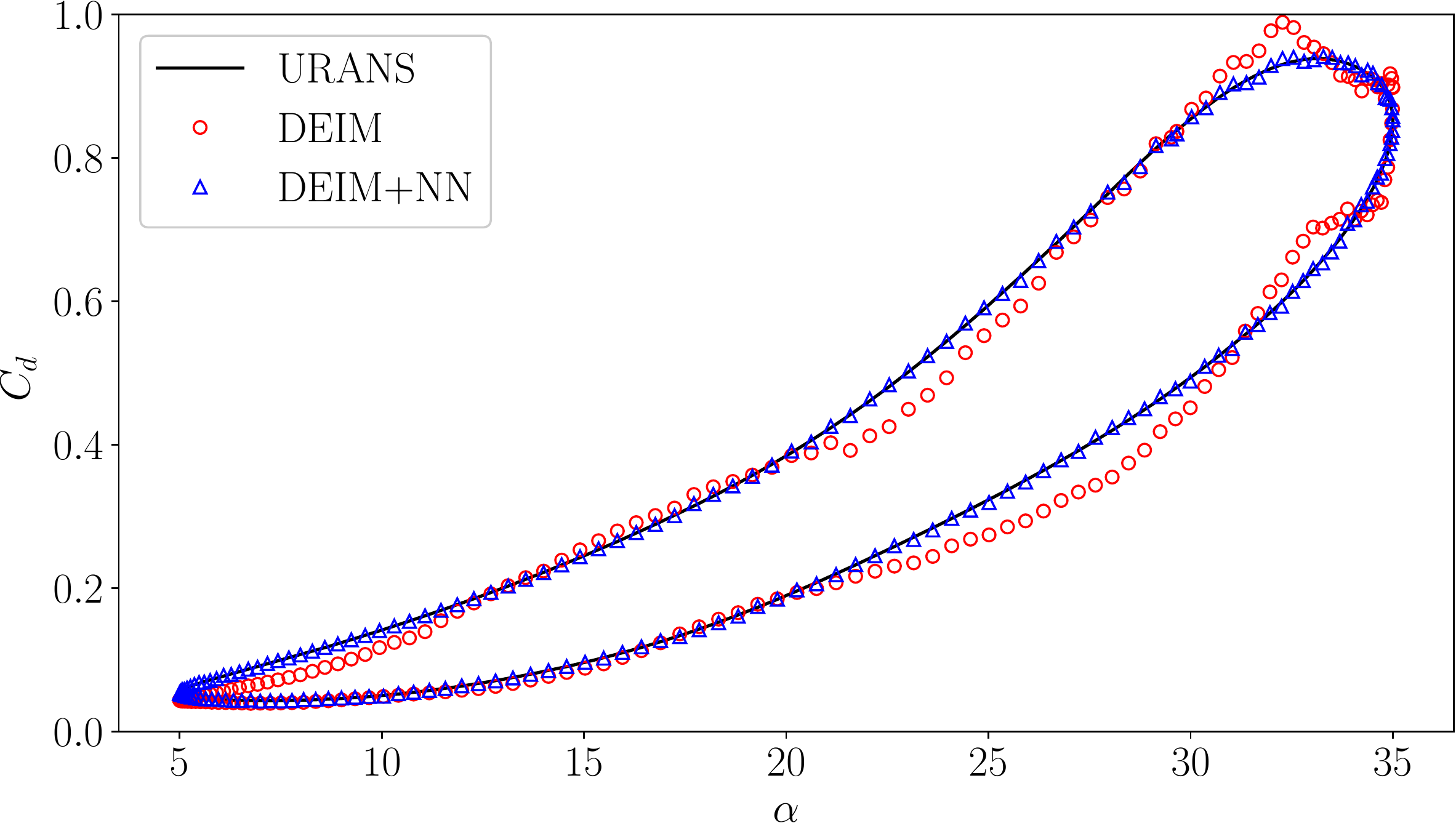}
        \caption{$n_s=10$, $C_d$}
    \end{subfigure}

    \begin{subfigure}[t]{0.49\textwidth}
        \centering
        \includegraphics[width=\linewidth]{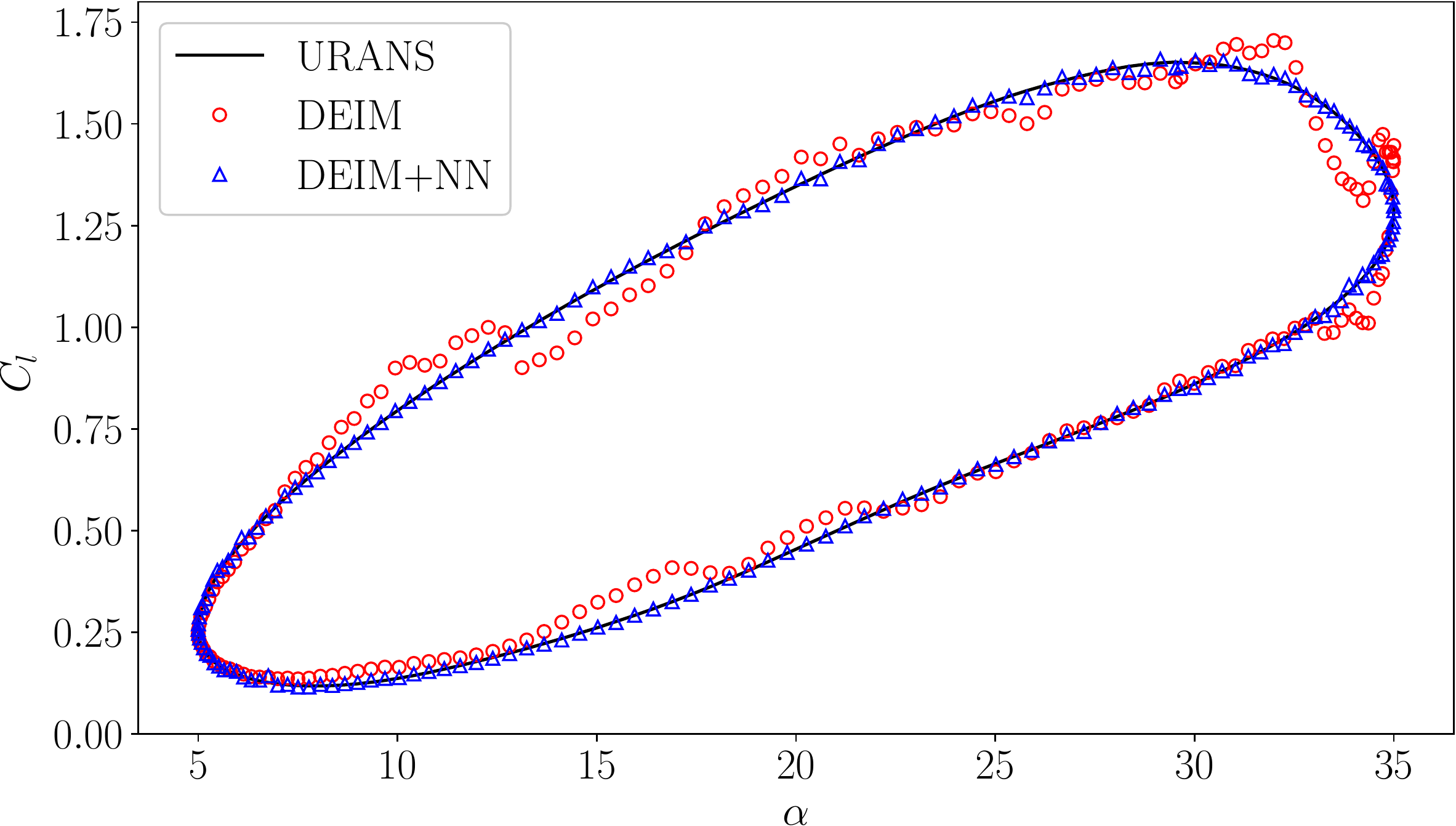}
        \caption{$n_s=15$, $C_l$}
    \end{subfigure}
    \begin{subfigure}[t]{0.49\textwidth}
        \centering
        \includegraphics[width=\linewidth]{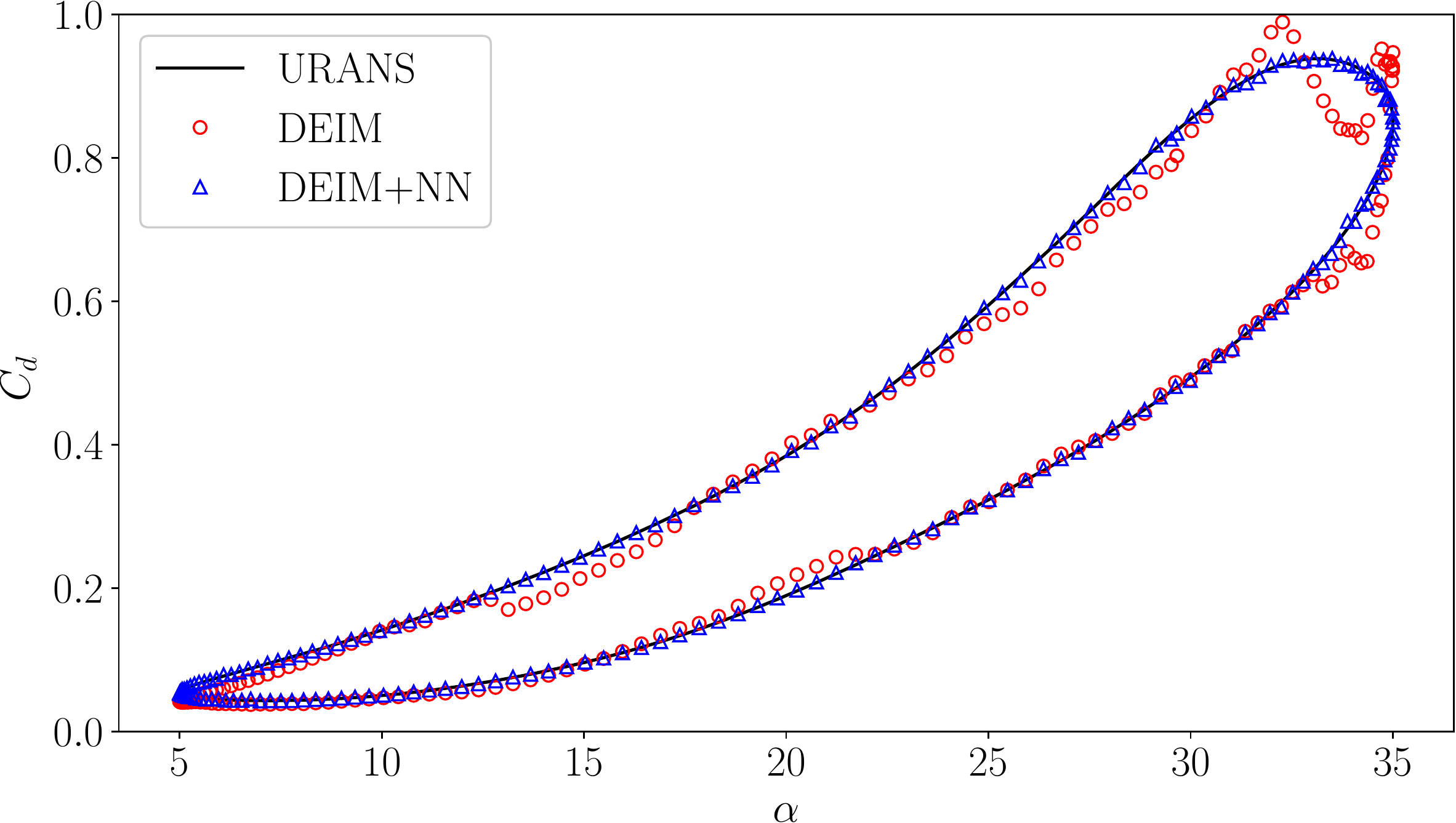}
        \caption{$n_s=15$, $C_d$}
    \end{subfigure}
\caption{3D drone: $C_l, C_d$ w.r.t. $\alpha$ for the testing data with $1.5\%$ noise in the pressure sensor inputs.
}
\label{fig:3DDrone_aero_coeff_aoa_noise=1.5}
\end{figure}

\begin{figure}[hbt!]
    \centering
    \begin{subfigure}[t]{0.49\textwidth}
        \centering
        \includegraphics[width=\linewidth]{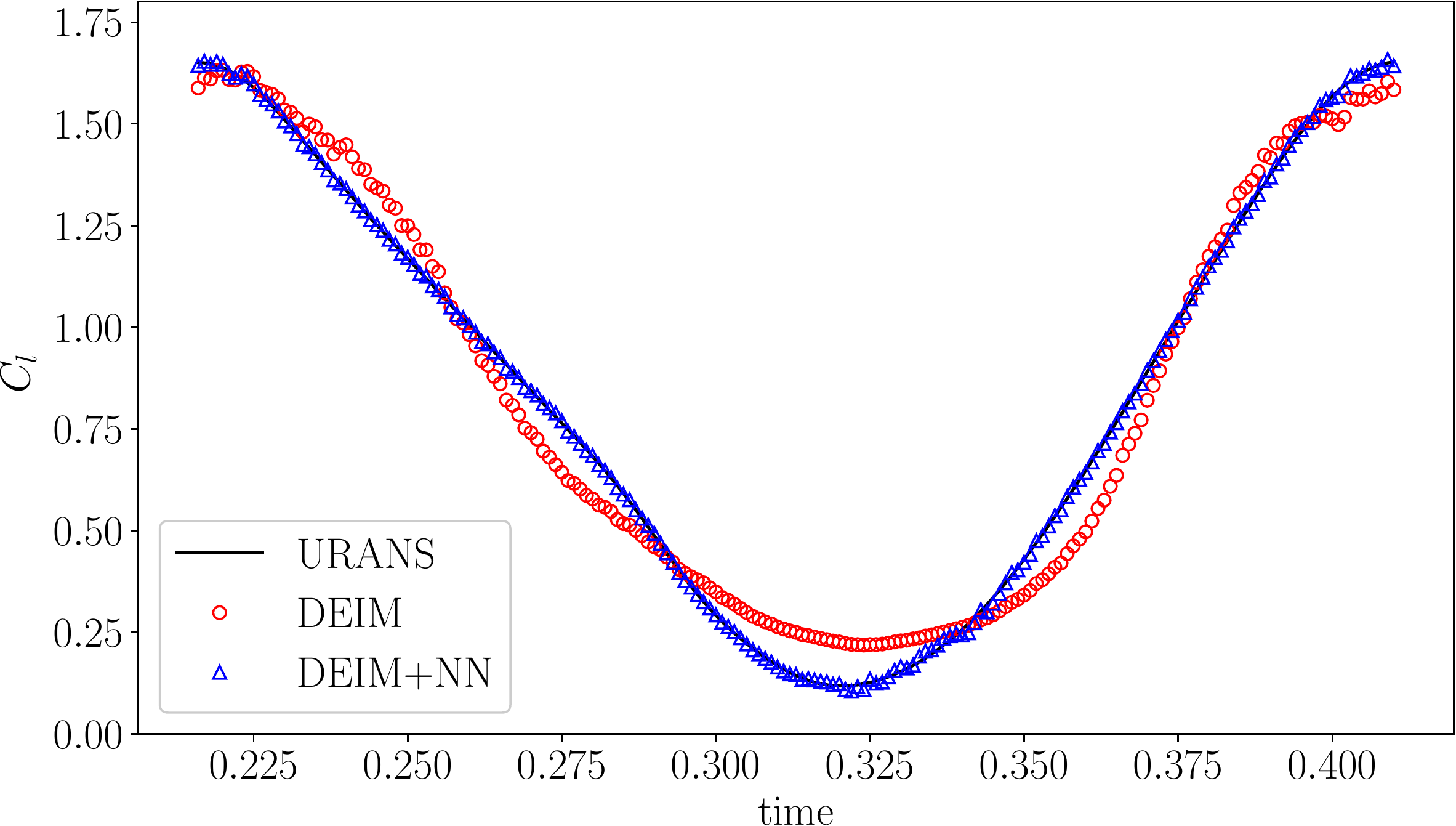}
        \caption{$n_s=5$, $C_l$}
    \end{subfigure}
    \begin{subfigure}[t]{0.49\textwidth}
        \centering
        \includegraphics[width=\linewidth]{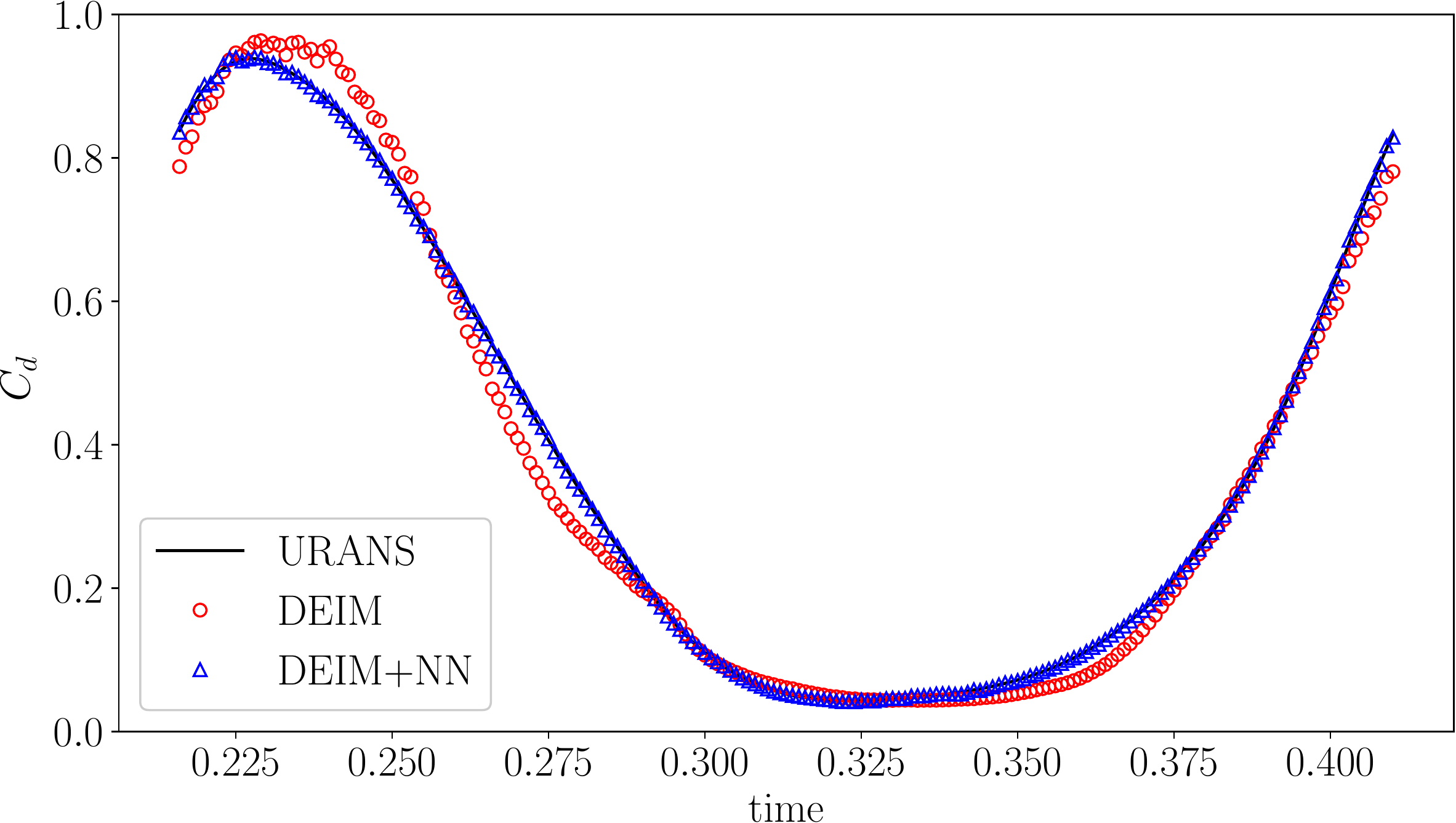}
        \caption{$n_s=5$, $C_d$}
    \end{subfigure}

    \begin{subfigure}[t]{0.49\textwidth}
        \centering
        \includegraphics[width=\linewidth]{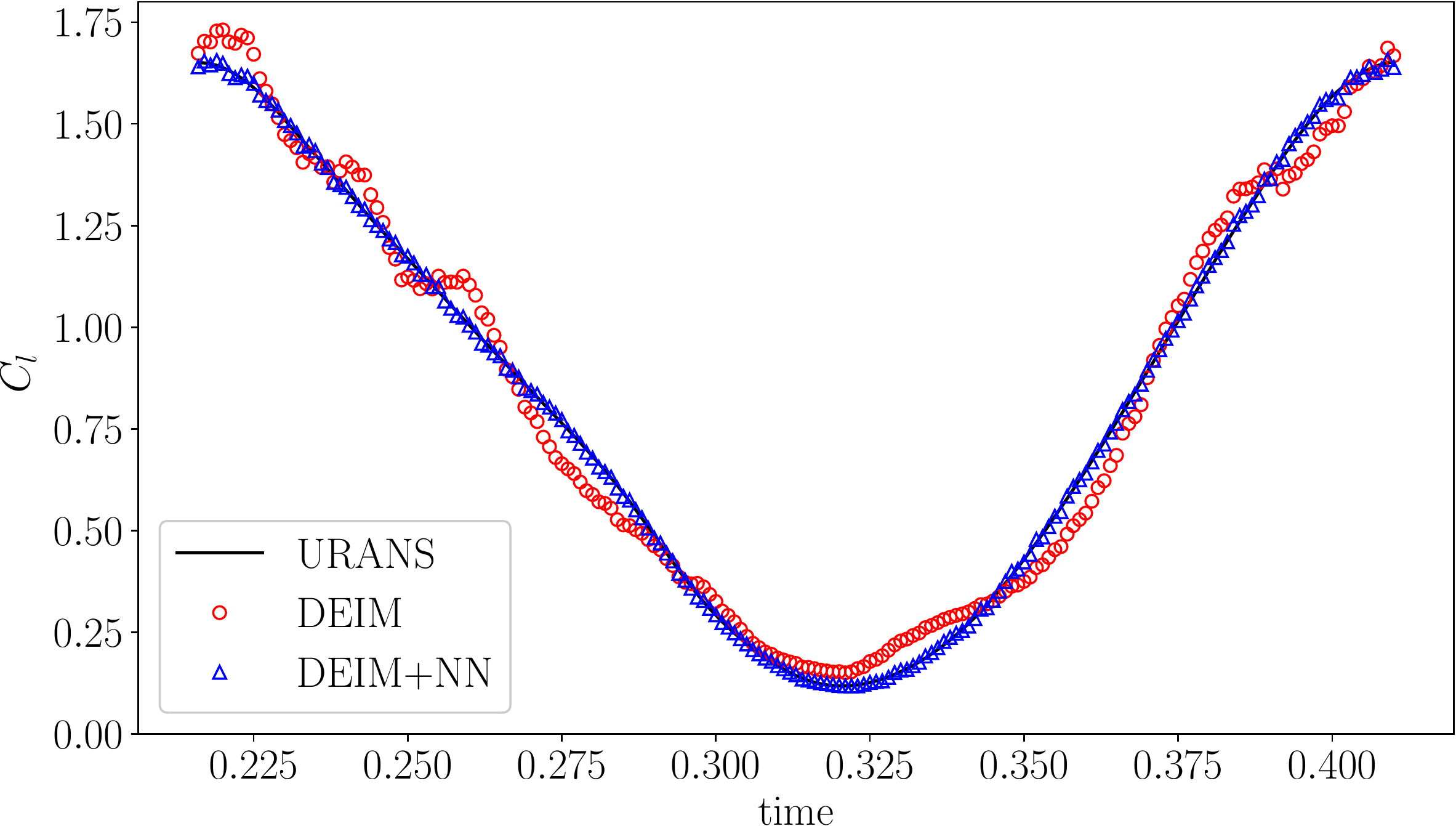}
        \caption{$n_s=10$, $C_l$}
    \end{subfigure}
    \begin{subfigure}[t]{0.49\textwidth}
        \centering
        \includegraphics[width=\linewidth]{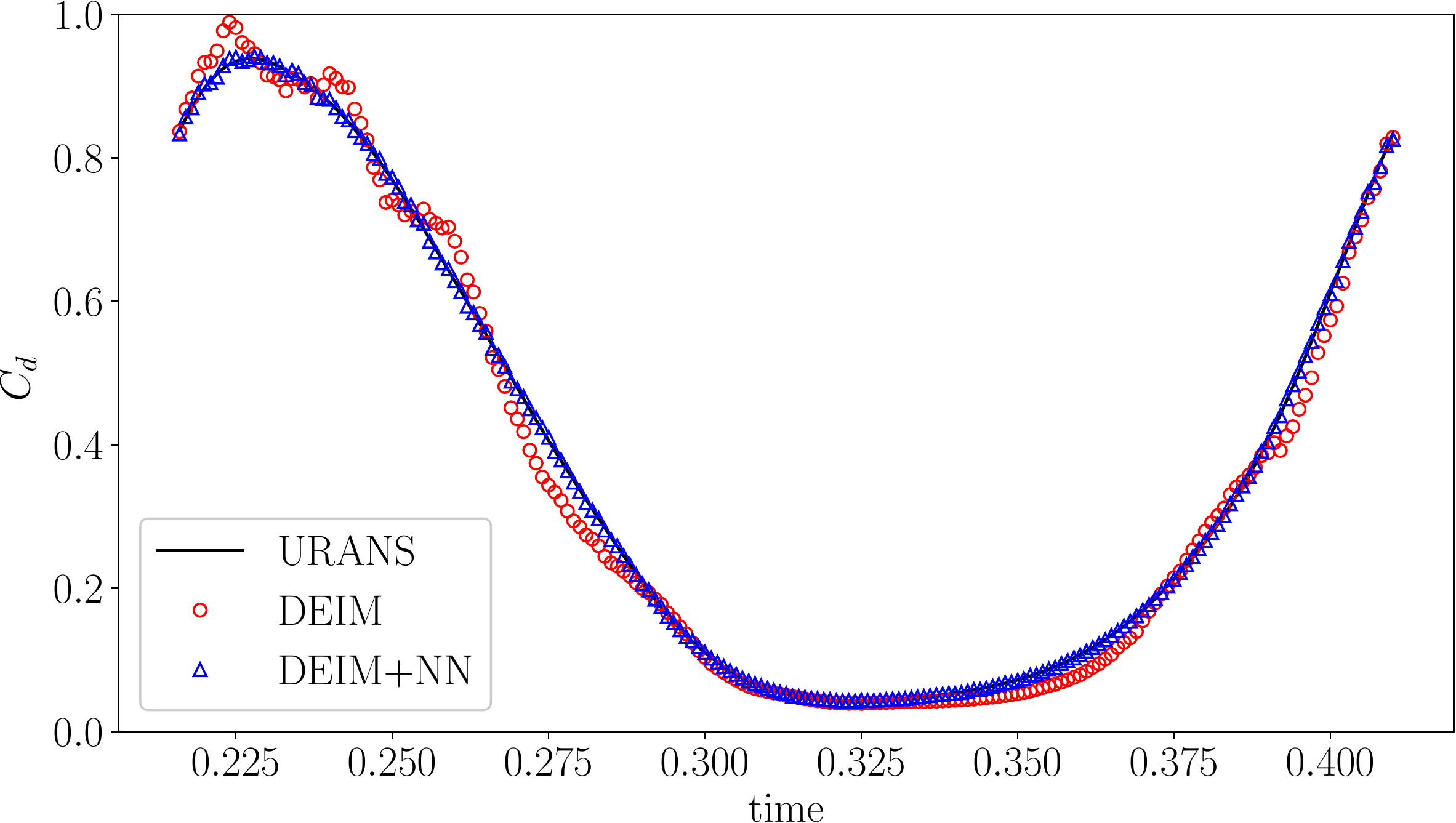}
        \caption{$n_s=10$, $C_d$}
    \end{subfigure}

    \begin{subfigure}[t]{0.49\textwidth}
        \centering
        \includegraphics[width=\linewidth]{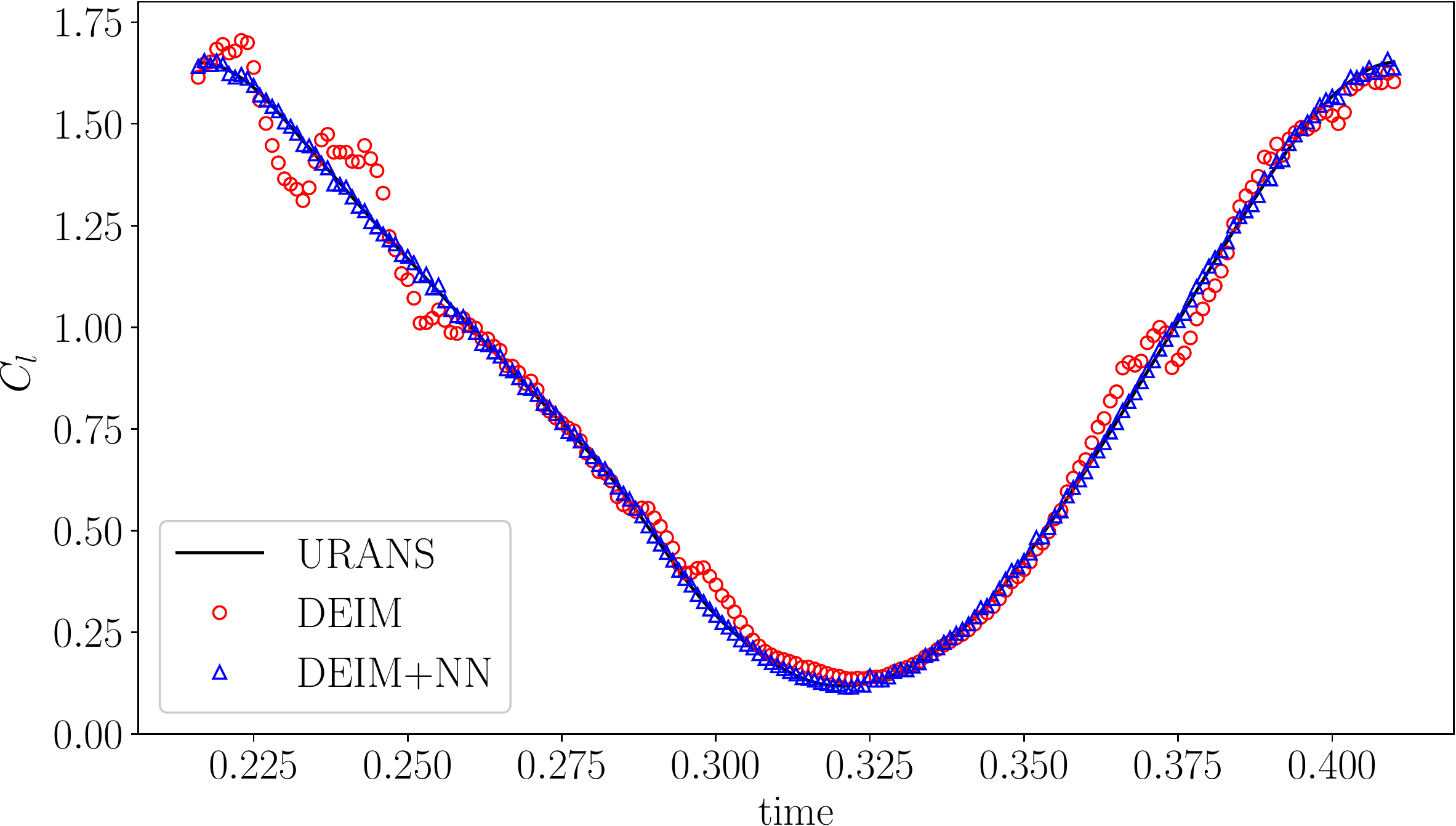}
        \caption{$n_s=15$, $C_l$}
    \end{subfigure}
    \begin{subfigure}[t]{0.49\textwidth}
        \centering
        \includegraphics[width=\linewidth]{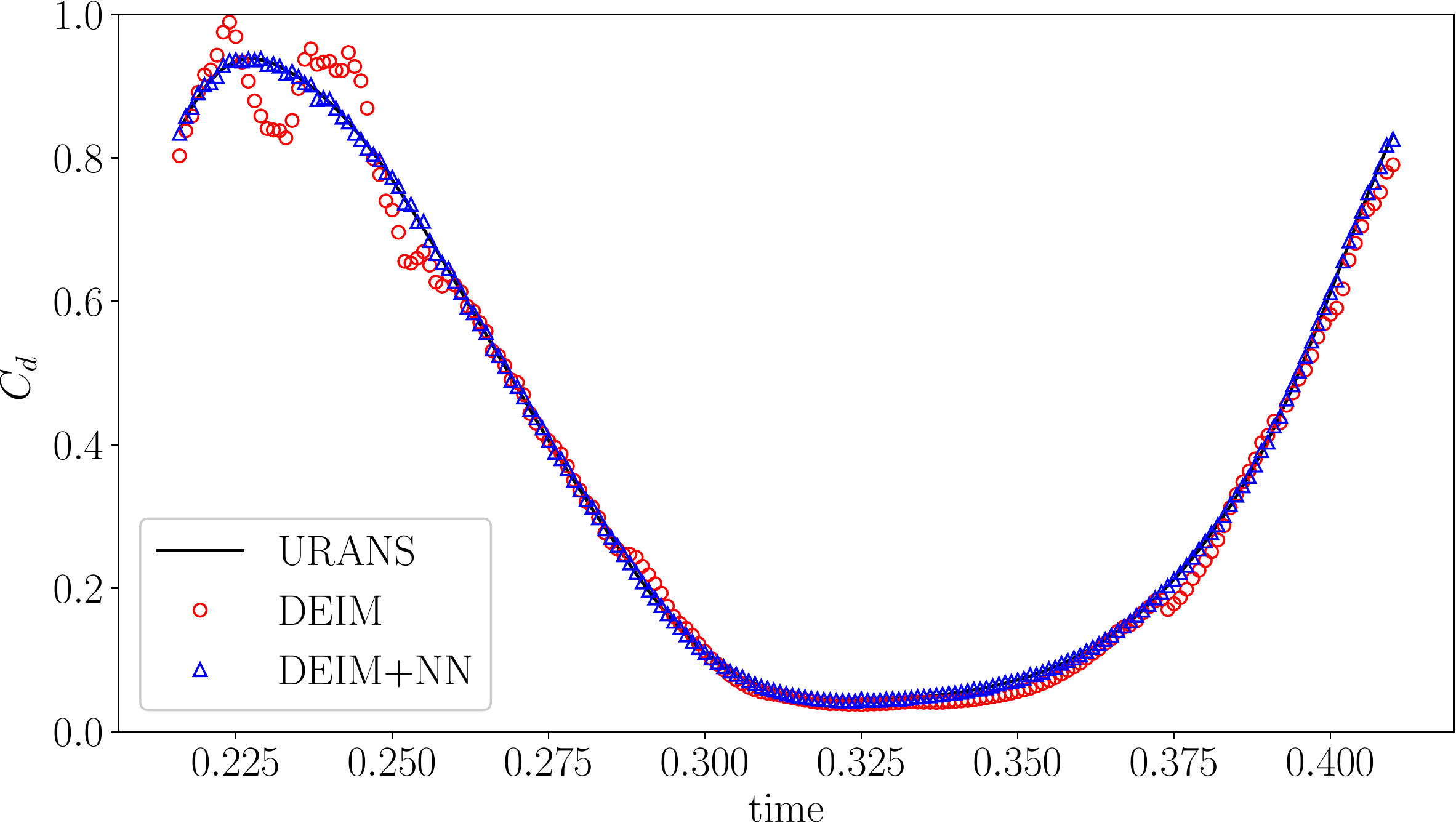}
        \caption{$n_s=15$, $C_d$}
    \end{subfigure}
\caption{3D drone: $C_l, C_d$ w.r.t. time for the testing data  with $1.5\%$ noise in the pressure sensor inputs.
}
\label{fig:3DDrone_aero_coeff_time_noise=1.5}
\end{figure}

In the training of the NN, the mini-batch size is $32$.
We perform a grid search to find a preferred architecture,
with $2$, $3$, $4$ hidden layers,
$10$, $20$, $30$, $40$ neurons each layer,
and $10^{-5}$, $10^{-6}$, $10^{-7}$ weight decay.
The best model is obtained with $n_s=10$, $4$ hidden layers with $40$ neurons in each layer and a weight decay as $10^{-6}$.
Figure \ref{fig:3DDrone_aero_coeff_aoa_noise=0} gives the lift and drag coefficients $C_l,C_d$ with respect to the angle of attack $\alpha$ for different $n_s$,
and Fig. \ref{fig:3DDrone_aero_coeff_time_noise=0} plots the evolution of $C_l, C_d$ with respect to time.
The numerical results are only shown during one whole period,
as the URANS simulation is periodic in time.
The maximal $C_l$ appears at $\alpha\approx \qty{30}{deg}$,
and the drone is near the region of dynamic stall with $\alpha$ in $\left[28,35\right] \text{deg}$.
The lift and drag coefficients $C_l, C_d$ predicted by the DEIM model deviate from the URANS simulation results,
and the results do not improve with larger $n_s$.
After adding the NN correction, the results are much better,
and very close to the URANS simulation,
thus the NN correction term is vital in the improvement of the accuracy.
The results with $1.5\%$ noise in the pressure sensor inputs are also shown in Figs. \ref{fig:3DDrone_aero_coeff_aoa_noise=1.5}-\ref{fig:3DDrone_aero_coeff_time_noise=1.5}.
One observes that the lift and drag coefficients are still well predicted by the DEIM+NN without obvious oscillations.

Similar to the 2D airfoil case,
the $\ell^2$ and $\ell^\infty$ errors in $C_l, C_d$ and corresponding $\alpha$ are computed to examine the performance of the proposed approach, listed in Tables \ref{tab:3DDrone_lift_err}-\ref{tab:3DDrone_drag_err}.
One observes that when no noise is added, the smallest errors in $C_l$ and $C_d$ by adding the NN correction are \num{1.67e-3} and \num{7.32e-4}, respectively,
and the DEIM+NN yields at least $30$ times more accurate $C_l$ and $C_d$ than the DEIM,
which verifies the high accuracy of the proposed approach.
When $1.5\%$ noise is added to the pressure sensor inputs,
the errors in the lift and drag coefficients are still below \num{1.0e-2},
suggesting that the approach is robust to noise.
The online CPU time costs in Table \ref{tab:CPU_times} confirm that the DEIM+NN model is efficient and can be used for real-time prediction.

\begin{table}[hbt!]
\caption{3D drone: the $\ell^2$ errors, $\ell^\infty$ errors in $C_l$, and the angles of attack corresponding to the $\ell^\infty$ errors for different $n_s$.}
\label{tab:3DDrone_lift_err}
\centering
\begin{tabular}{cccccccc}
\hline\hline
& & \multicolumn{3}{c}{DEIM} & \multicolumn{3}{c}{DEIM+NN} \\
\cmidrule(lr){3-5}\cmidrule(lr){6-8}
& $n_s$ & $\epsilon_{l}^{\tt DEIM}$ & $\epsilon_{l,\infty}^{\tt DEIM}$ & $\alpha(\epsilon_{l,\infty}^{\tt DEIM})$ & $\epsilon_{l}^{\tt NN}$ & $\epsilon_{l,\infty}^{\tt NN}$ & $\alpha(\epsilon_{l,\infty}^{\tt NN})$ \\
\hline
\multirow{3}{*}{without noise} & $5$ & \num{7.46e-02} & \num{1.47e-01} & \qty{8.0}{deg} & \num{2.25e-03} & \num{7.36e-03} & \qty{7.5}{deg}\\
& $10$ & \num{5.66e-02} & \num{1.10e-01} & \qty{28.1}{deg} & \num{1.67e-03} & \num{7.30e-03} & \qty{17.7}{deg}\\
& $15$ & \num{5.28e-02} & \num{1.55e-01} & \qty{35.0}{deg} & \num{2.78e-03} & \num{1.11e-02} & \qty{34.9}{deg}\\
\hline
\multirow{3}{*}{$1.5\%$ noise} & $5$ & \num{7.56e-02} & \num{1.51e-01} & \qty{8.0}{deg} & \num{6.27e-03} & \num{2.23e-02} & \qty{5.0}{deg}\\
& $10$ & \num{5.76e-02} & \num{1.10e-01} & \qty{32.3}{deg} & \num{6.35e-03} & \num{1.82e-02} & \qty{25.8}{deg}\\
& $15$ & \num{5.37e-02} & \num{1.62e-01} & \qty{35.0}{deg} & \num{6.18e-03} & \num{1.94e-02} & \qty{34.8}{deg}\\
\hline\hline
\end{tabular}
\end{table}

\begin{table}[hbt!]
\caption{3D drone: the $\ell^2$ errors, $\ell^\infty$ errors in $C_d$, and the angles of attack corresponding to the $\ell^\infty$ errors for different $n_s$.}
\label{tab:3DDrone_drag_err}
\centering
\begin{tabular}{cccccccc}
\hline\hline
& & \multicolumn{3}{c}{DEIM} & \multicolumn{3}{c}{DEIM+NN} \\
\cmidrule(lr){3-5}\cmidrule(lr){6-8}
& $n_s$ & $\epsilon_{d}^{\tt DEIM}$ & $\epsilon_{d,\infty}^{\tt DEIM}$ & $\alpha(\epsilon_{d,\infty}^{\tt DEIM})$ & $\epsilon_{d}^{\tt NN}$ & $\epsilon_{d,\infty}^{\tt NN}$ & $\alpha(\epsilon_{d,\infty}^{\tt NN})$ \\
\hline
\multirow{3}{*}{without noise} & $5$ & \num{3.22e-02} & \num{7.67e-02} & \qty{28.9}{deg} & \num{7.61e-04} & \num{3.71e-03} & \qty{33.7}{deg}\\
& $10$ & \num{2.61e-02} & \num{6.93e-02} & \qty{28.1}{deg} & \num{7.32e-04} & \num{4.19e-03} & \qty{34.8}{deg}\\
& $15$ & \num{2.88e-02} & \num{9.38e-02} & \qty{33.9}{deg} & \num{1.41e-03} & \num{7.97e-03} & \qty{34.9}{deg}\\
\hline
\multirow{3}{*}{$1.5\%$ noise} & $5$ & \num{3.26e-02} & \num{7.82e-02} & \qty{34.9}{deg} & \num{1.82e-03} & \num{8.35e-03} & \qty{32.3}{deg}\\
& $10$ & \num{2.63e-02} & \num{6.52e-02} & \qty{28.1}{deg} & \num{2.48e-03} & \num{8.95e-03} & \qty{34.8}{deg}\\
& $15$ & \num{2.92e-02} & \num{9.70e-02} & \qty{35.0}{deg} & \num{2.51e-03} & \num{1.16e-02} & \qty{34.8}{deg}\\
\hline\hline
\end{tabular}
\end{table}
 
\begin{table}[hbt!]
\caption{Averaged CPU times (\unit{s}) for the prediction of lift and drag coefficients.
The tests are performed on a Linux server with Intel\textsuperscript{\textregistered} Xeon\textsuperscript{\textregistered} Gold 6148 CPU @ 2.40GHz.}
\label{tab:CPU_times}
\centering
\begin{tabular}{cccccc}
\hline\hline
\multicolumn{3}{c}{2D airfoil} & \multicolumn{3}{c}{3D drone} \\
\cmidrule(lr){1-3}\cmidrule(lr){4-6}
$n_s$ & DEIM & NN & $n_s$ & DEIM & NN \\
\hline
$5$ & $2.41\times 10^{-6}$ & $7.89\times 10^{-5}$ & $5$ & $3.60\times 10^{-6}$ & $1.19\times 10^{-4}$ \\
$8$ & $2.28\times 10^{-6}$ & $7.05\times 10^{-5}$ & $10$ & $2.41\times 10^{-6}$ & $7.89\times 10^{-5}$ \\
$10$ & $2.86\times 10^{-6}$ & $7.30\times 10^{-5}$ & $15$ & $2.41\times 10^{-6}$ & $7.89\times 10^{-5}$ \\
\hline\hline
\end{tabular}
\end{table}

\section{Conclusion}\label{sec:Conclusion}
In the navigation and control of UAVs,
accurate and efficient real-time aerodynamic prediction based on sensor inputs plays an important role.
This paper presents a systematic approach for the construction of a data-driven aerodynamic model combined with discrete empirical interpolation method (DEIM) to predict aerodynamic coefficients.
Pressure coefficients on the aircraft surface from URANS simulations serve as snapshots
and are used to obtain a set of reduced basis.
The sensor locations are optimized by the DEIM,
and the basis coefficients are computed based on real-time pressure sensor inputs at the selected locations.
The aerodynamic forces are computed by integrating the reduced basis reconstruction of the surface pressure distribution.
The DEIM model captures main dynamics,
and a neural network is proposed to bridge the gap between the aerodynamic forces of the ground truth and the DEIM prediction,
resulting in a more accurate model.
The approach is tested on the URANS simulation and experimental data of the dynamic stall of a 2D NACA0015 airfoil,
and URANS simulation data of the dynamic stall of a 3D drone.
The results show that the approach yields fast and accurate real-time predictions of the aerodynamic coefficients,
and the neural network-based correction term improves the accuracy significantly.
Furthermore, we demonstrated that the approach is not sensitive to noise in the pressure sensor inputs.

% \section*{Appendix}

% An Appendix, if needed, appears \textbf{before} research funding information and other acknowledgments.

\section*{Funding Sources}
The first and second authors were partially supported by the Sense Dynamics Project (SD11, Collaborative Data Science Project, Swiss Data Science Center). The second author is also partially supported by Project U2230402 of NSFC.

\section*{Acknowledgments}
The authors thank Dr. Guosheng He and Prof. Karen Mulleners for providing the wind tunnel experiments data. We are grateful for the inspiring discussions with Dr. Nicol{\`{o}} Ripamonti. 

\bibliography{reference}

\end{document}